\RequirePackage{silence}
\newcommand{\thesisTheme}{ucph} % to colortheme and titlepage image

\documentclass[%
	table,
	paper=A4,					% paper size
	12pt,						% font size
	twoside=false,				% two-sided printing
	openright,					% doublepage cleaning ends up right side
	parskip=full,				% spacing value / method for paragraphs
	chapterprefix=true,			% prefix for chapter marks
	headings=normal,			% size of headings
	bibliography=totoc,			% include bib in toc
	listof=totoc,				% include listof entries in toc
	titlepage=on,				% own page for each title page
	captions=tableabove,		% display table captions above the float env
	draft=false,				% value for draft version
	abstract=on,                % abstract title on/off
]{scrreprt}
\usepackage [table]{xcolor}
\usepackage[utf8]{inputenc}
\usepackage[english]{babel}     % adjust the language
\DeclareOldFontCommand{\bf}{\normalfont\bfseries}{\mathbf}
\DeclareUnicodeCharacter{0301}{\'{e}}
\usepackage{rotating}
\newcommand{\specialcell}[2][c]{%
  \begin{tabular}[#1]{@{}c@{}}#2\end{tabular}}

\usepackage{amsmath}
\usepackage{euscript}
\usepackage{latexsym}
\usepackage{tabularx}
\usepackage{booktabs}       % professional-quality tables
\usepackage{graphicx}       %Required
\usepackage{amsfonts}       % blackboard math symbols
\usepackage{natbib}
\usepackage{microtype}
\newcolumntype{P}[1]{>{\centering\arraybackslash}p{#1}}
\usepackage{lscape}
\usepackage{arydshln}

\usepackage{multirow, makecell}
\usepackage{tikz}
\usepackage{color, colortbl}
\usepackage{comment}
\usepackage{soul}
\usetikzlibrary{calc}
\usetikzlibrary{decorations.pathmorphing}
\newcommand{\datasetname}{\textit{SufficientFacts}}
\newcommand{\taskname}{Evidence Sufficiency Prediction}

\usepackage{colortbl}
\definecolor{linksblue}{RGB}{2, 82, 189}

\definecolor{mygreen}{RGB}{179, 253, 179}
\DeclareRobustCommand{\hlviolet}[1]{{\sethlcolor{mygreen}\hl{#1}}}
\definecolor{myred}{RGB}{255, 205, 205}
\DeclareRobustCommand{\hlred}[1]{{\sethlcolor{myred}\hl{#1}}}
\definecolor{myblue}{RGB}{204, 229, 255}
\DeclareRobustCommand{\hlblue}[1]{{\sethlcolor{myblue}\hl{#1}}}
\definecolor{myyellow}{RGB}{253, 253, 153}
\DeclareRobustCommand{\hlyellow}[1]{{\sethlcolor{myyellow}\hl{#1}}}

\definecolor{mylgreen}{RGB}{181, 234, 245}
\DeclareRobustCommand{\hlightgreen}[1]{{\sethlcolor{mylgreen}\hl{#1}}}
\definecolor{mydblue}{RGB}{20, 136, 199}
\DeclareRobustCommand{\hldblue}[1]{{\sethlcolor{mydblue}\hl{#1}}}

\definecolor{bad_res}{HTML}{800000}

\definecolor{orange}{HTML}{D55E00}
\definecolor{blue_editing}{HTML}{56B4E9}
\definecolor{green_editing}{HTML}{009E73}
\definecolor{purple_editing}{HTML}{882255}
\newcommand{\politihop}{$\mathtt{PolitiHop}$} 
\usepackage{url}
\usepackage{amsthm}
\usepackage{algorithm}
\usepackage{algorithmic}
\usepackage{multicol}
\usepackage{paralist} 
\usepackage{amssymb} 
\usepackage{breqn}

\newcommand{\myul}[2][black]{
\setulcolor{#1}\setul{}{1.7pt}\ul{#2}
}\setulcolor{black}
\usepackage[super]{nth}
\newcommand{\topn}[1]{$\mathrm{Top^{#1}}$}
\newcommand{\leadn}[1]{$\mathrm{Lead^{#1}}$}
\newcommand{\topedit}[1]{$\mathrm{Top^{#1}}$+$\mathrm{Edits^{#1}}$}
\newcommand{\topeditpara}[1]{$\mathrm{Top^{#1}}$+$\mathrm{Edits^{#1}}$+$\mathrm{Par}$}
\definecolor{Blue}{rgb}{0, 0, 255}

\newcommand{\salmean}{$\textit{Saliency}^{\mu}$}
\newcommand{\salnorm}{$\textit{Saliency}^{\ell2}$}
\newcommand{\inputxmean}{$\textit{InputXGrad}^{\mu}$}
\newcommand{\inputxnorm}{$\textit{InputXGrad}^{\ell2}$}
\newcommand{\guidedmean}{$\textit{GuidedBP}^{\mu}$}
\newcommand{\guidednorm}{$\textit{GuidedBP}^{\ell2}$}
\newcommand{\occlusion}{\textit{Occlusion}}
\newcommand{\shapsamp}{\textit{ShapSampl}}
\newcommand{\lime}{\textit{LIME}}
\newcommand{\rand}{\textit{Random}}

\newcommand{\trans}{$\mathtt{Transformer}$}
\newcommand{\transrand}{$\mathtt{Transformer^{RI}}$}
\newcommand{\cnn}{$\mathtt{CNN}$}
\newcommand{\cnnrand}{$\mathtt{CNN^{RI}}$}
\newcommand{\lstm}{$\mathtt{LSTM}$}
\newcommand{\lstmrand}{$\mathtt{LSTM^{RI}}$}

\newcommand{\salscores}{$\omega_{x_i,c}^M$}
\newcommand{\property}[0]{diagnostic property}
\newcommand{\propertyplural}[0]{diagnostic properties}
\newcommand{\salmap}[0]{\ensuremath{\textrm{SD}}}

\newcommand{\faithfulness}{Faithfulness}
\newcommand{\confidence}{Confidence Indication}
\newcommand{\consistency}{Data Consistency}
\newcommand{\thesisTitle}{Accountable and Explainable Methods for Complex Reasoning over Text}
\newcommand{\thesisSubtitle}{}
\newcommand{\thesisName}{Pepa Kostadinova Atanasova}
\newcommand{\thesisDate}{Submission date: 16$^{th}$ September 2022}

\newcommand{\thesisInternalSupervisor}{Isabelle Augenstein, Jakob Grue Simonsen and Christina Lioma}
\newcommand{\thesisSubject}{\footnotesize This thesis has been submitted to the PhD School of The Faculty of Science, University of Copenhagen}
\usepackage[					% UCPH thesis style
    figuresep=colon,        
    sansserif=false,        
    hangfigurecaption=false,
    hangsection=true,       
    hangsubsection=true,    
    colorize=full,          
    colortheme={\thesisTheme},  % ucph, sund, science, hum, etc.?
    bibsys=biber,
    bibfile=references,       % defines your .bib file
    bibstyle=authoryear,        % refer to https://bit.ly/2YsvIJz
]{ucphthesis}
\hypersetup{					% setup the hyperref-package options
	pdftitle={\thesisTitle},	% 	- title (PDF meta)
	pdfsubject={\thesisSubject},% 	- subject (PDF meta)
	pdfauthor={\thesisName},	% 	- author (PDF meta)
	plainpages=false,			% 	-
	colorlinks=true,
	citecolor=linksblue,
	allcolors=linksblue,
	linkcolor=linksblue, % 	- colorize links
	pdfborder={0 0 0},			% 	-
	breaklinks=true,			% 	- allow line break inside links
	bookmarksnumbered=true,		%
	bookmarksopen=true			%
    }

\subject{\vspace{4.5cm} \thesisSubject}
\title{\thesisTitle}
\subtitle{\thesisSubtitle}
\author{\thesisName \\
        \small{Supervised by {\thesisInternalSupervisor}}
    }
\date{\thesisDate}

\begin{document}
\parskip=0pt
% -------------------------- 
% Front matter
% --------------------------
\pagenumbering{roman}
\pagestyle{empty}				            % no header or footers
\AddToShipoutPicture*{\TitleBackground}     % adding background picture
\maketitle                                  % making the title
\pagestyle{plain}
% \input{frontbackmatter/quotes.tex}        
% \input{frontbackmatter/preface.tex}   
% \clearpage
\pdfbookmark[0]{Acknowledgements}{Acknowledgements}
\chapter*{Acknowledgements}
\label{chap:acknowl}
This Ph.D. thesis is the ultimate outcome of the challenging yet rewarding Ph.D. journey, which would not have been possible without the support of many people. First and foremost, I would like to express my deepest appreciation to my supervisors, Isabelle, Jakob, and Christina. They have provided me with invaluable guidance, kindhearted support, much-needed critical advice, and the freedom to pursue my research interests. To my principal supervisor, Isabelle, I am grateful for being an inspiring role model, for establishing a stimulating lab environment, and for providing opportunities for professional growth.

I also sincerely thank the European Union’s Horizon 2020 research and innovation programme under the Marie Skłodowska-Curie grant agreement No 801199 that has funded my Ph.D. research. 

Furthermore, I could not have undertaken this journey without the initial guidance from my Master's degree supervisor, Preslav. I continue to be indebted to his support and introduction to research and academia. I am also grateful to all the bright and kindhearted lab mates from the CopeNLU group for filling my journey with many memorable moments and engaging discussions. 

Next, I would like to express my gratitude to all the remarkable collaborators I have been fortunate to work with. They have shared research interests with me and provided valuable perspectives and contributions to my line of research. I was also fortunate to conduct two research internships at Google and Meta, where Abe, Alyssa, Cong, Yashar, and Wenhan provided outstanding mentorship and made a lot of efforts to create an inclusive and productive environment despite the conditions during the pandemic.

On a personal note, this endeavor would not have been possible without the unfailing support of my family and friends. To Ludmila and Kostadin, I am forever indebted for their unconditional and loving encouragement that kept me motivated and confident. To Atanas, I am forever grateful for the enthusiasm to make a new home in a different country and for always being a continuous source of sanity, joy, and encouragement. % INCLUDE Acknowledgements

\pdfbookmark[0]{Abstract}{Abstract}
\chapter*{Abstract}
\label{chap:abstract}
A major concern of Machine Learning (ML) models is their opacity. They are deployed in an increasing number of applications where they often operate as black boxes that do not provide explanations for their predictions. Among others, the potential harms associated with the lack of understanding of the models' rationales include privacy violations, adversarial manipulations, and unfair discrimination. As a result, the accountability and transparency of ML models have been posed as critical desiderata by works in policy and law, philosophy, and computer science. 

In computer science, the decision-making process of ML models has been studied by developing accountability and transparency methods. Accountability methods, such as adversarial attacks and diagnostic datasets, expose vulnerabilities of ML models that could lead to malicious manipulations or systematic faults in their predictions. Transparency methods explain the rationales behind models' predictions gaining the trust of relevant stakeholders and potentially uncovering mistakes and unfairness in models' decisions. To this end, transparency methods have to meet accountability requirements as well, e.g., being robust and faithful to the underlying rationales of a model. 

% Overall, accountability and transparency methods facilitate the analysis of the reasons behind the outputs of ML models and assist in detecting and correcting for potential harms. 

This thesis presents my research that expands our collective knowledge in the areas of accountability and transparency of ML models developed for complex reasoning tasks over text.
First, this thesis contributes with two methods for accountable ML models. They generate adversarial inputs and a diagnostic dataset demonstrating significant model vulnerabilities and suggesting ways to correct those. In the area of transparency of ML models, this thesis advances the state-of-the-art with methods generating textual explanations that are further improved to be fluent, easy to read, and to contain logically connected multi-chain arguments. Finally, this thesis makes contributions in the area of diagnostics for explainability approaches with a set of properties for evaluating existing explainability techniques and methods for enhancing those further in produced explanations. All of the contributions are empirically tested on complex reasoning tasks over text, including fact checking, question answering, and natural language inference.      % INCLUDE Abstract
\pdfbookmark[0]{Resume}{Resume}
\chapter*{Resum{\'e}}
\label{chap:abstract_dk}

En væsentlig kilde til bekymring i forskning om maskinlæringsmodeller er modellernes uigennemskuelighed. Sådanne modeller benyttes i stigende grad til
anvendelser, hvor de opererer som ``black boxes'' som ikke forklarer deres forudsigelser eller beslutninger. Blandt de potentielle farer knyttet til mangel på forståelse for modellernes rationaler er krænkelser af privatlivet, fjendtlig  manipulation og uretfærdig forskelsbehandling. Som følge heraf er ansvarlighed (en.\ \emph{accountability}) og gennemskuelighed (en.\ \emph{transparency}) for maskinlæringsmodeller blevet foreslået som kritisk vigtige designkriterier af forskning i politik, jura, filosofi og datalogi.

I datalogi er maskinlæringsmodellers beslutningsprocesser blevet undersøgt ved udvikling af metoder for ansvarlighed og gennemskuelighed. Flere metoder for undersøgelse af ansvarlighed, herunder brugen af fjendtlige angreb og diagnostiske datasæt, har påvist eksistensen af sårbarhed for modeller, og at sådanne sårbarheder kan føre til ondsindet manipulation af resultater, eller systematiske fejl i modellernes forudsigelser. For at imødegå dette, er det nødvendigt, at metoder til at sikre gennemskuelighed tillige opfylder en række krav for ansvarlighed, f.eks. at udvise robusthed og være tro mod de underliggende rationaler i modellerne.

Denne afhandling fremlægger min forskning, som udvider den samlede viden inden for ansvarlighed og gennemskuelighed for maskinlæringsmodeller, som er udviklet til at løse opgaver, der involverer komplekst ræsonnement om tekstdata. For det første bidrager afhandlingen med to ny metoder for ansvarlige maskinlæringsmodeller, herunder skabelsen fjendtlige input og et diagnostisk datasæt, som påviser væsentlige sårbarheder i modeller, og foreslår metoder til at afhjælpe disse sårbarheder. Inden for gennemskuelighed af maskinlæringsmodeller, bidrager afhandlingen med metoder til automatisk skabelse af forklaringer på skriftform, som yderligere forbedres til at benytte flydende sprog, er letlæselige og indeholder logisk sammenhængende argumentationskæder. Sluttelig bidrager afhandlingen til diagnostik af metoder til forklarlighed af maskinlæringsmodellers forudsigelser ved at definere en række egenskaber med hvilke allerede eksisterende forklarlighedsmetoder - og metoder til forbedring af sådanne - kan evalueres. All afhandlingens bidrag er eksperimentelt afprøvet på problemer, som involverer komplekst ræsonnement om tekstdata, herunder faktatjek, automatisk besvarelse af spørgsmål, og følgeslutninger i naturligt sprog.      % INCLUDE Abstract in Danish
\pdfbookmark[0]{Publications}{Publications}
\chapter*{Publications}
\label{chap:publications}
This thesis includes the following papers as chapters, listed in the order of their appearance ($^*$ denotes equal contribution):

\begin{enumerate}
    \item \cite{10.1162/tacl_a_00486} Pepa Atanasova, Jakob Grue Simonsen, Christina Lioma, and Isabelle Augenstein. Fact Checking with Insufficient Evidence. 2022. Transactions of the Association for Computational Linguistics, pages 746–763.
    \item \cite{atanasova-etal-2020-generating} Pepa Atanasova$^*$, Dustin Wright$^*$, and Isabelle Augenstein. Generating Label Cohesive and Well-Formed Adversarial Claims. 2020. In Proceedings of the 2020 Conference on Empirical Methods in Natural Language Processing (EMNLP), pages 3168–3177, Online. Association for Computational Linguistics.
    \item \cite{atanasova-etal-2020-generating-fact} Pepa Atanasova, Jakob Grue Simonsen, Christina Lioma, and Isabelle Augenstein. Generating Fact Checking Explanations. 2020. In Proceedings of the 58th Annual Meeting of the Association for Computational Linguistics, pages 7352–7364, Online. Association for Computational Linguistics.
    \item \cite{DBLP:journals/corr/abs-2112-06924} Shailza Jolly, Pepa Atanasova, and Isabelle Augenstein. Generating Fluent Fact Checking Explanations with Unsupervised Post-Editing. 2022. In Information, Special Issue on Advances in Explainable Artificial Intelligence.
    \item \cite{ijcai2021-536} Ostrowski, Wojciech, Arnav Arora, Pepa Atanasova, and Isabelle Augenstein. Multi-hop fact checking of political claims. In Proceedings of the Thirtieth International Joint Conference on Artificial Intelligence, pages 3892-3898, Montreal, Canada.
    \item \cite{atanasova-etal-2020-diagnostic} Pepa Atanasova, Jakob Grue Simonsen, Christina Lioma, and Isabelle Augenstein. A Diagnostic Study of Explainability Techniques for Text Classification. 2020. In Proceedings of the 2020 Conference on Empirical Methods in Natural Language Processing (EMNLP), pages 3256–3274, Online. Association for Computational Linguistics.
    \item \cite{atanasova2021diagnostics} Pepa Atanasova, Jakob Grue Simonsen, Christina Lioma, and Isabelle Augenstein. Diagnostics-Guided Explanation Generation. 2022. Proceedings of the AAAI Conference on Artificial Intelligence 36 (10), pages 10445-10453.
\end{enumerate}

The following list of papers were also published during my Ph.D. studies. As they are unrelated to the topic of the thesis, they are not included in it. The topics of these publications include: dealing with limited labelled data (1), offensive language identification (2-3), fact checking and check-worthiness detection (4-6), dialogue system evaluation (7).

\begin{enumerate}
    \item \cite{de2022joint} Luna De Bruyne, Pepa Atanasova, and Isabelle Augenstein. Joint emotion label space modeling for affect lexica. 2022. Computer Speech \& Language 71, pages: 101257.
    \item \cite{zampieri-etal-2020-semeval} Marcos Zampieri, Preslav Nakov, Sara Rosenthal, Pepa Atanasova, Georgi Karadzhov, Hamdy Mubarak, Leon Derczynski, Zeses Pitenis, and Çağrı Çöltekin. SemEval-2020 Task 12: Multilingual Offensive Language Identification in Social Media (OffensEval 2020). 2020. In Proceedings of the Fourteenth Workshop on Semantic Evaluation, pages 1425–1447, Barcelona (online). 
    \item \cite{rosenthal-etal-2021-solid} Sara Rosenthal, Pepa Atanasova, Georgi Karadzhov, Marcos Zampieri, and Preslav Nakov. SOLID: A Large-Scale Semi-Supervised Dataset for Offensive Language Identification. 2021. In Findings of the Association for Computational Linguistics: ACL-IJCNLP 2021, pages 915–928, Online. Association for Computational Linguistics.
    \item \cite{atanasova2019overview} Pepa Atanasova, Preslav Nakov, Georgi Karadzhov, Mitra Mohtarami, and Giovanni Da San Martino. Overview of the CLEF-2019 CheckThat! Lab: Automatic Identification and Verification of Claims. Task 1: Check-Worthiness. 2019. International Conference of the Cross-Language Evaluation Forum for European Languages. Springer, Cham.
    \item \cite{vasileva-etal-2019-takes} Slavena Vasileva, Pepa Atanasova, Lluís Màrquez, Alberto Barrón-Cedeño, and Preslav Nakov. It Takes Nine to Smell a Rat: Neural Multi-Task Learning for Check-Worthiness Prediction. 2019. In Proceedings of the International Conference on Recent Advances in Natural Language Processing (RANLP 2019), pages 1229–1239, Varna, Bulgaria. 
    \item \cite{10.1145/3297722} Pepa Atanasova, Preslav Nakov, Lluís Màrquez, Alberto Barrón-Cedeño, Georgi Karadzhov, Tsvetomila Mihaylova, Mitra Mohtarami, and James Glass. Automatic Fact-Checking Using Context and Discourse Information. 2019. J. Data and Information Quality 11, 3, Article 12 (September 2019), 27 pages.
    \item \cite{10.1145/3331184.3331308} Pepa Atanasova, Georgi Karadzhov, Yasen Kiprov, Preslav Nakov, and Fabrizio Sebastiani. Evaluating Variable-Length Multiple-Option Lists in Chatbots and Mobile Search. 2019. In Proceedings of the 42nd International ACM SIGIR Conference on Research and Development in Information Retrieval (SIGIR'19). Association for Computing Machinery, New York, NY, USA, pages: 997–1000.
\end{enumerate}      % INCLUDE Publications

\clearpage

\setcounter{secnumdepth}{3}
\setcounter{tocdepth}{2}		% define depth of toc
{
\hypersetup{linkcolor=black}

\tableofcontents				% display table of contents
    \clearpage
\listoffigures
     \clearpage
\listoftables
     \clearpage
}
% -------------------------- 
% Main matter
% --------------------------
\pagenumbering{arabic}			% arabic page numbering
\setcounter{page}{1}			% set page counter
\pagestyle{maincontentstyle} 	% fancy header and footer
\part{Executive Summary}
\chapter{Executive Summary}
\label{chap:intro}

\section{Introduction}
Recent progress in the field of machine learning and specifically in natural language processing has been driven by the development of large models pre-trained on massive amounts of data \cite{vaswani2017attention,devlin-etal-2019-bert}. Notably, such models have been used to extend the state of the art in a broad range of tasks \cite{wang-etal-2018-glue} and have been deployed in an increasing number of downstream applications \cite{angwin2022machine,obermeyer2019dissecting,barocas2016big,lambrecht2019algorithmic}. On the other hand, the models' increased architectural complexity has raised concerns about the decreased ability of humans to understand the opaque decision-making processes of these models \cite{raji2020closing,bender2021dangers}. To this end, methods for accountable and transparent machine learning models have been developed that verify and unveil the reasons behind the models' predictions \cite{raji2020closing}. These methods assess critical aspects of machine learning models beyond the achieved task performance, such as vulnerability to adversarial decision manipulations \cite{kreps2022all,agarwal2019protecting}, unfair embedded biases towards certain groups and individuals \cite{kiritchenko-mohammad-2018-examining,raji2020closing,ntoutsi2020bias}, privacy violations \cite{carlini2019secret}, and generalisation to out-of-distribution samples \cite{koh2021wilds}. Accountability and transparency methods can further be used as means to engender trust in a model's decisions \cite{ribeiro2016should}, expand the knowledge about a downstream task \cite{forde2022concepts,ghandeharioun2022dissect} and debug and improve a model's decision-making process \cite{ANDERS2022261,abid2022meaningfully}.

This section introduces accountability and transparency methods for machine learning models from the perspective of computer science and, in particular, for complex reasoning tasks over text, such as fact checking, question answering, and natural language inference. The papers included in the following chapters of this Ph.D. thesis are cross-referenced where relevant. Section \ref{sec:contributions} provides a detailed overview of the contributions of the separate publications included in this Ph.D. thesis in the areas of accountable and transparent machine learning models. Section \ref{sec:summary} offers an introspective summary of the contributions and suggests prospects for future work.

\subsection{Accountability}
The accountability of a machine learning model is verified by methods that analyse the model's outputs on specifically crafted instances in order to detect and correct for flaws in its reasoning process, such as reliance on spurious correlations. To this end, accountability methods usually produce datasets used to inspect whether the model's outputs are the desired outcomes for the instances in the created dataset. The produced datasets can be challenging \cite{nie-etal-2020-adversarial,10.1162/tacl_a_00486} or adversarial in nature \cite{atanasova-etal-2020-generating,song-etal-2021-universal}. They can reveal specific model vulnerabilities such as a model's reliance on spurious features \cite{mccoy-etal-2019-right,schuster-etal-2019-towards}, vulnerability to maliciously manipulated inputs \cite{atanasova-etal-2020-generating,song-etal-2021-universal,guo-etal-2021-gradient}, lack of generalisation to out-of-distribution samples \cite{koh2021wilds}, and specific reasoning skills that a model failed to acquire \cite{dua-etal-2019-drop,talmor-etal-2020-olmpics}. Moreover, such datasets can steer the development of model architectures designed to handle the revealed model vulnerabilities \cite{zhao2020transformer-xh}. They can also provide additional training data points to enhance the performance of existing models on the challenges presented by these datasets \cite{nie-etal-2020-adversarial,schuster-etal-2021-get}.

\subsubsection{Diagnostic Challenge Datasets}
Owing to improvements in computational power and the development of effective machine learning models, such as models with the Transformer architecture \cite{vaswani2017attention,devlin-etal-2019-bert,liu2019roberta} and pre-trained language models \cite{howard-ruder-2018-universal, devlin-etal-2019-bert, liu2019roberta,NEURIPS2020_1457c0d6}, the time for achieving near-human performance on new tasks has decreased to a few months \cite{kiela-etal-2021-dynabench}. Existing work, however, has questioned whether a model's performance on a dataset indicates it has learned meaningful features required to solve the task underlying the dataset \cite{bowman-dahl-2021-will}. Studies have found that, on the contrary, machine learning models often learn to rely on spurious correlations located in the training data \cite{mccoy-etal-2019-right, ribeiro-etal-2020-beyond}. These findings have motivated the research on \textit{challenge datasets} that diagnose whether models learn specific meaningful features and do not overfit to correlations in the training set. Moreover, \citet{dua-etal-2019-drop,nie-etal-2020-adversarial,kiela-etal-2021-dynabench} propose dynamic benchmarks where challenge datasets are collected via an iterative human-and-model-in-the-loop procedure. In the first step of the procedure, human annotators collect a challenge dataset for which a given model cannot predict the correct labels. In the second step, the training split of the challenge dataset is used to train a better-performing model. The two steps can be applied repeatedly, creating a moving target, rather than a static benchmark that models quickly overfit to.

Predominantly, challenge datasets are designed around particular sets of reasoning skills such as logical reasoning \cite{ijcai2020-0501}, linguistic capabilities \cite{ribeiro-etal-2020-beyond,saha-etal-2020-conjnli}, and common-sense inference \cite{zellers-etal-2018-swag} (see Section \ref{sec:complex} for an overview of reasoning skills). There also exist challenge datasets that test a model's ability to detect when the provided input is insufficient to make the correct decision \cite{rajpurkar-etal-2018-know,10.1162/tacl_a_00486}. Other challenge datasets are contrastive in nature and verify whether a model can detect minimal changes in the input that lead to a change in the prediction \cite{gardner-etal-2020-evaluating,Kaushik2020Learning, sen-etal-2021-counterfactually}.

% compositional generalisation \cite{yanaka-etal-2021-sygns}, complex reasoning skills \cite{zellers-etal-2018-swag,talmor-etal-2020-olmpics}

%General linguistic capabilities that can be tested on a majority of NLP tasks include negation, named entities, coreferences, and semantic role labeling. Complex reasoning skills of interest in challenge datasets include common sense reasoning \cite{zellers-etal-2018-swag}, discrete reasoning over text, logical reasoning, and multi-hop reasoning. 

Challenge datasets can further be categorised as model-agnostic \cite{talmor-etal-2020-olmpics,saha-etal-2020-conjnli,schuster-etal-2019-towards} or created with a model in the loop \cite{nie-etal-2020-adversarial}. Model-agnostic datasets are usually produced manually, where expert knowledge is utilised to construct tests for different skills. There is, however, no guarantee that the resulting dataset will cover potential model flaws. On the other hand, challenge datasets created with a model in the loop could cover deficiencies only of the employed models. The latter can be produced manually, where data creators are incentivised to deceive a given model \cite{wallace-etal-2019-trick,nie-etal-2020-adversarial} or with automated data generation techniques \cite{10.1162/tacl_a_00486,le2020adversarial}. 

%Finally, automated model-in-the-loop techniques for creating challenge datasets generate instances in an automated way and can further select samples that models particularly struggle to recognise or that contain model biases and can be used for decreasing dependance on artifacts \cite{le2020adversarial}.

This thesis presents advances in the area of challenge datasets for model accountability with novel methodology, insights, and models' accountability improvements. \textit{Paper 1} (\S \ref{chap:insufficient_information}) presents a novel automated method for constructing a contrastive model-in-the-loop challenge dataset to study what information models consider sufficient for producing a prediction. In knowledge-intensive tasks such as fact checking, it is markedly crucial to make predictions only when the presented input information is sufficient and otherwise indicate it is not enough. The technique introduced in this publication employs three separate models in the loop to preserve validity beyond a single model. The latter also allows one to compare and contrast different models' deficiencies. The introduced method is further used to improve models' performance on instances with insufficient evidence information.

\textit{Paper 5} (\S \ref{chap:multihop}) presents a challenge dataset for multi-hop reasoning for the task of fact checking. The dataset contains real-world claims with manual annotations of sets of logically connected evidence pieces that lead to the final verdict of a claim. It enables accountability investigations of whether a model employs multi-hop reasoning by logically connecting evidence chunks needed for a prediction as opposed to predicting based on a single inference step. Based on the challenge dataset, the publication presents findings on the multi-hop reasoning capabilities of two existing models where an architecture designed specifically to conduct multi-hop reasoning performs the best.

\subsubsection{Adversarial Attacks}
Adversarial attacks reveal model vulnerabilities to changes in the input that manipulate the model to produce a target prediction, different from the correct one. Adversarial attacks can be performed at training or test time. Training-time adversarial attacks \cite{qi-etal-2021-mind,kurita-etal-2020-weight,wallace-etal-2021-concealed} manipulate either the model weights or the training data, assuming unrestricted access to the training process of a model. Test-time adversarial attacks unveil model vulnerabilities of already trained models and can assume access to the parameters of a model -- white-box attacks \cite{atanasova-etal-2020-generating,guo-etal-2021-gradient}, or no access to them -- black-box attacks \cite{chen-etal-2021-multi,berger-etal-2021-dont}. 
% As this thesis focuses on complex reasoning over text for tasks such as fact checking and question answering, what follows next is focused on methods applicable to language tasks. In natural language processing, inputs have a discrete nature, which poses additional challenges for developing methods for accountability, e.g., there is a reduced number of possible manipulations that preserve the validity of the input. In contrast, tasks where the input spaces are continuous, e.g., in image and time-series analysis, allow for numerous input perturbations that do not harm the overall realistic outlook of the input \cite{papernot2016limitations,https://doi.org/10.48550/arxiv.1412.6572,szegedy2013intriguing} and are sometimes even invisible to the human eye \cite{szegedy2013intriguing,moosavi2016deepfool}.

Adversarial attacks usually perform manipulations of the input, which are the smallest possible changes required to achieve a target prediction. For textual inputs, changes can be performed at character \cite{eger-etal-2019-text}, word \cite{mozes-etal-2021-contrasting,zang-etal-2020-word}, or sentence level \cite{jia-liang-2017-adversarial,iyyer-etal-2018-adversarial}. Due to the discrete nature of textual inputs, performing input manipulations is additionally challenging as there is a reduced number of possible manipulations that preserve the validity of the input. In contrast, in tasks where the input spaces are continuous, e.g., image and time-series analysis, it is possible to perform numerous input perturbations that do not harm the overall realistic outlook of the input \cite{papernot2016limitations,https://doi.org/10.48550/arxiv.1412.6572,szegedy2013intriguing} and are sometimes even invisible to the human eye \cite{szegedy2013intriguing,moosavi2016deepfool}.

The potency of adversarial attacks is commonly measured as the number of samples where a model's prediction can be manipulated. One example of a potent adversarial technique is the universal adversarial attack, which degrades the performance of a model by inserting a particular textual sequence, termed as a trigger, in all instances \cite{wallace-etal-2019-universal,song-etal-2021-universal}. For a more detailed overview of adversarial attacks, refer to \citet{xu2020adversarial,chakraborty2021survey}.
% data-independent

In \textit{Paper 2} (\S \ref{chap:adversarial_claims}) of this thesis, I present a novel method for generating test-time white-box adversarial attacks. It draws on the universal adversarial attack approach, which suffers from two precluding deficiencies when applied for inference tasks such as fact checking. First, universal adversarial attacks generate triggers that often invert the meaning of the instances they are inserted in, thus changing also their gold standard label. The method proposed in the paper mitigates this with a novel extension to the universal adversarial attack approach that generates triggers preserving the label of the original instance. Second, universal adversarial attacks produce semantically invalid inputs, as they simply concatenate triggers to existing samples. The method proposed in the paper alleviates this with a conditional language model trained to generate semantically valid statements, which include the found universal triggers. The method is empirically tested with a model for the task of fact checking, where the generated adversarial claims are highly effective in fooling the model and lead to a performance decrease of 23.1 $F_1$ score points compared to the model's performance on the original claims. At the same time, the generated attacks constitute valid fact checking instances as they have preserved the gold label and the semantic validity of the input.

\subsection{Explainability}
\label{sec:explainability}
Machine learning models have been heavily criticized for their opaque nature \cite{zarsky2016trouble,pasquale2015black}. 
% Moreover, many studies identify the flaws and biases of machine learning systems deployed in high-stake scenarios \cite{lambrecht2019algorithmic, angwin2022machine}. 
As a result, explanations of their decisions are increasingly needed for debugging, measuring bias and fairness, instilling trust, and making model behavior transparent in general. European law has also introduced a requirement for "the right \dots to obtain an explanation of the decision reached" \cite{goodman2017european}. Efforts to make models' decisions transparent have led to a growing influx of explainability approaches. What follows next is an overview of common types of explainability techniques. For more details, I refer the reader to \citet{ras2022explainable,molnar2022}.

\subsubsection{Post-hoc Saliency Explanations}
Post-hoc explanations reveal the decision process of an already trained model and can be applied to various types of models and different data modalities. Saliency explanations are the most prominent type of post-hoc explanations. They highlight regions of the input according to the region's importance for the prediction of a model. Saliency explanations can be gradient-based, such as the Vanilla Gradient \cite{Simonyan2013DeepIC}, which computes the gradient of the output w.r.t. the input. Follow-up gradient-based approaches \cite{Kindermans2016InvestigatingTI,springenberg2014striving} improve saturation and stability problems of the former. Saliency explanations can also be perturbation-based, e.g., Occlusion \cite{zeiler2014visualizing}, and Shapley Value Sampling \cite{shapley1953value}, which estimate the contribution of input regions to a model's prediction by occluding regions from the input and observing the corresponding changes in the model's prediction. Finally, simplification-based explanations such as LIME \cite{ribeiromodel} train a simple self-interpretable model to approximate the local decision boundary of the opaque model for each instance.

Contrastive explanations are another type of post-hoc explanations. They find small changes to the input that cause a change in the prediction of a model \cite{9321372}. Studies in social science \cite{lipton_1990} argue that contrastive explanations are more intuitive to end users as they unveil the causal factors that explain why an event occurred instead of an alternative event. Contrastive explanations can be constructed by manipulations either at the discrete textual input level or at a latent representation level of a given model. Methods for contrastive explanations at the input level are formulated as search problems where a model, e.g., a language generation model, is trained to apply edit, replace or delete operations at different positions of the input until a change in the prediction of a model is achieved \cite{wu-etal-2021-polyjuice,ross-etal-2021-explaining}. By contrast, \citet{jacovi-etal-2021-contrastive} observe changes in the predictions of a model caused by projecting its latent representations to a similar representation space where only a particular concept, such as gender, is removed.

\subsubsection{Natural Language Explanations}
\label{sec:nles}
Natural language explanations (NLEs) explain model predictions with free text, which contrary to other explanation types, is a natural means of communication that does not require a preliminary clarification phase. Furthermore, NLEs are not constrained to contain only input segments but can also contain generated text that provides more explanation of the model's rationales for the prediction. This gives them greater expressive power in terms of the reasoning they can convey, especially with complex reasoning tasks (see Section \ref{sec:complex}) involving rationales beyond what is explicitly stated in the input. Existing datasets with NLEs include the tasks of natural language inference \cite{NIPS2018_8163,https://doi.org/10.48550/arxiv.2004.03744}, common sense reasoning \cite{rajani-etal-2019-explain,8953217}, fact checking \cite{alhindi-etal-2018-evidence,kotonya-toni-2020-explainable-automated}, and relation extraction \cite{hancock-etal-2018-training,Wang*2020Learning}. For a longer discussion of datasets with NLEs, I refer the reader to \citet{wiegreffe2021teach}.

NLEs are typically produced in a supervised way where models generating NLEs can be trained jointly or separately from the models for the downstream task \cite{NIPS2018_8163}. Prior work has found that training the two tasks jointly leads to more label-informed explanations \cite{wiegreffe-etal-2021-measuring}. The explanations can also be conditioned on the predicted label \cite{hase-etal-2020-leakage,kumar-talukdar-2020-nile}. 

With respect to the input, produced NLEs can be extractive -- selected important portions of the input text that constitute a free text explanation, or abstractive -- generated text explaining the model's reasoning in words that do not necessarily appear in the input. Extractive NLE techniques still produce natural text, as they usually choose whole sentences from the input \cite{atanasova-etal-2020-generating-fact,thorne-etal-2018-fever}. In knowledge-intensive tasks such as question answering and fact checking, producing short extractive explanations from long input documents is often regarded as part of the task, and the performance on the explanation and the downstream task is judged jointly with a unified measure \cite{thorne-etal-2018-fever,jiang-etal-2020-hover,trivedi-etal-2019-repurposing,petroni-etal-2021-kilt}. Compared to abstractive NLEs, extractive NLEs require less training data, which makes them more suitable for low-resource scenarios. Extractive NLEs also produce explanations that are always factual w.r.t. the input, i.e., they contain only information that is correct given the input. The latter cannot be guaranteed for abstractive generated NLEs. On the other hand, abstractive NLEs can be more coherent, provide more information about the model's rationales, and have less redundant information compared to extractive ones \cite{DBLP:journals/corr/abs-2112-06924}.

This thesis makes important contributions to the field of explainability establishing methods and datasets for generating NLEs given limited resources and complex reasoning tasks such as fact checking. In \textit{Paper 3} (\S \ref{chap:generating_explanations}), I introduce the task of generating NLEs for fact checking veracity predictions. Generating veracity explanations is a challenging task, especially when considering real-world claims where training data is limited and claim verification requires constructing fact checking evidence of multiple arguments involving complex reasoning capabilities. In addition, most NLEs for existing tasks contain no more than one sentence per instance. At the same time, the explanations produced for real-world claims have several sentences, which also indicates the complex reasoning required for veracity prediction. Paper 3 proposes a method that generates fact checking explanations in an extractive way and jointly with the task at hand. I find that optimising explanation generation jointly with veracity prediction produces explanations that achieve better coverage and overall quality and are better suited for explaining the correct veracity label than explanations learned solely to mimic human justifications. 

As extractive NLEs can lack fluency and coherence and can contain redundant information, \textit{Paper 4} (\S \ref{chap:editing_explanations}) introduces a method to improve the fluency and readability of fact checking NLEs. The method performs post-editing of extracted NLEs and is the first to explore an iterative unsupervised edit-based algorithm using only phrase-level edits. The proposed method also leads to computationally feasible explanation generation solutions for long text inputs. More importantly, the resulting explanations are found to be fluent, easy to read, and concise.

Finally, \textit{Paper 5} (\S \ref{chap:multihop}) provides a supervised extractive dataset for producing fact checking explanations that form chains of logically connected arguments. Utilising the dataset, the paper documents the first study on how models construct rationales to verify political claims requiring multi-hop evidence reasoning. The main finding
of the study is that the best performance is achieved with an architecture that specifically models multi-hop reasoning over evidence pieces in combination with in-domain transfer learning.

\subsubsection{Self-Interpretable Models}
A simple solution for achieving model transparency is using self-interpretable models that produce predictions in a way that can be interpreted by a non-expert from their inner workings. Examples of self-interpretable models are linear regression \cite{ge2018interpretable}, decision trees \cite{prentzas2019integrating}, Bayesian models \cite{letham2015interpretable}, and general additive models \cite{hastie2017generalized}. These models usually have simple architectures, which struggle to achieve good performance, especially on non-structured input such as text. On the other hand, existing work has attributed self-interpretable capabilities to models using the attention mechanism \cite{wiegreffe-pinter-2019-attention,meister-etal-2021-sparse}, which achieve high performance on many tasks. However, the use of attention weights as explanations has also met criticism \cite{bastings-filippova-2020-elephant} as they are not always faithful to the prediction rationale of the underlying model \cite{jain-wallace-2019-attention,wiegreffe-pinter-2019-attention}. One can also often find a different set of attention weights resulting in the same prediction \cite{serrano-smith-2019-attention}.

In \textit{Paper 5} (\S \ref{chap:multihop}) of this thesis, I employ eXtra-hop attention to model the interaction between evidence sentences for fact checking claims. The eXtra-hop attention introduces a way to structure text where the important evidence sentences are linked in a logically connected set of arguments. One of the research questions of the paper explores whether eXtra-hop attention indicates which are the important evidence sentences from the input document for predicting the target task of fact checking. In fact, I find that the model assigns higher eXtra-hop attention weights to evidence sentences employed for the prediction as opposed to the remaining sentences in the input document.

% \subsubsection{Global Explanations}
% Concept-reasoning tasks
% probing

\subsection{Diagnostic Explainability Methods}
The overwhelming influx of explainability approaches increases the number of different explanations that can be produced for a model's prediction \cite{DBLP:journals/corr/abs-2105-03287}. An open question becomes which explainability approaches produce better-quality explanations and which faithfully relay the reasons behind the decisions of a model. Existing studies \cite{deyoung-etal-2020-eraser,Adebayo:2018:SCS:3327546.3327621,adebayo2022post,ding-koehn-2021-evaluating} point that explainability approaches can be unfaithful to the rationales used by a model and can cover potential flaws and biases in the model's reasoning. Such findings call for systematic diagnostics of explainability approaches to estimate their reliability and to motivate further progress in explainability approaches.

% They argue that some popular saliency methods should not be used for explainability purposes since the maps they produce are not sensitive to the underlying model that is to be explained.

\subsubsection{Explainability Diagnostics}
Explainability approaches can be assessed by human judges that estimate the utility of explanations, e.g., for guessing the label predicted by the model \cite{lertvittayakumjorn-toni-2019-human,narayanan2018humans}. However, human studies often suffer from low inter-annotator agreement as the evaluation protocols can be subjective and assess how appealing explanations are to human judges instead of evaluating their qualities.

Another way of evaluating the utility of explainability approaches is by using automated measures for various explanation properties. One commonly assessed explainability property is faithfulness. It estimates whether an explainability approach faithfully reflects the rationales used in the decision-making process of a model. While some existing work \cite{alvarez2018robustness,kindermans2019reliability} estimates the lack of faithfulness based on a few counter-examples, \citet{jacovi-goldberg-2020-towards} recommend the use of faithfulness evaluation measures, as in \citet{deyoung-etal-2020-eraser}, computing rather a degree of explanation faithfulness. Another common evaluation measure is the extent of the explanation's agreement with human rationales, which indicates the plausibility and appeal of the rationales to human judges. \citet{deyoung-etal-2020-eraser,ding-koehn-2021-evaluating} include measures of faithfulness, human agreement, and others in benchmarks for saliency-based explanations.

While some existing studies \cite{yin-etal-2022-sensitivity,arras-etal-2019-evaluating,guan2019towards} evaluate explainability approaches with various measures for explainability properties, most studies are limited in scope, exploring only one or a few properties, datasets, and models. In \textit{Paper 6} (\S \ref{chap:diagnostic_study_saliency}) of this thesis, I construct a comprehensive list of diagnostic properties tied with automated measures thereof. The study provides a broad overview and a unified comparison of different groups of common explainability approaches across three text classification tasks and three model architectures. Stemming from the individual property results, the central finding of this work is that gradient-based methods have the best performance across all of the models and downstream text classification tasks considered in this work. Other explainability techniques, such as Shapley Value Sampling \cite{castro2009polynomial}, LIME \cite{ribeiromodel}, and Occlusion \cite{zeiler2014visualizing} take more time to compute, are considerably less faithful to the models, and are less consistent for similar model rationales and similar instances. 

As saliency explanations provide a score for each input segment, there is a direct mapping between the explanation and the input. The latter enables explainability evaluation measures based on the mapping between the input and explanation. The same measures cannot be applied to NLEs as they contain words and rationales not explicitly present in the input. To this end, NLEs are usually evaluated using simulatability studies \cite{hase-etal-2020-leakage,https://doi.org/10.48550/arxiv.2207.00779}, where humans or models verify that the explanation indicates the label predicted by the model. \citet{wiegreffe-etal-2021-measuring} also evaluate whether models generating NLEs pay attention to the same input tokens as the prediction model and whether model predictions and generated explanations are equally robust to noise introduced in the input.

\subsubsection{Diagnostic-Guided Explainability}

Besides evaluating the qualities of existing explainability approaches, diagnostic properties of explanations can also motivate the development of explainability techniques where those properties are improved.
% , which is still an understudied area
In \textit{Paper 7} (\S \ref{chap:diagnostic_guided_explanations}) of this thesis, I present the first method that produces property-optimised explanations in an unsupervised way. As a result, the generated explanations have improved faithfulness to the underlying prediction method, they better indicate the confidence of the model's prediction and are more consistent across similar instances. A later study proposes an explainability approach that has improved sensitivity to adversarial perturbations of important tokens and is more consistent across similar instances \cite{yin-etal-2022-sensitivity}. In addition, \citet{https://doi.org/10.48550/arxiv.2208.03339} propose novel differentiable combinatorial solvers that encode property constraints for explainable multi-hop inference.

\subsection{Complex Reasoning in Natural Language Tasks}
\label{sec:complex}
There has been substantial progress in natural language processing of downstream tasks where the input has a short textual form and requires a shallow-level semantic understanding of literal cues \cite{wang-etal-2018-glue}. Notably, we have witnessed the emergence of efficient natural language processing models that can be employed to automate a wide range of these tasks \cite{devlin-etal-2019-bert,liu2019roberta}. While such models can reach near-human performance on these tasks, shallow-level semantic understanding of literal cues is insufficient for many real-world natural language processing application tasks. Many real-world tasks, such as fact checking and question answering, require a human to possess a broad range of complex reasoning skills. Consequently, the current prevailing hypothesis in the field of natural language processing is that models need to possess similar reasoning skills to automate these real-world tasks. To achieve progress on these tasks and natural language processing in general, new benchmarks with tasks that constitute a more rigorous test of language understanding have been proposed \cite{NEURIPS2019_4496bf24}.

Some examples of complex reasoning skills include reading comprehension \cite{rajpurkar-etal-2016-squad}, multi-hop composition \cite{yadav-etal-2019-quick,jiang-etal-2020-hover}, and logical reasoning \cite{ijcai2020-0501}. Reading comprehension is the ability to deeply understand long-form textual input and locate the relevant text spans needed for correct inference. In some real-world scenarios, reading comprehension also involves the ability to detect when the provided text is missing information pertinent to drawing an inference. Building on reading comprehension skills, multi-hop composition incorporates the requirement for a model to find arguments scattered across multiple paragraphs or documents and connect them logically into a meaningful structure of arguments, e.g., a graph, that results in a correct prediction. Logical reasoning requires the model to deduce the logical relationship between statements in two textual inputs. Examples of logical reasoning are comparison, negation, categorical reasoning, disjunctive reasoning, and conjunctive reasoning.

Complex reasoning tasks require a model to obtain a combination of different complex reasoning skills to draw a correct inference. What follows next is a brief introduction to complex reasoning tasks central to this thesis.

\subsubsection{Fact Checking} \label{sec:fc}
Fact checking is a time-consuming and elaborate task performed by human fact checkers. Automating the process is of pivotal importance for scaling the number of verified claims in accordance with the growing amount of misinformation and disinformation online. Most of the existing work on automating fact checking is concerned with predicting the veracity of a claim given evidence information \cite{Ma:2018:DRS:3184558.3188729, mohtarami-etal-2018-automatic, Xu2019AdversarialDA,augenstein-etal-2019-multifc}. 

The \textit{reasoning skills} required for automatic fact checking of claims depend on the nature of the employed dataset. Artificially constructed datasets \cite{thorne-etal-2018-fever,schuster-etal-2021-get}, where claims have been written based on Wikipedia evidence, can involve handling negations and simple lexical and semantic matching between the evidence and the claims. Some of them are designed to test for specific skills such as multi-hop reasoning \cite{jiang-etal-2020-hover} and tabular reasoning over structured evidence from tables in Wikipedia \cite{aly2021feverous}. Fact checking datasets containing real claims and evidence can require various complex reasoning skills, including multi-hop, logical and mathematical reasoning, but are limited in size \cite{alhindi-etal-2018-evidence,kotonya-toni-2020-explainable}. 

Existing work has explored the \textit{accountability} of fact checking models and pointed to the following model deficiencies. \citet{schuster-etal-2019-towards} were the first to reveal that fact checking models often make predictions based solely on the claim without consulting the provided evidence. \citet{schuster-etal-2021-get} show that fact checking models exhibit a bias for significantly higher word overlap in supporting evidence-claim pairs over refuting pairs. Finally, \citet{10.1162/tacl_a_00486} (\textit{Paper 1} \S\ref{chap:insufficient_information}) point that fact checking models are prone to make predictions based on insufficient information. 
% To address these deficiencies, challenge datasets have been collected to test and improve the models' robustness. 
\citet{thorne-etal-2019-evaluating} are the first to propose hand-crafted adversarial attacks for fact checking systems. In the subsequent FEVER 2.0 task \cite{thorne-etal-2019-fever2}, participants designed adversarial attacks for existing fact checking systems testing for multi-hop reasoning \cite{niewinski-etal-2019-gem, hidey-etal-2020-deseption} or generated various attacks manually \cite{kim-allan-2019-fever}. Finally, \citet{atanasova-etal-2020-generating} (\textit{Paper 2} \S\ref{chap:adversarial_claims}) propose a method to generate highly potent and semantically coherent adversarial attacks in an automated way.

Saliency explanations, abstractive, and extractive NLEs have been studied to enhance the \textit{transparency} of fact checking systems. In Wikipedia-based datasets, sentences from Wikipedia documents are extracted as explanations, and the retrieval performance of systems is measured jointly with the verification task -- a label prediction is considered correct only when the correct evidence is found \cite{thorne-etal-2018-fever}. In real-world datasets, where the claims are not artificially produced but occur naturally, summaries of long ruling comments justifying claims are used as explanations, and generated explanations are evaluated with ROUGE scores \cite{atanasova-etal-2020-generating, kotonya-toni-2020-explainable}. The implementation of the methods producing NLEs varies widely and can be grouped into models optimised separately or jointly with the task at hand \cite{malon-2018-team,atanasova-etal-2020-generating} (\textit{Papers 3} \S\ref{chap:generating_explanations}, \textit{4} \S\ref{chap:editing_explanations}, \textit{5} \S\ref{chap:multihop}). Existing work also proposes differentiable theorem proving approaches, which are self-interpretable models providing logical relations between the evidence and the claim and leading to a prediction \cite{DBLP:journals/corr/abs-2108-11357}.

\subsubsection{Question Answering}
Similar to fact checking, existing work has indicated that automatic question answering models have to learn a variety of complex reasoning skills, equivalent to human reasoning skills, in order to perform well on the task \cite{rogers2021qa,https://doi.org/10.48550/arxiv.2109.07102}. Challenge datasets are developed to audit question answering systems for complex reasoning skills such as multi-hop reasoning \cite{yadav-etal-2019-quick}, unanswerable questions \cite{rajpurkar-etal-2016-squad}, and logical reasoning \cite{ijcai2020-0501}. Explanations for question answering systems are produced by extracting supporting sentences from the provided document \cite{yadav-etal-2020-unsupervised,https://doi.org/10.48550/arxiv.2010.00389} (\textit{Paper 7} \S\ref{chap:diagnostic_guided_explanations}). Another way of producing explanations for question answering systems is by generating NLEs \cite{rajani-etal-2019-explain}, e.g., for common-sense multiple-choice question answering. 

\subsubsection{Natural Language Inference}
Natural language inference is the task of recognising textual entailment \cite{dagan2013recognizing} between two pieces of text, namely the premise and the hypothesis. Models have to predict the relation between the two parts, which could be entailment, contradiction, or neutral. Multiple challenge datasets have been developed to audit the reasoning capabilities of natural language inference models, including linguistic \cite{saha-etal-2020-conjnli} and logical reasoning \cite{tian-etal-2021-diagnosing}. Other studies produce challenge datasets to reveal flaws in the reasoning of automatic natural language inference models. \citet{gururangan-etal-2018-annotation} point that models can attain high performance based only on the premise without consulting the hypothesis. \citet{sanchez-etal-2018-behavior} find that natural language inference models are insensitive to small but semantically significant changes, and that their predictions can be manipulated with simple statistical correlations between words and training labels present in the training split. Explainability approaches for natural language inference models include post-hoc saliency explanations as well as abstractive NLEs \cite{NIPS2018_8163}, which usually are single sentences explaining the rationales of a model.

% \subsubsection{Long Document Understanding}

\subsection{Modelling Complex Reasoning Tasks}
Lately, language models (LMs) employing the Transformer architecture \cite{vaswani2017attention,devlin-etal-2019-bert} have become the core building blocks of architectures utilised to effectively automate many machine learning problems, including complex reasoning tasks. Such models have been the subject of rigorous studies inspecting their capabilities to learn complex reasoning skills. \citet{talmor-etal-2020-olmpics} find that different Transformer models exhibit qualitatively different reasoning abilities, such as that only RoBERTa-L \cite{liu2019roberta}, compared to BERT \cite{devlin-etal-2019-bert} and other RoBERTa model sizes, performs well at number comparison. \citet{kassner-schutze-2020-negated,nie-etal-2020-adversarial} find that LMs cannot detect the presence of negation in the input text. \citet{talmor-etal-2020-olmpics} discover that LMs are unable to learn multi-hop reasoning and even struggle to learn it with some supervision.

Several architectural improvements have been proposed to enhance the reasoning abilities of LMs. Graph attention networks \cite{liu-etal-2020-fine,zhou-etal-2019-gear} and eXtra-hop attention \cite{zhao2020transformer-xh} have been employed to improve the multi-hop reasoning abilities of LMs. Knowledge graphs have been incorporated into LMs to improve common-sense reasoning skills \cite{10.1007/978-3-030-77385-4_41}. Furthermore, contrastive learning techniques have been explored to improve a model's performance, especially given contrastive challenge datasets for particular skills \cite{schuster-etal-2021-get,10.1162/tacl_a_00486}. The publications in this thesis consider the Transformer architecture and its extensions to handle multi-hop reasoning (\textit{Paper 5}) as well as contrastive learning to improve a model's performance for instances with insufficient information (\textit{Paper 1}).

\section{Scientific Contributions}
\label{sec:contributions}

\subsection{Accountability for Complex Reasoning Tasks over Text}

\subsubsection{Paper 1: Fact Checking with Insufficient Evidence}

\begin{figure}[t]
    \centering
    \includegraphics[scale=0.95]{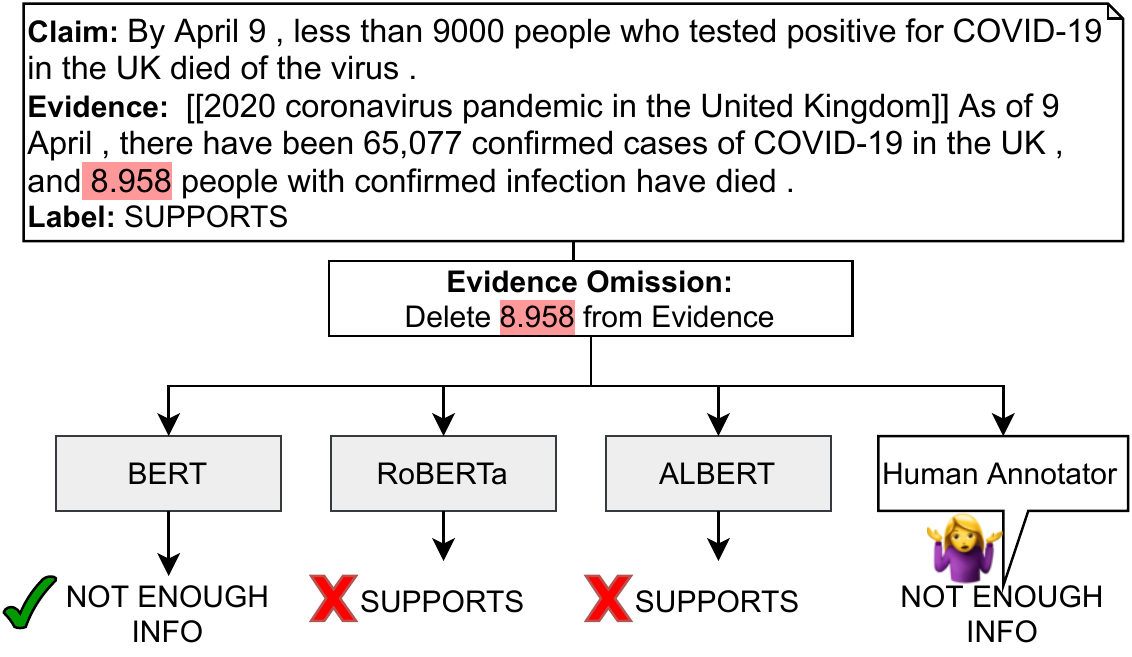}
    \caption{An example from the VitaminC test set, where the number modifier has been omitted from the evidence. This results in insufficient evidence for predicting its support for the claim as judged by human annotators. Two of the models still find the remaining evidence to be sufficient.}
    \label{fig:missing_intro}
\end{figure}

Automating the fact checking process relies on information obtained from external sources \cite{thorne-etal-2018-fever, diggelmann2020climate,Augenstein2021Doctoral} (see Section \ref{sec:fc}). However, the necessary information is not always available, either due to incomplete knowledge sources, or because the claim has newly emerged and the relevant facts are not documented yet. In this work, I posit that it is crucial for fact checking models to make veracity predictions only when there is sufficient evidence and otherwise indicate when it is not enough.

To this end, this work introduces the \textbf{novel task of \taskname\ illustrated in Figure\ \ref{fig:missing_intro}} \textbf{, which is defined as the task of identifying what information is sufficient for making a veracity prediction by fact checking models.}
I study the new task by, first, conducting a thorough empirical analysis of what models consider to be sufficient evidence for fact checking. For the \textbf{empirical analysis}, I propose \textbf{a new fluency-preserving method that occludes portions of the evidence}, automatically removing constituents or entire sentences, to create incomplete evidence. Secondly, I collect human annotations for sufficient evidence for fact checking, which results in a \textbf{novel challenge dataset, \datasetname}, for fact checking with omitted evidence. I observe that it is the hardest for fact checking models to detect when the evidence is missing information for the prediction that was removed from adverbial modifiers, followed by subordinate clauses. By contrast, it is easiest to detect missing information when it is a date modifier, followed by
number modifiers. Finally, I employ the information occlusion method introduced for the empirical analysis to \textbf{improve the performance of models on the new task of \taskname}. I show that considering it a component task of fact checking significantly improves fact checking performance. The performance for Evidence Sufficiency Prediction is improved by up to 17.8 $F_1$ score, which in turn improves fact checking performance by up to 2.6 $F_1$ score.

\subsubsection{Paper 2: Fact Checking with Insufficient Evidence}\begin{figure}[t]
\centering
\includegraphics[width=0.65 \columnwidth]{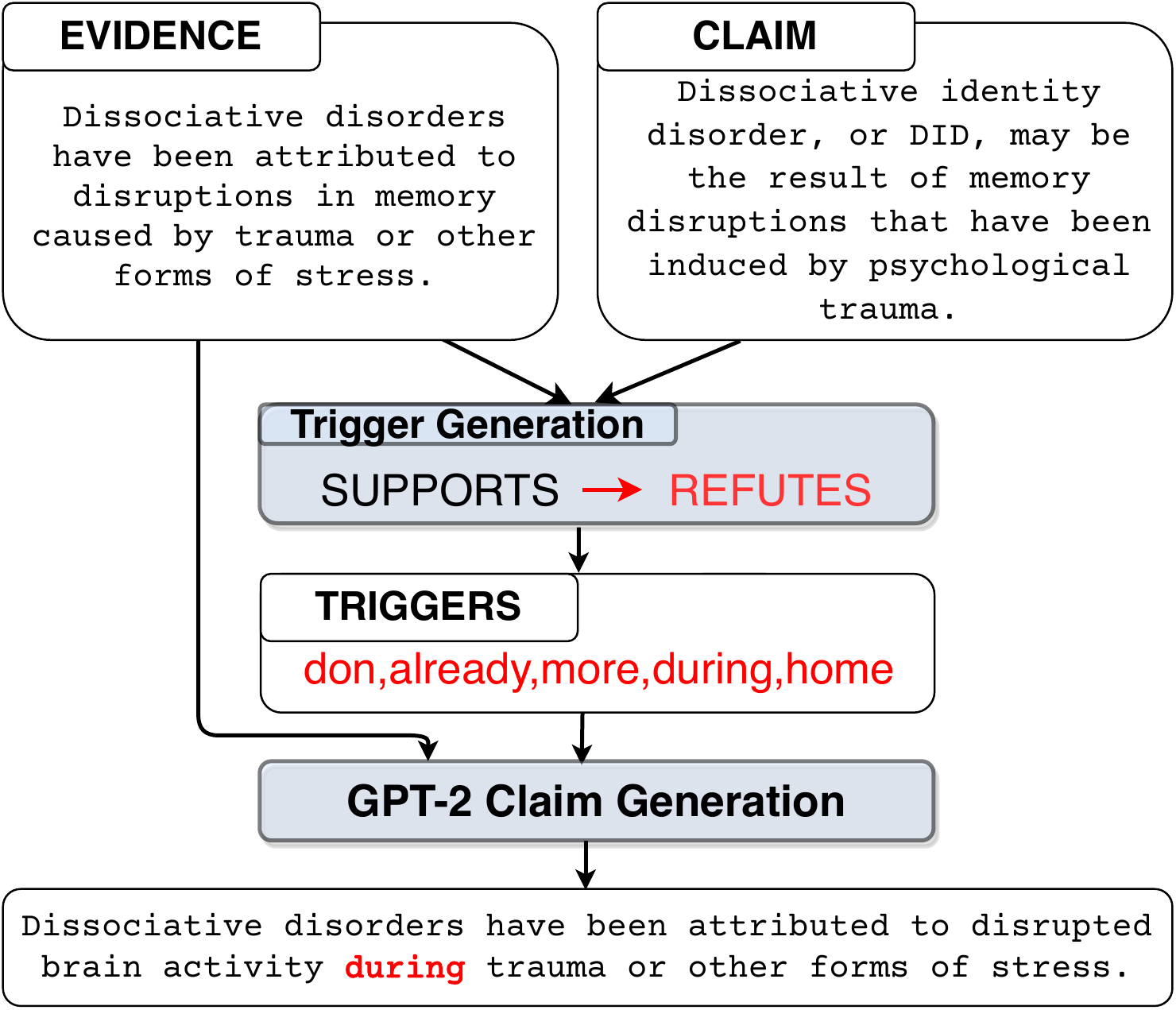}
\caption{High level overview of the method. First, universal adversarial triggers are discovered for flipping a source to a target label (e.g. SUPPORTS $\rightarrow$ REFUTES). These triggers are then used to condition the GPT-2 language model to generate novel claims with the original label, including at least one of the found triggers.}
\label{fig:puc_intro}
\end{figure}

Adversarial attacks reveal vulnerabilities and flaws of trained models \cite{https://doi.org/10.48550/arxiv.1412.6572, szegedy2013intriguing}. One attack type that has a high success rate in fooling a provided machine learning model is the universal adversarial triggers approach \cite{wallace-etal-2019-universal}. It produces individual n-grams, termed triggers, that, when appended to instances of a class under attack, can trick a model into predicting a target class, different from the instances' correct labels. However, for inference tasks such as fact checking, these triggers often invert the meaning of instances they are inserted into, thus also changing their gold labels. In addition, such attacks produce nonsensical inputs, as they simply concatenate triggers to existing samples. 
This paper proposes to address these two deficiencies of universal adversarial attacks, thus allowing for automatically generated adversarial attacks against fact checking systems that are both semantically valid and have correct gold labels.  
% investigates how to generate adversarial attacks against fact checking systems that preserve the ground truth meaning and the semantic validity of the input text. 
% preserve the ground truth meaning and are semantically valid.

The core contribution of the paper is a \textbf{method for automatically generating potent adversarial examples} that \textbf{preserve the meaning} of the source text and \textbf{improve the semantic validity} of universal adversarial triggers. This is accomplished via: 1) a \textbf{novel extension to the HotFlip attack}~\cite{ebrahimi-etal-2018-hotflip}, which jointly minimizes the target class loss of a fact checking model and the entailment class loss of a natural language inference model; 2) a \textbf{conditional language model} trained using GPT-2~\cite{radford2019language}, which takes %a set of 
trigger tokens and a piece of evidence, and generates a semantically coherent new claim containing at least one trigger. Figure \ref{fig:puc_intro} shows an overview of the method. %We demonstrate that the 
The resulting triggers maintain potency against a fact checking model while preserving the original claim label. Moreover, the conditional language model produces semantically coherent adversarial examples containing triggers, which lead to a decrease of 23.1 $F_1$ score points in the performance of the fact checking model when compared to its performance on the original claims. The resulting adversarial attacks unveil the vulnerability of fact checking models to particular trigger words present in the input, which require the development of appropriate defenses for ensuring robust fact checking performance.

\subsection{Explainability for Complex Reasoning Tasks over Text}
\subsubsection{Paper 3: Generating Fact Checking Explanations}
\begin{table}[t]
% \fontsize{8.4}{8.4}\selectfont
\begin{center}
\begin{tabular}{|p{425pt}|}
\toprule
\textbf{Claim}: The last major oil spill from a drilling accident in America happened over 40 years ago in 1969.\\ \midrule
\textbf{Ruling Comments}: 
(...) \hlyellow{The last major oil spill from a drilling accident in America happened over 40 years ago in 1969.} \\
% This item will deal with the claim about the date of the last major oil spill from a drilling accident. \\
\hspace*{3mm}(...) The largest in volume was the Santa Barbara spill of 1969 referenced by Murdock and Johnson, in which an estimated 100,000 barrels of oil spilled into the Pacific Ocean, according to the API. \hlblue{The Santa Barbara spill was so big it ranked seventh among the 10 largest oil spills caused by marine well blowouts in the world, the report states.} Two other U.S. spills, both in 1970, rank eighth and 10th. \hlred{Fourteen marine blowouts have taken place in the U.S. between 1969 and 2007.} Six of them took place after 1990 and spilled a total of nearly 13,700 barrels. \\
\hspace*{3mm}(...) We interviewed three scientists who said that the impact of a spill has little to do with its volume. \hlviolet{Scientists have proven that spills far smaller than Santa Barbara's have been devastating.} \\  \midrule
\textbf{Justification}: While the nation's largest oil well blowout did take place in 1969, it's not factually correct to call it the ``last major oil spill". First of all, two of the largest blowouts in the world took place in the U. S.  the following year. More importantly, experts agree that spills far smaller in volume to the 1969 disaster have been devastating. From a scientific perspective, Johnson's decision to single out the 1969 blowout as the last ``major" one makes no sense. \\ \midrule

\textbf{Ruling}: Half-True \\ \bottomrule
\end{tabular}
\end{center}
\caption{\label{tab:Example_intro} Example instance from the LIAR-PLUS dataset, with oracle sentences for generating the justification highlighted.}
\end{table}

Most existing work on automated fact checking is concerned with predicting the veracity of claims based on metadata, social network spread, language used in claims \cite{Ma:2018:DRS:3184558.3188729, mohtarami-etal-2018-automatic, Xu2019AdversarialDA}, and, more recently, evidence supporting or denying claims \cite{thorne-etal-2018-fever, stammbach-neumann-2019-team}. A crucial piece of the puzzle that is still missing is to understand how to automate the most elaborate part of the process -- generating justifications for verdicts on claims. 

In this publication, I present the \textbf{first study on generating natural language veracity explanations}, showing that they can successfully describe the reasons behind a veracity prediction as illustrated in Table \ref{tab:Example_intro}. This work frames fact checking explanations as extractive summarisation to address the challenges of the task stemming from the complex reasoning required for claim verification and the limited training data. I find that the veracity prediction model can utilise information from the detailed fact checking reports of professional journalists, resulting in a performance increase. The performance can be further improved by training veracity prediction and veracity explanation jointly. Notably, optimising the joint objective of veracity prediction and veracity explanation produces explanations that achieve better coverage and overall quality and serve better at explaining the correct veracity label than explanations learned solely to mimic human justifications. 
Overall, this work establishes important fundamentals in the area of transparency for models where explanations require complex reasoning and consist of multiple arguments spanning over several sentences and where the training resources are limited.

\subsubsection{Paper 4: Generating Fluent Fact Checking Explanations with Unsupervised Post-Editing}
\begin{figure}[t]
    \center
    \includegraphics[width=0.8\linewidth]{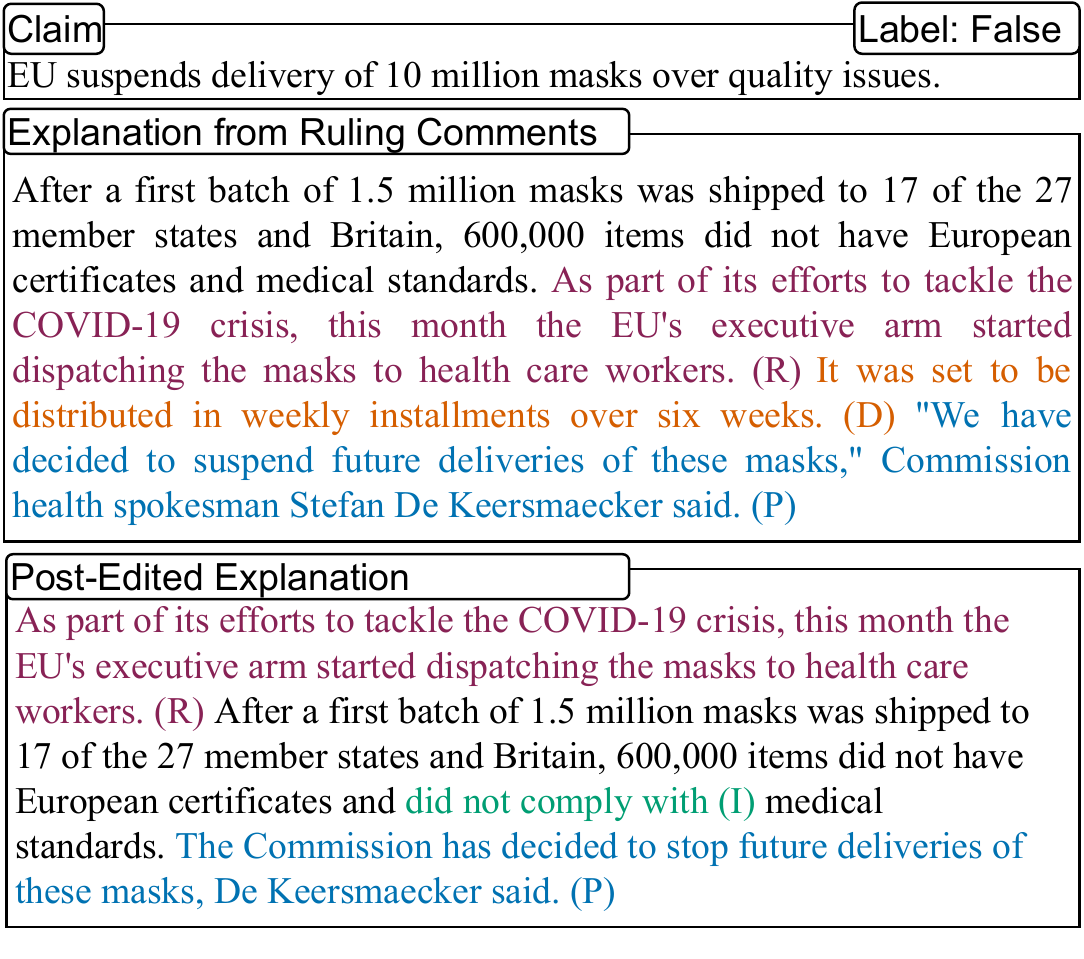}
    \caption{Example of a post-edited explanation from PubHealth that was initially extracted from RCs. It illustrates four post-editing steps: \textcolor{purple_editing}{reordering (R)},  \textcolor{green_editing}{insertion (I)}, \textcolor{orange}{deletion (D)}, and \textcolor{blue_editing}{paraphrasing (P)}.}
    \label{fig:introexp_intro}
\end{figure}

Fact-checking systems have become important tools to verify fake and misguiding news. These systems become more trustworthy when human-readable explanations accompany the veracity labels. Prior work \cite{atanasova-etal-2020-generating} (Paper 3 \S\ref{chap:generating_explanations}) has proposed to use automatic summarisation to select a subset of sentences from the long ruling comments (RCs) of professional journalists and used them as short layman explanations for fact checking veracity predictions. However, with a purely extractive approach, the sentences are cherry-picked from different parts of the corresponding RCs, and as a result, explanations are often disjoint and non-fluent. 

This work presents an \textbf{iterative edit-based algorithm that only uses phrase-level edits to perform unsupervised post-editing of disconnected extractive explanations} as illustrated in Figure \ref{fig:introexp_intro}. 
To the best of my knowledge, this work is the first to explore an iterative unsupervised edit-based algorithm using only phrase-level edits. The proposed algorithm also leads to the first computationally feasible solutions for unsupervised post-editing of long text inputs. A scoring function with components including fluency and semantic preservation is used to regulate the editing algorithm. Notably, combining the iterative post-editing algorithm with grammatical correction and paraphrasing-based post-processing leads to fluent and easy-to-read explanations. The paper presents extensive experiments on the LIAR-PLUS\ \cite{wang-2017-liar} and PubHealth\ \cite{kotonya-toni-2020-explainable} fact checking datasets. The automated evaluation confirms the success of the proposed method for preserving the semantics important to perform verification of the claim and enhancing the readability of the generated explanations. Finally, a manual evaluation confirms that the proposed approach improves the fluency and conciseness of the generated explanations.

\subsubsection{Paper 5: Multi-Hop Fact Checking of Political Claims}
\begin{figure}[t]
    \centering
    \includegraphics[scale=1]{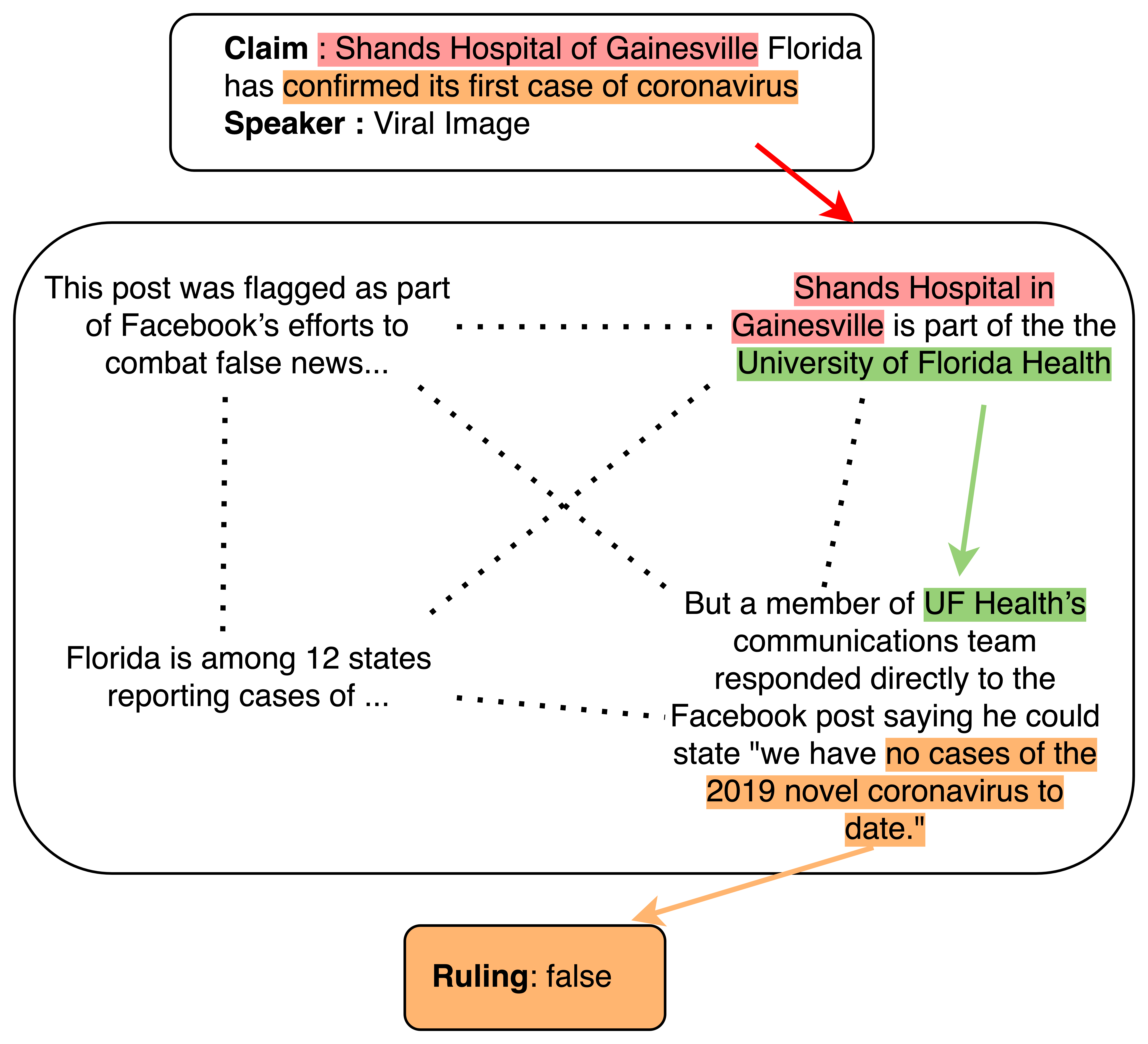}
    \caption{An illustration of multiple hops over an instance from \politihop. Each instance consists of a claim, a speaker% (author of the claim)%
    , a veracity label, and a PolitiFact article the annotated evidence sentences. The highlighted sentences represent the evidence sentences a model needs to connect to arrive at the correct veracity prediction.}
    \label{fig:example_4_intro}
\end{figure}

% Recent work has proposed multi-hop models and datasets for studying complex natural language reasoning \cite{liu-etal-2020-fine,zhou-etal-2019-gear,zhao2020transformer-xh}. One notable task requiring multi-hop reasoning is fact checking, where a set of connected evidence pieces leads to the final verdict of a claim as illustrated in Figure \ref{fig:example_4_intro}.
As noted in Section \ref{sec:fc}, one of the important reasoning skills required for fact checking is multi-hop reasoning, where a set of connected evidence pieces leads to the final verdict of a claim, as illustrated in Figure \ref{fig:example_4_intro}.
However, existing datasets do not provide annotations for gold evidence pages, except for FEVER \cite{thorne-etal-2018-fever}, where only 17\% of the claims require multi-hop reasoning and the claims are constructed artificially. 

This publication presents a study of more complex \textbf{claim verification with naturally occurring claims where rationales consist of multiple hops over interconnected evidence chunks}. To the best of my knowledge, this is the first work on multi-hop fact checking of political claims. To this end, this study constructs a small annotated dataset, PolitiHop, of evidence sentences for claim verification for the task. PolitiHop is employed to analyze to what degree existing multi-hop reasoning methods are suitable for the task. Furthermore, PolitiHop is used to investigate whether reasoning skills learned with a multi-hop model on similar datasets can be transferred to PolitiHop. The main finding of the study is that the task of multi-hop fact checking of real-world claims is complex and that the best performance is achieved with an architecture that specifically models multi-hop reasoning over evidence pieces in combination with in-domain transfer learning.

\subsection{Diagnostic Explainability Methods}
\subsubsection{Paper 6: A Diagnostic Study of Explainability Techniques for Text Classification}
\begin{figure}
\centering
\includegraphics[width=300pt]{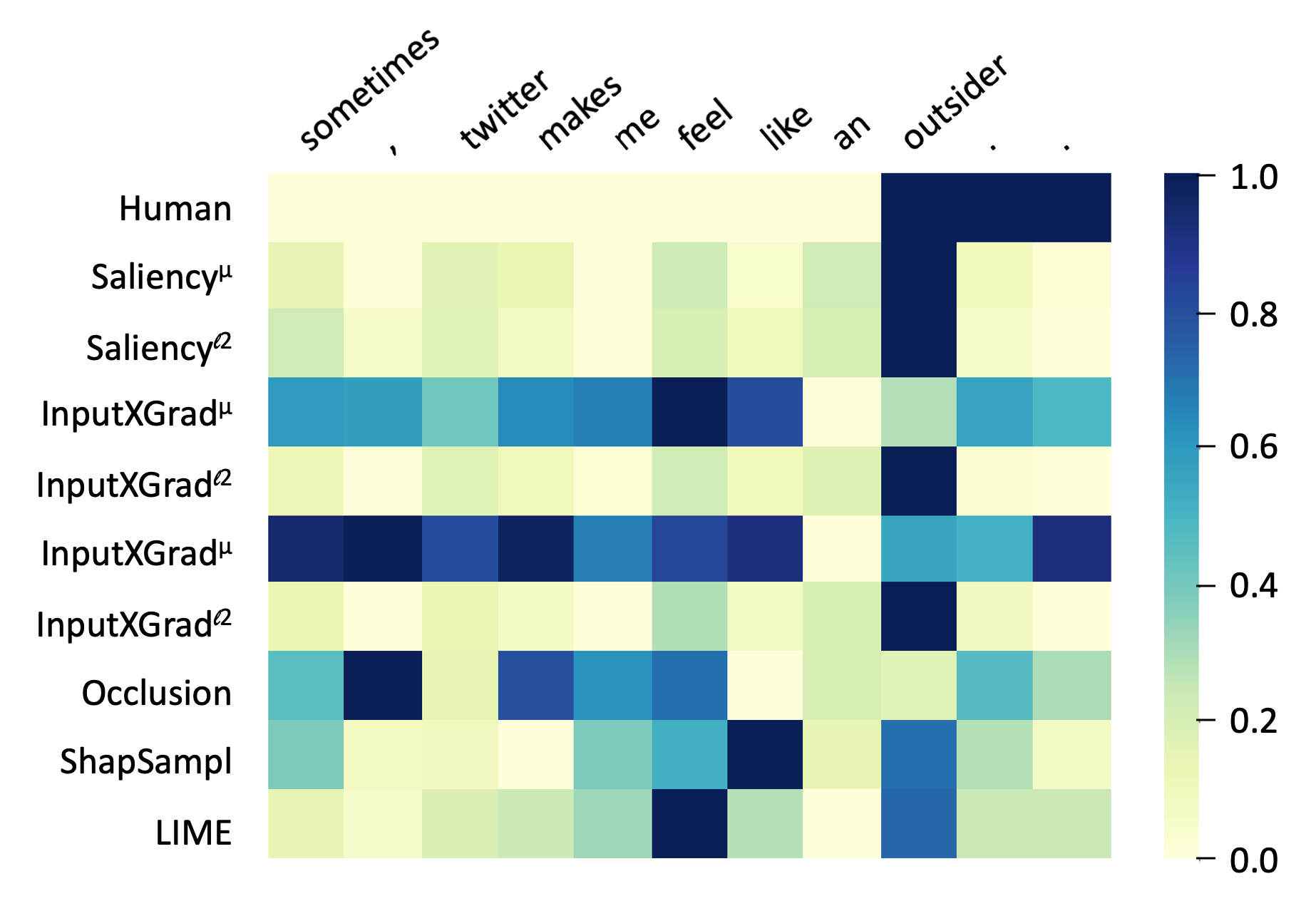}
\caption{Example of the saliency scores for the words (columns) of an instance from the Twitter Sentiment Extraction dataset. They are produced by the explainability techniques (rows) given a \trans{} model. The first row is the human annotation of the salient words. The scores are normalized in the range $[0, 1]$.}
\label{fig:example_intro}
\end{figure}

Recent developments in machine learning have introduced models that approach human performance at the cost of increased architectural complexity \cite{strubell-etal-2019-energy}. Efforts to make the rationales behind the models' predictions transparent have inspired an abundance of new explainability techniques. Provided with an already trained model, they compute saliency scores for the words of an input instance as illustrated in Figure \ref{fig:example_intro} (see Section \ref{sec:explainability}). However, there exists no definitive guide for: (i) how to choose such a technique given a particular application task and model architecture; and (ii) the benefits and drawbacks of using each such technique. In this paper, I develop a comprehensive list of diagnostic properties for evaluating existing explainability techniques. 

This work presents a \textbf{comprehensive list of \propertyplural{} for explainability and automatic measurement of them}, allowing for their effective assessment in practice. The proposed list of \propertyplural{} is used to study and compare the characteristics of different groups of explainability techniques in three different application tasks and three different model architectures. Furthermore, the list of \propertyplural{} is employed to study the attributions of the explainability techniques and human annotations of salient regions to compare and contrast the rationales of humans and machine learning models. Notably, the main finding of this diagnostic study of explainability techniques is that the investigated gradient-based explanation generation methods perform best across tasks and model architectures. This work also presents further detailed insights into the properties of the reviewed explainability techniques. 

\subsubsection{Paper 7: Diagnostics-Guided Explanation Generation}
\begin{figure}[t]
\centering
\includegraphics[width=0.75\columnwidth]{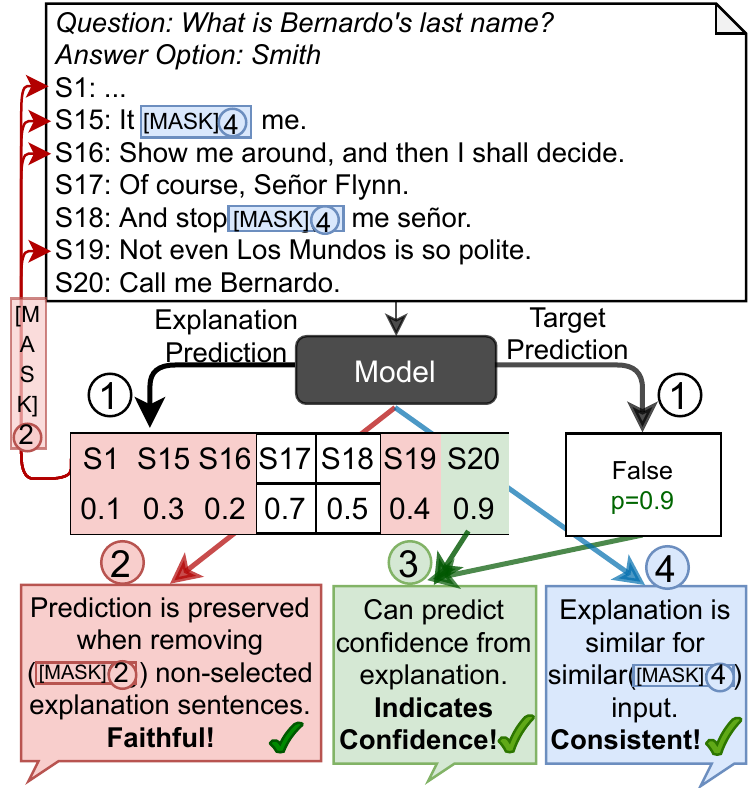}
\caption{Example instance from MultiRC with predicted target and explanation (Step 1), where sentences with confidence $\!\geq\!$ 0.5 are selected as explanations (S17, S18, S20). Steps 2-4 illustrate the use of \faithfulness, \consistency, and \confidence\ diagnostic properties as additional learning signals. `[MASK](2)' is used in Step 2 for sentences (in red) that are not explanations, and `[MASK](4)'--for random words in Step 4.}
\label{figure:example_intro} 
\end{figure}
% Extractive NLE techniques still produce natural text, as they usually choose whole sentences from the input

Extractive natural language explanation techniques shed light on a machine learning model's rationales by producing free text explanations, which are usually whole sentences extracted from the input (\S \ref{sec:nles}). 
% and can aid in identifying deficiencies in its reasoning process. 
Such techniques are typically constructed as models trained in a supervised way given human explanations. When such annotations are not available, explanations are often selected as those portions of the input that maximise a downstream task's performance, which corresponds to optimising an explanation's faithfulness to a given model. Faithfulness is one of several so-called diagnostic properties, which prior work has identified as useful for gauging the quality of an explanation without requiring annotations \cite{deyoung-etal-2020-eraser}. Other diagnostic properties are Data Consistency, which measures how similar explanations are for similar input instances, and Confidence Indication, which shows whether the explanation reflects the confidence of the model \cite{atanasova2021diagnostics} (Paper 6 \S\ref{chap:diagnostic_study_saliency}). 

The main contribution of this paper is a \textbf{novel method to learn the diagnostic properties -- Faithfulness, Data Consistency, and Confidence Indication, in an unsupervised way}, directly optimising for them to improve the quality of generated explanations as illustrated in Figure \ref{figure:example_intro}. I implement a joint task prediction and explanation generation model, which selects rationales at sentence level. Each property can then be included as an additional training objective in the joint model. With experiments on three complex reasoning tasks, I find that apart from improving the properties I optimised for, diagnostic-guided training also leads to explanations with higher agreement with human rationales and improved downstream task performance. Moreover, I find that jointly optimising for diagnostic properties leads to a reduced claim/question-only bias~\cite{schuster-etal-2019-towards} for the target prediction, which means that the model relies more extensively on the provided evidence. Importantly, I also find that optimising for diagnostic properties of explanations without supervision for explanation generation does not lead to good human agreement. This indicates the need for human rationales to train models that make the right predictions for the right reasons.
%\multirow{ 2}{*}{

\section{Summary of Contributions and Future Work}
\label{sec:summary}

\begin{table}[t]
\small
\centering
\begin{tabular}{lp{20pt}p{20pt}p{20pt}@{\hskip 0.05in \vline \hskip 0.05in}p{20pt}p{20pt}p{20pt}@{\hskip 0.05in \vline \hskip 0.05in}p{20pt}p{20pt}p{20pt}}
\toprule
% & \multicolumn{3}{c}{\textbf{Topic}} & \multicolumn{3}{c}{\textbf{Contribution}} \\
& \multicolumn{3}{c@{\hskip 0.05in \vline \hskip 0.05in}}{\rotatebox{0}{\textbf{\specialcell{Accountability}}}} & 
\multicolumn{3}{c@{\hskip 0.05in \vline \hskip 0.05in}}{\rotatebox{0}{\textbf{Explainability}}} & 
\multicolumn{3}{c}{\rotatebox{0}{\textbf{\specialcell{Explainability \\ Diagnostics}}}}\\ 
& M & D & A & M & D & A & M & D & A \\
\midrule
1. \citet{10.1162/tacl_a_00486} & \checkmark & \checkmark & \checkmark &  & & \\
2. \citet{atanasova-etal-2020-generating} & \checkmark & \checkmark & \checkmark & &  & \\
3. \citet{atanasova-etal-2020-generating-fact} & & & & \checkmark &  &  \\
4. \citet{DBLP:journals/corr/abs-2112-06924} & & & & \checkmark & &  \\
5. \citet{ijcai2021-536} & & \checkmark & \checkmark & & \checkmark & \checkmark  \\
6. \citet{atanasova-etal-2020-diagnostic} & & & & & & & \checkmark & & \checkmark \\
7. \citet{atanasova2021diagnostics} & & & & \checkmark & & & \checkmark \\
\bottomrule
\end{tabular}
\caption{Summary of contributions made by the publications in this thesis by topic -- Accountability Methods, Explainability and Explainability Diagnostics, and type of contribution -- Methodological (M), Dataset (D), Diagnostic Analysis (A).}
\label{tab:contributions}
\end{table}
The publications in this thesis collectively contribute to advancing the state of the art of accountable and transparent machine learning for complex reasoning tasks over text. In particular, they facilitate the analysis of the reasons behind the outputs of ML models and assist in detecting and correcting for potential harms. Table \ref{tab:contributions} maps the methodological, dataset, and analysis contributions of each paper along each of the accountability and transparency axes.  

Many of the proposed methods for auditing and explaining machine learning models in this thesis are empirically validated on the task of fact checking as it provides a rich test bed for testing complex reasoning over text. Fact checking is also considered in the publications in this thesis as it is particularly critical when developing models for this task that they are both accountable and transparent. I further test on other complex reasoning tasks, namely natural language inference and question answering, where appropriate datasets are available. Due to the fact-checking task's complexity and the methods proposed in this work being generally applicable, they could also be easily validated on other tasks requiring complex reasoning skills, given suitable benchmark datasets.

\subsection{Accountability for Complex Reasoning Tasks over Text}
The contributions made in the area of \textit{accountability} of machine learning models include challenge datasets for prediction with insufficient information (Paper 1 \S\ref{chap:insufficient_information}) and multi-hop reasoning (Paper 5 \S\ref{chap:multihop}) as well as a dataset showing a model's vulnerability to adversarial manipulations (Paper 2 \S\ref{chap:adversarial_claims}). The methodological contributions presented in my thesis enable the automated generation of these challenge and adversarial datasets, which extends the applicability of the proposed accountability audits for other complex reasoning tasks and machine learning models. The resulting resources reveal important insights about the models' capabilities and advance our understanding of the models' decision-making processes. Furthermore, they reveal model vulnerabilities that necessitate the development of appropriate complex reasoning models and defenses against adversarial attacks. To this end, this thesis also makes methodological contributions that improve the reasoning capabilities of machine learning models regarding the uncovered vulnerabilities, thus leading to enhancements in their accountability. 

While challenge and adversarial datasets for complex reasoning are developed mainly for natural language inference, I present studies improving the accountability of fact checking systems, which have longer textual inputs and require more complex reasoning skills with compositions thereof. Moreover, the accountability of fact checking systems in deployment is imperative and requires extensive research, even more so for critical domains such as fact checking of medical claims.

The current research landscape of challenge datasets and adversarial attacks for complex reasoning tasks is discordant -- separate studies investigate a limited number of complex reasoning skills and reasoning flaws over one or a few models and tasks. Hence, some prospects for future work include efforts to unify the studies and findings on complex reasoning skills and flaws in models' rationales across different tasks and models. This could potentially result in benchmarks designed around the notion of skills rather than downstream tasks, which could facilitate overall assessments of the accountability of machine learning models. Furthermore, future interdisciplinary synergies involving linguistics, cognitive science, and machine learning could lead to the design of comprehensive lists of complex reasoning skills.

\subsection{Explainability for Complex Reasoning Tasks over Text}

In regards to \textit{model transparency}, this thesis pushes the state-of-the-art for generated natural language explanations (Papers 3 \S \ref{chap:generating_explanations}, 4 \S \ref{chap:editing_explanations}, and 5 \S \ref{chap:multihop}, and 7 \S \ref{chap:diagnostic_guided_explanations}) for fact checking systems. Generating NLEs for the veracity predictions of real-world claims is a particularly challenging task as there are limited training resources, and it requires multiple connected arguments to be presented in a readable and accessible way. With this thesis, I lay the foundations for automatically generating such explanations. The produced explanations improve our understanding of fact checking models' decision-making processes. They can instill trust in the models' predictions, serve as a further means for auditing the accountability of machine learning models, and enable end users to expand their knowledge by leveraging information from the rationales of the models.

There is still a lack of sufficiently large datasets for generating complex reasoning explanations for real-world fact checking and, in general, for explanations consisting of multiple arguments spanning over several sentences. This limits the possible achieved quality of generated explanations and could be addressed in future work. Moreover, there is a need for datasets that could allow for automating real-world fact checking explainability fully -- from collecting evidence documents to producing veracity labels and explanations.

There are many prospects for enhancing the availability of datasets for natural language explanation generation in general. The explanations contained in existing natural language inference datasets are often based on templates making them rather structure-based explanations \cite{wiegreffe2021teach} and the quality of some common-sense NLE datasets is questioned in related work \cite{DBLP:journals/corr/abs-2004-14546,wiegreffe2021teach}. Moreover, there are currently no existing datasets with NLEs for tasks that require different complex reasoning skills. Such datasets could be developed in future work to generate explanations and study models' rationales for the different types of complex reasoning skills required for a task.

\subsection{Diagnostic Explainability Methods}
Explainability techniques aim to reveal the rationales of machine learning models. End users make decisions to trust and rely on models' predictions based on the explanations produced to reveal the rationales employed by the models for their predictions. Hence, explainability techniques have to be robust and faithful to the underlying model as well. This thesis moves forward our collective knowledge of the field of \textit{explanation diagnostics} with a diagnostic study of post-hoc saliency-based explainability techniques (Paper 6 \S \ref{chap:diagnostic_study_saliency}), which are further directly optimised for in generated explanations, thus improving explanation quality (Paper 7 \S \ref{chap:diagnostic_guided_explanations}). The insights gained from the analysis performed in my thesis reveal which explainability techniques perform better than others, as well as which tasks and models necessitate the development of more robust and appropriate explainability techniques. Finally, improving the quality of the produced explanations additionally enhances the understanding and trust in the model's rationales. Measuring and improving explanations' quality instills trust in the employed explainability approach as well.

Currently, there is limited work on measuring and ensuring the quality of NLEs \cite{wiegreffe-etal-2021-measuring,hase-etal-2020-leakage}. NLEs are generated by supervised systems, optimised to resemble human-annotated explanations, which does not guarantee that they convey the rationales used by a model. This calls for future studies examining the faithfulness and other properties of NLEs and improving these properties in generated NLEs.

% 

% \begin{table}[t]
% \small
% \centering
% \begin{tabular}{lcccc}
% \toprule
% & \textbf{\specialcell{Fact \\ Checking}} & \textbf{\specialcell{Natural \\ Language \\ Inference}} & \textbf{\specialcell{Question \\ Answering}} & \textbf{Long Text} \\ \midrule
% 1. \citet{10.1162/tacl_a_00486} & \checkmark &  \\
% 2. \citet{atanasova-etal-2020-generating} & \checkmark & \\
% 3. \citet{atanasova-etal-2020-generating-fact} & \checkmark & \\
% 4. \citet{DBLP:journals/corr/abs-2112-06924} & \checkmark & \\
% 5. \citet{ijcai2021-0536} & \checkmark &  \\
% 6. \citet{atanasova-etal-2020-diagnostic} & \checkmark & \checkmark & & \checkmark \\
% 7. \citet{atanasova2021diagnostics} & \checkmark & & \checkmark & \checkmark \\
% \bottomrule
% \end{tabular}
% \caption{Summary of the complex reasoning tasks used to validate the methodological contributions in this thesis.}
% \label{tab:contributions}
% \end{table}

\part{Accountability for Complex Reasoning Tasks over Text}
\chapter{Fact Checking with Insufficient Evidence}
\label{chap:insufficient_information}
\section{Introduction}
% Automating the time-consuming process of professional fact checking is an important task that has led to the creation of several fact checking (FC) datasets differing in size, complexity, and domain~\cite{thorne-etal-2018-fever, diggelmann2020climate, thorne2021evidence}. 
\begin{figure}[t]
    \centering
    \includegraphics[scale=0.95]{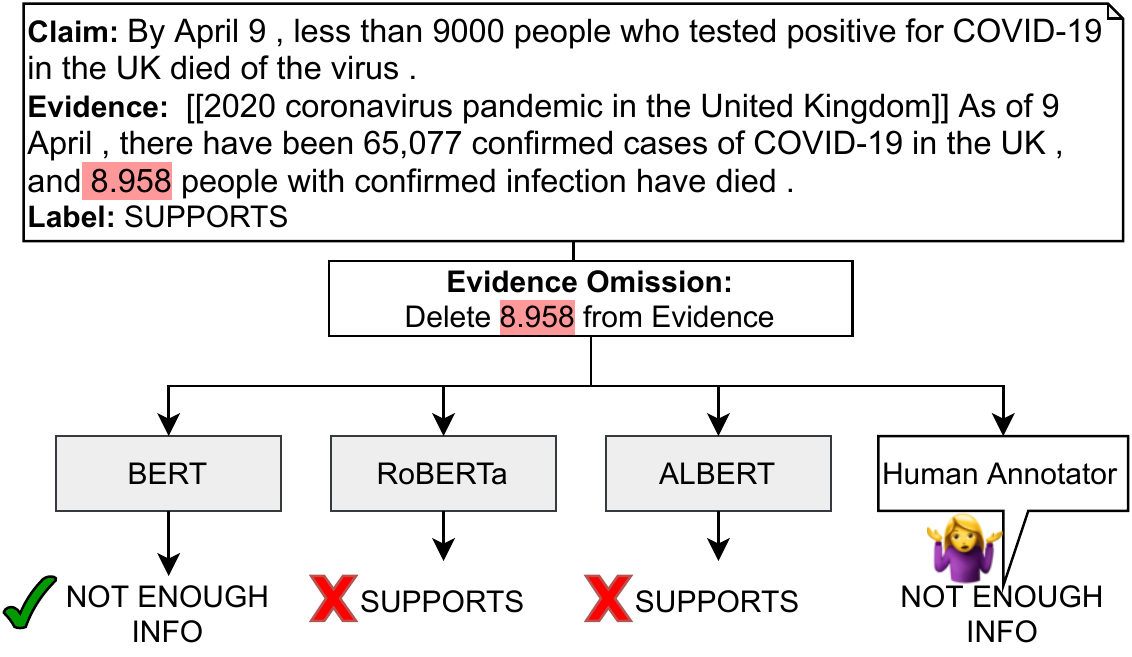}
    \caption{An example from the VitaminC test set, where the number modifier has been omitted from the evidence. This results in there not being enough evidence for predicting its support for the claim as judged by human annotators, while two of the models still find the remaining evidence to be sufficient.}
    \label{fig:missing}
\end{figure}

Computational fact checking approaches typically use deep learning models to predict the veracity of a claim given background knowledge~\cite{thorne-etal-2018-fever, diggelmann2020climate,Augenstein2021Doctoral}. However, the necessary evidence is not always available, either due to incomplete knowledge sources, or because the claim has newly emerged and the relevant facts are not documented yet. In such cases, FC models should indicate that the information available is insufficient to predict the label, as opposed to making a prediction informed by spurious correlations.
% In addition, FC models are typically based on language models pre-trained on textual data that can change its veracity over time~\cite{schuster-etal-2021-get}.

Prior work shows that FC models can sometimes predict the correct veracity based on just the claim, ignoring the evidence, and that they can overly rely on features such as the word overlap between the evidence and the claim~\cite{schuster-etal-2019-towards,schuster-etal-2021-get}, leading to biased predictions. 
%Previous work has shown that FC models can only rely on the claim as opposed to predicting based on the evidence documents paired with the claim, or can make biased predictions, e.g., based on word overlap between the evidence and the claim~\cite{schuster-etal-2019-towards,schuster-etal-2021-get}. 
However, there are no previous studies on what evidence a FC model considers to be enough for predicting a veracity label. To this end, this work introduces the \textbf{novel task of \taskname\ illustrated in Fig.\ \ref{fig:missing}} \textbf{, which we define as the task of identifying what information is sufficient for making a veracity prediction.} This task is related to FC and can operate on instances and models from FC datasets, but is focused on evaluating the capability of models to detect missing important information in the provided evidence for a claim. The latter is usually not evaluated explicitly in current FC benchmarks, where joint scores disregard a FC model's prediction when insufficient evidence is retrieved.

We study the new task by, first, conducting a thorough empirical analysis of what models consider to be sufficient evidence for FC. Secondly, we collect human annotations for the latter, which results in a novel diagnostic dataset, \datasetname, for FC with omitted evidence. Finally, we employ the method introduced for the empirical analysis to improve the performance of models on the new task of \taskname, and show that considering it a component task of FC significantly improves FC performance. %and FC in general.
% We demonstrate that these insights will lead to more robust fact checking models, avoiding fall-backs of models on spurious correlations. 

For the \textbf{empirical analysis}, we propose a new fluency-preserving method that occludes portions of evidence% texts
, automatically removing constituents or entire sentences, to create incomplete evidence. We provide those as input to an ensemble of Transformer-based FC models to obtain instances on which FC models agree vs. disagree to have (in)sufficient information. %To make sure our findings generalise beyond a single pre-trained Transformer model and dataset, w
We perform extensive experiments with three models -- BERT~\cite{devlin-etal-2019-bert}, RoBERTa~\cite{liu2019roberta}, ALBERT~\cite{Lan2020ALBERT:}, and three textual FC datasets with different types of claims -- FEVER~\cite{thorne-etal-2018-fever}, HoVer~\cite{jiang-etal-2020-hover}, VitaminC~\cite{schuster-etal-2021-get}.

To compare model behavior with human rationales for \taskname, we ask annotators to %manually annotate
indicate if the occluded evidence texts still provide enough information for a fact-check. This results in a \textbf{novel diagnostic test dataset, \datasetname}, which contains information about the type of the omitted information, allowing for in-depth analyses of model behavior. 

Finally, to improve model performance for detecting omitted important evidence and, in turn, FC, we propose to combine the proposed evidence omission method with tri-training~\cite{10.1109/TKDE.2005.186}, which utilises the agreement of three different machine learning models to label unlabeled training instances (\S\ref{sec:method}). This results in a \textbf{novel counterfactual data augmentation schema for learning of (in)sufficient information}. We find that the proposed approach is highly effective in improving model performance by up to 17.8 $F_1$ score on the newly introduced \datasetname. This also leads to improvements of up to 2.6 $F_1$ score on the standard FC test sets for the corresponding datasets. 

% Existing adversarial attacks usually perform insertions and changes of words or are manually generated by a human. In this work, as we are interested in finding what the model finds as important in the input, we perform removal of information pieces, which allows to measure the causal influence of the information piece on the prediction of the model. We analyse the effect of the occlusions by distinguishing between easy, hard and ambiguous instances, between in-domain and out-of-domain ones. We produce diverse input perturbations and provide them to a human annotator to verify which of the removed information pieces are not removing important information. 

% Following \citet{nie-etal-2020-adversarial}, we also use part of the annotated instances for improving a model's performance in a few iterations. Unlike \citet{nie-etal-2020-adversarial}, we propose a curriculum learning set-up, where the annotations are used during the training of the model with contrastive learning that indicates to the model which information piece is not/important. Thus, we alleviate the need for training multiple models, while still building an increasingly robust model. Apart from that, as human annotations are expensive, we provide a list of perturbations that already show the effect of a number of occlusions for the annotator to choose from.
\section{Related Work}
\label{sec:related}
Here, we study when models trained on existing FC datasets find evidence with omitted important information to still be sufficient for veracity prediction. Such cases might be considered vulnerabilities of the models and can be due to models' faulty reasoning, learned biases, etc. Hence, our work is mainly related to studies exploring potential biases learned by FC models and the vulnerabilities of FC models to adversarial attacks. We further propose a method for evidence omission, which creates counterfactual instances, which is related to studies on input-level instance re-writing. 
We also use the proposed evidence omission method to collect counterfactually augmented data (CAD) and compare that to using the collected data in a contrastive learning (CL) loss to improve performance on \taskname\ and FC more generally. We thus discuss the relationship between our work and prior studies on CAD and CL. Finally, we compare our work based on deep learning models to FC performed against knowledge bases (KBs), where fact triples can also be missing. % Hence, our study is also related to work on counterfactually augmented data (CAD).
% The evidence omission method is also used by us for learning to improve performance on \taskname. Hence, our study is also related to work on contrastive self-learning.

\textbf{Fact Checking Diagnostics.} Previous work has exposed various biases of FC models. 
While FEVER~\cite{thorne-etal-2018-fever} is one of the largest datasets for FC, \citet{schuster-etal-2019-towards} points out that models trained on it can verify a claim solely based on the text of the claim, without considering the evidence. To this end, \citet{schuster-etal-2019-towards} introduce a new diagnostic dataset, FeverSymmetric, of contrastively re-written claims and evidence. They show that the models fail to detect the contrastive changes in the text, leading to a drop of up to 57.46 $F_1$-score, compared to 85.85 $F_1$-score on the original FEVER development set. 
Furthermore, the claims in FEVER were manually written based on Wikipedia article sentences, and thus have a large token overlap between the evidence and the claim, especially for supporting evidence. Hence, \citet{schuster-etal-2021-get} construct a new FC dataset, VitaminC, where they instruct the annotators to avoid using the same words as in the evidence. \citet{ijcai2021-536} further create PolitiHop -- a dataset for claim verification of naturally occurring claims with evidence comprised of multiple hops over interconnected evidence chunks. They study how multi-hop vs. single inference architectures reason over the evidence sets in PolitiHop. In addition, several works~\cite{thorne-etal-2019-evaluating,niewinski-etal-2019-gem, hidey-etal-2020-deseption} explored the vulnerability of FC models to adversarial attacks, e.g., by discovering universal trigger words that fool a model into wrongly changing its prediction~\cite{atanasova-etal-2020-generating}. In contrast, we are interested in how much evidence is enough for veracity prediction, studying this with three different FC models trained on three different datasets by omitting information at the constituent and sentence levels and comparing it to human judgments.

\textbf{Instance Re-Writing.} The above studies mainly perform re-writing or insertion operations for FC evidence. Here,
we employ causal interventions on the evidence by omission to study when information is (in)sufficient for a model's prediction. \citet{elazar2021amnesic} also use causal interventions that estimate the importance of a property by removing it from a representation. By comparison, even though text-level causal interventions are more intricate due to the discrete nature of text, we perform them on the text itself, by following linguistic rules for optional constituents to preserve the semantics and the fluency of the text. \citet{thorne-vlachos-2021-evidence} perform re-writing of claims by masking and then correcting separate words. They thus generate claims supported by the evidence, particularly for claims not supported before the factual correction. In similar vein, \citet{wright-etal-2022-generating} decompose long, scientific claims into shorter, atomic claims. They then generate negative instances for those by masking single words in claims and replacing them with antonyms retrieved from a scientific knowledge base. In contrast, we perform omissions of evidence information at the sentence and constituent levels and for the new task of \taskname.

\textbf{Contrastive Learning (CL) and Counterfactual Data Augmentation (CAD).} Most existing work of CL in NLP employs contrastive self-learning for model pre-training~\cite{rethmeier2021primer}. Contrary to this, \citet{rethmeier2020long} propose for CL to be performed jointly with the supervised objective. We follow the latter to improve the performance of FC models in detecting when important information is missing from the evidence, by using the original evidence texts paired with evidence texts with omitted information as contrastive data points. We perform contrastive self-training jointly with the supervised objective, as we use the contrastive loss as an unsupervised training for \taskname. In contrast, using it for pre-training followed by supervised training could lead to the models forgetting the information learned during pre-training%by the self-supervised contrastive pre-training
, which is needed to improve the performance on \datasetname. 
An important factor for CL is the augmentation of negative and positive instances, which can be challenging due to the discrete nature of text. Related work explores augmentation through back-translation ~\cite{sennrich-etal-2016-improving}, masked word substitution with an LM~\cite{wu2019conditional}, graph neighbourhood sampling~\cite{ostendorff2022neighborhod}, mix-up~\cite{chen-etal-2020-mixtext}, or a combination thereof~\cite{qu2021coda}. In a similar vein, automated approaches for CAD in NLP include paraphrasing~\cite{iyyer-etal-2018-adversarial}, and controlled~\cite{madaan2021generate} text generation, which do not necessarily change the target label of an instance. CAD is found to improve model robustness to data artifacts~\cite{Kaushik2020Learning,cadteney} and to perform better out of domain~\cite{samory2021sexism,sen-etal-2021-counterfactually}. In contrast, we use evidence omission, combined with tri-training for contrastive negative evidence mining (\S\ref{sec:method}).

\textbf{Knowledge-Base Fact Checking.} A relevant line of work conducts FC against knowledge bases (KBs) by finding fact triple chains that are (in)consistent with the claim~\cite{kimfact}. Discovering such missing triples could also be used to detect insufficient evidence information. As KBs can contain an incomplete set of fact triples, related work completes KBs from unstructured textual data on the Web~\cite{trisedya-etal-2019-neural} or with graph embedding techniques~\cite{kim2018kbcnn}. This work uses machine learning models that use textual evidence as input instead of performing an intermediate step of completing a knowledge base with needed fact triples.

\begin{table*}[t]
% \fontsize{10}{8}\selectfont
\centering
\begin{tabular}{p{2.5cm}p{11cm}}
\toprule
\textbf{Dataset/Size}& \textbf{Example} \\
\midrule
FEVER\newline145,449 train \newline999,999 dev\newline 999,999 test & \textbf{Label}: REFUTES {($\in$ \{SUPPORTS, REFUTES, NOT ENOUGH INFO\})}\newline\textbf{Claim}: Sindh borders Indian states and is in India.\newline\textbf{Evidence}: [Sindh] Sindh is home to a large portion of Pakistan's industrial sector and contains two of Pakistan's commercial seaports -- Port Bin Qasim and the Karachi Port. \\ \midrule
Vitamin C\newline370,653 train \newline63,054 dev \newline55,197 test & \textbf{Label}: SUPPORTS {($\in$ \{SUPPORTS, REFUTES, NOT ENOUGH INFO\})}\newline\textbf{Claim}: Westlife sold more than 1 m. video albums and made over 23.5 m. sales in the UK.\newline\textbf{Evidence}: [Westlife] According to the British Phonographic Industry (BPI), Westlife has been certified for 13 m. albums, 1.3 m. video albums, and 9.8 m. singles, with a total of more than 24 m. combined sales in the UK. \\ \midrule
HoVer\newline18,171 train \newline1818 dev \newline4,000 test & \textbf{Label}: NOT SUPPORTED { ($\in$ \{SUPPORTS, NOT SUPPORTS=(REFUTES+NOT ENOUGH INFO)\}}\newline\textbf{Claim}: Reason Is Treason is the second single release from a British rock band that are not from England. The band known for the early 90's album Novelty are not from England either.\newline\textbf{Evidence}: [Kasabian] Kasabian are an English rock band formed in Leicester in 1997. [Jawbox] Jawbox was an American alternative rock band from Washington, D.C., United States. [Reason Is Treason] "Reason Is Treason" is the second single release from British rock band Kasabian. [Novelty (album)] Novelty is an album from the early 90's by Jawbox. \\
\bottomrule
\end{tabular}
\caption{Sizes and examples instances for the studied fact checking datasets (see \S \ref{sec:datasets}).}
\label{tab:datasets}
\end{table*}

\section{Datasets}
\label{sec:datasets}

We employ three fact checking datasets (see Table\ \ref{tab:datasets}) %shows information about the datasets, including example instances. 
and use the gold evidence documents, i.e., we do not perform document or sentence retrieval (apart from for the ablation experiment in Section \ref{sec:data:irrelevant}). Thus, we avoid potential enforced biases for the veracity prediction models if they had to learn to predict the correct support of the evidence for the claim given wrong evidence sentences. Hence, each of the three fact checking datasets $D=\{(x_i, y_i)| x_i=(c_i, e_i), i \in [1,|D|] \}$ consists of instances with input $x_i$ and veracity labels $y_i$. The input is comprised of a claim $c_i$ and gold evidence $e_i$. The veracity label $y_{i}\in$ \{0=SUPPORTS, 1=REFUTES, 2=NEI\} for FEVER and VitamiC, and $y_{i} \in$ \{0=SUPPORTING, 1=NOT SUPPORTING\} for HoVer.

\textbf{FEVER~\cite{thorne-etal-2018-fever}} contains claim-evidence pairs, where the evidence consists of sentences from Wikipedia pages, and the claims are written manually based on the content of those Wikipedia pages. 87\% of the claims have evidence consisting of one sentence. The dataset has a high ratio of token overlap between the claim and the evidence, where the overlap is naturally higher for claims that are supporting (69\%), than refuting (59\%) and NEI (54\%) claims.
The high overlap ratio can create a bias for learning from token overlap, which can further prevent generalisation, as also noted in related work~\cite{schuster-etal-2021-get}.
% For motivation? In our work (Sec.~\ref{sec:overlap}), we also study the extent to which models can detect omitted evidence depending on the overlap of words between the claim and the evidence. 
% Apart from that, the dataset contains a large amount of similar claims such as ``X is a person.". In such cases, the model can learn to associate certain claim patterns with labels or the model could already contain from pre-training the needed information. In both cases, the model can learn to ignore the evidence, which could lead to wrong predictions as facts can change. Our study of evidence omission can reveal when the model does or does not consult the relevant information in the evidence needed to verify the claim.

\textbf{Vitamin C~\cite{schuster-etal-2021-get}} is a collection of sentences from Wikipedia containing factual edits. For each factual edit, annotators construct a claim that is SUPPORTED and one that is REFUTED with the old and the new version of the evidence. When the factual edit introduces/removes facts from the evidence, claims are constructed so that there is NOT ENOUGH INFORMATION (NEI) to support them. Due to its contrastive nature and reduced claim-evidence overlap, the authors demonstrate that models trained on the dataset gain a 10\% accuracy improvement on adversarial fact verification.
% TODO: add info about NEI and num of sentences

\textbf{HoVer~\cite{jiang-etal-2020-hover}} is designed to collect claims that need several hops over Wikipedia evidence sentences to verify a claim. The evidence contains between two and four sentences from different Wikipedia articles. As the test dataset is blind and we use the gold evidence, we use the development set for testing purposes and randomly select 10\% of the training dataset for development.

\section{Evidence Omission}
\label{sec:omission}
To study what types of information the evidence models consider important, we propose to conduct causal interventions for the evidence by omitting information from it. %at the constituent and sentence levels. 
We hypothesise that removing information important for the model to predict the support of evidence for a claim will cause a change in its original prediction, leading to the model indicating that there is missing information. If the removed information is not important for the model though, removing it would not change the model's prediction. We then ask whether the information that is important for a model when predicting the support of the evidence text for a claim, is actually important as judged by human annotators. The human annotations allow for a systematic study of common model errors, i.e., when the models still predict the correct label even if important evidence information has been removed and when they consider the information to be insufficient if unrelated evidence has been removed.

\subsection{Evidence Omission Generation}
\label{sec:omission:gen}
\begin{table*}[t]
% \fontsize{10}{8}\selectfont
\centering
\begin{tabular}{llp{4.7cm}p{7cm}}
\toprule
\textbf{Type} & \textbf{L} & \textbf{Claim} & \textbf{Evidence} \\
\midrule
S & R & The Endless River is an album by a band formed in 1967. & [[The Endless River]] The Endless River is a studio album by Pink Floyd. \textcolor{red}{[[Pink Floyd]] Pink Floyd were founded in 1965 by students \dots} \\
PP & R & Uranium-235 was discovered by Arthur Jeffrey Dempster in 2005. & [[Uranium-235]] It was discovered in 1935 \textcolor{red}{by Arthur Jeffrey Dempster}.\\
NOUNM & S & Vedam is a drama film. & [[Vedam (film)]] Vedam is a 2010 Indian \textcolor{red}{drama} film written and directed by Radhakrishna Jagarlamudi \dots\\
ADJM & S & Christa McAuliffe taught social studies. & [[Christa McAuliffe]] She took a teaching position as a \textcolor{red}{social} studies teacher at Concord High School\dots \\
ADVM & S & Richard Rutowski heavily revised the screenplay for Natural Born Killers. & [[Natural Born Killers]] The film is based on an original screenplay that was \textcolor{red}{heavily} revised by writer David Veloz , associate producer Richard Rutowski \dots \\
NUMM & S & Being sentenced to federal prison is something that happened to Efraim Diveroli. & [[Efraim Diveroli]] Diveroli was sentenced to \textcolor{red}{four} years in federal prison .\\
DATEM & R & Colombiana was released 1st October 2001. & [[Colombiana]] Colombiana is a French action film from \textcolor{red}{1st October} 2011  \dots\\
SBAR & R & North Vietnam existed from 1945 to 1978. & [[North Vietnam]] North Vietnam, was a state in Southeast Asia \textcolor{red}{which existed from 1945 to 1976}.\\
\bottomrule
\end{tabular}
\caption{Examples from the FEVER dataset of constituent types (\S\ref{sec:omission-types}) removed from the evidence for a claim with Label (L) one of SUPPORTS (S) or REFUTES (R).}
\label{tab:omm:examples}
\end{table*}
\label{sec:omission-types}

We omit information from the evidence text at the sentence and constituent level. Particularly, we aim to remove information from the evidence such that it does not change its stance towards the claim from SUPPORTS to REFUTES, or vice-versa, while preserving the grammatical correctness and fluency of the evidence. Following studies of linguistic sentence structure~\cite{burton2016analysing,borjars2019introducing}, illustrated with examples in Table\ \ref{tab:omm:examples}, we collect prepositional phrases, modifiers and other optional sentence constructs -- i.e. those constructs that can be removed from the sentence without impairing its grammatical correctness, and where the remaining text is semantically identical to the original one, except for the additional information from the removed construct \citep{garvin1958syntactic}. We use the following optional sentence constructs:

%\begin{itemize}[nosep]
%\item 
\textbf{Sentences (S).} In FEVER and HoVer, the evidence can consist of more than one sentence. The separate sentences are supposed to contain information important for the fact check, which we further verify with manual annotations as explained in Section\ \ref{sec:manual_annotations}. VitaminC consists of single sentences only, and we thus only perform constituent-level omissions for it, as described next. %Note that, for the VitaminC dataset, we only perform constituent-level omission, as its evidence consists of single sentences.

%\item 
\textbf{Prepositional Phrases (PP)} %Looking at the constituency parse tree of a text, PPs that 
are optional phrases that are not part of a Verb Phrase (VP), but are child nodes of the root sentence in the constituent tree \citep{brown1991syntax}. These usually function as adverbs of place and consist of more than one word.

%\item 
\textbf{Noun Modifiers (NOUNM)} %In general, modifiers 
are optional elements of a phrase or clause structure \citep{huddleston2005cambridge}. NOUNM can be a single or a group of nouns that modify another noun.

%\item 
\textbf{Adjective Modifiers (ADJM)} are a single or a group of adjectives that modify a noun.

%\item 
\textbf{Adverb Modifiers (ADVM)} are a single or a group of adverbs that modify verbs, adjectives, or other adverbs and typically express manner, place, time, etc.

%\item 
\textbf{Number Modifiers (NUMM)} are a single or a group of words denoting cardinality that quantify a noun phrase.

%\item 
\textbf{Date Modifiers (DATEM)} are a single or a group of words that express temporal reference. To preserve fluency, from a date expression consisting of a day, a month, and a year, we omit either the date, the date and the month, the year, or the year and the date.

%\item 
\textbf{Subordinate Clauses (SBAR)} are introduced by a subordinating conjunction. Subordinate clauses depend on the main clause and complement its meaning. SBARs can be adverb clauses, adjective clauses, and noun clauses.

%\end{itemize}

For the omission process, we use two pre-trained models with high performance from the Spacy library\footnote{\url{https://spacy.io/}} -- a part-of-speech (PoS) tagger with an accuracy of 97.2 and a constituency parser~\cite{kitaev-klein-2018-constituency} with an $F_1$-score of 96.3 on the revised WSJ test set~\cite{bies2015english}.
During the omission process, we use the PoS tags to find nouns, adjectives, adverbs, and numbers and use the constituency tags to select only the modifiers. Thus, we find the NOUNM, ADJM, ADVM, and NUMM constructs. We collect SBAR and PP constructs by finding their corresponding tags in the constituent dependency tree. Finally, for the date, we use two regular expressions that are common date templates used in Wikipedia articles -- <month name, date, year> or <date, month name, year>, and remove parts from the templates that preserve the coherency -- <date>, <year>, <month name and date>, or <year and date>.

Overall, in this work, we perform a study of insufficient evidence for FC by removing information from the gold evidence. As explained in Section\ \ref{sec:related}, we perform causal interventions on the evidence by omission to study when information is (in)sufficient for a model's prediction. Replacement of words is another operation that can be applied to the evidence. We can, for example, replace different types of named entities with pronouns, and different parts of the speech with demonstrative pronouns to induce insufficient information. However, the replacement operation does not allow for direct causal conclusions as any change of a word with another could potentially lead to confounding factors of the newly introduced word and the model's predictions. Note that, there are some pronouns used in the evidence when they refer to the person/object of the article. We do not treat such cases as insufficient information as the title of the page with the name of the person/object is always prepended to the sentence, which allows for coreference resolution. Finally, another possible operation is the insertion of new information, which would lead to insufficient evidence when performed on the claim. The latter, however, requires the insertion of text that preserves the grammatical correctness and meaning of the claim, which is hard to achieve in an automated way.

\subsection{Manual Annotations.}
\label{sec:manual_annotations}
\textbf{Models.} We train three Transformer-based FC models -- BERT~\cite{devlin-etal-2019-bert}, RoBERTa~\cite{liu2019roberta}, and ALBERT~\cite{Lan2020ALBERT:}. BERT is pre-trained with masked language modeling and next sentence prediction objectives on the Toronto Book Corpus~\cite{kiros2015skip} and the English Wikipedia.\footnote{\url{https://en.wikipedia.org}} It is also the most widely used pre-trained Transformer model.\footnote{\url{https://huggingface.co/models}} RoBERTa improves upon BERT by optimising key hyper-parameters, and is trained without the next sentence prediction objective. RoBERTa is one of the top-performing models on the GLUE~\cite{wang-etal-2018-glue} and SuperGLUE~\cite{NEURIPS2019_4496bf24} benchmarks comprised of various NLP tasks. The latter also holds for ALBERT, another Transformer architecture that improves upon BERT. It does so with parameter-reduction techniques, which lower the memory consumption of the model. ALBERT also employs a self-supervised pre-training loss for inter-sentence coherence. The latter is found to be beneficial for tasks with multiple sentences, and \citet{schuster-etal-2021-get} report improved FC robustness with it on VitaminC compared to BERT. 

We train each model on the respective training splits of each dataset with the claim $c$ and the gold evidence $e$ as input to predict the gold veracity label $y$: $f(c,x) = \hat{y}$. We optimise the supervised cross-entropy loss:
\begin{equation}
\label{crossentropy:general}
\mathcal{L}^{S}=-\frac{1}{m} \sum_{j=1}^{m} y^{j} \cdot \log(\hat{y}^{j})
\end{equation}
\noindent where $m$ is the label space size.

We then use an ensemble of these three different Transformer-based FC models to collect predictions for our new task \taskname, as we want to find instances with omitted information that are more broadly applicable (e.g., those on which the models agree). The (dis)agreements between the models also allow us to study the differences between them in detecting omitted information. Transformer Language Models are pre-trained on large datasets, the veracity of which can change over time~\cite{schuster-etal-2021-get}. This makes it important that the FC models take into account the facts in the given evidence. When provided with differences and similarities in the three FC models' predictions, future work could then also investigate the degree to which different Transformer-based FC models encode FC-relevant world knowledge they default to in their predictions.

\textbf{Annotation Task.} Next, we collect evidence with removed information as described above. We then use the models to find which of the omitted evidence they consider important, resulting in a prediction change to NEI. We consider instances from the original test splits of each of the datasets, where all models predicted the veracity correctly before the evidence omission was performed, as these are the cases where we can observe whether evidence omission causes the veracity prediction to change to NEI. We collect instances with omitted evidence information where the models: (1) agree that the evidence is still enough vs. (2) insufficient; and where they (3) disagree in their prediction. We collect a total of 400 instances at the sentence, and 600 instances at the constituent level from the test splits of the corresponding datasets, distributed equally among the above three groups.

We employ annotators on Amazon Mechanical Turk\footnote{\url{https://www.mturk.com/}}. We first train potential annotators, presenting them with annotation guidelines and illustrative examples. We then select annotators using a qualification test with nine test annotations for our task. Each annotation had the cost of 0.10\$, and annotators were paid 10\$ on average per hour. %In the annotation task, they are asked to 
The annotation task is to determine whether the evidence is still sufficient for predicting the label without the omitted information. If the remaining evidence is still sufficient, we ask them for the reason -- %If the evidence is still sufficient, we ask them to choose 
whether this is because the removed evidence is repeated in the remaining text or because the removed evidence is not relevant to the veracity of the claim. Following the annotation guidelines for FEVER and HoVer, we ask the annotators not to use any world knowledge or knowledge they might have about the claim. For more details on the annotation task and the guidelines, we will release the dataset with a detailed README file.
%\footnote{\url{https://bit.ly/3lplNCy}}
  
The final dataset \datasetname\ = $\{(x_i', y_i')| x_i'=(c_i, e_i'), i \in [1, |\datasetname|]\}$ consists of test instances $x_i'$ with labels $y_i'$. All of the instances in \datasetname\ are a subset of the instances in the test datasets of FEVER, VitaminC, and HoVer with the following changes. The input $x_i'$ is comprised of the original claim $c_i$ and the evidence with omitted information $e_i'$. The tokens of $e_i'$ are a subset of the tokens of the original gold evidence $e_i$ of the instance. To re-iterate, the label of the originally selected instances is either SUPPORTS or REFUTES, i.e. they have sufficient gold evidence information, where after omitting information from the evidence, the new label $y_i'$ becomes either NEI if the majority of the annotators selected that important information was removed, and otherwise remains the original label -- SUPPORTS and REFUTES for FEVER and VitamiC, or SUPPORTING for HoVer.

The resulting inter-annotator agreement (IAA) for \datasetname\ is 0.81 Fleiss' $\kappa$ from three annotators. Due to the novelty of the introduced task of \taskname, we do not have direct points of comparison for IAA. However, we point as a reference the IAA reported for the related task of fact checking for the HoVer dataset -- 0.63 Fleiss' $\kappa$, and for the FEVER dataset -- 0.68 Fleiss' $\kappa$, where, for both datasets, the annotators were thoroughly trained and highly paid. The biggest challenges for our annotators, judging by their errors during the qualification test, were not to use common knowledge and assumptions in their annotations, and the general complexity of the task. 
% We believe these challenges to be the primary detractors from the IAA scores.

\subsection{\datasetname\ Analysis.} 
\label{sec:analysis}

\begin{table}[t!]
\fontsize{10}{10}\selectfont
\centering
\begin{tabular}{m{1.6cm}lrrr}
\toprule
\textbf{Dataset} & \textbf{Model Pred} & \textbf{EI\_I} & \textbf{EI\_R} & \textbf{NEI} \\ \midrule
\multirow{4}{0pt}{FEVER SENT} & EI Agree & \cellcolor{mydblue} 61 & \cellcolor{mydblue} 20 & \cellcolor{mylgreen} 119\\ 
& NEI Agree & \cellcolor{mylgreen} 13 & \cellcolor{mylgreen} 9 & \cellcolor{mydblue} \underline{178}\\ 
& Disagree & 39 & 24 & 137\\ 
& Total & 113 & 53 & 434\\ 
\midrule
\multirow{4}{0pt}{FEVER CONST} & EI Agree & \cellcolor{mydblue} 146 & \cellcolor{mydblue} 3 & \cellcolor{mylgreen} 51\\ 
&NEI Agree & \cellcolor{mylgreen} 0 & \cellcolor{mylgreen} 0 & \cellcolor{mydblue} 200\\ 
&Disagree & 43 & 1 & 156\\ 
&Total & 189 & 4 & 407\\
\midrule
\multirow{4}{0pt}{HoVer SENT} & EI Agree & \cellcolor{mydblue} \underline{32} & \cellcolor{mydblue} \underline{12} & \cellcolor{mylgreen} 156\\ 
& NEI Agree & \cellcolor{mylgreen} 4 & \cellcolor{mylgreen} 1 & \cellcolor{mydblue} 195\\ 
& Disagree & 7 & 1 & 192\\ 
& Total & 43 & 14 & 543\\
\midrule
\multirow{4}{0pt}{HoVer CONST} & EI Agree & \cellcolor{mydblue} 139 & \cellcolor{mydblue} 6 & \cellcolor{mylgreen} 55\\ 
& NEI Agree & \cellcolor{mylgreen} 1 &\cellcolor{mylgreen} 0 & \cellcolor{mydblue} 199\\ 
& Disagree & 48 & 1 & 151\\ 
& Total & 188 & 7 & 405\\ 
\midrule
\multirow{4}{0pt}{VitaminC CONST} & EI Agree & \cellcolor{mydblue} \textbf{146}  & \cellcolor{mydblue} \textbf{5} & \cellcolor{mylgreen} 49\\ 
& NEI Agree & \cellcolor{mylgreen} 0 & \cellcolor{mylgreen} 0 & \cellcolor{mydblue} \textbf{200}\\ 
& Disagree & 13 & 0 & 187\\ 
& Total & 159 & 5 & 436\\ 
\midrule \midrule
\multirow{4}{0pt}{Total} & EI Agree & \cellcolor{mydblue} 524 & \cellcolor{mydblue} 46 & \cellcolor{mylgreen} 430\\ 
& NEI Agree & \cellcolor{mylgreen} 18 & {\cellcolor{mylgreen} 10} & \cellcolor{mydblue} 972\\ 
& Disagree & 150 & 27 & 823\\ 
& Total & 692 & 83 & 2225\\ 
\bottomrule
\end{tabular}
\caption{Statistics of \datasetname\ presenting the predictions of the models in the ensemble (Model Pred: Agree Enough Information (EI Agree), Agree Not Enough Information (NEI Agree), Disagree, and Total) vs human annotations of the same (EI -- Irrelevant (EI\_I), EI -- Repeated (EI\_R),  NEI). We present sentence (SENT) and constituent omission (CONST) dataset splits separately.
We embolden/underline results of the datasets for predictions where the three models agree (NEI Agree, EI Agree) and have the highest/lowest agreement with human annotations about EI\_I, EI\_R and NEI predictions. We use \hlightgreen{light blue}/\hldblue{dark blue} to denote where lower/higher results are better.}
\label{tab:dataset}
\end{table}

\textbf{Overall Agreement with Annotators.} The statistics of the resulting dataset, \datasetname, are presented in Table\ \ref{tab:dataset}. We find that all three models agree that the remaining evidence is still sufficient (EI Agree) even when it has become insufficient after omitting information needed for verifying the claim (NEI) in 430 out of 1000 instances. We assume that these failures of all three models to detect missing information for FC point to the models making predictions based only on patterns observed in claims, or to the models defaulting to world knowledge encoded in the pre-trained Transformer models. We further find that when the models disagree about whether the remaining information is still sufficient (Disagree), they disagree mostly about instances where the omitted evidence information is needed for veracity prediction (NEI) -- in 823 out of 1000 instances. By contrast, when the models agree that the remaining evidence is insufficient, they are correct in 972 out of 1000 of the instances. 

\textbf{Separate Dataset Agreement with Annotators.} Looking at the separate datasets, it is the hardest for the models to identify missing evidence information needed for the fact check (EI Agree vs. NEI) for HoVer, particularly with sentence omissions, and the easiest for the VitaminC dataset with constituent omissions. We hypothesise that the latter is due to the HoVer dataset having more complex claims and requiring cross-sentence reasoning, whereas VitaminC contains contrastive instances which, during training, guide the models to identify the parts of the evidence needed for FC. Overall, the models fail to detect missing information more from sentences rather than from constituents. 
We hypothesise that this effect can be observed partly because models struggle to conduct multi-hop reasoning over them. Another possible reason for that is that the models could be better at verifying the type of information removed from a sentence constituent rather than from a sentence.

% We assume that when evidence consists of multiple sentences, models fail to conduct multi-hop reasoning over them and instead find single-hop evidence patterns to conduct the fact-check, which leads to them not being able to detect missing multi-hop evidence information.
\begin{figure}[t]
    \centering
    \includegraphics[scale=2]{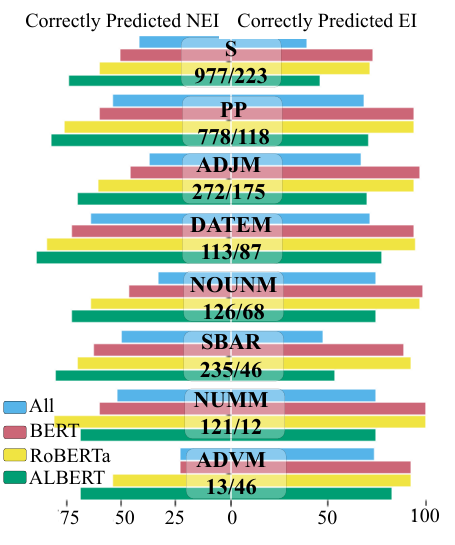}
    \caption{\datasetname\ -- fine-grained analysis by type of removed evidence inftype (\S\ref{sec:omission:gen}) vs. proportion of correct predictions of NEI/EI instances. The proportion is computed for the separate models -- BERT, RoBERTa, ALBERT, and for all three models agreeing on the correct NEI/EI label (All). The total number of NEI/EI instances of each type is provided under each of the types of removed evidence information. \textit{A higher} proportion of correct predictions is \textit{better}.}
    \label{fig:types}
\end{figure}
\textbf{Performance by Omitted Evidence Type and Model.} Figure\ \ref{fig:types} provides a fine-grained analysis of the performance of the models for different types of omitted constituents. We observe that it is the hardest to detect when the evidence is missing information for the prediction (Correctly Predicted NEI) that was removed from adverbial modifiers (ADVM), followed by subordinate clauses (SBAR). By contrast, it is easiest to detect missing information when it is a date modifier (DATEM), followed by number modifiers (NUMM). BERT has the lowest rate of correctly detecting insufficient evidence from the three models, followed by RoBERTa, whereas ALBERT performs best. We conjecture that this is due to RoBERTa being an optimisation of BERT, and due to ALBERT including pre-training with an inter-sentence coherence objective, which has been shown to make the model more robust for factual verification~\cite{schuster-etal-2021-get}. Even though ALBERT contains fewer parameters than BERT, it still detects better when the evidence is insufficient. Finally, we see a natural trade-off between correctly detecting sufficient and correctly detecting insufficient information. In particular, some models such as ALBERT have a higher number of correct predictions on instances without enough information (Fig.\ \ref{fig:types}, left). However, on instances with sufficient evidence information (Fig.\ \ref{fig:types}, right), ALBERT has the lowest number of correct predictions. In contrast, BERT has the worst performance on the NEI instances, but the best performance on EI instances.

\section{Evidence Omission Detection}
\label{sec:method}
\begin{figure}[t]
    \centering
    \includegraphics[scale=1]{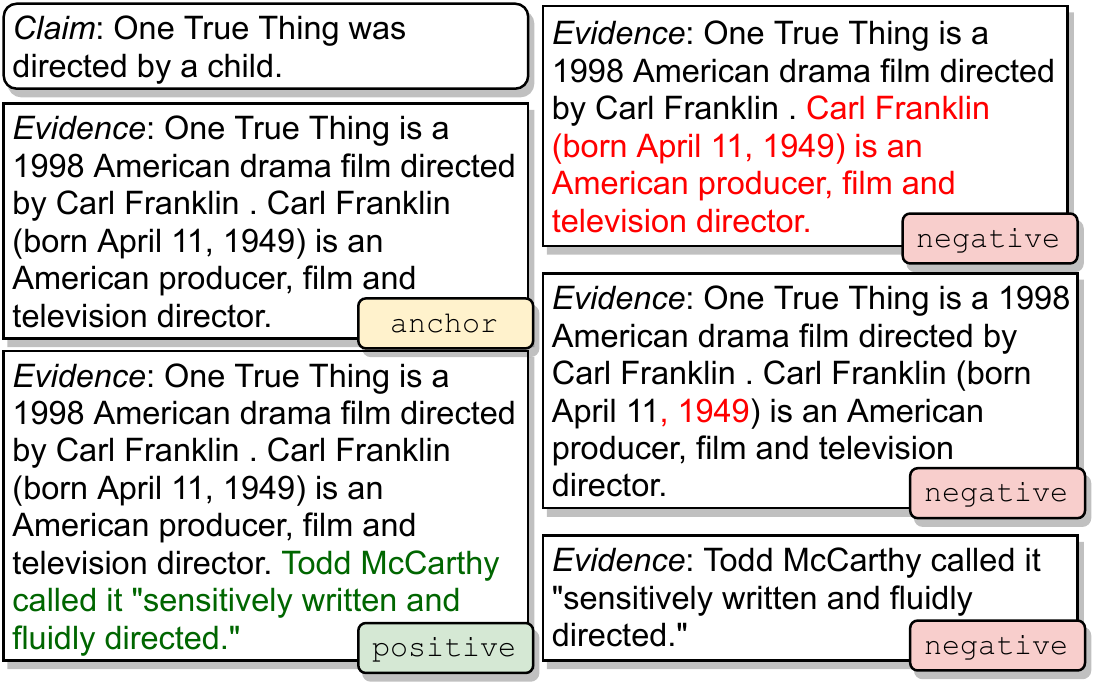}
    \caption{Example of augmented contrastive instances for the original (anchor) instance. \textcolor{red}{Red} designates removed evidence information, where the models agree that the remaining evidence is not sufficient, producing a negative contrastive instance. \textcolor{green_editing}{Green} designates an added distractor sentence, producing a positive instance. The distractor sentence, selected to have high overlap with the claim but with insufficient information, is used as another negative instance.}
    \label{fig:contrastive}
\end{figure}

To improve the performance of models in recognising when the evidence is not enough for verifying a claim, we experiment with CAD (\S\ref{sec:method:cad}) and a CL loss (\S\ref{sec:method:cl}). Both methods use contrastive data augmented with the proposed evidence omission method (\S\ref{sec:omission:gen}) in combination with tri-training, as illustrated in Fig.\ \ref{fig:contrastive}.
We omit information from the original (anchor) evidence to collect potential negative instances with missing important evidence information compared to the original evidence (Fig.\ \ref{fig:contrastive}, right). From the resulting candidates, we select as negative only those predicted as having insufficient information by the other two supervised models from the ensemble (\S\ref{sec:omission}) (e.g., RoBERTa and ALBERT predict NEI when we are training a model with a BERT Transformer architecture). We also collect positive instances that still have sufficient evidence information after applying a data augmentation operation. For each instance $x_i$, we find one distractor sentence from the document of the gold evidence that is the most similar to the claim by word overlap. We append the distractor sentence to the original evidence, which serves as a positive instance (Fig.\ \ref{fig:contrastive}, left). Finally, we include only the distractor sentence as a negative instance as it does not have enough evidence contrasted both with the positive and the anchor instances. We conjecture that the latter would serve as a training signal for avoiding the bias for overlap between the claim and the evidence.

\subsection{Contrastive Learning}
\label{sec:method:cl}
We study self-supervised learning to train FC models that recognise when the evidence is not enough for verifying a claim. In particular, we propose to use self-supervised contrastive learning (CL) jointly with the supervised learning of the model to predict the support of the evidence for a claim. Given an anchor instance $x_i$, a positive instance $x_i^+$, and $K^-$ negative instances $x_{i,k}^-$, $k \in [1, K^-]$, the objective of CL is to make the anchor and the positive instance closer in the representation space, and the anchor and the negative instances further apart. The anchor, positive, and negative instances are collected and/or augmented from the training splits of the corresponding datasets as described above.
% As the original CL employs one positive instance, we include the two positive instances by having a triple consisting of the anchor, the negatives, and one positive repeated two times for each positive.
Each model, $g(x) = l(h(x)) = l(e) = \hat{y}$, uses 12 encoding layers to encode an input instance $h(x) = e$ and uses the encoding $e$ of the last encoding layer to predict the veracity label with a linear layer: $l(e) = \hat{y}$. We encode the anchor, the positive, and the negative instances with the corresponding model $g$, resulting in the anchor $e_i$, the positive  $e_{i}^+$, and the negative $e_{i,j}^-$ representations, and minimise the following CL loss:
\begin{equation}
% \small
\mathcal{L}^{\mathrm{CL}}\!=\!\log \sigma(s(e_{i}, e_{i}^{+}\!;\!\tau)\!+\!\sum_{k=1}^{K^{-}}\\log\sigma(1\!-\!s(e_{i}, e_{i,k}^{-}\!;\!\tau))
\end{equation}
% \begin{equation}
% \mathcal{L}^{CL}=-\log \frac{e^{s(e_{i}, e_{i}^{+})/ \tau} } {\sum_{j=1}^{K^{-}}e^{s(e_{i}, e_{i,j}^{-})/ \tau} }
% \end{equation}
\noindent where $s$ is a similarity function between the representation of the two instances -- cosine similarity in our case, $\tau$ is a temperature parameter subtracted from the cosine similarity~\cite{ma-collins-2018-noise}, and $K^{-}$ is the number of negatives. Note that the CL loss is the same as Noise Contrastive Estimation~\cite{ma-collins-2018-noise} expressed as a binary objective loss. The representation of each instance is obtained by mean pooling of the word representations of the instance in the last layer of the model M. We include the contrastive self-learning loss for those instances that are not annotated as NEI, as we cannot construct contrastive negative evidence with insufficient information for the instances that already do not have enough information for verification. Finally, the CL loss is optimised jointly with the supervised loss:
\begin{equation}
\label{crossentropy}
\mathcal{L}^{S}=-\frac{1}{m} \sum_{j=1}^{m} y^{j} \cdot \log(\hat{y}^{j})
\end{equation}
\begin{equation}
\mathcal{L}=\mathcal{L}^{S} + \mathcal{L}^{\mathrm{CL}}
\end{equation}
\noindent where $\hat{y}_{i}$ is the label prediction of model M, $m$ the label space size, $y_{i}$ is the gold label for instance $x_i$, $y_{i}\in$ \{0=SUPPORTS, 1=REFUTES, 2=NEI\} for FEVER and VitamiC, and $y_{i} \in$ \{0=SUPPORTING, 1=NOT SUPPORTING\} for HoVer.

\subsection{Counterfactual Data Augmentation}
\label{sec:method:cad}
We also experiment with counterfactually augmented evidence%to train FC models that recognise when the evidence is not enough for verifying a claim. 
%We use
, using the negative and positive instances constructed as described above (\S\ref{sec:method} and Fig.\ \ref{fig:contrastive}). As the models have high accuracy when they agree that a piece of evidence with omitted information is not sufficient (see agreement with human annotations in Table\ \ref{tab:dataset}), we conjecture that the counterfactually augmented instances would serve as a good training signal for detecting (in)sufficient evidence information without incurring annotation costs for training data. 
The counterfactually augmented data is thus simply combined with the training instances of each dataset.  In particular, we include in the training set the claim and the original evidence (anchor) with the corresponding  gold label $y_i$. We include the positive instance -- original evidence with distractor sentence appended to it, with the original gold label $y_i$. The negative instances, i.e., with insufficient evidence information, are included with a gold label $y_i=$ NEI for FEVER and VitaminC, and $y_i=$ NOT SUPPORTING for HoVer. Each model, $h(c,e)=\hat{y}$, receives as input the original claim $c$ and the augmented or the original evidence $e$ and predicts the veracity label $\hat{y}$. We optimise a supervised cross-entropy loss as per Equation~\ref{crossentropy}.
\subsection{Baseline Ensemble}
\label{sec:method:ensemble}
We include a simple ensemble, consisting of the three models -- BERT, RoBERTa, and ALBERT. Each ensemble contains only supervised models (\S\ref{sec:manual_annotations}), models trained with CAD (\S\ref{sec:method:cad}), or models trained with CL loss (\S\ref{sec:method:cl}). We employ majority voting, where the final prediction is the most common class among the predictions of the three models on an instance, defaulting to the class with the highest predicted probability if there is no most common class.% (e.g., each model predicts a different class)
%, we take the class with the highest predicted probability.}

\subsection{Experimental Details}
\label{sec:experimental}
All models are trained on the respective training splits of each dataset. We select the checkpoint with the highest macro $F_1$-score on the dev sets and provide results on the test sets. We note that for the newly introduced task \taskname, we have an annotated test dataset \datasetname, but no training dataset. The training is performed on the original training splits of the corresponding datasets, which have a different label distribution from the introduced diagnostic test set. Hence, it is possible that some of the instances in \datasetname\ are out of the original training distribution, which would make this diagnostic dataset of rather adversarial nature.

We select the learning rate = $1e\!-\!5$ and the temperature parameters $\tau\!=\!1.5$ by grid search over the performance on the dev sets from $[1e\!-\!5, 2e\!-\!5, 3e\!-\!5]$ and $[0, 0.5, 1, 1.5, 2]$ respectively. We use the batch sizes for corresponding models from prior work -- 8 for HoVeR, 32 for FEVER, and 16 for VitaminC.

\section{Results and Discussion}
\begin{table*}[t]
\fontsize{10}{9}\selectfont
\centering
\begin{tabular}{llrrrrrrrr}
\toprule
\multirow{2}{*}{\textbf{Dataset}} & \multirow{2}{*}{\textbf{Model}} & \multicolumn{4}{c}{\textbf{Veracity Pred. / Orig.Test}}  & \multicolumn{4}{c}{\textbf{Evidence Sufficiency / Suff.Facts}} \\
 & & \textbf{\scriptsize BERT} & \textbf{\scriptsize RoBERTa} & \textbf{\scriptsize ALBERT} & \textbf{{\scriptsize E}ns.}
 & \textbf{\scriptsize BERT} & \textbf{\scriptsize RoBERTa} & \textbf{\scriptsize ALBERT} & \textbf{{\scriptsize E}ns.} \\ \midrule
% \multicolumn{3}{l}{\textbf{FEVER}} \\
\multirow{3}{*}{FEVER}&Supervised & 87.16 & 88.69 & 86.67 & 88.81 & 59.51 & 59.10 & 63.00 & 61.36 \\
&\hspace{1.5mm} + CL & 87.62 & 88.81 & 86.62 & 89.02 & 65.79 & 67.98 & \underline{\textbf{70.83}} & \textbf{69.90}\\
&\hspace{1.5mm} + CAD & \textbf{87.86} & \underline{\textbf{89.23}} & \textbf{87.31} & \textbf{89.14} & \textbf{67.18} & \textbf{69.58} & 68.56 & 69.25 \\
\midrule
% \multicolumn{3}{l}{\textbf{HoVer}} \\
\multirow{3}{*}{HoVer}&Supervised & 80.75 & 83.37 & 76.88 & 82.73 & 58.15 & 64.81 & 66.28 & 65.88  \\ 
&\hspace{1.5mm} + CL & 81.82 & 83.38 & 77.62 & 83.08 & 74.91 & 75.41 & 72.83 & 78.05 \\
&\hspace{1.5mm} + CAD & \textbf{81.87} & \underline{\textbf{83.65}} & \textbf{79.44} & \textbf{83.65} & \textbf{74.98} & \underline{\textbf{77.14}} & \textbf{76.12} & \textbf{79.07} \\
\midrule
% \multicolumn{3}{l}{\textbf{VitaminC}} \\
\multirow{3}{*}{VitaminC}&Supervised & 82.26 & 84.98 & 83.38 & 86.01 & 58.51 & 69.07 & 66.57 & 66.76 \\
&\hspace{1.5mm} + CL & 83.00 & \underline{\textbf{85.54}} & 83.48 & \textbf{86.22}  & 62.34 & 72.18 & 68.13 & 70.42 \\
&\hspace{1.5mm} + CAD & \textbf{83.56} & 85.65 & \textbf{83.82} & 86.14 & \textbf{72.93} & \underline{\textbf{75.79}} & \textbf{75.13} &  \textbf{78.60} \\
\bottomrule

\end{tabular}
\caption{Macro $F_1$-score test performance of models and an ensemble (Ens.) (\S\ref{sec:method:ensemble}) trained on the supervised training splits of each dataset (Supervised), and in addition with the contrastive objective (+CL) (\S\ref{sec:method:cl}) and the counterfactually augmented data (+CAD) (\S\ref{sec:method:cad}). Results are the average of three different seed runs. The highest results for a test dataset and a model are in bold, and the overall highest result of a model for a test dataset are additionally underlined.}
\label{tab:test}
\end{table*}

\subsection{Supervised Model Performance} We start by discussing the performance of models trained on the supervised splits of the corresponding datasets to predict labels for claims based on the newly created dataset \datasetname\ for \taskname, presented in Table\ \ref{tab:test}. 
% In Table\ \ref{tab:test}, we see that the performance of the supervised models on \datasetname\ is lower that on the original test splits with a difference of up to 30 $F_1$-score for the RoBERTa model on FEVER. 
Recall that the instances in \datasetname\ had correct predictions from all models before the evidence omission was performed (\S \ref{sec:manual_annotations}), i.e., the performance of the models on the instances in \datasetname\ had 100 $F_1$-score before the evidence omission. Hence, the omission of information from the evidence results in a performance decrease from 100 to 58 $F_1$-score (BERT model for the HoVer dataset), i.e. a decrease of up to 42 $F_1$-score. Out of the three FC models, BERT has the lowest performance on \datasetname, whereas ALBERT has the highest. The latter corroborates that ALBERT is a more robust model for fact verification, as explained in more detail in Section\ \ref{sec:manual_annotations}.

Further, we observe the worst performance on \datasetname\ for the HoVer dataset -- down to 58 $F_1$-score, followed by FEVER, and with the best performance on VitaminC. We suggest that the contrastive nature of the instances in VitaminC that contain factual edits of the evidence, changing the support of the evidence for the claim, as described in Section\ \ref{sec:datasets}, can indeed provide a better learning signal for the models about which parts of the evidence are important for verifying the claim.

\subsection{Contrastive Loss and Augmented Model Performance}
Including a CL loss or CAD results in improvements for all models and datasets on \datasetname\, by up to 17.2 $F_1$-score. Note that the proposed technique does not incur additional annotation costs for training data for \taskname. This corroborates that our proposed evidence omission approach combined with tri-training improves the recognition of (in)sufficient evidence. This, in turn, improves the performance on the original test sets by up to 3.6 $F_1$-score. Comparing the CL loss with counterfactually augmented data, we see that CAD improves the model performance in more cases on \datasetname, except for ALBERT for the FEVER dataset. This could be because the augmented data uses raw labels obtained with tri-learning, while the CL loss only drives apart the negative instances from the anchor in the representation space. 

Finally, we compare the performance of CAD and CL loss that rely on the agreement predictions of the supervised models with the simple majority voting ensembles (\S\ref{sec:method:ensemble}). Single models trained with CAD and CL loss still outperform the ensembles of the supervised models. A majority voting classifier from the models trained with CAD and CL loss improves the performance on the original and \datasetname\ sets even further.% compared to the ensemble consisting of the supervised models only.

%\section{Discussion}
%In this section, we further compare the performance of our models to existing baselines and state-of-the-art (SOTA) results on the corresponding dataset. We also explore how our evidence omission method can also be applied to existing text input rationales, particularly in VitaminC, to study the performance of models on inputs with omitted information. 
\subsection{Comparison to Related Work}
\begin{table}[!t]
% \fontsize{10}{9}\selectfont
\centering
\begin{tabular}{llr}
\toprule
\textbf{Dataset} & \textbf{Model} & $\mathbf{F_1}$ \\ \midrule
% \multicolumn{2}{l}{\textbf{FEVER}} \\
\multirow{4}{*}{FEVER} & DA\textit{\footnotesize ~\cite{thorne-etal-2018-fever}} & 83.84 \\
&RoBERTa Supervised & 88.69  \\ 
&\hspace{1.5mm} + CL & 88.68 \\
&\hspace{1.5mm} + Augmented & \textbf{89.23} \\
\midrule
% \multicolumn{2}{l}{\textbf{HoVer}} \\
\multirow{4}{*}{HoVer}&BERT\textit{\footnotesize ~\cite{jiang-etal-2020-hover}} & \textit{81.20} \\
&BERT Supervised & 80.75 \\ 
&\hspace{1.5mm} + CL & 81.82   \\
&\hspace{1.5mm} + Augmented & \textbf{81.87}  \\
\midrule
% \multicolumn{2}{l}{\textbf{VitaminC}} \\
\multirow{4}{*}{VitaminC}&ALBERT\textit{\footnotesize ~\cite{schuster-etal-2021-get}} & 82.76  \\
&ALBERT Supervised & 83.38 \\
&\hspace{1.5mm} + CL & 83.48  \\
&\hspace{1.5mm} + Augmented & \textbf{83.82} \\
\bottomrule
\end{tabular}
\caption{Macro $F_1$-score on the original test set compared to baseline (FEVER) and SOTA (HoVer, VitaminC) oracle results. Highest results for a dataset are in bold.}
\label{tab:sota}
\end{table}

We further compare the performance of our models to existing systems on the used datasets (see Table\ \ref{tab:sota}). %to existing baselines and state-of-the-art (SOTA) results on the corresponding datasets
%Table~\ref{tab:sota} presents a comparison of our model to existing systems on the used datasets. 
Note that we are particularly interested in veracity prediction to study what evidence models consider as sufficient for factuality prediction. Thus, in the base setting, we do not conduct evidence retrieval, as typically performed for the HoVer and FEVER datasets, but train models using gold evidence (oracle). For FEVER, existing systems report results on both tasks, hence we can only compare to the veracity prediction results with oracle evidence available in the FEVER dataset paper with a Decomposable Attention (DA) model~\cite{parikh-etal-2016-decomposable}. For HoVer and VitaminC, the presented results are also from the dataset papers of models trained with oracle evidence. As there are no other reported results on these datasets, they also represent the state-of-the-art for these two datasets. To compare to them, we pick those of our models with the same Transformer architecture as used in the respective dataset papers, and the best-performing model architecture for FEVER. %, as the best model in the FEVER dataset paper is not Transformer-based. 
Note that we use the same training setting as in related work (\S\ref{sec:experimental}) for all models and datasets. We find that our supervised models are close in performance to prior reported results. Furthermore, including counterfactual data augmentation and contrastive learning leads to improvements over prior results for all three datasets, by up to 2.6 $F_1$-score. 

\subsection{Incorrect Evidence}
\label{sec:data:irrelevant}
So far, we studied model performance on instances with omitted information from the gold evidence. We now probe how well the models detect missing information given retrieved incorrect evidence, which does not contain sufficient information. The latter is possible in real-world scenarios. The evidence we feed to the fact checking model depends on the preceding evidence retrieval step, which can retrieve gold evidence with varying performance. While the fact checking model is possibly trained on gold evidence to avoid learning spurious correlations, we want to evaluate its capability to recognise when the retrieval system has discovered incorrect evidence as well. Note that current FC benchmarks do not consider the prediction of a veracity model if the correct evidence is not retrieved. However, in realistic situations, we do not know whether the evidence is correct, and FC models would still provide a veracity for a claim. Hence, we further study the performance of models on incorrect evidence. For each instance in the original test splits, we retrieve incorrect evidence by selecting the closest evidence of another claim in the dataset by word overlap between the claim and the evidence candidates. We then use the retrieved instead of the original evidence. This results in a test set of claims with incorrect evidence of the same size as the original test split.
\begin{table}[t]
% \fontsize{10}{9}\selectfont
\centering
\begin{tabular}{lrrrr}
\toprule
\textbf{Model} & \textbf{\scriptsize BERT} & \textbf{\scriptsize RoBERTa} & \textbf{\scriptsize ALBERT} & \textbf{\scriptsize Ens.}\\ \midrule
\multicolumn{5}{l}{\textbf{FEVER}} \\
Supervised & 82.18 & 81.88 & 85.03 & 84.24 \\ 
\hspace{1.5mm} + CL  & 87.63 & 93.53 & \textbf{95.18} & \textbf{91.60}  \\ 
\hspace{1.5mm} + CAD & \textbf{89.50} & \textbf{94.73} & 90.89 & 90.95 \\
\midrule
\multicolumn{5}{l}{\textbf{HoVer}} \\
Supervised & 97.27 & 78.64 & 97.65 & 88.57 \\ 
\hspace{1.5mm} + CL  & 99.58 & \textbf{99.71} & \textbf{99.45} & \textbf{99.98} \\ 
\hspace{1.5mm} + CAD  & \textbf{99.65} & 98.52 & 99.30 & 99.97\\ 
\midrule
\multicolumn{5}{l}{\textbf{VitaminC}} \\
Supervised  & 69.99 & 80.36 & \textbf{80.69} & 78.33  \\ 
\hspace{1.5mm} + CL  & 75.77 & 79.32 & 78.95 & 78.90 \\ 
\hspace{1.5mm} + CAD & \textbf{80.71} & \textbf{82.69} & 75.69 & \textbf{80.78} \\ 
\bottomrule
\end{tabular}
\caption{Accuracy of models trained on the supervised training splits of each dataset (Supervised), the contrastive objective in addition to training with Supervised (+CL), and the counterfactually augmented data (+CAD). The models are evaluated on the task of Evidence Sufficiency Prediction on datasets with extracted unrelated evidence information (\S\ref{sec:data:irrelevant}).}
\label{tab:unrelated}
\end{table}

Table\ \ref{tab:unrelated} reports results on the test datasets incorrect evidence. As all instances in the dataset have the new gold label of NEI, we report accuracy, which corresponds to the ratio of the instances with a predicted NEI label. We find that the performance of the models is improved by as much as 27 accuracy points after training with CAD or CL, which is another indication for the effectiveness of the proposed training methods. We also find that CAD again brings larger performance gains than CL, except for HoVer, where the two approaches achieve very similar accuracy scores. 

The extended evaluation of incorrect evidence is an important complement to the study of missing evidence. However, the two are not necessarily directly comparable. First, in Table\ \ref{tab:test}, the two test datasets -- the Original Test and SufficientFacts, both have instances with and without sufficient evidence. The extended study on incorrect evidence in this section only has instances that do not have sufficient evidence. This also results in our use of different measures to report results -- accuracy in Table\ \ref{tab:unrelated}, which is the percentage of detected incorrectly retrieved evidence, and macro $F_1$ score in Table\ \ref{tab:test}, which combines the performance on up to three classes in a balanced way.

However, it is worth addressing the high performance of the models on the irrelevant evidence dataset. We employ evidence that has word overlap with the claim, but is not necessarily semantically similar to the claim. If the models were to only rely on features of the claim or on surface word overlap between the claim and the evidence, the models would have low performance on the irrelevant evidence dataset. We train models to avoid such spurious correlations with CAD and CL loss, which make discovering missing evidence information in irrelevant evidence easy, leading to the observed high performance in Table\ \ref{tab:unrelated}.

\subsection{Error Analysis}
Lastly, we conduct an error analysis on the newly introduced \datasetname\ to understand whether known biases in models trained on FC datasets (\S\ref{sec:related}) also affect predictions on \datasetname. 

\textbf{Claim-Only Prediction.} \citet{schuster-etal-2019-towards} found that FC models often learn spurious correlations and can predict the correct label even when no evidence is provided, as they learn only features of the claim. We investigate whether it is also among the reasons for incorrect predictions of the models on the \datasetname\ dataset. We compute the percentage of instances in \datasetname\ where the models do not predict when provided with evidence. We find that for the HoVer dataset, the supervised BERT model does not predict an NEI label for 36\% of the instances in \datasetname\, whereas the respective number for RoBERTa is 23\% and 14\% for ALBERT. This indicates that supervised models trained on HoVer learn claim-only features for some instances. After training the models with CAD (\S\ref{sec:method:cad}) and CL loss (\S\ref{sec:method:cl}), fewer than 1\% of instances from \datasetname\ are predicted as having enough information by each of thee models when given only the claim. This indicates that training with CAD and CL loss decreases the claim-only bias for the HoVer dataset. For FEVER and VitaminC, we find a lower percentage of instances (fewer than 4\%) in the corresponding \datasetname\ splits that the supervised models predict as having enough information when given only the claim. We hypothesise that this is due to the larger amount of training data in both datasets and due to the contrastive nature of VitaminC, which requires the models to learn features from the evidence as well. The percentage is again decreased after training with CAD and CL (fewer than 1\%). Finally, we find that the instances that are still not detected as having insufficient evidence after training with CAD/CL loss are those that the model could have gained world knowledge about during pre-training. One example of such a claim is given in Table\ \ref{tab:examples}, row 3.
\begin{table}[t]
% \fontsize{10}{8}\selectfont
\small
\centering
\begin{tabular}{p{415pt}}
\toprule
\textbf{1.} \textit{Claim:} Unison (Celine Dion album) was originally released by Atlantic Records. \\
\textit{Evidence:} [Unison (Celine Dion album)] The album was originally released on 2 April 1990.\\
\textit{Dataset:} FEVER, \textit{Model:} BERT \textit{Gold:} NEI, \textit{Sup.:} SUPPORTS, \textit{+CAD:} NEI, \textit{+CL:} NEI \\
\midrule
\textbf{2.} \textit{Claim:} Jean-Jacques Dessalines was born on October 2nd, 2017.\\
\textit{Evidence:} [Jean-Jacques Dessalines] He defeated a French army at the Battle of Vertières. \\
\textit{Dataset:} FEVER, \textit{Model:} RoBERTa, \textit{Gold:} NEI, \textit{Sup.:} SUPPORTS, \textit{+CAD:} NEI, \textit{+CL:} SUPPORTS \\
\midrule
\textbf{3.} \textit{Claim:} The Times is a website. \textit{Evidence:} N/A \\
\textit{Dataset:} FEVER, \textit{Model:} RoBERTa, \textit{Gold:} NEI, \textit{Sup.:}REFUTES, \textit{+CAD:} REFUTES, \textit{+CL:} REFUTES \\
\midrule
% \textit{Dataset: HoVer, Model : BERT, supervised: SUPPORTS, +CAD: SUPPORTS, +CL: SUPPORTS, Gold: NEI} \\
% \textit{Claim:} Rudolf Christoph Eucken received the 1908 Nobel Prize for Literature. The author of The Hessian did not. \\
% \textit{Evidence:} [Howard Fast] Howard Melvin Fast was an American novelist and television writer. [Rudolf Christoph Eucken]  Rudolf Christoph Eucken was a German philosopher.
% \midrule
\textbf{4.} \textit{Claim:} The Bragg–Gray cavity theory was developed by Louis Harold Gray, William Lawrence Bragg, and a man %that was 
knighted in the year 1920. \\
\textit{Evidence:} [William Henry Bragg] He was knighted in 1920. \\
\textit{Dataset: HoVer, Model : RoBERTa, Gold: NEI, supervised: SUPPORTS, +CAD: SUPPORTS, +CL: SUPPORTS} \\
\bottomrule
\end{tabular}
\caption{Example model predictions before (Sup.) and after including CAD/CL loss training.}

\label{tab:examples}
\end{table}

\textbf{Claim-Evidence Overlap.} \citet{schuster-etal-2021-get} also find that FC models are biased in predicting the SUPPORT class when the overlap between the claim and the evidence is high. We conjecture that this is another possible reason that the instances in \datasetname\ are hard for the models to distinguish as having missing important evidence information as their evidence still has a high overlap with the claim. To probe this, we compute the average overlap between the claim and the evidence, disregarding stop words, of instances in the \datasetname\ that are predicted as having insufficient information by the supervised models and by the models trained with CAD and CL loss. For FEVER and HoVer, the instances predicted as NEI by the supervised models have low overlap with the claim that increases after training with CAD and CL loss (61\% to 68\% for HoVer and 63\% to 65\% for FEVER). An example instance where the evidence has high overlap with the claim and is predicted as NEI only after training with CAD and CL loss can be found in Table\ \ref{tab:examples}, row 1. The latter is an indication that training with CAD and CL loss also reduces the overlap bias of FC models. We do not observe a change in the overlap ratio for VitaminC, where we assume that training with contrastive instances already prevents learning biases, including the overlap bias.

\textbf{Spurious Patterns.}
Finally, we investigate whether the models learn other spurious patterns that could lead to low results on \datasetname. We already observed that for some instances, the supervised models predict that the evidence is not sufficient after removing irrelevant information (Table\ \ref{tab:dataset}), which is one indication of learned spurious patterns. 
Further, when removing important information, the supervised models still predict the same label for some instances, as they rely on other parts of the input, which might not be important. 
Table~\ref{tab:examples} shows one example where the supervised models did not recognise that the evidence is missing important information (row 1), but after training with CAD or CL loss, it was detected as NEI. However, there are still possible spurious correlations that the models learn even after training with CAD or CL loss, e.g. the example in row 4. Another such example is in row 3, where even after training with CAD and CL loss, the models still find the claim without any provided evidence sufficient for predicting a refuted claim. As this example relies on knowledge of common facts, we assume that the models rely on knowledge obtained during pre-training or fine-tuning instead. 
Finally, we find that CAD can prevent the model from learning spurious correlations more than the CL loss. This leads to more instances having the correct prediction only after training with CAD, as in the example in row 2.

\section{Conclusion}
We propose a new task related to fact checking, namely detecting when evidence with omitted information is (in)sufficient. To this end, we conducted an in-depth empirical analysis with a newly introduced fluency-preserving method for omitting evidence information. We compared what Transformer-based models and humans find to be sufficient information for FC, resulting in a novel dataset, \datasetname. Finally, we showed that the proposed evidence omission method can be used for collecting contrastive examples for CL and CAD, which improved the performance of the studied models on the \taskname\ task and on veracity prediction.

The resulting models could be applied to detect emergent false claims, which gain popularity before any reputable source can refute them, as our proposed models can indicate when the provided input is insufficient for making a decision and whether to provide the user with the veracity prediction. 
% In contrast, in current fact checking benchmarks, the predictions on incorrect evidence are not considered. 
Such models could also be used for detecting knowledge or evidence gaps that need to be filled to refute or support popular claims.
Another possible future research direction would be to build FC models that indicate the particular part of the claim that they are missing supporting evidence for. Moreover, our proposed analysis and methods could be applied to other knowledge-intensive tasks, such as question answering.

\section*{Acknowledgments}

$\begin{array}{l}\includegraphics[width=1cm]{euflag2.png} \end{array}$ The research documented in this paper has received funding from the European Union's Horizon 2020 research and innovation programme under the Marie Sk\l{}odowska-Curie grant agreement No 801199. Isabelle Augenstein's research is further partially funded by a DFF Sapere Aude research leader grant. The authors would like to thank the anonymous reviewers and action editors for their helpful comments and suggestions.

\chapter{Generating Label Cohesive and Well-Formed Adversarial Claims}
\label{chap:adversarial_claims}
\section{Introduction}
\begin{figure}[t]
\centering
\includegraphics[width=0.65 \columnwidth]{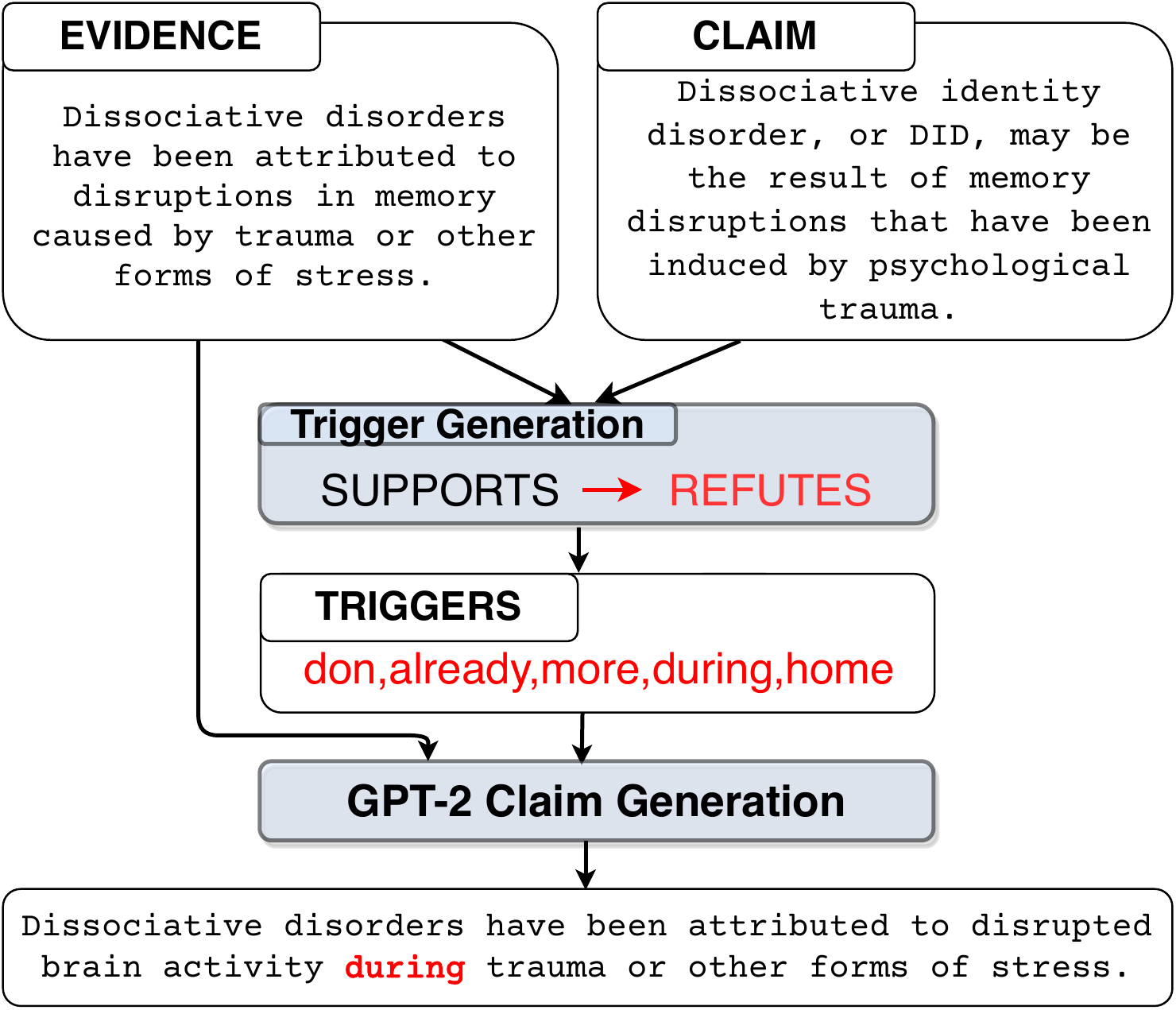}
\caption{High level overview of our method. First, universal triggers are discovered for flipping a source to a target label (e.g. SUPPORTS $\rightarrow$ REFUTES). These triggers are then used to condition the GPT-2 language model to generate novel claims with the original label, including at least one of the found triggers.}
\label{fig:puc}
\end{figure}

Adversarial examples~\cite{https://doi.org/10.48550/arxiv.1412.6572, szegedy2013intriguing} are deceptive model inputs designed to mislead an ML system into making the wrong prediction. They expose regions of the input space that are outside the training data distribution where the model is unstable. 
%Slight, sometimes imperceptible perturbations to a model's input, can change its prediction, which can be a significant security threat ~\cite{biggio2013evasion}. 
It is important to reveal such vulnerabilities and correct for them, especially for tasks such as fact checking (FC). %A shared task~\cite{thorne-etal-2019-fever2} has been proposed to find adversarial attacks that `break' existing fact checking systems as well as to `fix' the models to be more robust to attacks. 

In this paper, we explore the vulnerabilities of FC models trained on the FEVER dataset~\cite{thorne-etal-2018-fever}, where the inference between a claim and evidence text is predicted. We particularly construct \textit{universal adversarial triggers}~\cite{wallace-etal-2019-universal} -- single n-grams appended to the input text that can shift the prediction of a model from a source class to a target one. Such adversarial examples are of particular concern, as they can apply to a large number of input instances. 
% Universal adversarial triggers are generated with the Hotflip algorithm~\cite{ebrahimi2018hotflip} that finds triggers maximizing the loss for the correct prediction by using directional derivatives.

However, we find that the triggers also change the meaning of the claim such that the true label is in fact the target class. For example, when attacking a claim-evidence pair with a `SUPPORTS' label, a common unigram found to be a universal trigger when switching the label to `REFUTES' is `none'. Prepending this token to the claim drastically changes the meaning of the claim such that the new claim is in fact a valid 
`REFUTES' claim as opposed to an adversarial `SUPPORTS' claim. %, which can be due to the natural language inference (NLI) nature of the task. 
%The adversarial examples instead remain to be correct predictions. 
Furthermore, we find adversarial examples constructed in this way to be nonsensical, as a new token is simply being attached to an existing claim. 
% In this work, we suggest to preserve the meaning of the source text by utilizing an auxiliary NLI model. To improve the semantical validity of the adversarial claims, we use the produced triggers and the text to generate claims with semantic validity. 
%

Our \textbf{contributions} are as follows. We \textit{preserve the meaning} of the source text and \textit{improve the semantic validity} of universal adversarial triggers to automatically construct more potent adversarial examples. This is accomplished via: 1) a \textit{novel extension to the HotFlip attack}~\cite{ebrahimi-etal-2018-hotflip}, where we jointly minimize the target class loss of a FC model and the attacked class loss of a natural language inference model; 2) a \textit{conditional language model} trained using GPT-2~\cite{radford2019language}, which takes %a set of 
trigger tokens and a piece of evidence, and generates a semantically coherent new claim containing at least one trigger. %We demonstrate that the 
The resulting triggers maintain potency against a FC model while preserving the original claim label. Moreover, the conditional language model produces semantically coherent adversarial examples containing triggers, on which a FC model performs 23.8\% worse than with the original FEVER claims. The code for the paper is publicly available.\footnote{\url{https://github.com/copenlu/fever-adversarial-attacks}}

%Our contributions are as follows:
%\begin{itemize}[noitemsep]
%\item Preserve the meaning of the source text by jointly minimizing the loss of the target class and the entailment class of an auxiliary NLI model.
%\item Train a conditional Language Model (LM) to generate adversarial examples with improved semantical validity and using the produced universal adversarial triggers.
%\item Provide an analysis of the generated adversarial examples with insights of fact checking model vulnerabilities.
%\end{itemize}

\section{Related Work}
% Initial methods for constructing adversarial are designed for models performing image analysis tasks~\cite{szegedy2013intriguing}. However, most of them are not applicable in the Natural Language Processing (NLP) subfield due to the discrete nature of the input. 
\subsection{Adversarial Examples}
Adversarial examples for %Natural Language Processing (NLP) 
NLP systems can be constructed as automatically generated text~\cite{ren-etal-2019-generating} or perturbations of existing input instances~\cite{jintextfool,ebrahimi-etal-2018-hotflip}. For a %n overview of adversarial attacks in NLP
detailed literature overview, see~\citet{zhang2019adversarial}.

One potent type of adversarial techniques are universal adversarial attacks~\cite{gao2019universal, wallace-etal-2019-universal} -- single perturbation changes that can be applied to a large number of input instances and that cause significant performance decreases of the model under attack. 
~\citet{wallace-etal-2019-universal} find universal adversarial triggers that can change the prediction of the model using the HotFlip algorithm~\cite{ebrahimi-etal-2018-hotflip}. 

However, for NLI tasks, they also change the meaning of the instance they are appended to, and the prediction of the model remains correct. ~\citet{michel-etal-2019-evaluation} %manage the trade-off between decreasing a model's performance and decreasing the semantic similarity between the original and the perturbed instance. They achieve that by exploring only perturbed instances in the neighborhood of the original one.
address this by exploring only perturbed instances in the neighborhood of the original one.
Their approach is for instance-dependent attacks, whereas we suggest finding \textit{universal} adversarial triggers that also preserve the original meaning of input instances. 
Another approach to this 
are rule-based perturbations of the input~\cite{ribeiro-etal-2018-semantically} or imposing adversarial constraints on the produced perturbations~\cite{dia2019semantics}. %using a Gram-Schmidt Basis Sign Method to
By contrast, we extend the HotFlip method by including an auxiliary Semantic Textual Similarity (STS) objective. We additionally use the extracted universal adversarial triggers to generate adversarial examples with low perplexity.

\subsection{Fact Checking}

Fact checking systems consist of components to identify check-worthy claims \cite{atanasova2018overview,hansen2019neural,wright-augenstein-2020-claim}, retrieve and rank evidence documents \cite{yin-roth-2018-twowingos,allein2020timeaware}, determine the relationship between claims and evidence documents \cite{bowman-etal-2015-large,augenstein-etal-2016-stance,baly-etal-2018-integrating}, and finally predict the claims' veracity \cite{thorne-etal-2018-fever,augenstein-etal-2019-multifc}.
As this is a relatively involved task, models easily overfit to shallow textual patterns, necessitating the need for adversarial examples to evaluate the limits of their performance.

%For fact checking,~\citet{thorne2019evaluating} are the first to propose hand-crafted adversarial attacks. %for the FEVER dataset. 
\citet{thorne-etal-2019-evaluating} are the first to propose hand-crafted adversarial attacks. %for the FEVER dataset. 
They follow up on this with the FEVER 2.0 %shared 
task~\cite{thorne-etal-2019-fever2}, where participants design adversarial attacks for existing FC systems. The first two winning systems~\cite{niewinski-etal-2019-gem, hidey-etal-2020-deseption} produce claims requiring multi-hop reasoning, which has been shown to be challenging for fact checking models \cite{ijcai2021-536}. The other remaining system~\cite{kim-allan-2019-fever} generates adversarial attacks manually. We instead find universal adversarial attacks that can be applied to most existing inputs while markedly decreasing fact checking performance.
\citet{niewinski-etal-2019-gem} additionally feed a pre-trained GPT-2 model with the target label of the instance along with the text for conditional adversarial claim generation. Conditional language generation has also been employed by \citet{keskar2019ctrl} to control the style, content, and the task-specific behavior of a Transformer.

% is a GPT-2 language model fine-tuned to generate claims that require multi-hop reasoning. 
% from the text of two hyperlinked Wikipedia pages . 
% In addition, the team manually generated claims with SUPPORTS labels to ensure class balance in their submission
% Team CUNLP \" multi-hop reasoning claims by augmenting existing claims with conjunctions or relative clauses sourced from linked Wikipedia articles. For temporal reasoning adversarial examples they use hand-written rules to manipulate claims containing dates, for example changing “in 2001” to “4 years before 2005” or “between 1999 and 2003”. Finally, they create noisy versions of existing claims by using entities that have a disambiguation page in Wikipedia and by using the lexical substitution method of Alzantot et al. (2018). \\
% \cite{kim-allan-2019-fever} manual rules
% \cite{schuster2019towards} models in fever based solely on the claim perform on par when they are also exposed to the evidence. The authors collect a new dataset, where they re-write claim-evidence pairs to have the same relation, but to state the opposite facts, 20\% decrease in performance. They also provide a list of phrases associated with the different labels. 
\section{Methods}
% \begin{figure*}[t]
%   \centering
%     \includegraphics[width=0.95\textwidth]{images/architecture.pdf}
%     \caption{High level approach used in this paper. First, universal triggers are discovered for flipping a given label to a selected label (e.g. SUPPORTS to REFUTES). These triggers are then used to condition the GPT-2 language model to generate novel claims with the original label including at least one of the found triggers.}
%       \label{fig:puc}
% \end{figure*}

\subsection{Models}
We take a RoBERTa~\cite{liu2019roberta} model pretrained with a LM objective and fine-tune it to classify claim-evidence pairs from the FEVER dataset as SUPPORTS, REFUTES, and NOT ENOUGH INFO (NEI). The evidence used is the gold evidence, available for the SUPPORTS and REFUTES classes. For NEI claims, we use the system of \citet{malon-2018-team} to retrieve evidence sentences. 
To measure the semantic similarity between the claim before and after prepending a trigger, we use a large RoBERTa model fine-tuned on the Semantic Textual Similarity Task.\footnote{\url{https://huggingface.co/SparkBeyond/roberta-large-sts-b}} For further details, we refer the reader to \S\ref{sec:appendixA}.

\subsection{Universal Adversarial Triggers Method}
The Universal Adversarial Triggers method is developed to find n-gram trigger tokens $t_{\alpha}$, which, appended to the original input $x$,  $f(x) = y$, cause the model to predict a target class $\widetilde{y}$ : $f(t_{\alpha}, x) = \widetilde{y}$. In our work, we generate unigram triggers, as generating longer triggers would require additional objectives to later produce well-formed adversarial claims. We start by initializing the triggers with the token `a'. Then, we update the embeddings of the initial trigger tokens $\mathbf{e}_{\alpha}$ with embeddings $\mathbf{e}_{w_i}$ of candidate adversarial trigger tokens $w_i$ that minimize the loss $\mathcal{L}$ for the target class $\widetilde{y}$. Following the HotFlip algorithm, we reduce the brute-force optimization problem using a first-order Taylor approximation around the initial trigger embeddings:
\begin{equation}
\underset{\mathbf{w}_{i} \in \mathcal{V}}{\arg \min }\left[\mathbf{e}_{w_i}-\mathbf{e}_{\alpha}\right]^{\top} \nabla_{\mathbf{e}_{\alpha}} \mathcal{L}
\end{equation}
where $\mathcal{V}$ is the vocabulary of the RoBERTa model and $\nabla_{\mathbf{e}_{\alpha}} \mathcal{L}$ is the average gradient of the task loss accumulated for all batches. This approximation allows for a $\mathcal{O}(|\mathcal{V}|)$ space complexity of the brute-force candidate trigger search.

While HotFlip %algorithm 
finds universal adversarial triggers that successfully fool the model for many instances, we find that the most potent triggers are often negation words, e.g., `not', `neither', `nowhere'. Such triggers change the meaning of the text, making the prediction of the target class correct. Ideally, adversarial triggers would preserve the original label of the claim. To this end, we propose to include an auxiliary STS model objective when searching for candidate triggers. The additional objective is used to minimize the loss $\mathcal{L'}$ for the maximum similarity score (5 out of 0) between the original claim and the claim with the prepended trigger. Thus, we arrive at the combined optimization problem:
\begin{equation}
% \small
\underset{\mathbf{w}_{i} \in \mathcal{V}}{\arg \min }([\mathbf{e}_{w_i}-\mathbf{e}_{\alpha}]^{\top} \nabla_{\mathbf{e}_{\alpha}} \mathcal{L} + [\mathbf{o}_{w_i}-\mathbf{o}_{\alpha}]^{\top} \nabla_{\mathbf{o}_{\alpha}} \mathcal{L'})
\end{equation}
where $\mathbf{o}_w$ is the STS model embedding of word $w$. For the initial trigger token, we use ``[MASK]'' as STS selects candidates from the neighborhood of the initial token.

\subsection{Claim Generation}
\label{sec:claim_generation}
In addition to finding highly potent adversarial triggers, it is also of interest to generate coherent statements containing the triggers. To accomplish this, we use the HuggingFace implementation of the GPT-2 language model~\cite{radford2019language,Wolf2019HuggingFacesTS}, a large transformer-based language model trained on 40GB of text. 
%The model used in this work is the base GPT-2 model from HuggingFace~\cite{Wolf2019HuggingFacesTS}. 
The objective is to generate a coherent claim, which either entails, refutes, or is unrelated a given piece of evidence, while also including trigger words.

The language model is first fine tuned on the FEVER FC corpus with a specific input format. FEVER consists of claims and evidence with the labels \texttt{SUPPORTS}, \texttt{REFUTES}, or \texttt{NOT ENOUGH INFO} (NEI). We first concatenate evidence and claims with a special token. %, so that the model can learn to predict a claim given the evidence. 
Next, to encourage generation of claims with certain tokens, a sequence of tokens separated by commas is prepended to the input. For training, the sequence consists of a single token randomly selected from the original claim, and four random tokens from the vocabulary. %The noise tokens are to %encourage the model to learn to ignore tokens which do not fit in the claim,
%discourage the model from selecting unsuitable tokens,
This encourages the model to only select the one token most likely to form a coherent and correct claim. The final input format is \texttt{[trigger tokens]}\textbar\textbar\texttt{[evidence]}\textbar\textbar\texttt{[claim]}.
Adversarial claims are then generated by providing an initial input of a series of five comma-separated trigger tokens plus evidence, and progressively generating the rest of the sequence. Subsequently, the set of generated claims is pruned to include only those which contain a trigger token, %This is then filtered down to the best claims by first checking if one of the trigger tokens is indeed present in the claim, 
and constitute the desired label. The latter is ensured by passing both evidence and claim through an external NLI model trained on SNLI \cite{bowman-etal-2015-large}. %to ensure that the predicted label is correct.

\section{Results}
We present results for universal adversarial trigger generation and coherent claim generation. 
%Results are measured in terms of the performance of the original fact checking model on claims with added triggers as well as generated claims (macro $F_1$). 
Results are measured using the original FC model on claims with added triggers and generated claims (macro $F_1$). We also measure how well the added triggers maintain the claim's original label (semantic similarity score), the perplexity (PPL) of the claims with prepended triggers, and the semantic quality of generated claims (manual annotation). PPL is measured with a pretrained RoBERTa LM.

\subsection{Adversarial Triggers}

\begin{table}[t]
% \small
\centering
\begin{tabular}{l@{\hspace{1.2\tabcolsep}}l@{\hspace{1.2\tabcolsep}}l@{\hspace{1.2\tabcolsep}}l}
\toprule
\textbf{Class} & \textbf{$\mathbf{F_1}$} & \textbf{STS} & \textbf{PPL}\\ \midrule
\multicolumn{4}{c}{\textbf{No Triggers}} \\
All & .866 & 5.139 & 11.92 ($\pm$45.92) \\
S & .938 & 5.130 & 12.22 ($\pm$40.34) \\
R & .846 & 5.139 &  12.14 ($\pm$37.70) \\
NEI & .817 & 5.147 & 14.29 ($\pm$84.45) \\
\midrule
\multicolumn{4}{c}{\textbf{FC Objective}} \\
All & .602 ($\pm$.289) & 4.586 ($\pm$.328) & 12.96 ($\pm$55.37) \\
S$\rightarrow$R & .060 ($\pm$.034) & 4.270 ($\pm$.295) & 12.44 ($\pm$41.74) \\
S$\rightarrow$NEI & .611 ($\pm$.360) & 4.502 ($\pm$.473) & 12.75 ($\pm$40.50) \\
R$\rightarrow$S & .749 ($\pm$.027) & 4.738 ($\pm$.052) & 11.91 ($\pm$36.53) \\
R$\rightarrow$NEI & .715 ($\pm$.026) & 4.795 ($\pm$.094) & 11.77 ($\pm$36.98) \\
NEI$\rightarrow$R & .685 ($\pm$.030) & 4.378 ($\pm$.232) & 14.20 ($\pm$83.32) \\
NEI$\rightarrow$S & .793 ($\pm$.054) & 4.832 ($\pm$.146) & 14.72 ($\pm$93.15) \\
\midrule
\multicolumn{4}{c}{\textbf{FC+STS Objectives}} \\
All & .763 ($\pm$.123) & 4.786 ($\pm$.156) & 12.97 ($\pm$58.30) \\
S$\rightarrow$R & .702 ($\pm$.237) & 4.629 ($\pm$.186) & 12.62 ($\pm$41.91) \\
S$\rightarrow$NEI & .717 ($\pm$.161) & 4.722 ($\pm$.152) & 12.41 ($\pm$39.66) \\
R$\rightarrow$S & .778 ($\pm$.010) & 4.814 ($\pm$.141) & 11.93 ($\pm$37.04) \\
R$\rightarrow$NEI & .779 ($\pm$.009) & 4.855 ($\pm$.098) & 12.20 ($\pm$37.67) \\
NEI$\rightarrow$R & .780 ($\pm$.078) & 4.894 ($\pm$.115) & 15.27 ($\pm$111.2) \\
NEI$\rightarrow$S & .821 ($\pm$.008) & 4.800 ($\pm$.085) & 13.42 ($\pm$82.30) \\
\bottomrule
\end{tabular}
\caption{Universal Adversarial Trigger method performance. Triggers are generated given claims from a source class to fool the classifier to predict a target class (column \textit{Class}, with SUPPORTS (S), REFUTES (R), NEI). %(we show summaries for the two remaining target classes). 
The results are averaged over the top 10 triggers.}
\label{tab:eval}
\end{table}
Table~\ref{tab:eval} presents the results of applying universal adversarial triggers to claims from the source class.
The top-performing triggers for each direction are found in \S\ref{sec:appendixC}. 
% The first group of results presents the performance of the FC model on the original claim, without any adversarial triggers. 
The adversarial method with a single FC objective successfully deteriorates model performance by a margin of 0.264 $F_1$ score overall. The biggest performance decrease is when the adversarial triggers are constructed to flip the predicted class from SUPPORTS to REFUTES. We also find that 8 out of 18 triggers from the top-3 triggers for each direction, are negation words such as  `nothing', `nobody', `neither', `nowhere' (see Table~\ref{tab:evalonetrig} in the appendix). The first of these triggers decreases the performance of the model to 0.014 in $F_1$. While this is a significant performance drop, these triggers also flip the meaning of the text. The latter is again indicated by the decrease of the semantic similarity between the claim before and after prepending a trigger token, which is the largest for the SUPPORTS to REFUTES direction. We hypothesise that the success of the best performing triggers is partly due to the meaning of the text being flipped.
%Finding triggers with high potency for the other directions is less effective, as it is harder to fool the model that ``refuting'' or ``not sufficient evidence'' claims hold a different stance. 
% When the evidence refutes the claim, we find that the best triggers are ``some'' ($\rightarrow$SUPPORTS) and ``Recommend''($\rightarrow$NEI) decreasing the F1 score to 0.687 and 0.686 respectively. 
% We observe that it is the hardest to find triggers in the NEI$\rightarrow$SUPPORTS direction, where the best performing trigger ``existed'' manages to reduce the F1 score of the fact checking model to 0.800. 
%Finally, we note that the better the performance of the adversarial triggers method is for a particular direction, the smaller the similarity of the adversarial claim and the original claim is. 
%Finally, we note a strong relationship between lower performance and a reduction in similarity score. We hypothesise that the success of the best performing triggers is partially due to the meaning of the text being flipped.

%The second group of adversarial triggers shows that 
Including the auxiliary STS objective increases the similarity between the claim before and after prepending the trigger for five out of six directions. Moreover, we find that now only one out of the 18 top-3 triggers for each direction are negation words. Intuitively, these adversarial triggers are worse at fooling the FC model as they also have to preserve the label of the original claim. Notably, for the SUPPORTS to REFUTES direction the trigger performance is decreased with a margin of 0.642 compared to the single FC objective.
% Notably, the performance of the triggers for the SUPPORTS to REFUTES direction is the worst, while with the single FC objective, this was the most successful direction. 
% The best label preserving trigger here is ``Netflix'', which decreases the F1 score of the model to 0.789 -- a margin of 0.149 F1 score from the performance of the model without the triggers. 
% The easiest direction for flipping the prediction of the FC model in this group of experiments is REFUTES $\rightarrow$ NEI, where using the token ``Trump'',  the performance is decreased to 0.731 F1 score. 
We conclude that including the STS objective for generating Universal Adversarial triggers helps to preserve semantic similarity with the original claim, but also makes it harder to both find triggers preserving the label of the claim while substantially decreasing the performance of the model.

\subsection{Generation}
\begin{table}[t!]
% \fontsize{10}{10}\selectfont
\centering
\begin{tabular}{lccc}
\toprule
\textbf{Target} & \textbf{$\mathbf{F_1}$} & \textbf{Avg Quality} & \textbf{\# Examples}\\ \midrule
%FEVER-DEV & 0.600 & 5 & 603 \\ 
%Ours & 0.644 & 4.84 & 138\\
%\midrule
\multicolumn{4}{c}{\textbf{FC Objective}} \\
Overall& 0.534& 4.33&156\\
SUPPORTS& 0.486& 4.79& 39\\
REFUTES& 0.494& 4.70&32\\
NEI& 0.621& 3.98 &85\\
\midrule
\multicolumn{4}{c}{\textbf{FC+STS Objectives}} \\
Overall& 0.635& 4.63&156\\
SUPPORTS& 0.617& 4.77&67\\
REFUTES& 0.642& 4.68&28\\
NEI& 0.647& 4.44&61\\
\bottomrule
\end{tabular}
\caption{FC performance for generated claims.}
\label{tab:generation_eval}
\end{table}
We use the method described in \S\ref{sec:claim_generation} to generate 156 claims using triggers found with the additional STS objective, and 156 claims without. 52 claims are generated for each class (26 flipping to one class, 26 flipping to the other). A different GPT-2 model is trained to generate claims for each specific class, with triggers specific to attacking that class used as input. The generated claims are annotated manually (see \S\ref{app:B3} for the procedure). The overall average claim quality is 4.48, indicating that most generated statements are highly semantically coherent. The macro $F_1$ of the generative model w.r.t. the intended label is 58.9 overall. For the model without the STS objective, the macro $F_1$ is 56.6, and for the model with the STS objective, it is 60.7, meaning that using triggers found with the STS objective helps the generated claims to retain their intended label.

We measure the performance of the original FC model on generated claims (Table \ref{tab:generation_eval}). We compare between using triggers that are generated with the STS objective (Ex2) and without (Ex1). In both cases, the adversarial claims effectively fool the FC model, which performs 38.4\% worse and 23.8\% worse on Ex1 and Ex2, respectively.  Additionally, the overall sentence quality increases when the triggers are found with the STS objective (Ex2). The FC model's performance is higher on claims using triggers generated with the STS objective but still significantly worse than on the original claims. We provide examples of generated claims with their evidence in Table \ref{tab:generation_examples}.

Comparing FC performance with our generated claims vs. those from the development set of adversarial claims from the FEVER shared task %2.0~\cite{thorne-etal-2019-fever2} 
%versus our high quality generated claims
, we see similar drops in performance (0.600 and 0.644 macro $F_1$, respectively). While the adversarial triggers from FEVER cause a larger performance drop, they were manually selected to meet the label coherence and grammatical correctness requirements. Conversely, we automatically generate claims that meet these requirements.

\begin{table*}[!ht]%[h!]
\fontsize{9}{9}\selectfont
\renewcommand{\arraystretch}{1.4}
\centering
\begin{tabular}{p{6cm} p{3.5cm} p{4.4cm}}
\toprule
\textbf{Evidence} & \textbf{Triggers} & \textbf{Generated Claim}\\ 
\midrule

\multicolumn{3}{c}{\textbf{SUPPORTS Claims}} \\
Since the 19th century, some Romani have also migrated to the Americas. & don, already, more, \textcolor{red}{during}, home & Romani have moved to the Americas \textcolor{red}{during} the 19th century. \\

Cyprus is a major tourist destination in the Mediterranean. & foreign, biggest, \textcolor{red}{major}, every, friends & Cyprus is a \textcolor{red}{major} tourist destination. \\

The first Nobel Prize in Chemistry was awarded in 1901 to Jacobus Henricus van't Hoff, of the Netherlands, ``for his discovery of the laws of chemical dynamics and osmotic pressure in solutions.'' & later, \textcolor{red}{already}, quite, altern, whereas & Henricus Van't Hoff was \textcolor{red}{already} awarded the Nobel Prize. \\
\midrule
\multicolumn{3}{c}{\textbf{REFUTES Claims}} \\

California Attorney General Kamala Harris defeated Sanchez , 61.6\% to 38.4\%. & phys, incarn, \textcolor{red}{not}, occasionally,
something & Kamala Harris did \textcolor{red}{not} defeat Sanchez, 61.6\% to 38.4\%. \\

Uganda is in the African Great Lakes region. & unless, endorsed, picks, pref, \textcolor{red}{against} & Uganda is \textcolor{red}{against} the African Great Lakes region. \\

Times Higher Education World University Rankings is an annual publication of university rankings by Times Higher Education (THE) magazine. & interested, reward, visit, \textcolor{red}{consumer}, conclusion & Times Higher Education World University Rankings is a \textcolor{red}{consumer} magazine. \\

\midrule
\multicolumn{3}{c}{\textbf{NOT ENOUGH INFO Claims}} \\
The KGB was a military service and was governed by army laws and regulations, similar to the Soviet Army or MVD Internal Troops. & nowhere, \textcolor{red}{only}, none, no, nothing & The KGB was \textcolor{red}{only} controlled by a military service. \\

The series revolves around Frank Castle, who uses lethal methods to fight crime as the vigilante ``the Punisher'', with Jon Bernthal reprising the role from Daredevil. & says, said, \textcolor{red}{take}, say, is & \textcolor{red}{Take} Me High is about Frank Castle's use of lethal techniques to fight crime. \\

The Suite Life of Zack \& Cody is an American sitcom created by Danny Kallis and Jim Geoghan. & whilst, interest, applic, \textcolor{red}{someone}, nevertheless & The Suite Life of Zack \& Cody was created by \textcolor{red}{someone} who never had the chance to work in television. \\
\bottomrule
\end{tabular}
\caption{Examples of generated adversarial claims. These are all claims which the FC model incorrectly classified.}
\label{tab:generation_examples}
\end{table*}

\section{Conclusion}
%We present a method for automatically generating highly potent adversarial claims for fact checking with universal adversarial triggers, which creates label coherent claims with an STS model, and enforces well-formed language using GPT-2. 
We present a method for automatically generating highly potent, well-formed, label cohesive claims for FC. 
%While previous work on universal adversarial triggers simply prepends or appends trigger words, we take this a step further by determining how to construct a valid claim containing a trigger word. 
We improve upon previous work on universal adversarial triggers by determining how to construct valid claims containing a trigger word. 
%Additionally, previous work on generating claims for fact checking is generally rule based or requires manual intervention, whereas ours is fully automatic. 
Our method is fully automatic, whereas previous work on generating claims for fact checking is generally rule-based or requires manual intervention. As FC is only one test bed for adversarial attacks, it would be interesting to test this method on other NLP tasks requiring semantic understanding such as question answering %and natural language inference in order 
to better understand shortcomings of models. %trained on such datasets.

\section*{Acknowledgements}
$\begin{array}{l}\includegraphics[width=1cm]{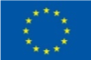} \end{array}$ This project has received funding from the European Union's Horizon 2020 research and innovation programme under the Marie Sk\l{}odowska-Curie grant agreement No 801199. 
\section{Appendices}
\subsection{Implementation Details}
\label{sec:appendixA}
\textbf{Models}. The RoBERTa FC model (125M parameters) is fine-tuned with a batch size of 8, learning rate of 2e-5 and for a total of 4 epochs, where the epoch with the best performance is saved. We used the implementation provided by HuggingFace library. We performed a grid hyper-parameter search for the learning rate between the values 1e-5, 2e-5, and 3e-5. The average time for training a model with one set of hyperparameters is 155 minutes ($\pm3$). The average accuracy over the different hyperparameter runs is 0.862($\pm$ 0.005) $F_1$ score on the validation set.

For the models that measure the perplexity and the semantical similarity we use the pretrained models provided by HuggingFace-- RoBERTa large model (125M parameters) fine tuned on the STS-b task and RoBERTa base model (355M parameters) pretrained on a LM objective.

We used the HuggingFace implementation of the small GPT-2 model, which consists of 124,439,808 parameters and is fine-tuned with a batch size of 4, learning rate of 3e-5, and for a total of 20 epochs. We perform early stopping on the loss of the model on a set of validation data. The average validation loss is 0.910. The average runtime for training one of the models is 31 hours and 28 minutes.

We note that, the intermediate models used in this work and described in this section, are trained on large relatively general-purpose datasets. While, they can make some mistakes, they work well enough and using them, we don't have to rely on additional human annotations for the intermediate task.

\textbf{Adversarial Triggers.} The adversarial triggers are generated based on instances from the validation set. We run the algorithm for three epochs to allow for the adversarial triggers to converge. At each epoch the initial trigger is updated with the best performing trigger for the epoch (according to the loss of the FC or FC+STS objective). At the last step, we select only the top 10 triggers and remove any that have a negative loss. We choose the top 10 triggers as those are the most potent ones, adding more than top ten of the triggers preserves the same tendencies in the results, but smooths them as further down the list of adversarial attacks, the triggers do not decrease the performance of the model substantially. This is also supported by related literature~\cite{wallace-etal-2019-universal}, where only the top few triggers are selected.

The adversarial triggers method is run for 28.75 ($\pm$ 1.47) minutes for the FC objective and 168.6 ($\pm$ 28.44) minutes for the FC+STS objective. We perform the trigger generation with a batch size of four. We additionally normalize the loss for each objective to be in the range [0,1] and also re-weight the losses with a weight of 0.6 for the FC loss and a weight of 0.4 for the SST loss as when generated with an equal weight, the SST loss tends to preserve the same initial token in all epochs.

\textbf{Datasets.} 
The datasets used for training the FC model consist of 161,249 SUPPORTS, 60,227 REFUTES, and 69,885 NEI claims for the training split; 6,207 SUPPORTS, 6,235 REFUTES, and 6,554 NEI claims for the dev set; 6,291 SUPPORTS, 5,992 REFUTES, and 6522 NEI claims. The evidence for each claim is the gold evidence provided from the FEVER dataset, which is available for REFUTES and SUPPORTS claims. When there is more than one annotation of different evidence sentences for an instance, we include them as separate instances in the datasets. For NEI claims, we use the system of \citet{malon-2018-team} to retrieve evidence sentences. 

\subsection{Top Adversarial Triggers}
Table~\ref{tab:evalonetrig} presents the top adversarial triggers for each direction found with the Universal Adversarial Triggers method. It offers an additional way of estimating the effectiveness of the STS objective by comparing the number of negation words generated by the basic model (8) and the STS objective (2) in the top-3 triggers for each direction.
\label{sec:appendixC}
\begin{table*}[t]
% \scriptsize
\fontsize{9}{9}\selectfont
\centering
\begin{tabular}{l@{\hspace{1.2\tabcolsep}}l@{\hspace{1.2\tabcolsep}}l@{\hspace{1.2\tabcolsep}}l@{\hspace{1.2\tabcolsep}}l}
\toprule
\textbf{Class} & \textbf{Trigger} & \textbf{$\mathbf{F_1}$} & \textbf{STS} & \textbf{PPL}\\ \midrule
\multicolumn{5}{c}{\textbf{FC Objective}} \\
S$\rightarrow$R & only &  0.014 &  4.628 &  11.660 (36.191) \\
S$\rightarrow$R & nothing &  0.017 &  4.286 &  13.109 (56.882) \\
S$\rightarrow$R & nobody &  0.036 &  4.167 &  12.784 (37.390) \\
S$\rightarrow$NEI & neither &   0.047 &  3.901 &  11.509 (31.413) \\
S$\rightarrow$NEI & none &  0.071 &  4.016 &  13.136 (39.894) \\
S$\rightarrow$NEI & Neither &  0.155 &  3.641 &  11.957 (44.274) \\
R$\rightarrow$S & some &  0.687 &  4.694 &  11.902 (33.348) \\
R$\rightarrow$S & Sometimes &  0.724 &  4.785 &  10.813 (32.058) \\
R$\rightarrow$S & Some &  0.743 &  4.713 &  11.477 (37.243) \\
R$\rightarrow$NEI & recommended &  0.658 &  4.944 &  12.658 (36.658) \\
R$\rightarrow$NEI & Recommend &  0.686 &  4.789 &  10.854 (32.432) \\
R$\rightarrow$NEI & Supported &  0.710 &  4.739 &  11.972 (40.267) \\
NEI$\rightarrow$R & Only &  0.624 &   4.668 &  12.939 (57.666) \\
NEI$\rightarrow$R & nothing &  0.638 &  4.476 &   11.481 (48.781) \\
NEI$\rightarrow$R & nobody & 0.678 &  4.361 &  16.345 (111.60) \\
NEI$\rightarrow$S & nothing &  0.638 &  4.476 &  18.070 (181.85) \\
NEI$\rightarrow$S & existed &  0.800 &  4.950  &  15.552 (79.823) \\
NEI$\rightarrow$S & area &  0.808 &  4.834  &  13.857 (93.295) \\

\midrule
\multicolumn{5}{c}{\textbf{FC+STS Objectives}} \\
S$\rightarrow$R & never & 0.048 & 4.267 & 12.745 (50.272) \\
S$\rightarrow$R & every & 0.637 & 4.612 & 13.714 (51.244) \\
S$\rightarrow$R & didn & 0.719 & 4.986 & 12.416 (41.080) \\
S$\rightarrow$NEI & always  & 0.299 &  4.774 &  11.906 (35.686) \\
S$\rightarrow$NEI & every & 0.637 & 4.612 & 12.222 (38.440) \\
S$\rightarrow$NEI & investors & 0.696 & 4.920 & 12.920 (42.567) \\
R$\rightarrow$S & over &  0.761 &  4.741 &  12.139 (33.611) \\
R$\rightarrow$S & about &  0.765 &   4.826 &  12.052 (37.677) \\
R$\rightarrow$S & her &   0.774 &   4.513 &   12.624 (41.350) \\
R$\rightarrow$NEI & top &  0.757 &  4.762 &  12.787 (39.418) \\
R$\rightarrow$NEI & also &   0.770 &   5.034 &   11.751 (35.670) \\
R$\rightarrow$NEI & when &   0.776 &   4.843 &   12.444 (37.658) \\
NEI$\rightarrow$R & only &  0.562 &  4.677 &  14.372 (83.059) \\
NEI$\rightarrow$R & there &   0.764 & 4.846 &    11.574 (42.949) \\
NEI$\rightarrow$R & just &   0.786 & 4.916 &   16.879 (135.73) \\
NEI$\rightarrow$S & of&   0.802 & 4.917 &  11.844 (55.871) \\
NEI$\rightarrow$S & is &   0.815 & 4.931 & 17.507 (178.55) \\
NEI$\rightarrow$S & A &   0.818 & 4.897 & 12.526 (67.880) \\

\bottomrule
\end{tabular}
\caption{Top-3 triggers found with the Universal Adversarial Triggers methods. The triggers are generated given claims from a source class (column \textit{Class}), so that the classifier is fooled to predict a different target class. The classes are SUPPORTS (S), REFUTES (R), NOT ENOUGH INFO (NEI).}
\label{tab:evalonetrig}
\end{table*}

\subsection{Computing Infrastructure}
All experiments were run on a shared cluster. Requested jobs consisted of 16GB of RAM and 4 Intel Xeon Silver 4110 CPUs. We used two NVIDIA Titan RTX GPUs with 12GB of RAM for training GPT-2 and one NVIDIA Titan X GPU with 8GB of RAM for training the FC models and finding the universal adversarial triggers.

\subsection{Evaluation Metrics}
The primary evaluation metric used was macro $F_1$ score. We used the sklearn implementation of \texttt{precision\_recall\_fscore\_support}, which can be found here: \url{https://scikit-learn.org/stable/modules/generated/sklearn.metrics}. Briefly:
\begin{equation*}
   p = \frac{tp}{tp + fp} 
\end{equation*}
\begin{equation*}
   r = \frac{tp}{tp + fn} 
\end{equation*}
\begin{equation*}
   F_1 = \frac{2*p*r}{p+r} 
\end{equation*}
where $tp$ are true positives, $fp$ are false positives, and $fn$ are false negatives.

\subsection{Manual Evaluation}
\label{app:B3}
After generating the claims, two independent annotators label the overall claim quality (score of 1-5) and the true label for the claim. The inter-annotator agreement for the quality label using Krippendorff's alpha is 0.54 for the quality score and 0.38 for the claim label. Given this, we take the average of the two annotator's scores for the final quality score and have a third expert annotator examine and select the best label for each contested claim label.

\part{Explainability for Complex Reasoning Tasks over Text}
\chapter{Generating Fact Checking Explanations}
\label{chap:generating_explanations}

\section{Introduction}
\noindent When a potentially viral news item is rapidly or indiscriminately published by a news outlet, the responsibility of verifying the truthfulness of the item is often passed on to the audience. To alleviate this problem, independent teams of professional fact checkers manually verify the veracity and credibility of common or particularly check-worthy statements circulating the web. However, these teams have limited resources to perform manual fact checks, thus creating a need for automating the fact checking process.
\setlength{\tabcolsep}{0.2em}

\begin{table}[t]
% \fontsize{8.4}{8.4}\selectfont
\begin{center}
\begin{tabular}{p{425pt}}
\toprule
\textbf{Claim}: The last major oil spill from a drilling accident in America happened over 40 years ago in 1969.\\ \midrule
\textbf{Ruling Comments}: 
(...) \hlyellow{The last major oil spill from a drilling accident in America happened over 40 years ago in 1969.} \\
% This item will deal with the claim about the date of the last major oil spill from a drilling accident. \\
\hspace*{3mm}(...) The largest in volume was the Santa Barbara spill of 1969 referenced by Murdock and Johnson, in which an estimated 100,000 barrels of oil spilled into the Pacific Ocean, according to the API. \hlblue{The Santa Barbara spill was so big it ranked seventh among the 10 largest oil spills caused by marine well blowouts in the world, the report states.} Two other U.S. spills, both in 1970, rank eighth and 10th. \hlred{Fourteen marine blowouts have taken place in the U.S. between 1969 and 2007.} Six of them took place after 1990 and spilled a total of nearly 13,700 barrels. \\
\hspace*{3mm}(...) We interviewed three scientists who said that the impact of a spill has little to do with its volume. \hlviolet{Scientists have proven that spills far smaller than Santa Barbara's have been devastating.} \\  \midrule
\textbf{Justification}: While the nation's largest oil well blowout did take place in 1969, it's not factually correct to call it the ``last major oil spill". First of all, two of the largest blowouts in the world took place in the U. S.  the following year. More importantly, experts agree that spills far smaller in volume to the 1969 disaster have been devastating. From a scientific perspective, Johnson's decision to single out the 1969 blowout as the last ``major" one makes no sense. \\ \midrule

\textbf{Ruling}: Half-True \\ \bottomrule
\end{tabular}
\end{center}
\caption{\label{tab:Example} Example instance from the LIAR-PLUS dataset, with oracle sentences for generating the justification highlighted.}
\end{table}

The current research landscape in automated fact checking is comprised of systems that estimate the veracity of claims based on available metadata and evidence pages. Datasets like LIAR \cite{wang-2017-liar} and the multi-domain dataset MultiFC \cite{augenstein-etal-2019-multifc} provide real-world benchmarks for evaluation. There are also artificial datasets of a larger scale, e.g., the FEVER \cite{thorne-etal-2018-fever} dataset based on Wikipedia articles. As evident from the effectiveness of state-of-the-art methods for both real-world -- 0.492 macro $F_1$ score \cite{augenstein-etal-2019-multifc}, and artificial data -- 68.46 FEVER score (label accuracy conditioned on evidence provided for `supported' and `refuted' claims) \cite{stammbach-neumann-2019-team},  the task of automating fact checking remains a significant and poignant research challenge.

A prevalent component of existing fact checking systems is a stance detection or textual entailment model that predicts whether a piece of evidence contradicts or supports a claim \cite{Ma:2018:DRS:3184558.3188729, mohtarami-etal-2018-automatic, Xu2019AdversarialDA}. Existing research, however, rarely attempts to directly optimise the selection of relevant evidence, i.e., the self-sufficient explanation for predicting the veracity label \cite{thorne-etal-2018-fever, stammbach-neumann-2019-team}.
On the other hand, \citet{alhindi-etal-2018-evidence} have reported a significant performance improvement of over 10\% macro $F_1$ score when the system is provided with a short human explanation of the veracity label. Still, there are no attempts at automatically producing explanations, and automating the most elaborate part of the process - producing the \emph{justification} for the veracity prediction - is an understudied problem.

In the field of NLP as a whole, both explainability and interpretability methods have gained importance recently, because most state-of-the-art models are large, neural black-box models. Interpretability, on one hand, provides an overview of the inner workings of a trained model such that a user could, in principle, follow the same reasoning to come up with predictions for new instances. However, with the increasing number of neural units in published state-of-the-art models, it becomes infeasible for users to track all decisions being made by the models.
Explainability, on the other hand, deals with providing local explanations about single data points that suggest the most salient areas from the input or are generated textual explanations for a particular prediction.

Saliency explanations have been studied extensively \cite{Adebayo:2018:SCS:3327546.3327621, arras-etal-2019-evaluating, poerner-etal-2018-evaluating}, however, they only uncover regions with high contributions for the final prediction, while the reasoning process still remains behind the scenes. An alternative method explored in this paper is to generate textual explanations. In one of the few prior studies on this, the authors find that feeding generated explanations about multiple choice question answers to the answer predicting system improved QA performance \cite{rajani-etal-2019-explain}.

Inspired by this, we research how to generate explanations for veracity prediction. We frame this as a summarisation task, where, provided with elaborate fact checking reports, later referred to as \textit{ruling comments}, the model has to generate \textit{veracity explanations} close to the human justifications as in the example in Table~\ref{tab:Example}. We then explore the benefits of training a joint model that learns to generate veracity explanations while also predicting the veracity of a claim.\\
In summary, our \textbf{contributions} are as follows:
\begin{enumerate}%[noitemsep]
    \item We present the first study on generating veracity explanations, showing that they can successfully describe the reasons behind a veracity prediction.
    \item We find that the performance of a veracity classification system can leverage information from the elaborate ruling comments, and can be further improved by training veracity prediction and veracity explanation jointly.
    \item We show that optimising the joint objective of veracity prediction and veracity explanation produces explanations that achieve better coverage and overall quality and serve better at explaining the correct veracity label than explanations learned solely to mimic human justifications.
\end{enumerate}

\section{Dataset}

Existing fact checking websites publish claim veracity verdicts along with ruling comments to support the verdicts. Most ruling comments span over long pages and contain redundancies, making them hard to follow. Textual explanations, by contrast, are succinct and provide the main arguments behind the decision. PolitiFact~\footnote{\url{https://www.politifact.com/}} provides a summary of a claim's ruling comments that summarises the whole explanation in just a few sentences. 

We use the PolitiFact-based dataset LIAR-PLUS \cite{alhindi-etal-2018-evidence}, which contains 12,836 statements with their veracity justifications. The justifications are automatically extracted from the long ruling comments, as their location is clearly indicated at the end of the ruling comments. Any sentences with words indicating the label, which \citet{alhindi-etal-2018-evidence} select to be identical or similar to the label, are removed. We follow the same procedure to also extract the ruling comments without the summary at hand.

We remove instances that contain fewer than three sentences in the ruling comments as they indicate short veracity reports, where no summary is present. The final dataset consists of 10,146 training, 1,278 validation, and 1,255 test data points. A claim's ruling comments in the dataset span over 39 sentences or 904 words on average, while the justification fits in four sentences or 89 words on average. 

\section{Method}
We now describe the models we employ for training separately (1) an explanation extraction and (2) veracity prediction, as well as (3) the joint model trained to optimise both.

The models are based on DistilBERT \cite{sanh2019distilbert}, which is a reduced version of BERT \cite{devlin-etal-2019-bert} performing on par with it as reported by the authors. For each of the models described below, we take the version of DistilBERT that is pre-trained with a language-modelling objective and further fine-tune its embeddings for the specific task at hand. 

\subsection{Generating Explanations}\label{sec:explanationGen}

\begin{figure*}[t]
\centering
\includegraphics[width=\linewidth]{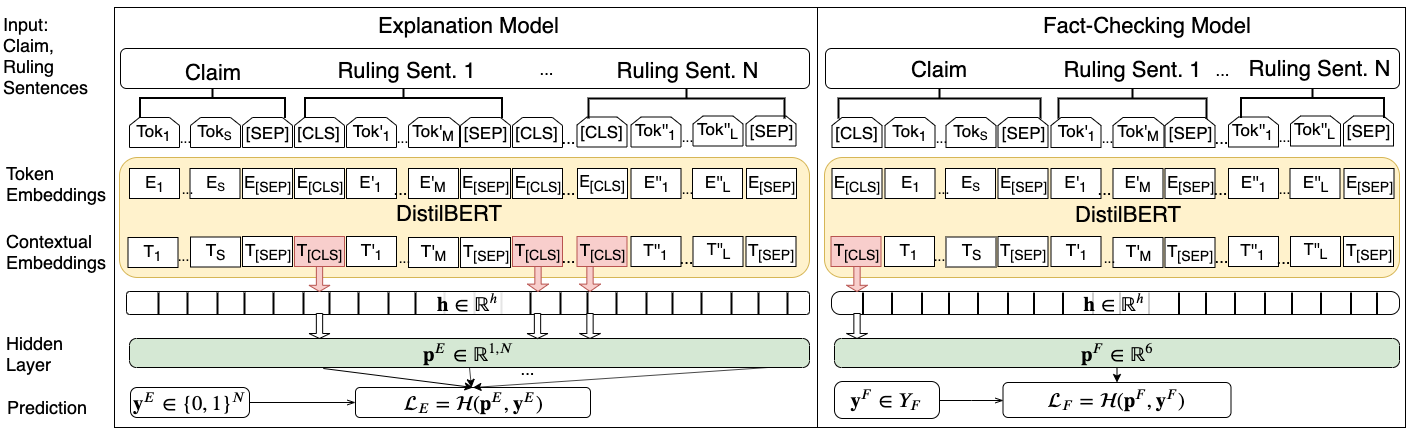}
\caption{Architecture of the \textit{Explanation} (left) and \textit{Fact-Checking} (right) models that optimise separate objectives.} %that optimise their separate objectives.
\label{figure:separateModels}
\end{figure*}

Our explanation model, shown in Figure~\ref{figure:separateModels} (left) is inspired by the recent success of utilising the transformer model architecture for extractive summarisation \cite{liu-lapata-2019-text}. It learns to maximize the similarity of the extracted explanation with the human justification.  

We start by greedily selecting the top $k$ sentences from each claim's ruling comments that achieve the highest ROUGE-2 $F_1$ score when compared to the gold justification. We choose $k = 4$, as that is the average number of sentences in veracity justifications. The selected sentences, referred to as oracles, serve as positive gold labels - $\mathbf{y}^E \in \{0,1\}^N $, where $N$ is the total number of sentences present in the ruling comments. Appendix~\ref{appendix:a} provides an overview of the coverage that the extracted oracles achieve compared to the gold justification. Appendix~\ref{appendix:o} further presents examples of the selected oracles, compared to the gold justification. 

At training time, we learn a function $f(X) = \mathbf{p}^E$, $\mathbf{p}^E \in \mathbb{R}^{1, N}$ that, based on the input $X$, the text of the claim and the ruling comments, predicts which sentence should be selected - \{0,1\}, to constitute the explanation. At inference time, we select the top $n = 4$ sentences with the highest confidence scores.

Our extraction model, represented by function $f(X)$, takes the contextual representations produced by the last layer of DistilBERT and feeds them into a feed-forward task-specific layer - $\mathbf{h} \in \mathbb{R}^{h}$. It is followed by the prediction layer $\mathbf{p}^{E} \in \mathbb{R}^{1,N}$ with sigmoid activation. The prediction is used  to optimise the cross-entropy loss function $\mathcal{L}_{E}=\mathcal{H}(\mathbf{p}^{E}, \mathbf{y}^{E})$.
% The first task-specific layer reduces the dimension of the transformer's output.

\subsection{Veracity Prediction}\label{sec:veracityPred}
For the veracity prediction model, shown in Figure~\ref{figure:separateModels} (right), we learn a function $g(X) = \mathbf{p}^F$ that, based on the input X, predicts the veracity of the claim $\mathbf{y}^{F} \in Y_{F}$, $Y_F =$ \textit{\{true, false, half-true, barely-true, mostly-true, pants-on-fire\}}. 
% We conduct different experiments, where X is the text of the ruling comments, the ruling oracles of the justification.

The function $g(X)$ takes the contextual token representations from the last layer of DistilBERT and feeds them to a task-specific feed-forward layer $\mathbf{h} \in \mathbb{R}^{h}$. It is followed by the prediction layer with a softmax activation $\mathbf{p}^{F} \in \mathbb{R}^{6}$. We use the prediction to optimise a cross-entropy loss function $\mathcal{L}_{F}= \mathcal{H}(\mathbf{p}^{F}, \mathbf{y}^{F})$.
% The input to the transformer model is the `[CLS]' token, the text of the claim and the text of the evidence. From the contextual representations that are produced from DistilBERT, we take only the embedding of the `[CLS]' token and pass it to task-specific layers.

\subsection{Joint Training}\label{sec:jointTraining}
\begin{figure}[t]
\centering
\includegraphics[width=200pt]{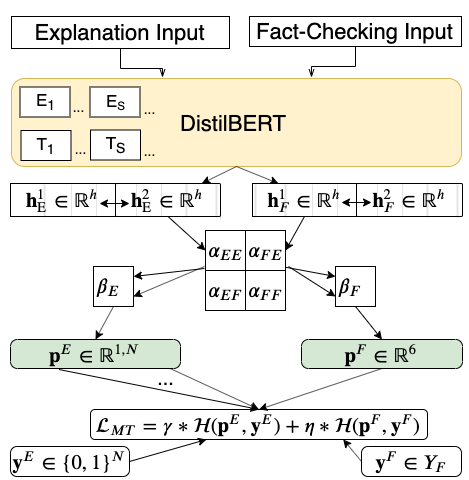}
\caption{Architecture of the \textit{Joint} model learning Explanation (E) and Fact-Checking (F) at the same time.}
\label{figure:jointmodel} 
\end{figure}

Finally, we learn a function $h(X) = (\mathbf{p}^E, \mathbf{p}^F)$ that, given the input X - the text of the claim and the ruling comments, predicts both the veracity explanation $\mathbf{p}^E$ and the veracity label $\mathbf{p}^F$ of a claim. The model is shown Figure~\ref{figure:jointmodel}. The function $h(X)$ takes the contextual embeddings $\mathbf{c}^E$ and $\mathbf{c}^F$ produced by the last layer of DistilBERT and feeds them into a cross-stitch layer \cite{misra2016cross,ruder122019latent}, which consists of two layers with two shared subspaces each - $\mathbf{h}_{E}^1$ and $\mathbf{h}_{E}^2$ for the explanation task and $\mathbf{h}_F^1$ and $\mathbf{h}_F^2$ for the veracity prediction task. In each of the two layers, there is one subspace for task-specific representations and one that learns cross-task representations. The subspaces and layers interact trough $\alpha$ values, creating the linear combinations $\widetilde{h}^i_E$ and $\widetilde{h}^j_F$, where i,j$\in \{1,2\}$:
%
% The two tasks are optimized by minimizing the weighted average of the two task-specific losses.
\begin{equation}
\centering
\begin{bmatrix}
\widetilde{h}^i_E\\ 
\widetilde{h}^j_F
\end{bmatrix}
=
\begin{bmatrix}
\alpha_{EE} & \alpha_{EF}\\ 
\alpha_{FE} & \alpha_{FF}
\end{bmatrix}
\begin{bmatrix}
{h^i_E}^T & {h^j_F}^T\\ 
\end{bmatrix}  
\end{equation}

We further combine the resulting two subspaces for each task - $\widetilde{h}^i_E$ and $\widetilde{h}^j_F$ with parameters $\beta$ to produce one representation per task:
\begin{equation}
\centering
\widetilde{h}^T_P
=
\begin{bmatrix}
\beta_P^1\\ 
\beta_P^2
\end{bmatrix}^T
\begin{bmatrix}
\widetilde{h}^1_P & \widetilde{h}^2_P\\ 
\end{bmatrix}^T
\end{equation}
where P $\in \{E, F\}$ is the corresponding task.

Finally, we use the produced representation to predict $\mathbf{p}^{E}$ and $\mathbf{p}^{F}$, with feed-forward layers followed by sigmoid and softmax activations accordingly. We use the prediction to optimise the joint loss function $\mathcal{L}_{MT}= \gamma*\mathcal{H}(\mathbf{p}^{E}, \mathbf{y}^{E}) + \eta * \mathcal{H}(\mathbf{p}^{F}, \mathbf{y}^{F})$, where $\gamma$ and $\eta$ are used for weighted combination of the individual loss functions.

\section{Automatic Evaluation}\label{sec:automaticEval}
We first conduct an automatic evaluation of both the veracity prediction and veracity explanation models. 

\subsection{Experiments}\label{subsec:automaticExperiments}
In Table~\ref{tab:results:explanation}, we compare the performance of the two proposed models for generating extractive explanations. \textit{Explain-MT} is trained jointly with a veracity prediction model, and \textit{Explain-Extractive} is trained separately. We include the \textit{Lead-4} system \cite{Nallapati:2017:SRN:3298483.3298681} as a baseline, which selects as a summary the first four sentences from the ruling comments. The \textit{Oracle} system presents the best greedy approximation of the justification with sentences extracted from the ruling comments. It indicates the upper bound that could be achieved by extracting sentences from the ruling comments as an explanation. The performance of the models is measured using ROUGE-1, ROUGE-2, and ROUGE-L $F_1$ scores.

In Table~\ref{tab:results:fact-checking}, we again compare two models - one trained jointly - \textit{MT-Veracity@Rul}, with the explanation generation task and one trained separately - \textit{Veracity@Rul}. As a baseline, we report the work of \citet{wang-2017-liar}, who train a model based on the metadata available about the claim. It is the best known model that uses only the information available from the LIAR dataset and not the gold justification, which we aim at generating. 

We also provide two upper bounds serving as an indication of the approximate best performance that can be achieved given the gold justification. The first is the reported system performance from \citet{alhindi-etal-2018-evidence}, and the second - \textit{Veracity@Just}, is our veracity prediction model but trained on gold justifications. The \citet{alhindi-etal-2018-evidence} system is trained using a BiLSTM, while we train the \textit{Veracity@Just} model using the same model architecture as for predicting the veracity from the ruling comments with \textit{Veracity@Rul}. 

Lastly, \textit{Veracity@RulOracles} is the veracity model trained on the gold oracle sentences from the ruling comments. It provides a rough estimate of how much of the important information from the ruling comments is preserved in the oracles. The models are evaluated with a macro $F_1$ score.

\subsection{Experimental Setup}

Our models employ the base, uncased version of the pre-trained DistilBERT model. The models are fed with text depending on the task set-up - claim and ruling sentences for the explanation and joint models; claim and ruling sentences, claim and oracle sentences or claim and justification for the fact-checking model. We insert a `[CLS]' token before the start of each ruling sentence (explanation model), before the claim (fact-checking model), or at the combination of both for the joint model. The text sequence is passed through a number of Transformer layers from DistilBERT. We use the `[CLS]' embeddings from the final contextual layer of DistilBERT and feed that in task-specific feed-forward layers $\mathbf{h} \in \mathbb{R}^{h}$, where h is 100 for the explanation task, 150 for the veracity prediction one and 100 for each of the joint cross-stitch subspaces. Following are the task-specific prediction layers ${p}^E$. 

The size of $h$ is picked with grid-search over \{50, 100, 150, 200, 300\}. We also experimented with replacing the feed-forward task-specific layers with an RNN or Transformer layer or including an activation function, which did not improve task performance.

The models are trained for up to 3 epochs, and, following \citet{liu-lapata-2019-text}, we evaluate the performance of the fine-tuned model on the validation set at every 50 steps, after the first epoch. We then select the model with the best ROUGE-2 $F_1$ score on the validation set, thus, performing a potential early stopping. The learning rate used is 3e-5, which is chosen with a grid search over \{3e-5, 4e-5, 5e-5\}. We perform 175 warm-up steps (5\% of the total number of steps), after also experimenting with 0, 100, and 1000 warm-up steps. Optimisation is performed with AdamW \cite{loshchilov2017fixing}, and the learning rate is scheduled with a warm-up linear schedule \cite{goyal2017accurate}. The batch size during training and evaluation is 8.

The maximum input words to DistilBERT are 512, while the average length of the ruling comments is 904 words. To prevent the loss of any sentences from the ruling comments, we apply a sliding window over the input of the text and then merge the contextual representations of the separate sliding windows, mean averaging the representations in the overlap of the windows. The size of the sliding window is 300, with a stride of 60 tokens, which is the number of overlapping tokens between two successive windows. The maximum length of the encoded sequence is 1200. We find that these hyper-parameters have the best performance after experimenting with different values in a grid search.

We also include a dropout layer (with 0.1 rate for the separate and 0.15 for the joint model) after the contextual embedding provided by the transformer models and after the first linear layer as well.

The models optimise cross-entropy loss, and the joint model optimises a weighted combination of both losses. Weights are selected with a grid search - 0.9 for the task of explanation generation and 0.1 for veracity prediction. The best performance is reached with weights that bring the losses of the individual models to roughly the same scale.

\subsection{Results and Discussion}\label{subsec:automaticResults}

\begin{table}
% \fontsize{10}{10}\selectfont
\centering
\begin{tabular}{lll}
\toprule
\textbf{Model} & \textbf{Val} & \textbf{Test}  \\ 
\midrule
\citet{wang-2017-liar}, all metadata & 0.247 & 0.274 \\
\midrule
Veracity@RulOracles & 0.308 & 0.300 \\ 
Veracity@Rul & 0.313 & 0.313 \\ 
MT-Veracity@Rul & \textbf{0.321} & \textbf{0.323}  \\ 
\midrule
\citet{alhindi-etal-2018-evidence}@Just & 0.37 & 0.37 \\ 
Veracity@Just & \textbf{0.443}& \textbf{0.443} \\ 
\bottomrule
\end{tabular}
\caption{Results (macro $F_1$ scores) of the veracity prediction task on all of the six classes. The models are trained using the text from the ruling oracles (@RulOracles), ruling comment (@Rul), or the gold justification (@Just).}
\label{tab:results:fact-checking}
\end{table}

\begin{table*}[t]
% \fontsize{10}{10}\selectfont
\centering
\begin{tabular}{l@{\hskip 0.05in \vline \hskip 0.05in}ccc@{\hskip 0.05in \vline \hskip 0.05in}ccc}
\toprule
\multirow{2}{*}{\textbf{Model}} & \multicolumn{3}{c@{\hskip 0.05in \vline \hskip 0.05in}}{\textbf{Validation}}&  \multicolumn{3}{c}{\textbf{Test}} \\ %\cmidrule{2-7}
& \textbf{\small ROUGE-1} & \textbf{\small ROUGE-2} & \textbf{\small ROUGE-L} & \textbf{\small ROUGE-1} & \textbf{\small ROUGE-2} & \textbf{\small ROUGE-L} \\ \midrule
% Ruling & 8.65 & 78.65 & 14.84 & 3.53 & 33.76 & 6.16 & 8.10 & 74.14 & 13.92 \\
Lead-4 & 27.92 & 6.94 & 24.26 & 28.11 & 6.96 & 24.38 \\
Oracle & 43.27 & 22.01 & 38.89 & 43.57 & 22.23 & 39.26 \\
\midrule
Explain-Extractive & \textbf{35.64} & \textbf{13.50} & \textbf{31.44} & \textbf{35.70} & \textbf{13.51} & \textbf{31.58} \\
Explain-MT & 35.18 & 12.94 & 30.95 & 35.13 & 12.90 & 30.93 \\
\bottomrule 		
\end{tabular}
\caption{Results of the veracity explanation generation task. The results are ROUGE-N $F_1$ scores of the generated explanation w.r.t. the gold justification.} 
\label{tab:results:explanation}
\end{table*} 

For each claim, our proposed joint model (see \S\ref{sec:jointTraining}) provides both (i) a veracity explanation and (ii) a veracity prediction. We compare our model's performance with models that learn to optimise these objectives \emph{separately}, as no other joint models have been proposed.
Table~\ref{tab:results:fact-checking} shows the results of veracity prediction, measured in terms of macro $F_1$. %The label proportion for each of the labels in the test set is: 0.20-false;0.13-true;0.09- pants-on-fire;0.19-barely-true;0.19-half-true;0.20-mostly-true.
% As a baseline, we use the system of \cite{wang2017liar} that is trained only on metadata. 
% We also include two systems - \cite{alhindi-etal-2018-evidence} and \textit{Fact-check, justification}, that indicate the approximate upper bound of the performance that could be achieved by predicting the veracity of the claim based on the gold justifications. 

Judging from the performance of both \textit{Veracity@Rul} and \textit{MT-Veracity@Rul}, we can assume that the task is very challenging. Even given a gold explanation (\citet{alhindi-etal-2018-evidence} and \textit{Veracity@Just}), the macro $F_1$ remains below 0.5. This can be due to the small size of the dataset and/or the difficulty of the task even for human annotators. We further investigate the difficulty of the task in a human evaluation, presented in Section~\ref{sec:manualEval}. 

% The \textit{Veracity@Rul} model presents the performance of the system when trained on the oracle veracity explanation, while the \textit{Fact-check, ruling comments} is trained with the whole text of the ruling comments. 

Comparing \textit{Veracity@RulOracles} and \textit{Veracity@Rul}, the latter achieves a slightly higher macro $F_1$ score, indicating that the extracted ruling oracles, while approximating the gold justification, omit information that is important for veracity prediction. Finally, when the fact checking system is learned jointly with the veracity explanation system - \textit{MT-Veracity@Rul}, it achieves the best macro $F_1$ score of the three systems. The objective to extract explanations provides information about regions in the ruling comments that are close to the gold explanation, which helps the veracity prediction model to choose the correct piece of evidence.

In Table~\ref{tab:results:explanation}, we present an evaluation of the generated explanations, computing ROUGE $F_1$ score w.r.t. gold justification. 
% The \textit{Oracle} system presents the best greedy approximation of the justification with sentences extracted from ruling comments. It indicates the upper bound that could be achieved by extracting sentences from the ruling comments as an explanation.
% The \textit{Lead-4} system serves as a baseline. 
Our first model, the \textit{Explain-Extractive} system, optimises the single objective of selecting explanation sentences. It outperforms the baseline, indicating that generating veracity explanations is possible.

\textit{Explain-Extractive} also outperforms the \textit{Explain-MT} system. While we would expect that training jointly with a veracity prediction objective would improve the performance of the explanation model, as it does for the veracity prediction model, we observe the opposite. This indicates a potential mismatch between the ruling oracles and the salient regions for the fact checking model. We also find a potential indication of that in the observed performance decrease when the veracity model is trained solely on the ruling oracles compared to the one trained on all of the ruling comments. We hypothesise that, when trained jointly with the veracity extraction component, the explanation model starts to also take into account the actual knowledge needed to perform the fact check, which might not match the exact wording present in the oracles, thus decreasing the overall performance of the explanation system. We further investigate this in a manual evaluation of which of the systems - Explain-MT and Explain-Extractive, generates explanations with better qualities and with more information about the veracity label.

Finally, comparing the performance of the extractive models and the \textit{Oracle}, we can conclude that there is still room for improvement of explanation systems when only considering extractive summarisation.

\subsection{A Case Study}\label{section:case}
Table~\ref{tab:example} presents two example explanations generated by the extractive vs. the multi-task model.
In the first example, the multi-task explanation achieves higher ROUGE scores than the extractive one. The corresponding extractive summary contains information that is not important for the final veracity label, which also appears to affect the ROUGE scores of the explanation. On the other hand, the multi-task model, trained jointly with a veracity prediction component, selects sentences that are more important for the fact check, which in this case is also beneficial for the final ROUGE score of the explanation.

In the second example, the multi-task explanation has lower ROUGE scores than the extractive one. We observe that the gold justification contains some sentences that are not relevant to the fact check, and the extractive summary is fooled to select explanation sentences that are close to the gold summary. As a result, the explanation does not provide enough information about the chosen veracity label. The multi-task model, on the other hand, selects sentences that are also contributing to the prediction of the veracity labels. Thus, its explanation turns out to be more beneficial for the final fact check even though it has a lower ROUGE score compared to the gold justification.

% \begin{landscape}
\begin{table*}
\centering
% \scriptsize
\begin{tabular}{p{0.98\textwidth}}
\toprule

\textbf{Label: }Mostly-true \textbf{Claim:} Says opponent U.S. Rep. Gwen Moore is Wisconsin's most absent member of Congress, missing nearly 17 percent of the House votes in the second quarter of 2012. \\
\textbf{Just:} Sebring says Moore is the state's "most absent" member of Congress and that she missed 17 percent of the House votes in the second quarter of 2012. Indeed, Moore missed nearly 17 percent of the votes in the second quarter. That is Moore's worst quarter, but it also amounts to Sebring doing some cherry-picking of data. A longer period shows a lower percentage, but even that places Moore at the top of the state's eight-member delegation. \\
\textbf{Explain-Extr:} Then he turns to the ad's main point: "Gwen Moore is Wisconsin's most absent member of Congress, missing nearly 17 percent of the House votes in the second quarter of 2012." That, again, compares to the median of 2.5 percent, according to GovTrack. Overall, the state delegation missed an average of 2.53 percent of the votes this session. But what about the other part of Sebring's claim, that Moore is the "most absent" member of the state delegation? \\
\textbf{Explain-MT:} Then he turns to the ad's main point: "Gwen Moore is Wisconsin's most absent member of Congress, missing nearly 17 percent of the House votes in the second quarter of 2012." If you take out the first two quarters of 2012, Moore missed an average of 3.4 percent of the votes in the other five quarters of this session. In the second quarter, the one cited by Sebring, the missed votes easily put Moore at the top of the state's eight-member House delegation. Looking at the full session, Moore missed 112 votes, or 6.98 percent. \\
\midrule
\textbf{Label:} Half-true \textbf{Claim:} Of the more than 1.3 million temporary mortgage modifications, over half have defaulted. \\
\textbf{Just:} In the final full week of the U.S. Senate race, how did Rubio fare on his numbers about the mortgage modification program? Rubio said "over 1.3 million temporary work-outs, over half have now defaulted,'' referring to a temporary mortgage modification program. \\
\textbf{Explain-Extr:} Over 1.3 million temporary work-outs, over half have now defaulted,'' he said. There have been 500,000 permanent (modifications). Rubio also said that more than half of those 1.3 million had defaulted." Rubio: "The temporary modifications." \\
\textbf{Explain-MT:} Rubio also said that more than half of those 1.3 million had "defaulted." "Over 1.3 million temporary work-outs, over half have now defaulted,'' he said. Of those permanent modifications, the majority survived while almost 29,000 were cancelled. The chart states that 699,924 trials had been cancelled -- so note that is slightly more than half.\\
\bottomrule
\end{tabular}
\caption{Examples of the generated explanation of the extractive (Explain-Extr) and the multi-task model (Explain-MT) compared to the gold justification (Just).}
\label{tab:example}
\end{table*}

\section{Manual Evaluation}\label{sec:manualEval}
As the ROUGE score only accounts for word-level similarity between gold and predicted justifications, we also conduct a manual evaluation of the quality of the produced veracity explanations.

\subsection{Experiments}\label{subsec:manualExperiments}
\textbf{Explanation Quality}. We first provide a manual evaluation of the properties of three different types of explanations - gold justification, veracity explanation generated by the  \textit{Explain-MT}, and the ones generated by \textit{Explain-Extractive}. We ask three annotators to rank these explanations with the ranks 1, 2, 3, (first, second, and third place) according to four different criteria:

\begin{enumerate}%[noitemsep]
    \item \textbf{Coverage.} The explanation contains important, salient information and does not miss any important points that contribute to the fact check.
    \item \textbf{Non-redundancy.} The summary does not contain any information that is redundant/repeated/not relevant to the claim and the fact check.
    \item \textbf{Non-contradiction.} The summary does not contain any pieces of information that are contradictory to the claim and the fact check. 
    \item \textbf{Overall.} Rank the explanations by their overall quality.
\end{enumerate}

We also allow ties, meaning that two veracity explanations can receive the same rank if they appear the same. %according to a criterion. 

For the annotation task set-up, we randomly select a small set of 40 instances from the test set and collect the three different veracity explanations for each of them. We did not provide the participants with information of the three different explanations and shuffled them randomly to prevent easily creating a position bias for the explanations. The annotators worked separately without discussing any details about the annotation task.

\textbf{Explanation Informativeness}. In the second manual evaluation task, we study how well the veracity explanations manage to address the information need of the user and if they sufficiently describe the veracity label. We, therefore, design the annotation task asking annotators to provide a veracity label for a claim based on a veracity explanation coming from the justification, the \textit{Explain-MT}, or the \textit{Explain-Extractive} system. The annotators have to provide a veracity label on two levels - binary classification - true or false, and six-class classification - true, false, half-true, barely-true, mostly-true, pants-on-fire. Each of them has to provide the label for 80 explanations, and there are two annotators per explanation. 

\subsection{Results and Discussion}\label{sec:manualResults}

\begin{table}[t]
% \fontsize{10}{10}\selectfont
\centering
\begin{tabular}{lccc}
\toprule
\textbf{Annotators} & \textbf{Just} & \textbf{Explain-Extr} & \textbf{Explain-MT}\\ \midrule
\multicolumn{4}{c}{Coverage} \\ \midrule
All & \textbf{1.48} & 1.89 & \cellcolor{myblue}1.68 \\
1st & \textbf{1.50} & 2.08 & \cellcolor{myblue}1.87 \\
2nd & \textbf{1.74} & 2.16 & \cellcolor{myblue}1.84 \\
3rd & \textbf{1.21} & 1.42 & \cellcolor{myblue}1.34 \\ \midrule
\multicolumn{4}{c}{Non-redundancy} \\ \midrule
All & \textbf{1.48} & \cellcolor{myblue}1.75 & 1.79 \\
1st & \textbf{1.34} & 1.84 & \cellcolor{myblue}1.76 \\
2nd & \textbf{1.71} & \cellcolor{myblue}1.97 & 2.08 \\
3rd & \textbf{1.40} & \cellcolor{myblue}1.42 & 1.53 \\ \midrule
\multicolumn{4}{c}{Non-contradiction} \\ \midrule
All & 1.45 & \cellcolor{myblue}\textbf{1.40} & 1.48 \\
1st & \textbf{1.13} & 1.45 & \cellcolor{myblue}1.34 \\
2nd & 2.18  & \cellcolor{myblue}\textbf{1.63} & 1.92 \\
3rd & \textbf{1.03} & \cellcolor{myblue}1.13  & 1.18 \\ \midrule
\multicolumn{4}{c}{Overall} \\ \midrule
All & \textbf{1.58} & 2.03 & \cellcolor{myblue}1.90 \\
1st & \textbf{1.58} & 2.18 & \cellcolor{myblue}1.95 \\
2nd & \textbf{1.74} & 2.13 & \cellcolor{myblue}1.92 \\
3rd & \textbf{1.42} & \cellcolor{myblue}1.76  & 1.82 \\
\bottomrule 		
\end{tabular}

\caption{Mean Average Ranks (MAR) of the explanations for each of the four evaluation criteria. The explanations come from the gold justification (Just), the generated explanation (Explain-Extr), and the explanation learned jointly (Explain-MT) with the veracity prediction model. The lower MAR indicates a higher ranking, i.e., a better quality of an explanation. For each row, the best results are in bold, and the best results with automatically generated explanations are in blue.} 
\label{tab:results:man1}
\end{table} 

\textbf{Explanation Quality}. Table~\ref{tab:results:man1} presents the results from the manual evaluation in the first set-up, described in Section~\ref{sec:manualEval}, where annotators ranked the explanations according to four different criteria. 

We compute Krippendorff's $\alpha$ inter-annotator agreement (IAA, \citet{hayes2007answering}) as it is suited for ordinal values. The corresponding alpha values are 0.26 for \textit{Coverage}, 0.18 for \textit{Non-redundancy}, -0.1 for \textit{Non-contradiction}, and 0.32 for \textit{Overall}, where $0.67<\alpha <0.8$ is regarded as significant, but vary a lot for different domains. 

We assume that the low IAA can be attributed to the fact that in ranking/comparison tasks for manual evaluation, the agreement between annotators might be affected by small differences in one rank position in one of the annotators as well as by the annotator bias towards ranking explanations as ties. Taking this into account, we choose to present the mean average recall for each of the annotators instead. Still, we find that their preferences are not in a perfect agreement and report only what the majority agrees upon. We also consider that the low IAA reveals that the task might be ``already too difficult for humans''. This insight proves to be important on its own as existing machine summarisation/question answering studies involving human evaluation do not report IAA scores \cite{liu-lapata-2019-text}, thus, leaving essential details about the nature of the evaluation tasks ambiguous. 

% might be due to the fact that some of the annotators ranked more explanations as ties, which creates a mismatch between the ranking positions of some explanations. 
% We, therefore, compute the mean average recall and also present the individual average ranking of the separate systems.
% Looking at the individual annotations and the mean over them, we can tell that there is a full majority vote for the Coverage criterion  

% We also didn't find any manual evaluation of summary or explanation generations that provides inter-annotator agreement scores that we could compare to. 
We find that the gold explanation is ranked the best for all criteria except for \textit{Non-contradiction}, where one of the annotators found that it contained more contradictory information than the automatically generated explanations, but Krippendorff's $\alpha$ indicates that there is no agreement between the annotations for this criterion. 

Out of the two extractive explanation systems, \textit{Explain-MT} ranks best in Coverage and Overall criteria, with 0.21 and 0.13 corresponding improvements in the ranking position. These results contradict the automatic evaluation in Section~\ref{subsec:automaticResults}, where the explanation of \textit{Explain-MT} had lower ROUGE $F_1$ scores. This indicates that an automatic evaluation might be insufficient in estimating the information conveyed by the particular explanation.

On the other hand, \textit{Explain-Extr} is ranked higher than \textit{Explain-MT} in terms of Non-redundancy and Non-contradiction, where the last criterion was disagreed upon, and the rank improvement for the first one is only marginal at 0.04. %The small difference in the mean average rank 

This implies that a veracity prediction objective is not necessary to produce natural-sounding explanations (\textit{Explain-Extr}), but that the latter is useful for generating better explanations overall and with higher coverage (\textit{Explain-MT}). %the explanations are not substantially different. 

\textbf{Explanation Informativeness}.
Table~\ref{tab:results:man2} presents the results from the second manual evaluation task, where annotators provided the veracity of a claim based on an explanation from one of the systems. We here show the results for binary labels, as annotators struggled to distinguish between 6 labels. %the 6-class variant had lower rates of correct labels due to the fine-grained labels. 
The latter follows the same trends and are shown in Appendix~\ref{appendix:q}. 

\begin{table}[t]
% \fontsize{10}{10}\selectfont
\centering
\begin{tabular}{clrrr}
\toprule
 & & \textbf{Just} & \textbf{Explain-Extr} & \textbf{Explain-MT} \\ \midrule
$\nwarrow$ & Agree-C & \textbf{0.403} & 0.237 & \cellcolor{myblue}0.300 \\
$\searrow$ & Agree-NS & \textbf{0.065} &  0.250 & \cellcolor{myblue}0.188 \\
$\searrow$ & Agree-NC & \textbf{0.064} & 0.113 & \cellcolor{myblue}0.088 \\
$\searrow$ & Disagree & 0.468 & \cellcolor{myblue}\textbf{0.400} & 0.425\\
\bottomrule 		
\end{tabular}
\caption{Manual veracity labelling, given a particular explanation from the gold justification (Just), the generated explanation (Explain-Extr), and the explanation learned jointly (Explain-MT) with the veracity prediction model. Percentages of the dis/agreeing annotator predictions are shown, with agreement percentages split into: \emph{correct} according to the gold label (Agree-C), \emph{incorrect} (Agree-NC) or \emph{insufficient information} (Agree-NS). The first column indicates whether higher ($\nwarrow$) or lower ($\searrow$) values are better. For each row, the best results are in bold, and the best results with automatically generated explanations are in blue.}
\label{tab:results:man2}
\end{table} 

The Fleiss' $\kappa$ IAA for binary prediction is: \textit{Just} -- 0.269,  \textit{Explain-MT} -- 0.345, \textit{Explain-Extr} -- 0.399. The highest agreement is achieved for \textit{Explain-Extr}, which is supported by the highest proportion of agreeing annotations from Table~\ref{tab:results:man2}. Surprisingly, the gold explanations from \textit{Just} were most disagreed upon. 
% which are the gold standard explanations. 
Apart from that, looking at the agreeing annotations, gold explanations were found most sufficient in providing information about the veracity label and also were found to explain the correct label most of the time. They are followed by the explanations produced by \textit{Explain-MT}. This supports the findings of the first manual evaluation, where the \textit{Explain-MT} ranked better in coverage and overall quality than \textit{Explain-Extr}.

\section{Related Work}
 
\textbf{Generating Explanations.}
Generating textual explanations for model predictions is an understudied problem. The first study was \citet{NIPS2018_8163}, who generate explanations for the task of natural language inference. The authors explore three different set-ups: prediction pipelines with explanation followed by prediction, and prediction followed by explanation, and a joint multi-task learning setting. They find that first generating the explanation produces better results for the explanation task, but harms classification accuracy. % as the model learns to predict based on the correct reasons, not spurious signals. \citet{rajani-etal-2019-explain} provide another study of generating explanations, this time for multiple-choice question answering. They find that generating explanations for the prediction model results in an increase of nearly 10\% in accuracy. 

We are the first to provide a study on generating veracity explanations. We show that the generated explanations improve veracity prediction performance, and find that jointly optimising the veracity explanation and veracity prediction objectives improves the coverage and the overall quality of the explanations.

\textbf{Fact Checking Interpretability.} Interpreting fact checking systems has been explored in a few studies. \citet{shu2019defend} study the interpretability of a system that fact checks full-length news pages by leveraging user comments from social platforms. They propose a co-attention framework, which selects both salient user comments and salient sentences from news articles. \citet{yang2019xfake} build an interpretable fact-checking system XFake, where shallow student and self-attention, among others, are used to highlight parts of the input. This is done solely based on the statement without considering any supporting facts. 
In our work, we research models that generate human-readable explanations, and directly optimise the quality of the produced explanations instead of using attention weights as a proxy. We use the LIAR dataset to train such models, which contains fact checked single-sentence claims that already contain professional justifications.
As a result, we make an initial step towards automating the generation of professional fact checking justifications.
 
% including linguistic features, attributes of a statement, and a neural network based on the text of the statement. 
% The authors do not, however, provide any information about the performance of the explanation model. 
\textbf{Veracity Prediction.}
Several studies have built fact checking systems for the LIAR dataset \cite{wang-2017-liar}. The model proposed by \citet{karimi-etal-2018-multi} reaches 0.39 accuracy by using metadata, ruling comments, and justifications. \citet{alhindi-etal-2018-evidence} also train a classifier, that, based on the statement and the justification, achieves 0.37 accuracy. To the best of our knowledge, \citet{long2017fake} is the only system that, without using justifications, achieves a performance above the baseline of \citet{wang-2017-liar}, an accuracy of 0.415---the current state-of-the-art performance on the LIAR dataset. Their model learns a veracity classifier with speaker profiles. 
While using metadata and external speaker profiles might provide substantial information for fact checking, they also have the potential to introduce biases towards a certain party or a speaker. 

In this study, we propose a method to generate veracity explanations that would explain the reasons behind a certain veracity label independently of the speaker profile. 
Once trained, such methods could then be applied to other fact checking instances without human-provided explanations or even to perform end-to-end veracity prediction and veracity explanation generation given a claim.
%The justifications in the LIAR dataset, on the other hand, should not be regarded as something that's always at our disposal as they are generated by human annotators. We can, however, learn to generate them, which is the goal of this work. 

Substantial research on fact checking methods exists for the FEVER dataset~\cite{thorne-etal-2018-fever}, which comprises rewritten claims from Wikipedia.
%where only 16.82\% of the cases require more than one evidence sentence for veracity classification. 
Systems typically perform document retrieval, evidence selection, and veracity prediction. Evidence selection is performed using keyword matching \cite{malon-2018-team,yoneda-etal-2018-ucl}, supervised learning \cite{hanselowski-etal-2018-ukp, chakrabarty-etal-2018-robust} or sentence similarity scoring \cite{Ma:2018:DRS:3184558.3188729, mohtarami-etal-2018-automatic, Xu2019AdversarialDA}. More recently, the multi-domain dataset MultiFC \cite{augenstein-etal-2019-multifc} has been proposed, which is also distributed with evidence pages. Unlike FEVER, it contains real-world claims, crawled from different fact checking portals.

While FEVER and MultiFC are larger datasets for fact checking than LIAR-PLUS, they do not contain veracity explanations and can thus not easily be used to train joint veracity prediction and explanation generation models, hence we did not use them in this study.
%In our study, we are interested in real-world claims that require up to four sentences for fact-checking and optimize the explanation of the veracity prediction to resemble the gold human explanation and do that jointly while predicting the veracity label.

\section{Conclusions}

We presented the first study on generating veracity explanations, and we showed that veracity prediction can be combined with veracity explanation generation and that the multi-task set-up improves the performance of the veracity system. A manual evaluation shows that the coverage and the overall quality of the explanation system is also improved in the multi-task set-up.

For future work, an obvious next step is to investigate the possibility of generating veracity explanations from evidence pages crawled from the Web. Furthermore, other approaches of generating veracity explanations should be investigated, especially as they could improve fluency or decrease the redundancy of the generated text.

\section*{Acknowledgments}
% \begin{wrapfigure}{L}{0.10\columnwidth}
% \vspace{-13pt}
% \includegraphics[width=0.17\columnwidth]{euflag2.png}
% \vspace{-25pt}
% \end{wrapfigure}
$\begin{array}{l}\includegraphics[width=1cm]{euflag2.png} \end{array}$ This project has received funding from the European Union’s Horizon 2020 research and innovation programme under the Marie Skłodowska-Curie grant agreement No 801199.

% \clearpage
% \appendix
\section{Appendices}\label{appendices}

\subsection{Comparison of different sources of evidence}\label{appendix:a}
Table \ref{tab:evidence} provides an overview of the ruling comments and the ruling oracles compared to the justification. The high recall in both ROUGE-1 and ROUGE-F achieved by the ruling comments indicates that there is a substantial coverage, i.e. over 70\% of the words and long sequences in the justification can be found in the ruling comments. On the other hand, there is a small coverage for the bi-grams. Selecting the oracles from all of the ruling sentences increases ROUGE-$F_1$ scores mainly by improving the precision. 

\begin{table*}[t] 
\centering
\begin{tabular}{l@{\hskip 0.05in \vline \hskip 0.05in}rrr@{\hskip 0.05in \vline \hskip 0.05in}rrr@{\hskip 0.05in \vline \hskip 0.05in}rrr}
\toprule
\multirow{2}{*}{\textbf{Evidence Source}}& \multicolumn{3}{c@{\hskip 0.05in \vline \hskip 0.05in}}{\textbf{ROUGE-1}}& \multicolumn{3}{c@{\hskip 0.05in \vline \hskip 0.05in}}{\textbf{ROUGE-2}} & \multicolumn{3}{c}{\textbf{ROUGE-L}} \\ %\cmidrule{2-10}
& P & R & $F_1$ & P & R & $F_1$ & P & R & $F_1$ \\ \midrule
Ruling & 8.65 & 78.65 & 14.84 & 3.53 & 33.76 & 6.16 & 8.10 & 74.14 & 13.92 \\
Ruling Oracle & 43.97 & 49.24 & 43.79 & 22.45 & 24.50 & 22.03 & 39.70 & 44.10 & 39.37 \\ \bottomrule
% Evidence Pages & 4.75 & 59.80 & 6.35 & 0.84 & 21.22 & 1.38 & 4.32 & 57.04 & 5.83 \\
% Evidence Oracles & 27.74 & 32.26 & 27.85 & 11.13 & 12.57 & 11.07 & 24.66 & 28.49 & 24.67 \\ \hline
%  Google Pages & 1.24 & 88.97 & 2.27 & 0.49 & 44.12 & 0.92 & 1.18 & 86.94 & 2.17 \\
% Google Oracles & 39.60 & 47.28 & 40.84 & 22.09 & 25.66 & 22.52 & 36.23 & 42.99 & 37.26 \\
% \hline
\end{tabular}
\caption{Comparison of sources of evidence - Ruling Comments and Ruling Oracles comapred to the target justification summary.}
\label{tab:evidence}
\end{table*}

\subsection{Extractive Gold Oracle Examples}
\label{appendix:o}
Table~\ref{tab:oracle-example} presents examples of selected oracles that serve as gold labels during training the extractive summarization model. The three examples represent oracles with different degrees of matching the gold summary. The first row presents an oracle that matches the gold summary with a ROUGE-L $F_1$ score of 60.40 compared to the gold summary. It contains all of the important information from the gold summary and even points precise, not rounded, numbers. The next example has a ROUGE-L $F_1$ score of 43.33, which is close to the average ROUGE-L $F_1$ score for the oracles. The oracle again conveys the main points from the gold justification, thus, being sufficient for the claim's explanation. Finally, the third example is of an oracle with a ROUGE-L $F_1$ score of 25.59. The selected oracle sentences still succeed in presenting the main points from the gold justification, which is at a more detailed level presenting specific findings. The latter might be found as a positive consequence as it presents the particular findings of the journalist that led to selecting the veracity label.

\begin{table*}
\centering
% \scriptsize
\begin{tabular}{p{0.98\textwidth}}
\toprule
\textbf{Claim: }``The president promised that if he spent money on a stimulus program that unemployment would go to 5.7 percent or 6 percent. Those were his words.'' \\
\textbf{Label: }Mostly-False  \\ 
\textbf{Just:} Bramnick said ``the president promised that if he spent money on a stimulus program that unemployment would go to 5.7 percent or 6 percent.
Those were his words.''
Two economic advisers estimated in a 2009 report that with the stimulus plan, the unemployment rate would peak near 8 percent before dropping to less than 6 percent by now.
Those are critical details Bramnick’s statement ignores.
To comment on this ruling, go to NJ.com. \\
\textbf{Oracle:}  ``The president promised that if he spent money on a stimulus program that unemployment would go to 5.7 percent or 6 percent.
Those were his words,'' Bramnick said in a Sept. 7 interview on NJToday.
But with the stimulus plan, the report projected the nation’s jobless rate would peak near 8 percent in 2009 before falling to about 5.5 percent by now.
So the estimates in the report were wrong.\\

\midrule
\textbf{Claim: }The Milwaukee County bus system has ``among the highest fares in the nation.''  \\
\textbf{Label:} False \\
\textbf{Just:} Larson said the Milwaukee County bus system has ``among the highest fares in the nation.''
But the system’s’ \$2.25 cash fare wasn’t at the top of a national comparison, with fares reaching as high as \$4 per trip.
And regular patrons who use a Smart Card are charged just \$1.75 a ride, making the Milwaukee County bus system about on par with average costs. \\
\textbf{Oracle:} Larson said the Milwaukee County bus system has ``among the highest fares in the nation.''
Patrons who get a Smart Card pay \$1.75 per ride.
At the time, nine cities on that list charged more than Milwaukee’s \$2.25 cash fare.
The highest fare -- in Nashville -- was \$4 per ride.\\
\midrule
 \textbf{Claim: }``The Republican who was just elected governor of the great state of Florida paid his campaign staffers, not with money, but with American Express gift cards.''  \\
 \textbf{Label: }Half-True \\
\textbf{Just:} First, we think many people might think Maddow was referring to all campaign workers, but traditional campaign staffers -- the people working day in and day out on the campaign -- were paid by check, like any normal job.
A Republican Party official said it was simply an easier, more efficient and quicker way to pay people.
And second, it's not that unusual.
In 2008, Obama did the same thing. \\
\textbf{Oracle:}
``It's a simpler and quicker way of compensating short-term help.''
Neither Conston nor Burgess said how many temporary campaign workers were paid in gift cards.
When asked how he was paid, Palecheck said: ``Paid by check, like any normal employee there.''
In fact, President Barack Obama's campaign did the same thing in 2008. \\
\bottomrule
\end{tabular}
\caption{Examples of the extracted oracle summaries (Oracle) compared to the gold justification (Just).}
\label{tab:oracle-example}
\end{table*}

\subsection{Manual 6-way Veracity Prediction from explanations}\label{appendix:q}

The Fleiss' $\kappa$ agreement for the 6-label manual annotations is: 0.20 on the \textit{Just} explanations, 0.230 on the \textit{Explain-MT} explanations, and 0.333 on the \textit{Explain-Extr} system. Table~\ref{tab:results:man2:6-way} represent the results of the manual veracity prediction with six classes.

\begin{table}[t]
\centering
\begin{tabular}{clrrr}
\toprule
 & & Just & Explain-Extr & Explain-MT \\ \midrule
$\nwarrow$ & Agree-C & \textbf{0.208} & 0.138 & \cellcolor{myblue}0.163 \\
$\searrow$ & Agree-NS & \textbf{0.065} &  0.250 & \cellcolor{myblue}0.188 \\
$\searrow$ & Agree-NC & \textbf{0.052} & 0.100 & \cellcolor{myblue}0.075 \\
$\searrow$ & Disagree & 0.675 & \cellcolor{myblue}\textbf{0.513} & 0.575\\
\bottomrule 		
\end{tabular}
\caption{Manual classification of veracity label - true, false, half-true, barely-true, mostly-true, pants-on-fire, given a particular explanations from the gold justification (Just), the generated explanation (Explain-Extr) and the explanation learned jointly with the veracity prediction model (Explain-MT). Presented are percentages of the dis/agreeing annotator predictions, where the agreement percentages are split to: correct according to the gold label (Agree-C) , incorrect (Agree-NC) or with not sufficient information (Agree-NS). The first column indicates whether higher ($\nwarrow$) or lower ($\searrow$) values are better. At each row, the best set of explanations is in bold and the best automatic explanations are in blue.}
\label{tab:results:man2:6-way}
\end{table} 

\chapter{Generating Fluent Fact Checking Explanations with Unsupervised Post-Editing}
\label{chap:editing_explanations}

\section{Introduction}
In today's era of social media, the spread of news is a click away, regardless of whether it is fake or real. However, the quick propagation of fake news has repercussions on peoples' lives. To alleviate them, professional fact checkers manually verify the veracity and credibility of news, which is time and labor-intensive, making the process expensive and less scalable. %Moreover, the manual process cannot keep pace with the information generated by billions of people online.
Therefore, the need for accurate, scalable, and explainable automatic fact checking (FC) systems is inevitable \cite{kotonya-toni-2020-explainable-automated}.

Current automatic fact checking systems perform veracity prediction for given claims based on evidence documents \cite{thorne-etal-2018-fever,augenstein-etal-2019-multifc},
% \textit{inter alia}
or based on long lists of supporting ruling comments (RCs, \citet{wang-2017-liar,alhindi-etal-2018-evidence}). RCs are in-depth explanations for predicted veracity labels, but they are challenging to read and not useful as explanations for human readers due to their sizable content. 
% Recently, \citet{atanasova-etal-2020-generating} used extractive summarization to select a subset of sentences from long RCs and used them as short layman explanations. However, since these sentences are cherry-picked from different parts of their corresponding RCs, they are often disjoint and non-fluent. 
%\citet{mishra-etal-2020-generating} %use an attention-based mechanism to generate summaries of evidence documents and 
%show that summaries of evidence documents perform better than using original evidence.
%Recent work has thus proposed to automatically generate short summaries of evidence documents and RCs, functioning as concise explanations.
Recent work \cite{atanasova-etal-2020-generating, kotonya-toni-2020-explainable} has thus proposed to use automatic summarisation to select a subset of sentences from long RCs and used them as short layman explanations. However, with a purely extractive approach \cite{atanasova-etal-2020-generating}, the sentences are cherry-picked from different parts of the corresponding RCs, and as a result, explanations are often disjoint and non-fluent. 

\begin{figure}[t!]
\centering
    \includegraphics[width=0.8\linewidth]{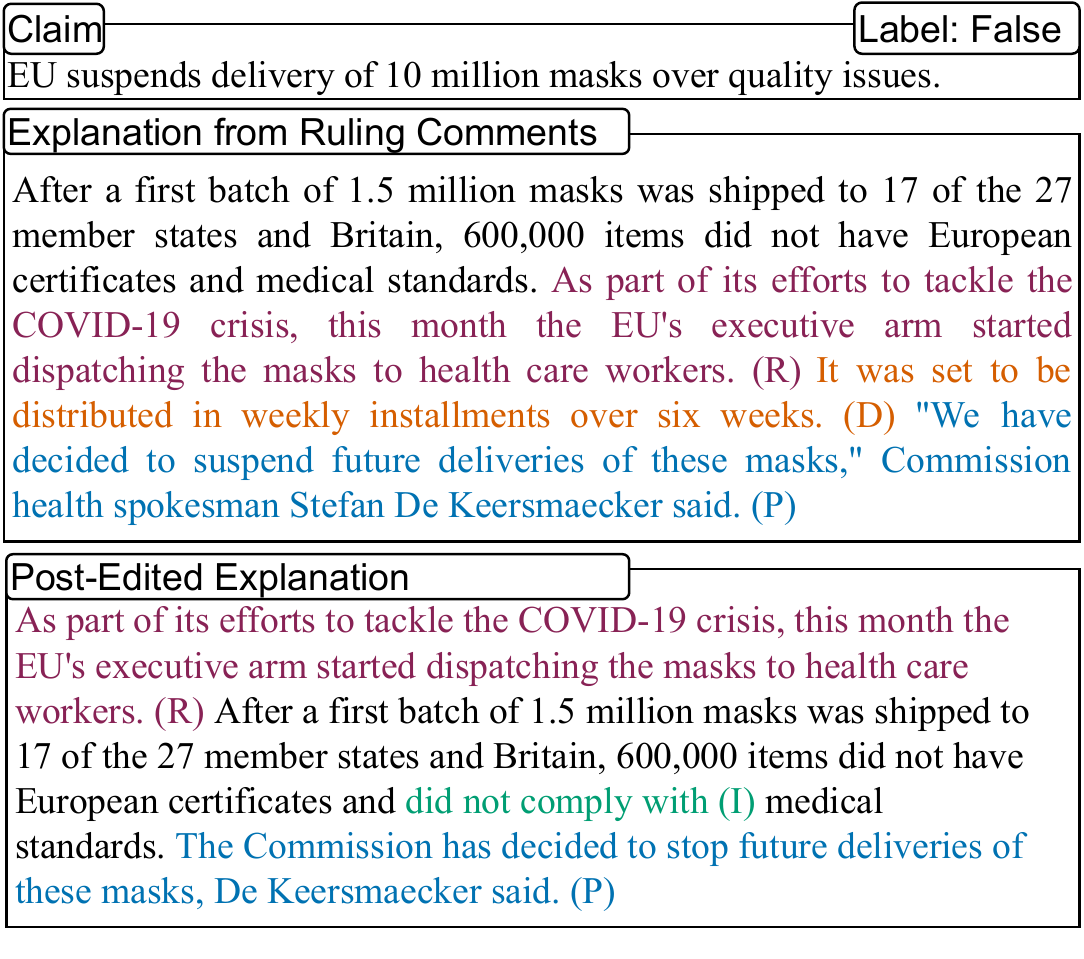}
    \caption{Example of a post-edited explanation from PubHealth that was initially extracted from RCs. We illustrate four post-editing steps: \textcolor{purple_editing}{reordering (R)},  \textcolor{green_editing}{insertion (I)}, \textcolor{orange}{deletion (D)}, and \textcolor{blue_editing}{paraphrasing (P)}.}
    \label{fig:introexp}
\end{figure}

While a sequence-to-sequence model trained on parallel data can partially alleviate these problems, as \citet{kotonya-toni-2020-explainable} propose, it is an expensive affair due to the large amount of data and compute required to train these models. Therefore, in this work, we focus on unsupervised post-editing of explanations extracted from RCs. Recently, researchers have leveraged unsupervised post-editing to generate paraphrases \cite{liu-etal-2020-unsupervised} and sentence simplifications \cite{kumar-etal-2020-iterative}. However, they use short single sentences and perform a combination of exhaustive word and phrase-level edits, which has limited applicability for longer text with multiple sentences, e.g., FC explanations, due to prohibitive convergence times.
% While a Seq2Seq model trained on parallel data can solve these problems, it is an expensive affair in terms of a large amount of data and compute required to train these models. Therefore, in this work, we focus on unsupervised post-editing of sentences obtained from RCs. Our work is motivated by \citet{kumar-etal-2020-iterative} who present an iterative, edit-based unsupervised sentence simplification approach at both word and phrase levels. However, unlike \citet{kumar-etal-2020-iterative} our input sentences are longer and, therefore, word-level edits will lead to longer convergence times. 

Hence, we present a \textit{novel iterative edit-based algorithm} performing three edit operations (insertion, deletion, reorder), all at the phrase level. 
Fig.~\ref{fig:introexp} %presents a qualitative example from PubHealth dataset \cite{kotonya-toni-2020-explainable}, which shows 
illustrates how each post-editing step contributes to creating more concise, readable, fluent, and coherent candidate explanations. %, while also preserving the information important for the fact check.
Our proposed method finds the best post-edited explanation candidate according to a scoring function, ensuring its fluency and readability, semantic preservation, and conciseness quality (\S \ref{sec:scoringfunction}). To ensure that the candidate explanations are grammatically correct, we also perform grammar checking (\S\ref{sec:gramcorr}). As a second step, we apply paraphrasing to improve further the conciseness and human readability of the explanations (\S\ref{sec:para}). Our approach is generic and can be applied to any other application where the objective is to generate a fluent and coherent summary. 

%Moreover, unlike recent editing approaches, we use Transformer-based architectures to edit a given explanation and score candidate explanations.

%Figure~\ref{fig:introexp} presents a qualitative example from PubHealth dataset \cite{kotonya-toni-2020-explainable}. It shows how each post-editing step contributes to creating explanations that are more readable, fluent, and create a coherent story, while also preserving the information important for the fact check.   
% As shown in Figure~\ref{fig:introexp}, we generate connected explanations using three edit operations (insertion, deletion, reordering), with all operations operating at phrase-level. 
% However, in their approach they deal with sentences to performing both word and phrase-level edits is well suited for smaller sentences, however, it becomes costly in time for longer sentences.  % Motivated by \citet{kumar-etal-2020-iterative}, in this work, we present an unsupervised post-editing approach that takes disconnected RCs and converts them to more readable, precise, and fluent explanations using only phrase-level edits. 
% We control our algorithm with a scoring function that measures the quality of our explanations in terms of fluency, semantic similarity, and semantic preservation. Unlike \citet{kumar-etal-2020-iterative}, we use transformer-based architectures for both editing a given explanation and scoring the candidate explanations. We generate connected explanations using three edit operations (insertion, deletion, reordering), with all operations operating at phrase-level. 

In summary, our main \textbf{contributions} are:
\begin{itemize}
    \item To the best of our knowledge, we are the first to explore an iterative unsupervised edit-based algorithm using only phrase-level edits that leads to feasible solutions for long text inputs.
    \item We show how combining an iterative algorithm with grammatical corrections, and paraphrasing-based post-processing leads to fluent and easy-to-read explanations.
    \item We conduct extensive experiments on the LIAR-PLUS\ \cite{wang-2017-liar} and PubHealth\ \cite{kotonya-toni-2020-explainable} FC datasets. Our automated evaluation confirms the success of our proposed method in preserving the semantics important for the fact check and enhancing the readability of the generated explanations. Our manual evaluation confirms that our approach improves the generated explanations' fluency and conciseness.
\end{itemize}

\section{Related Work}
The most closely related work are explainable FC, generative approaches to explainability, and post-editing for language generation.
\subsection{Explainable Fact Checking}
Recent work has produced fact-checking explanations by highlighting words in tweets using neural attention \cite{lu-li-2020-gcan}. 
% However, their explanations are used only to evaluate and compare the proposed model with other baselines without neural attention.
\citet{wu-etal-2020-dtca} propose to model evidence documents with decision trees, which are inherently interpretable ML models. Recently, \citet{atanasova-etal-2020-generating} propose to generate free-text explanations for political claims jointly with predicting the veracity of claims. They formulate an extractive summarisation task to select a few important sentences from a long FC report. Training the summarisation task jointly with veracity prediction results in summaries that better explain the correct veracity label. \citet{atanasova2021diagnostics} also perform extractive explanation generation guided by a set of diagnostic properties of explanations and evaluate on the FEVER~\cite{thorne-etal-2018-fever} FC dataset, where explanation sentences are extracted from Wikipedia documents.

In the domain of public health claims, \citet{kotonya-toni-2020-explainable} propose to generate explanations separately from the task of veracity prediction. \citet{mishra-etal-2020-generating} generate summaries of evidence documents from the Web using an attention-based mechanism. Their summaries perform better than using the original evidence documents directly. 
Similarly to \citet{atanasova-etal-2020-generating,kotonya-toni-2020-explainable}, we present a generative approach for creating FC explanations. In contrast to related work, we propose an unsupervised post-editing approach to improve the fluency and readability of previously extracted FC explanations.

\subsection{Generative Approaches to Explainability}
%Explainable AI \cite{gunning2017explainable} is important to encourage trust of blackbox model's decisions and increase their acceptability among users. 
While most work on explanation generation propose to highlight portions of the input \cite{deyoung-etal-2020-eraser},
% , \textit{inter alia})
some work studies generative approaches to explainability. \citet{NIPS2018_8163} combine explanation generation and target prediction in a pipeline or a joint model for Natural Language Inference with free-text label explanations.
% They find that first explaining and then predicting based on the explanation achieves better trust as the prediction is based on the right reasons.
\citet{stammbach2020fever} propose few-shot training for GPT-3~\cite{9321372} to explain a fact check from retrieved evidence snippets. GPT-3, however, is limited-access and has high computational costs. As in our work, \citet{kotonya-toni-2020-explainable} first extract evidence sentences, which an abstractive summarisation model then summarises. 
% They use the PubHealth dataset. 
In contrast, we are the first to perform unsupervised post-editing of explanations produced using automatic summarisation.

% \citet{liu2019towards} propose an approach to generate abstractive explanations both in terms of texts and numerical scores. 

\subsection{Post-Editing for Language Generation}
Previous work has addressed unsupervised post-editing for multiple tasks like paraphrase generation \cite{liu-etal-2020-unsupervised}, sentence simplification \cite{kumar-etal-2020-iterative} or sentence summarisation \cite{schumann-etal-2020-discrete}. However, all these tasks handle inputs shorter than the long multi-sentence extractive explanations that we have. Furthermore, they perform exhaustive edit operations at the word level and sometimes additionally at the phrase level, which increase computing complexity.
Therefore, we present a novel method that performs a fixed number of edits only at the phrase level followed by grammar correction and paraphrasing.

\section{Method}\label{sec:method4}
Our method is comprised of two steps. First, we select sentences from RCs that serve as extractive FC explanations (\S\ref{sec:method4:selection}). We then apply unsupervised post-editing on the extractive explanations to improve their fluency and coherence (\S\ref{sec:method4:postediting}). 

\subsection{Selecting Sentences for Post-Editing}\label{sec:method4:selection}
\textbf{Supervised Selection.} 
To produce supervised extractive explanations, we build models based on DistilBERT~\cite{sanh2019distilbert} for LIAR-PLUS, and SciBERT~\cite{beltagy-etal-2019-scibert} for PubHealth to allow for direct comparison with \citet{atanasova-etal-2020-generating,kotonya-toni-2020-explainable}. %(LIAR-PLUS) and \citet{kotonya-toni-2020-explainable} (PubHealth). %. We choose DistilBERT following \citet{atanasova-etal-2020-generating} as it is a reduced version of BERT~\cite{devlin-etal-2019-bert}, and we choose SciBERT following \citet{kotonya-toni-2020-explainable}, as it is more suitable for the health domain of the claims in PubHealth.
We supervise explanation generation by $k$ greedily selected sentences from a claim's RCs with the highest ROUGE-2 $F_1$ score w.r.t. the gold justification. We choose $k\!=\!4$ for LIAR-PLUS and $k\!=\!3$ for PubHealth, the average number of sentences in the gold justifications in the corresponding dataset. The selected sentences are positive gold labels, $\mathbf{y}^E\!\in\!\{0,1\}^N $, where $N$ is the number of RC sentences. We also use the veracity labels $\mathbf{y}^{F}\!\in\!Y_{F}$ for supervision. Following~\citet{atanasova-etal-2020-generating-fact}, we learn a multi-task model $g(X)\!=\!(\mathbf{p}^E, \mathbf{p}^F)$. Given input X, comprised of a claim and the RCs, it predicts jointly the veracity explanation $\mathbf{p}^E$ and the veracity label $\mathbf{p}^F$, where $\mathbf{p}^E \in \mathbb{R}^{1, N}$ selects sentences for explanation, i.e. \{0,1\}, and  $\mathbf{p}^{F}\!\in\!\mathbb{R}^{m}$, with $m\!=\!6$ for LIAR-PLUS, and $m\!=\!4$ for PubHealth. Finally, we optimise the joint cross-entropy loss $\mathcal{L}_{MT}\!=\!\mathcal{H}(\mathbf{p}^{E}, \mathbf{y}^{E}) + \mathcal{H}(\mathbf{p}^{F}, \mathbf{y}^{F})$.

\textbf{Unsupervised Selection.} 
We experiment with unsupervised sentence selection to test the possibility of constructing fluent FC explanations in an entirely unsupervised way. We use Longformer~\cite{Beltagy2020Longformer}, which was introduced for tasks with longer input, instead of the sliding-window approach used in \citet{atanasova-etal-2020-generating-fact}, which is without cross-window attention. 
We train a model $h(X)\!=\!\mathbf{p}^F$ to predict the veracity of a claim. We optimise cross-entropy loss $\mathcal{L}_{F}\!=\!\mathcal{H}(\mathbf{p}^{F}, \mathbf{y}^{F})$ and select $k$ sentences $\mathbf{p}^{E'}\!\in\!\mathbb{R}^{1, N}$, \{0, 1\}, with the highest saliency scores. The saliency score of a sentence is the sum of the saliency scores of its tokens. The saliency of a token is the gradient of the input token w.r.t. the output~\cite{Simonyan2013DeepIC}. We select sentences using the raw gradients as \citet{atanasova-etal-2020-diagnostic} show that different gradient-based methods yield similar results. As the selection could be noisy~\cite{kindermans2019reliability}, we consider these experiments as only complementary to the main supervised results.
\subsection{Post-Editing}\label{sec:method4:postediting}
Our post-editing is completely unsupervised and operates on sentences obtained in \S\ref{sec:method4:selection}. It is a search algorithm that evaluates the candidate sequences $\mathbf{p}^{C}$ for a given input sequence -- $\mathbf{p}^{E}$ for supervised selection or $\mathbf{p}^{E'}$ for unsupervised selection. Below, we use $\mathbf{p}^{E}$ to denote both.

Given $\mathbf{p}^{E}$, we iteratively generate multiple candidates by performing phrase-level edits (\S\ref{sec:cadidategeneration}). To evaluate a candidate explanation, we define a scoring function as a product of multiple scorers, also known as a product-of-experts model \cite{hinton2002training}. Our scoring function includes fluency and semantic preservation, and controls the length of the candidate explanation (\S\ref{sec:scoringfunction}). 
We repeat the process for $n$ steps and select the last best-scoring candidate as our final output. We then use grammar correction (\S\ref{sec:gramcorr}) and paraphrasing (\S\ref{sec:para}) to further ensure conciseness and human readability.

\subsubsection{Candidate sequence generation}\label{sec:cadidategeneration}
We generate candidate sequences by phrase-level edits. %Similarly to \citet{kumar-etal-2020-iterative}, 
We use the off-the-shelf syntactic parser from CoreNLP \cite{manning-etal-2014-stanford} to obtain the constituency tree of a candidate sequence $\mathbf{p}^{C}$. As $\mathbf{p}^{C}$ is long, we perform all operations at the phrase level. 
At each step $t$, our algorithm first randomly picks one operation -- insertion, deletion, or reordering, and then randomly selects a phrase. 

For \textbf{insertion}, our algorithm inserts a <MASK> token before the randomly selected phrase, and uses RoBERTa to evaluate the posterior probability of a candidate word \cite{NEURIPS2020_7a677bb4}. This allows us to leverage the pre-training capabilities of RoBERTa and insert high-quality words that support the context of the overall explanation. 
Furthermore, inserting a <MASK> token before a phrase prevents breaking other phrases within the explanation, thus preserving their fluency. 
%By contrast, \citet{kumar-etal-2020-iterative} use a forward and backward language model to evaluate the top-k candidate words.

The \textbf{deletion} operation deletes the randomly selected phrase.
For the \textbf{reorder} operation %\citet{kumar-etal-2020-iterative} performs a computationally expensive exhaustive search over all possible reordering candidates. In contrast, 
we randomly select one phrase, 
which we call \textit{reorder phrase}, and randomly select $m$ phrases, which we call \textit{anchor phrases}. We \textbf{reorder} each \textit{anchor} with a \textit{reorder phrase} and obtain $m$ candidate sequences. The candidates are fed to GPT-2 to select the most fluent one with the fluency score given by Eq.\ \ref{eq:fluency}.

\subsubsection{Scoring Functions}\label{sec:scoringfunction}
The \textbf{fluency score} ($f_{flu}$) measures the language fluency of a candidate sequence. 
%Unlike \citet{kumar-etal-2020-iterative}, who use RNNs for fluency evaluation, 
We use a pre-trained GPT-2 model \cite{radford2019language}. We use the joint likelihood of candidate $\mathbf{p}^{C}$:
\begin{equation}
f_{flu}(\mathbf{p}^{C}) = \prod\nolimits_{i = 1}^{n} P(\mathbf{p}^{C}_i|\mathbf{p}^{C}_1, ...., \mathbf{p}^{C}_{i-1})
\label{eq:fluency}
\end{equation}

In \S\ref{sec:results:manual:quality} we evaluate the achieved fluency of the generated explanations through human evaluation. Additionally, as the fluency score measures the likelihood of the text according to GPT-2, which is trained on 40GB of Internet text, we assume that complex text that is not common or is not likely to appear on the Internet, would also have lower fluency score. Hence, we expect that improving the fluency of an explanation, would lead to more easily understood explanations. We evaluate the latter in \S\ref{sec:result:readability} through automated readability scores.

\textbf{Length score ($f_{len}$)} This score encourages the generation of shorter sentences. We assume that reducing the length of the generated explanation is also beneficial for improving the readability of the explanation as it promotes shorter sentences, which are easier to read. The score is the inverse of the sequence length -- longer candidate sentence have a lower scores. To control over-shortening, we reject explanations with fewer than 40 tokens. The number of tokens is a hyper-parameter chosen after fine-tuning on the validation split.

For \textbf{semantic preservation}, we compute similarities at both word and explanation level between our source explanation ($\mathbf{p}^{E}$) and candidate sequence ($\mathbf{p}^{C}$) at time-step $t$. The word-level semantic scorer evaluates the preserved amount of keyword information in the candidate sequence. Similarly to \citet{NEURIPS2020_7a677bb4}, we use RoBERTa (R) \cite{liu2019roberta}, a pre-trained masked language model, to compute a contextual representation of $\mathrm{word_{i}}$ in an explanation as R$(\mathbf{p}^{E}_i, \mathbf{p}^{E})$. Here, $\mathbf{p}^{E}\!=\!(\mathbf{p}^{E}_1\dots \mathbf{p}^{E}_m)$ is an input sequence of words. We then extract keywords from $\mathbf{p}^{E}$ using Rake \cite{rose2010automatic} and compute a \textbf{keyword-level semantic similarity score}:
\begin{equation}
% \small
f_{w}(\mathbf{p}^{E}, \mathbf{p}^{C})\!=\!\min_{k \in kw(\mathbf{p}^{E})}
\max_{\mathbf{p}^{C}_i \in \mathbf{p}^{C}} R(k, \mathbf{p}^{E})^\intercal R(\mathbf{p}^{C}_i, \mathbf{p}^{C})
\label{eq:wordsemantics}
\end{equation}
which is the lowest cosine similarity among all keywords i.e. the least matched keyword of $\mathbf{p}^{E}$.
%between a keyword from $\mathbf{p}^{E}$ and candidate sequence $\mathbf{p}^{C}$.

The keyword-level semantic similarity preserves the semantic information of the separate keywords in the text. It is, thus, not affected by changes in words that do not bear significant meaning for the overall explanation. However, as this semantic similarity is performed at keyword-level it does not account for preserving the overall meaning of the text and the context that the keywords are used in. 

Hence, we also employ an \textbf{explanation-level semantic preservation scorer}. It measures the cosine similarity of two explanation vectors, which are explanation encodings that contain the overall semantic meaning of the explanation: 
% using Eq: \ref{eq:sentsemantics}.  
\begin{equation}
f_{e}(\mathbf{p}^{E}, \mathbf{p}^{C}) = \frac{(\mathbf{p}^{C})^\intercal  \mathbf{p}^{E}}{||\mathbf{p}^{C}||\mathbf{p}^{E}||}
\label{eq:sentsemantics}
\end{equation}
\noindent We use SBERT \cite{reimers-gurevych-2019-sentence} for obtaining embeddings for both $\mathbf{p}^{E}$, $\mathbf{p}^{C}$. Our overall semantic score is the product of the word level and the explanation level semantics scores:
\begin{equation}
% \small
f_{sem}(\mathbf{p}^{E}, \mathbf{p}^{C}) = f_{w}(\mathbf{p}^{E}, \mathbf{p}^{C})^\beta . \\ f_{e}(\mathbf{p}^{E}, \mathbf{p}^{C})^\eta
\label{eq:overallsemantics}
\end{equation}
\noindent where $\beta$, and $\eta$ are hyper-parameter weights for the separate scores. We evaluate the semantic preservation of the post-edited explanations with automated ROUGE scores (\S\ref{sec:result:rouge}) and manual human annotations (\S\ref{sec:results:manual:quality}, \S\ref{sec:results:manual:info}).

Lastly, \textbf{Named Entity (NE) score ($f_{ent}$)} is an additional  measure for meaning preservation, since NEs hold the key information within a sentence. We identify NEs using an off-the-shelf entity tagger \cite{spacy2} and count their number in a given explanation. 

Our \textbf{overall scoring} function is the product of individual scores, where $\alpha$, $\gamma$, and $\delta$ are hyper-parameter weights for the different scores:
\begin{equation}
% \small
f_(\mathbf{p}^{C}) = f_{flu}(\mathbf{p}^{C})^\alpha .  f_{sem}(\mathbf{p}^{E}, \mathbf{p}^{C}). \\ f_{len}(\mathbf{p}^{C})^\gamma .  f_{ent}(\mathbf{p}^{C})^\delta 
\label{eq:overallscore}
\end{equation}

\subsubsection{Iterative Edit-based Algorithm}\label{sec:iterativeedit}
Given input explanations, our algorithm iteratively performs edit operations for $n$ steps to search for a highly scored candidate ($\mathbf{p}^{C}$). At each step, it computes scores for the previous ($\mathbf{p}^{C-1}$) and candidate sequence (Eq.\ \ref{eq:overallscore}). It selects $\mathbf{p}^{C}$ if its score is larger than $\mathbf{p}^{C-1}$ by a multiplicative factor $r_{op}$:
\begin{equation}
\frac{f_{\mathbf{p}^{C}}}{f_{\mathbf{p}^{C-1}}} > r_{op}
\label{eq:selectop}
\end{equation}
\noindent For each edit operation, we use a separate threshold value $r_{op}$. 
$r_{op}$ allows controlling specific operations where $r_{op} < 1$ allows the selection of candidates ($\mathbf{p}^{C}$) which have lower scores than $\mathbf{p}^{C-1}$.
% like for reorder operation, if $\mathbf{p}^{C}$ gets a lower score than $\mathbf{p}^{C-1}$ then a lower value of $r_{op}$ will enable selection of $\mathbf{p}^{C}$. 
We tune all hyper-parameters, including $r_{op}$, $n$, etc., using the validation split of the LIAR-PLUS dataset.

\subsubsection{Grammatical Correction}\label{sec:gramcorr}
Once the best candidate explanation is selected, we feed it to the \citet{languagetool} toolkit, which detects grammatical errors like capitalization and irrelevant punctuation, and returns a corrected version of the explanation. Furthermore, to ensure that we have no incomplete sentences, we remove sentences without verbs in the explanation. These two steps further ensure that the generated explanations are fluent (further evaluated in \S\ref{sec:results:manual:quality}). %(auxiliary) 

\subsubsection{Paraphrasing}\label{sec:para}
Finally, to improve fluency and readability further, we use Pegasus \cite{zhang2020pegasus}, a model pre-trained for abstractive text summarisation. It focuses on relevant input parts to summarise the input semantics in a concise and readable way. Since we want both fluent and human-readable explanations, we leverage Pegasus without fine-tuning on downstream tasks. 
This way, after applying our iterative edit-based algorithm with grammatical error correction and paraphrasing, we obtain fluent, coherent, and human-readable explanations. %non-redundant. 

% We first split our filtered candidate sequence $(\mathbf{p}^{C})$ into separate sentences, and then generate their paraphrases using Pegasus with a Beam Search of size 10 and pick a paraphrase with maximum similarity with the candidate sequence's sentence. This way, after iterative edit-based algorithm and post-processing we obtain fluent, connect, and non-redundant justifications. 
\section{Experiments}
\subsection{Datasets}\label{sec:dataset}
We use two FC datasets, LIAR-PLUS~\cite{wang-2017-liar} and PubHealth~\cite{kotonya-toni-2020-explainable}. These are the only real-world FC datasets that provide short veracity justifications along with claims, RCs, and veracity labels. We provide the size of the splits in Tab.\ \ref{tab:datasets4}, app. 
% provides the size for each of the splits in the corresponding dataset.
The LIAR-PLUS labels are \{true, false, half-true, barely-true, mostly-true, pants-on-fire\}, and in PubHealth, \{true, false, mixture, unproven\}.
% The claims in LIAR-PLUS are only from PolitiFact and PubHealth contains claims from eight FC sources. 
PubHealth is manually curated, e.g., to exclude poorly defined claims. Finally, the claims in PubHealth are more challenging to read than those in LIAR-PLUS and other real-world FC datasets.

\subsection{Models}
Our experiments include the following models; their hyper-parameters are given in Appendix \ref{sec:experiments}.

%\noindent
\textbf{(Un)Supervised $\mathrm{\mathbf{Top^{N}}}$} extracts RC sentences in an (un)supervised way (\S\ref{sec:method4:selection}), which are later used as input to our method.

%\noindent
\textbf{(Un)Supervised $\mathrm{\mathbf{Top^{N}}}$+$\mathrm{\mathbf{Edits^{N}}}$} generates explanations with the iterative editing (\S\ref{sec:iterativeedit}) and grammar correction (\S\ref{sec:gramcorr}). The inputs are sentences extracted with (Un)Supervised \topn{N}.

%\noindent 
\textbf{(Un)Supervised $\mathrm{\mathbf{Top^{N}}}$+$\mathrm{\mathbf{Edits^{N}}}$+$\mathrm{\mathbf{Para}}$} generates explanations by paraphrasing the explanations from (Un)Supervised \topedit{N} (\S\ref{sec:para}).
%dits-N - (Un)Supervised (see Sec.~\ref{sec:para}).

%\noindent  \textbf{$Edits^{N}$ + Gram - (Un)Supervised} generates explanations by post-processing the explanations produced by $Edits^{N}$ - (Un)Supervised using grammatical tool (see Sec.~\ref{sec:postproc}).

%\noindent  \textbf{$Edits^{N}$ + Gram + Pegasus - (Un)Supervised} generates explanations by post-processing the explanations produced by $Edits^{N}$ + Gram - (Un)Supervised using Pegasus model (see Sec.~\ref{sec:postproc}).
%\noindent
\textbf{MTSum} -- \citet{atanasova-etal-2020-generating-fact}, is a reference model that trains a multi-task system to predict veracity labels and extract explanation N sentences, where N is the average number of the sentences in the justifications of each dataset.
%\noindent

\textbf{AbstrSum} -- \citet{kotonya-toni-2020-explainable}, is a baseline model that generates abstractive explanations with an average sentence length of 3.

%\noindent 
$\mathrm{\mathbf{Lead^{K}}}$~\cite{Nallapati:2017:SRN:3298483.3298681} is a common lower-bound baseline for summarisation models. It selects the first K sentences of the RCs.
\subsection{Evaluation Overview}
We perform both automatic and manual evaluations of the models above. We include automatic measures for readability (\S\ref{sec:result:readability}). While the latter was not included in prior work, we consider readability an essential quality of an explanation, and thus report it. We further include automatic ROUGE $F_1$ scores (overlap of the generated explanations with the gold ones, \S\ref{sec:result:rouge}) for compatibility with prior work and to ensure that our generated explanations don't shift much from the gold ones. In particular, we are interested whether the ROUGE scores for the post-edited explanations are not significantly different from the ROUGE scores of the non-edited explanations, which would indicate preservation of the content important for FC. We note, however, that the automatic measures are limited as they are based on word-level statistics. Especially ROUGE scores should be taken with a grain of salt, as only exact word matches are scored higher and paraphrases or synonyms of words in the gold summary are not scored. 
Hence, we conduct a manual evaluation
% following~\citet{atanasova-etal-2020-generating-fact} 
to further assess the quality of the generated explanations with a user study. As manual evaluation is expensive to obtain, the latter is, however, usually estimated based on small samples.

\section{Automatic Evaluation}\label{sec:result:automated}
We use ROUGE $F_1$ scores to compute overlap between the generated explanations and the gold ones, and compute readability scores to assess how challenging the produced explanations are to read.

\subsection{Readability Results}\label{sec:result:readability}
\textbf{Metrics.} Readability is desirable for FC explanations, as a challenging to read explanation would fail to convey the reasons for the chosen veracity label and would not improve the trust of end-users. %if the explanations generated by our algorithm have improved 
We evaluate readability with Flesch Reading Ease~\cite{kincaid1975derivation} and Dale-Chall Readability Score~\cite{powers1958recalculation}. Flesch Reading Ease gives a text a score $\in\![1, 100]$, where a score $\in\![30, 50]$ requires college education and is difficult to read, a score $\in\!(50, 60]$ requires a 10-\nth{12} school grade and is fairly difficult to read, a score $\in\!(60, 70]$ is regarded as plain English, easily understood by 13-15-year-old students. Dale-Chall Readability Score uses a curated list of words familiar to lower-grade students to assess the complexity of a text. It gives a text a score $\in\![9.0, 9.9]$ when it is easily understood by a 13-\nth{15}-grade (college) student, a score $\in\![8.0, 8.9]$ when it is easily understood by an 11-\nth{12}-grade student, a score $\in\![7.0, 7.9]$ when it is easily understood by a 9-\nth{10}-grade student. We take the mean of the readability scores for the separate instances in the test split (see validation in Tab.\ \ref{tab:results:readability:val}, app.). For 95\% confidence intervals for the scores, see Tab.\ \ref{tab:results:readability:full}, app.

\begin{table}[t]
\centering
\fontsize{10}{10}\selectfont
\begin{tabular}{llrr}
\toprule
& \bf Method &  {\bf Flesch$\uparrow$}  & {\bf DC$\downarrow$}  \\
\midrule
\multicolumn{4}{c}{\bf LIAR-PLUS}\\ 
\multirow{2}{*}{\bf Base} & \leadn{4} & 51.70  & 8.73  \\
& \leadn{6} &  53.24  & 8.43   \\
\cdashline{1-4}
\multirow{3}{*}{\bf Sup.} & \topn{6} (Sup.) & 58.82   & 7.88  \\
& \topedit{6} & 60.21  & 7.75   \\
& \topeditpara{6} & 66.34  & 7.42   \\
\cdashline{1-4}
\multirow{3}{*}{\bf Uns.} & \topn{6} (Uns.) & 53.33   & 8.50   \\
& \topedit{6} & 55.25  & \textcolor{purple}{8.46}  \\
& \topeditpara{6} & 62.13  & 8.11 \\
\cdashline{1-4}
 &$\mathrm{MTSum}^4$ \citeyearpar{atanasova-etal-2020-generating} & 58.56  & 7.99  \\
& Justification & 58.81  & 8.23  \\
\midrule
\multicolumn{4}{c}{\bf PubHealth} \\
\multirow{2}{*}{\bf Base} & \leadn{3} & 44.44 & 9.12   \\
& \leadn{5} & 45.96  & 8.85   \\
\cdashline{1-4}
\multirow{3}{*}{\bf Sup.} & \topn{5} (Sup.) & 48.63   & 8.67   \\
& \topedit{5} & 53.79 & 8.37   \\
& \topeditpara{5}& 61.39  & 7.97  \\
\cdashline{1-4}
\multirow{3}{*}{\bf Uns.} & \topn{5} (Uns.) & 45.20   & 8.94  \\
& \topedit{5} & 50.74  & 8.63  \\
& \topeditpara{5}& 60.07  & 8.15  \\
\cdashline{1-4}
&$\mathrm{MTSum}^3$ \citeyearpar{atanasova-etal-2020-generating} & 48.73  & 8.88   \\
& Justification & 49.29 & 9.17  \\
\bottomrule
\end{tabular}
\caption{Readability measures -- Flesch and Dale-Chall (DC) (\S\ref{sec:result:readability}), over the test splits. We report baseline (Base), supervised (Sup.), and unsupervised (Uns.) results (\S\ref{sec:method4:selection}). We also report prior work results -- MTSum$^{N}$, where we have the outputs to compute readability, and results given the gold Justification. Readability scores for \topedit{N} and \topeditpara{N} are statistically significant ($p$<0.05) compared to \topn{N} and MTSum$^{N}$, except for the score in \textcolor{purple}{purple}. }
\label{tab:results:readability:main}
\end{table}

\textbf{Results.} Table~\ref{tab:results:readability:main} presents the readability results, where our iterative edit-based algorithm consistently improves the reading ease of the explanations by up to 5.16 points, and reduces the grade requirement by up to 0.30 points. The improvements are statistically significant ($p$<0.05) in both supervised and unsupervised explanations, except for the Dale-Chall score for the LIAR unsupervised explanations.
% , where the 95\% confidence interval is still decreased compared to the non-edited explanations
Paraphrasing further improves significantly ($p$<0.05) the text's reading ease by up to 9.33 points, and reduces the grade requirement by up to 0.48 points. Importantly, MTSum \cite{atanasova-etal-2020-generating-fact} explantions and the gold justifications are fairly difficult to read and can require even college education to grasp the explanation, while the explanations generated by our algorithm can be easily understood by 13-15-year-old students according to the Flesch Reading Ease score.

\textbf{Overall observations.} Our results show that our method makes FC explanations less challenging to read and makes them accessible to a broader audience of up to \nth{10}-grade students.
% \bgroup
% \def\arraystretch{1.2}

\subsection{Automatic ROUGE Scores}\label{sec:result:rouge}
% \bgroup
% \def\arraystretch{1}

\textbf{Metrics.} To evaluate the generated explanations w.r.t. the gold justifications, we follow~\citet{atanasova-etal-2020-generating-fact, kotonya-toni-2020-explainable} and use measures from automatic text summarisation -- ROUGE-1/2/L scores. These account for n-gram (1/2) and longest (L) overlap between generated and gold justification. The scores are recall-oriented, i.e., they calculate how many of the n-grams in the gold text appear in the generated one.

\textbf{Caveats.} Here, we use ROUGE scores to verify that the generated explanations \textit{preserve information important for the fact check}, as opposed to generating completely unrelated text. Thus, we are interested in whether the ROUGE scores of the post-edited explanations are \textit{close but not necessarily higher} than those of the input sentences selected from RCs. Notably, we include paraphrasing and new word insertion to improve the explanation's readability, which, while bearing the same meaning, necessarily results in lower ROUGE scores.

\begin{table}[t]
\centering
\fontsize{10}{10}\selectfont
\begin{tabular}{llrrr}
\toprule
& \bf Method & \textbf{R-1}$\uparrow$   & \textbf{R-2}$\uparrow$  &  \textbf{R-L}$\uparrow$  \\
\midrule
\multicolumn{5}{c}{\bf LIAR-PLUS} \\ 
\multirow{2}{*}{\bf Base} &\small \leadn{4} & \textit{28.11} & \textit{6.96} & \textit{24.38}  \\
&\small \leadn{6} &  29.15 & 8.28 & 25.84  \\
\cdashline{1-5}
\multirow{3}{*}{\bf Sup.} &\small \topn{6} (Sup.) & 34.42 &12.36  & 30.58 \\
&\small \topedit{6} & 33.92 & 11.73 & 30.01  \\
&\small \topeditpara{6} & 33.94 & \underline{11.25} &  30.08 \\
\cdashline{1-5}
\multirow{3}{*}{\bf Uns.} &\small \topn{6} (Uns.) & 29.63 & 7.58  & 25.86 \\
&\small \topedit{6} & 28.93 &  \underline{7.06}  & 25.14  \\
&\small \topeditpara{6} & 28.98 & \underline{6.84} &  25.39 \\
\cdashline{1-5}
 &\small $\mathrm{MTSum}^4$ \scriptsize \citeyearpar{atanasova-etal-2020-generating}  & \textit{35.70} & \textit{13.51} & \textit{31.58} \\
\midrule

\multicolumn{5}{c}{\bf PubHealth} \\
\multirow{3}{*}{\bf Base} &\small \leadn{3} & \textit{29.01}  & \textit{10.24}  & \textit{24.18}  \\
&\small \leadn{3} & 23.05  & 6.28 & 19.27  \\
&\small \leadn{5} & 23.73 & 6.86 & 20.67 \\
\cdashline{1-5}
\multirow{3}{*}{\bf Sup.} &\small \topn{5} (Sup.) & 29.93 & 12.42  & 26.24   \\
&\small \topedit{5} & 29.38  & 11.16 & 25.41  \\
&\small \topeditpara{5}& 28.40 & \underline{9.56} & \underline{24.37}  \\
\cdashline{1-5}
\multirow{3}{*}{\bf Uns.} &\small \topn{5} (Uns.) & 23.52  & 6.12 & 19.93 \\
% $Edits^{5}$ - UnSupervised & 18.09 & 4.41 & 15.48 & 42.29 {\scriptsize $\pm$17.34} & 7.36{\scriptsize $\pm$ 0.80 } \\
&\small \topedit{5} & 23.09  & 5.56 & 19.44 \\
&\small \topeditpara{5}& 23.35 &  \underline{5.38}  & 19.56  \\
\cdashline{1-5}
&\small $\mathrm{AbstrSum^3}$\ \scriptsize \citeyearpar{kotonya-toni-2020-explainable} & \textit{32.30} & \textit{13.46} & \textit{26.99} \\
&\small $\mathrm{MTSum}^3$ \scriptsize \citeyearpar{atanasova-etal-2020-generating}  & 33.55 & 13.12 & 29.41 \\
\bottomrule
\end{tabular}
\caption{ROUGE-1/2/L $F_1$ scores (\S\ref{sec:result:rouge}) of baseline (Base), supervised (Sup.) and usupervised (Uns.) methods over the test splits. In \textit{italics}, are results reported in prior work. \underline{Underlined} scores of \topedit{N} and \topeditpara{N} are statistically significant ($p<0.05$) compared to \topn{N} scores, N=\{5,6\}. For validation and ablations (Tab.\ \ref{tab:results:rouge:val}), and for confidence intervals (Tab.\ \ref{tab:results:rouge:full}), see appendix.}
\label{tab:results:rouge:main}
\end{table}
\textbf{Results.} Table~\ref{tab:results:rouge:main} presents the ROUGE score results. First, comparing the results for the input \topn{N} sentences with the intermediate and final explanations generated by our system, we see that, while very close, the ROUGE scores tend to decrease. For PubHealth, we also see that the intermediate explanations always have higher ROUGE scores than our system's final explanations. These observations corroborate two main assumptions about our system. First, our system preserves a large portion of the information important for explaining the veracity label, which is also present in the justification. This is further corroborated by observing that the decrease in the ROUGE scores is often not statistically significant ($p<0.05$, except for some ROUGE-2 and one ROUGE-L score). Second, the iterative editing and the subsequent paraphrasing allow for the introduction of novel n-grams, which preserve the meaning of the text, but are not explicitly present in the gold justification, which affects the word-level ROUGE scores. We further discuss this in \S\ref{sec:discussion} and the appendix.

The ROUGE scores of the explanations generated by our post-editing algorithm when fed with sentences selected in an unsupervised way are considerably lower than with the supervised models. Hence, supervision for extracting the most important sentences is important to obtain explanations close to the gold ones. Finally, the systems' results are mostly above \leadn{N}, with a few exceptions for the unsupervised explanations for LIAR-PLUS.   

\textbf{Overall observations.} We note that while automatic measures can serve as sanity checks and point to major discrepancies between generated explanations and gold ones, related work in generating FC explanations~\cite{atanasova-etal-2020-generating-fact} has shown that the automatic scores to some extent disagree with human evaluation studies, as they only capture word-level overlap and cannot reflect improvements of explanation quality. Human evaluations are therefore conducted for most summarisation models~\cite{chen-bansal-2018-fast,tan-etal-2017-abstractive}, which we include in \S\ref{sec:results:manual}.

\section{Manual Evaluation}\label{sec:results:manual}
As automated ROUGE scores only account for word-level similarity between the generated and the gold explanation, and the readability scores account only for surface-level characteristics of the explanation, we further conduct a manual evaluation of the quality of the produced explanations.

\subsection{Explanation Quality}\label{sec:results:manual:quality}
We manually evaluate two explanations: our baseline method (the input \topn{N} sentences) and our best approach (the final explanations produced after paraphrasing (\topeditpara{N})). 
We perform a manual evaluation of the test explanations obtained from supervised selection for both datasets with two annotators for each. Both annotators have a university-level education in English. 

\textbf{Metrics.} We show a claim, veracity label, and two explanations to each annotator and ask them to rank the explanations according to the following criteria. \textbf{Coverage} means the explanation contains important and salient information for the fact check. \textbf{Non-redundancy} implies the explanation does not contain any redundant/repeated/not relevant information to the claim. \textbf{Non-contradiction} checks if there is information contradictory to the fact check. \textbf{Fluency} measures the grammatical correctness of the explanation and if there is a coherent story. \textbf{Overall} measures the overall explanation quality. We allow annotators to give the same rank to both explanations~\cite{atanasova-etal-2020-generating-fact}. We randomly sample 40 instances\footnote{Due to the increased cost and execution time of the complex annotation task, and following related work that manually evaluates FC explanations~\cite{atanasova-etal-2020-generating-fact} and machine-generated summaries~\cite{liu-lapata-2019-text}.} and do not provide the annotators the explanation type.

\textbf{Results.} Table \ref{tab:results:humanevals_task1} presents the human evaluation results for the first task. Each row indicates the annotator number and the number of times they ranked an explanation higher for one criterion. 
% \textit{Both} refers to both explanations being equal. %After adding values for \textit{Both} to numbers for input and output, 
Our system's explanations achieve higher acceptance for non-redundancy and fluency for LIAR-PLUS. The results are more pronounced for the PubHealth dataset, where our system's explanations were preferred in almost all metrics by both annotators. We hypothesise that PubHealth being a manually curated dataset leads to overall cleaner post-editing explanations, which annotators prefer. %explains that applying our approach on a clean dataset is highly preferred by annotators. 

\subsection{Explanation Informativeness}\label{sec:results:manual:info}
\textbf{Metrics.} We also perform a manual evaluation for veracity prediction. We ask annotators to provide a veracity label for a claim and an explanation where, as for Explanation Quality, the explanations are either our system's input or output. The annotators provide a veracity label for three-way classification: true, false, and insufficient (see map to original labels in app.). We use 30 instances of each explanation type and perform evaluation with two annotators for each dataset and instance.

\textbf{Results.} %In task two of veracity label prediction f
For LIAR-PLUS, one annotator gave the correct label 80\% times for the input and 67\% times for the output explanations. The second annotator chose the correct label 56\% \& 44\% times correspondingly (Tab.\  \ref{tab:results:humanevals_task2} in app.). 
% Both annotators found at least 16\% explanations to be insufficient for veracity prediction.
For PubHealth, both annotators found each explanation useful for the task. The first annotator chose the correct label 50\% \& 40\% of the times for the input and output explanations. The second annotator chose the correct label for 70\% of both explanations. This corroborates that for a clean dataset like PubHealth our explanations help for the task of veracity prediction.
\begin{table}[t]
\centering
\setlength{\tabcolsep}{3pt}
% \fontsize{10}{9}\selectfont
% \resizebox{.48\textwidth}{!}
{\begin{tabular}{llrrrrrr} 
\toprule
\multirow{2}{*}{\textbf{Criterion}} & & \multicolumn{3}{c}{\bf LIAR-PLUS}  & \multicolumn{3}{c}{\bf PubHealth} \\
% \midrule
& \# & \bf $T^{6}$ & \bf $T^{6}$+$E^{6}$+$P$ & \bf Both & \bf $T^{5}$ & \bf $T^{5}$+$E^{5}$+$P$ & \bf Both \\
\midrule
%  \multicolumn{7}{c}{\bf Coverage}  \\
%  \midrule
\multirow{2}{*}{Coverage} & 1 & \textbf{42.5} & 0.0 & 57.5 & 27.5 & \textbf{60.0} & 12.5 \\
& 2 & \textbf{40.0} & 5.0  & 55.0 &  \textbf{22.5} & 20.0 & 57.5 \\
\midrule
% \multicolumn{7}{c}{\bf Non-redundancy} \\
% \midrule
\multirow{2}{*}{Non-redundancy} & 1 & 10.0 & \textbf{87.5}  & 2.5 &  10.0 & \textbf{82.5}  & 7.5 \\
& 2 & 7.5  & \textbf{10.0}  & 82.5 &  7.5 & \textbf{75.0} & 17.5 \\
\midrule
% \multicolumn{7}{c}{\bf Non-contradictory}   \\
% \midrule
\multirow{2}{*}{Non-contradictory} & 1 & \textbf{32.5}  & 5.0 & 62.5 & 7.5 & \textbf{10.0} & 82.5 \\
& 2 &  \textbf{10.0}  & 7.5 & 82.5 & \textbf{20.0} & 15.0 & 65.0 \\
\midrule
% \multicolumn{7}{c}{\bf Fluency}  \\
% \midrule
\multirow{2}{*}{Fluency} & 1 & 40.0 & \textbf{57.5} & 2.5 & 35.0 & \textbf{52.5} & 12.5 \\
& 2 &  \textbf{77.5}  & 15.0 & 7.5 & 20.0 & \textbf{72.5} & 7.5 \\
\midrule
% \multicolumn{7}{c}{\bf Overall Quality}   \\
% \midrule
\multirow{2}{*}{Overall} & 1 & \textbf{57.5} & 42.5 & 0.0 & 35.0 & \textbf{62.5} & 2.5 \\
& 2 &  \textbf{62.5} & 15.0 & 22.5 & 25.0  & \textbf{67.5} & 7.5 \\
\bottomrule
\end{tabular}}
\caption{Manual annotation results of explanation quality.
% with two annotators for both datasets
Each value is the proportion of the times an annotator preferred a justification for a criterion. The preferred method, out of the input \topn{N} (Supervised) and the output of our method, \topeditpara{N}, is emboldened, Both indicates no preference.} %After adding value of Both, i.e., when annotators found both explanations better, to the values when annotators found either input or output better, we made the higher values bold. 
\label{tab:results:humanevals_task1}
\end{table}
\section{Discussion and Conclusion}\label{sec:discussion}
Our automatic and manual evaluation results suggest two main implications of our post-editing algorithm. 
First, the automatic ROUGE evaluation confirmed that the post-editing
%used to select from the generated candidate explanations, 
preserves a large portion of important information contained in the gold explanation and important for FC.
Our manual veracity predictions further supports this -- the post-edited explanations are most useful for predicting the correct label (see also Tab.\ \ref{tab:qualexamples}, app. for examples). Hence, we conjecture that our post-editing can be applied more generally for automated summarisation for knowledge-intensive tasks, such as FC and question answering, where the information needed for prediction has to be preserved. 

Second, with both the automatic and manual evaluation, we corroborate that our proposed post-editing method improves several qualities of the generated explanations -- fluency, conciseness, and readability. The latter supports the usefulness of the length and fluency scores as well as the grammatical correction and the paraphrasing steps promoting these particular qualities of the generated explanations. Fluency, conciseness, and readability are important prerequisites for building trust in automated FC predictions especially for systems used in practice as \citet{thagard1989explanatory} find that people generally prefer simpler, more general explanations with fewer causes.  They can also contribute to reaching a broader audience when conveying the claim's veracity. Conciseness and readability are also the downsides of current professional long and in-depth RCs, which some leading FC organisations, e.g., PolitiFact,
% \footnote{https://www.politifact.com/}
have slowly started addressing by including short overview sections.

% \section{Conclusion}\label{sec:conclusion}
% In this work, we present an unsupervised post-editing approach to improve extractive explanations for fact-checking. Our novel approach is based on an iterative edit-based algorithm and rephrasing-based post-processing. In our experiments on two fact checking benchmarking datasets, we observe, in both the manual and automatic evaluation, that our approaches generate fluent, coherent, and semantics-preserving explanations. 


% \clearpage
% \begin{verbatim}
%   \appendix
% \end{verbatim}
% \appendix
\label{sec:appendix}
\section{Appendices}
\subsection{Manual Evaluation}

As explained in the Section \ref{sec:results:manual} of the main paper, we mapped user inputs (TRUE / FALSE) for task two to the original labels for each dataset. For Liar, we map "true", "mostly-true", "half-true" to TRUE and "false", "pants-fire", and "barely-true" to FALSE. In the PubHealth dataset, we map "true" to TRUE, "false" to FALSE.
% and map both values to "mixture". 
The "insufficient" label is mapped to UNPROVEN. This way, once the mapping is done, we then compute the number of matches and non-matches to get an overall accuracy for this subset.

We appointed annotators with a university-level education in English. 

Additional human evaluation results are presented in Table \ref{tab:results:humanevals_task2}.
%Second Task Results for Human Evaluation
\begin{table}[t]
\centering
% \fontsize{10}{10}\selectfont
% \resizebox{.48\textwidth}{!}{
\begin{tabular}{llrrrrrr}
\toprule
\multirow{2}{*}{\bf \#} & \multirow{2}{*}{\bf Explanation Type} & \multicolumn{3}{c}{\bf LIAR-PLUS}  &  \multicolumn{3}{c}{\bf PubHealth} \\
 &  & \bf M & \bf NM & \bf I & \bf M & \bf NM & \bf I  \\
\midrule
1 & \topn{N} (Supervised) & 20 & 5 & 5 & 15 & 15 & 0 \\
1 & \topeditpara{6} & 14 & 7 & 9 & 12 & 18 & 0 \\
\midrule
2 & \topn{N} (Supervised) & 11 & 14 & 5 & 21 & 9 & 0 \\
2 & \topeditpara{5} & 13 & 10 & 7 & 21 & 9 & 0 \\
\bottomrule
\end{tabular}
% }
\caption{Manual evaluation results for predicting a veracity label. \# refers to annotator number, M/NM refers to number of matches/non-matches between annotator and original labels, I refers to number of explanations that were found to be insufficient to predict a label.}
\label{tab:results:humanevals_task2}
\end{table}

% \section{Examples} Table~\ref{tab:qualexamples} shows a qualitative example from the PubHealth dataset. We find that the final post-processed explanation is more readable, fluent, and concise in comparison to the originally selected explanation from RCs.

\subsection{Experimental Setup}
\label{sec:experiments}
\begin{table}[t]
\centering
\begin{tabular}{lrrr}
\toprule
\textbf{Dataset} & \textbf{Train size} & \textbf{Dev size} & \textbf{Test size} \\ \midrule
\small LIAR-PLUS & 10,146 & 1,278 & 1,255 \\
\small PubHealth & 9,817 & 1,227 & 1,235 \\
\bottomrule
\end{tabular}
\caption{Size of the fact checking datasets used in this work (\S\ref{sec:dataset}).}
\label{tab:datasets4}
\end{table}

Table \ref{tab:datasets4} presents split size information for the used datasets.

\subsubsection{Selection of Ruling Comments}

\begin{table*}[t]
\centering
% \fontsize{10}{10}\selectfont
\begin{tabular}{lrrrrrr}
\toprule
& \multicolumn{3}{c}{\bf Validation} & \multicolumn{3}{c}{\bf Test}\\
\textbf{Method} & \textbf{R-1}  & \textbf{R-2} & \textbf{R-L} & \textbf{R-1}  & \textbf{R-2} & \textbf{R-L}\\
\midrule
SciBERT, w-1, l-1200 & 26.00 & 7.29 & 21.41 & 25.78 & 7.71 & 21.42 \\
SciBERT, w-1, l-1500 & 27.78 & 9.81 & 23.32 & 27.37 & 9.62 & 23.07 \\
SciBERT, w-1, l-1700 & 28.73 & 11.27 & 24.42 & 28.45 & 11.32 & 24.21\\
SciBERT, w-2, l-1700 & 30.15 & 12.32 & 25.66 & 29.71 & 12.04 & 25.35\\
SciBERT, w-5, l-1700 & 30.96 & 12.59 & 26.54 & 30.79 & 12.31 & 26.38\\
\bottomrule
\end{tabular}
\caption{Fine-tuning for PubHealth supervised multi-task model over positive sentence loss weight, base model and maximum length.}
\label{tab:finetune:ph:sup}
\end{table*}

\begin{table*}[t]
\centering
% \fontsize{10}{10}\selectfont
\begin{tabular}{lrrrrrr}
\toprule
& \multicolumn{3}{c}{\bf Validation} & \multicolumn{3}{c}{\bf Test}\\
\textbf{Method} & \textbf{R-1}  & \textbf{R-2} & \textbf{R-L} & \textbf{R-1}  & \textbf{R-2} & \textbf{R-L}\\
\midrule
\multicolumn{7}{l}{\bf LIAR-PLUS Unsup}\\
\topn{6}  & 29.26 & 7.98 & 25.83 & 29.62 & 7.94 & 26.04\\ 
Filtered \topn{6} & 29.52 & 7.90 & 25.98 & 29.60 & 7.96 & 25.94 \\
\midrule
\multicolumn{7}{l}{\bf LIAR-PLUS SUP}\\
\topn{6}  & 34.42 & 12.35 & 30.64 & 34.49 & 12.54 & 30.67 \\ 
Filtered \topn{6} & 34.30 & 12.20 & 30.51 & 34.42 & 12.36 &  30.58 \\
\midrule
\multicolumn{7}{l}{\bf PubHealth Unsup}\\
\topn{5} & 23.78 & 6.23 & 19.95 & 23.13 & 6.08 & 19.63 \\
Filtered \topn{5} & 23.94 & 6.13 & 20.04 & 23.52 & 6.12 & 19.93 \\
\midrule
\multicolumn{7}{l}{\bf PubHealth SUP}\\
\topn{5} & 30.24 & 12.61 & 26.36 & 29.78 & 12.50 & 26.18 \\
Filtered \topn{5} & 30.35 & 12.63 & 26.43 & 29.93 & 12.42 & 26.24\\
\bottomrule
\end{tabular}
\caption{Sentence clean-up of long sentences for LIAR-PLUS and PubHealth.}
\label{tab:finetune:sentence-scores}
\end{table*}

For the supervised selection of RCs, as described in Section~\ref{sec:method4:selection}, we follow the implementation of the multi-task model of \citet{atanasova-etal-2020-generating-fact}. For LIAR-PLUS, we don't conduct fine-tuning as the model is already optimised for the dataset. For PubHealth, we change the base model to SciBERT, as the claims in PubHealth are from the health domain and previous work~\cite{kotonya-toni-2020-explainable} SciBERT outperforms BERTs for the domain. Table~\ref{tab:finetune:ph:sup} presents the results for the fine-tuning we performed over the multi-task architecture with a grid-search over the maximum length limit of the text and the weight for the positive sentences in the explanation extraction training objective. We finally select and use explanations generated with the multi-task model with a maximum text length of 1700, and a positive sentence weight of 5.

For the unsupervised selection of explanation sentences, we employ a Longformer model. We construct the Longformer model with BERT as a base architecture and conduct 2000 additional fine-tuning steps for the newly added cross-attention weights to be optimised. We then train models for both datasets supervised by veracity prediction. The most salient sentences are selected as the sentences that have the highest sum of token saliencies. 

Finally, we remove long sentences and questions from the RCs, where the ROUGE score changes after filtering are illustrated in Table~\ref{tab:finetune:sentence-scores}, which results in the \topn{N} sentences, that are used as input for the post-editing method.

These experiments were run on a single NVIDIA TitanRTX GPU with 24GB memory and 4 Intel Xeon Silver 4110 CPUs. Model training took $\sim 3$ hours.

\subsubsection{Iterative Based Algorithm}
We used the validation split of LIAR-PLUS to select the best hyper-parameters for both datasets. We use the weight of 1.5, 1.2, 1.4, 0.95 for $\alpha$, $\eta$, $\gamma$, $\delta$ and 1.0 for $\beta$ in our scoring function. We set the thresholds as 0.94 for reordering, 0.97 for deletion, and 1.10 for insertion.  We keep all models -- GPT-2, RoBERTa, and Pegasus, fixed and do not finetune them on any in-house dataset. We run our search algorithm on a single V100-32 GB GPU for 220 steps, which takes around 13 hours for each split for both datasets. 
\subsection{Novelty and Copy Rate}

\begin{table*}[t]
\centering
% \fontsize{10}{10}\selectfont
\begin{tabular}{llll}
\toprule
\textbf{Method} & \textbf{Copy Rate} & \textbf{Novelty} & \textbf{Gold Coverage}\\ \midrule
\multicolumn{4}{l}{\bf LIAR-PLUS}\\
% Justification & 70.7 $\pm$12.4 & 29.3 $\pm$12.4 \\
% $Edits^{6}$ Sup.  & 98.5 $\pm$1.7 & 1.5 $\pm$1.7 \\ 
% $Edits^{6}$ + PP Sup. & 94.8 $\pm$3.4 & 5.2 $\pm$3.4\\
\topn{6} Sup. & 100 & 0 & 29.2 $\pm$11.4\\
Justification & 41.4 $\pm$13.0 & 58.6 $\pm$13.0 & 100 \\ 
\topedit{6} Sup. & 98.5 $\pm$1.8 & 1.5 $\pm$1.8 & 30.7 $\pm$12.1\\
\topeditpara{6} Sup. &90.8 $\pm$4.8 & 9.2 $\pm$4.8 & 32.5 $\pm$12.6 \\
\midrule
\multicolumn{4}{l}{\bf PubHealth}\\
% Justification & 69.7 $\pm$19.2 & 30.3 $\pm$19.2\\
% $Edits^{6}$ Sup.  & 98.1 $\pm$3.4 & 1.8 $\pm$2.0 \\ 
% $Edits^{6}$ + PP Sup. & 93.6 $\pm$5.0 & 6.3 $\pm$4.2\\
\topn{5} Sup.& 100 & 0 & 26.3 $\pm$21.2 \\
Justification &  47.1 $\pm$21.0 & 52.9 $\pm$21.0 & 100 \\
\topedit{5} Sup.  & 98.1 $\pm$3.4 & 1.8 $\pm$2.0 & 27.8 $\pm$21.3\\
\topeditpara{5} Sup. & 90.4 $\pm$5.8  & 9.5 $\pm$5.2 & 28.5 $\pm$20.2\\
\bottomrule
\end{tabular}
\caption{Copy rate from the Ruling Comments, Novelty w.r.t. the Ruling comments, and Coverage \% of words in the explanation that are found in the justification.}
\label{tab:results:novelty}
\end{table*}

Table~\ref{tab:results:novelty} presents additional statistics for the generated explanations from the test sets of both datasets. First, we compute how many of the words from the input \topn{N} RCss are preserved in the final explanation. We find that with the final step of the post-editing process, up to 8\% of the tokens from the RCs are not found in the final explanation. On the other hand, our post-editing approach generates up to 10\% novel words that are not previously found in the RCs. This could explain the lower results for the ROUGE scores, which account only for exact token overlaps. Finally, while ROUGE scores are recall-oriented, i.e., they compute how many of the words in the gold explanation can be found in the candidate one, we compute a precision-oriented statistic of the words in the candidate that can be found in the gold explanation. Surprisingly, while ROUGE scores of our generated explanations decrease after post-processing, the reverse score increases, pointing to improvements in the precision-oriented overlap with our method.

In addition, in LIAR/PubHealth, the average summary length is 136/142 tokens for the extracted RCs, 89/86 for the gold justifications, 118.7/117.3 after iterative editing, and 98.5/94.7 after paraphrasing.

% However, we observe that some sentences lead to difficult interpretation like ``An article from the previous week
% said deaths decreased, but that’s what the latest CDC statement says''. It is because of the deletion and reordering operations of our post-editing approach. We can, however, control these operations by finding the optimal hyper-parameters which we leave as a future scope of this work. 

\subsection{Automatic Evaluation}
In Table~\ref{tab:results:readability:full} and Table~\ref{tab:results:rouge:full}, we provide more detailed results over the test splits including confidence intervals.
In Table~\ref{tab:results:readability:val} and Table~\ref{tab:results:rouge:val}, we provide results over the validation split of the datasets for the ROUGE and readability automatic evaluation. We additionally provide ablation results for components of our approach. First, applying Pegasus directly on the extracted sentences preserves a slightly larger amount of information when compared to applying Pegasus on top of the iterative editing approach -- up to 0.96 ROUGE-L scores, but the readability scores are still lower -- up to 4.28 Flesch Reading Ease points. We also show results of the two parts included in the Edits step -- the iterative editing and the grammar correction. We find that the grammar correction improves the ROUGE scores with up to 8 ROUGE-L score points and up to 8 Flesch Reading Ease points.

\begin{table*}[t]
\centering
\fontsize{10}{10}\selectfont
\begin{tabular}{llrrrr}
\toprule
& \bf Method &  {\bf Flesch $\nearrow$} &  {\bf Flesch  CI} & {\bf Dale-Chall $\searrow$}  & {\bf Dale-Chall CI}  \\
\midrule
\multicolumn{6}{c}{\bf LIAR-PLUS}\\ 
\multirow{2}{*}{\bf Baselines} & \leadn{4} & 51.70 & { [50.93-52.53]} & 8.73 & { [8.67-8.78]} \\
& \leadn{6} &  53.24 &  { [52.58-53.92]} & 8.43  & { [8.38-8.47]} \\
\cdashline{1-6}
\multirow{3}{*}{\bf Supervised} & \topn{6} (Supervised) & 58.82  & { [58.13-59.54]} & 7.88 &  { [8.17-8.28]} \\
& \topedit{6} & 60.21 &  { [59.51-60.95]} & 7.75  &   { [7.70-7.80]} \\
& \topeditpara{6} & 66.34 &  { [65.73-66.98]} & 7.42  & { [7.37-7.47]} \\
\cdashline{1-6}
\multirow{3}{*}{\bf Unsupervised} & \topn{6} (Unsupervised) & 53.33  & { [52.70-53.92]} & 8.50 &  { [8.46-8.54]}  \\
& \topedit{6} & 55.25 &  { [54.60-55.88]} & \textcolor{purple}{8.46} &  { [8.42-8.51]} \\
& \topeditpara{6} & 62.13 &  { [61.56-62.71]} & 8.11 &  { [8.06-8.15]} \\
\cdashline{1-6}
 & $\mathrm{MTSum}^4$ \scriptsize \citeyearpar{atanasova-etal-2020-generating} & 58.56 &  { [57.75-59.31]} & 7.99  &  {[7.94-8.03]}\\
& Justification & 58.81 &  { [58.22-59.41]} & 8.23 &  { [7.93-8.04]} \\
\midrule
\multicolumn{6}{c}{\bf PubHealth} \\
& \leadn{3} & 44.44 &  { [43.05-45.68]} & 9.12  & { [9.05-9.19]} \\
& \leadn{5} & 45.96 &  { [44.80-46.98]} & 8.85  & { [8.79-8.91]} \\
\cdashline{1-6}
\multirow{3}{*}{\bf Supervised} & \topn{5} (Supervised) & 48.63  & { [47.91-49.44]} & 8.67 &  { [8.62-8.72]}  \\
& \topedit{5} & 53.79 &  { [53.01-54.56]} & 8.37  & { [8.31-8.42]} \\
& \topeditpara{5}& 61.39 &  { [60.71-62.10]} & 7.97  & { [7.92-8.03]} \\
\cdashline{1-6}
\multirow{3}{*}{\bf Unsupervised} & \topn{5} (Unsupervised) & 45.20  & { [44.41-46.04]} & 8.94 &  { [8.89-8.98]} \\
& \topedit{5} & 50.74 &  { [49.89-51.53]} & 8.63 &  { [8.57-8.68]} \\
& \topeditpara{5}& 60.07 &  { [59.37-60.77]} & 8.15 &  { [8.09-8.20]} \\
\cdashline{1-6}
& $\mathrm{MTSum}^3$ \scriptsize \citeyearpar{atanasova-etal-2020-generating} & 48.73 &  { [47.81-49.66]} & 8.88  & { [8.82-8.94]} \\
& Justification & 49.29 &  { [48.27-50.40]} & 9.17  & { [9.08-9.26]} \\
\bottomrule
\end{tabular}
\caption{Readability measures (\S\ref{sec:result:readability}) over the test splits. Readability measures include 95\% confidence intervals based on 1000 random re-samples from the corresponding split (\S\ref{sec:result:readability}, Metrics.). We report results reported in MTSum, where we have the outputs to compute readability. We also report results given the gold explanation -- Justification. Readability scores for \topedit{N} and  \topedit{N}+Para are statistically significant ($p$<0.05) compared to \topn{N}, and to $\mathrm{MTSum}$, except for the score in \textcolor{purple}{purple}.}
\label{tab:results:readability:full}
\end{table*}

\begin{table*}[t]
% \centering
\fontsize{9}{9}\selectfont
\begin{tabular}{llrrrrrr}
\toprule
& \bf Method & \textbf{R-1}$\nearrow$ & \textbf{R-1 CI}  & \textbf{R-2}$\nearrow$ & \textbf{R-2 CI} &  \textbf{R-L}$\nearrow$ & \textbf{R-L CI}  \\
\midrule
\multicolumn{8}{c}{\bf LIAR-PLUS} \\ 
\multirow{2}{*}{\bf Baselines} & \leadn{4} & \textit{28.11} & {\small [27.39-28.39]}& \textit{6.96} &{\small  [6.52\hspace{1.25mm}-\hspace{1.25mm}7.33]}& \textit{24.38} & {\small [23.73-24.68]}  \\
& \leadn{6} &  29.15 & {\small [28.66-29.69]} & 8.28 & {\small [7.85\hspace{1.25mm}-\hspace{1.25mm}8.67]} & 25.84 & {\small [25.35-26.30]} \\
\cdashline{1-6}
\multirow{3}{*}{\bf Sup.} & \topn{6} (Supervised) & 34.42 & {\small [33.78-35.00]}&12.36 & {\small [11.85-12.84]} & 30.58 & {\small [30.01-31.13]}\\
& \topedit{6} & 33.92 & {\small [33.31-34.53]} & 11.73 & {\small [11.29-12.24]} & 30.01 & {\small [29.43-30.60]} \\
& \topeditpara{6} & 33.94 &{\small  [33.37-34.47] }& \underline{11.25} & {\small [10.81-11.73] }& 30.08 & {\small [29.49-30.59] }\\
\cdashline{1-6}
\multirow{3}{*}{\bf Unsup.} & \topn{6} (Unsupervised) & 29.63 & {\small [29.03-30.07] }& 7.58 &{\small  [7.53\hspace{1.25mm}-\hspace{1.25mm}8.25]} & 25.86 &{\small  [25.52-26.47]} \\
& \topedit{6} & 28.93 & {\small [28.42-29.42]} & \underline{7.06} & {\small [6.74\hspace{1.25mm}-\hspace{1.25mm}7.43]} & 25.14 & {\small [24.69-25.61]} \\
& \topeditpara{6} & 28.98 & {\small [28.50-29.52]}& \underline{6.84} & {\small [6.51\hspace{1.25mm}-\hspace{1.25mm}7.16]} &  25.39 & {\small [24.95-25.87]}\\
\cdashline{1-6}
 & $\mathrm{MTSum}^4$ \scriptsize \citeyearpar{atanasova-etal-2020-generating} & \textit{35.70} & {\small [34.23-35.39]} & \textit{13.51} & {\small [12.47-13.67]} & \textit{31.58} & {\small [30.07-31.21]} \\
\midrule

\multicolumn{8}{c}{\bf PubHealth} \\
\multirow{3}{*}{\bf Baselines} & \leadn{3} & \textit{29.01} & - & \textit{10.24} & - & \textit{24.18} & - \\
& \leadn{3} & 23.05 & {\small [22.53-23.59]} & 6.28 & {\small [5.48\hspace{1.25mm}-\hspace{1.25mm}6.37]} & 19.27 & {\small [18.42-19.40]} \\
& \leadn{5} & 23.73 & {\small [22.50-23.62]} & 6.86 & {\small [5.81\hspace{1.25mm}-\hspace{1.25mm}6.58]} & 20.67 & {\small [19.08-20.18]}\\
\cdashline{1-8}
\multirow{3}{*}{\bf Sup.} & \topn{5} (Supervised) & 29.93 & {\small [28.87-30.97]} & 12.42 & {\small [11.44-13.63]} & 26.24 & {\small [25.21-27.44]}  \\
& \topedit{5} & 29.38 & {\small [28.45-30.33]} & 11.16 & {\small [10.17-12.15]} & 25.41 & {\small [24.48-26.41]} \\
& \topeditpara{5}& 28.40 & {\small [27.55-29.17]} & \underline{9.56} & {\small [8.89\hspace{0.7mm}-\hspace{0.7mm}10.23]} & \underline{24.37} &{\small  [23.52-25.10]} \\
\cdashline{1-8}
\multirow{3}{*}{\bf Unsup.} & \topn{5} (Unsupervised) & 23.52 & {\small [22.95-24.12]} & 6.12 & {\small [5.76\hspace{1.25mm}-\hspace{1.25mm}6.46]}& 19.93 &{\small  [19.37-20.45]} \\
% $Edits^{5}$ - UnSupervised & 18.09 & 4.41 & 15.48 & 42.29 {\scriptsize $\pm$17.34} & 7.36{\scriptsize $\pm$ 0.80 } \\
& \topedit{5} & 23.09 & {\small [22.55-23.64]} & 5.56 & {\small [5.24\hspace{1.25mm}-\hspace{1.25mm}5.92]} & 19.44 &{\small  [18.93-19.93]}  \\
& \topeditpara{5}& 23.35 & {\small [22.85-23.86] }& \underline{5.38} & {\small [5.09\hspace{1.25mm}-\hspace{1.25mm}5.71]} & 19.56 & {\small [19.08-20.03]} \\
\cdashline{1-8}
&$\mathrm{AbstrSum^3}$\ \scriptsize \citeyearpar{kotonya-toni-2020-explainable} & \textit{32.30} & - & \textit{13.46} & - & \textit{26.99} & -\\
&$\mathrm{MTSum}^3$ \scriptsize \citeyearpar{atanasova-etal-2020-generating} & 33.55 & {\small [29.79-31.65]} & 13.12 &{\small  [11.17-13.42]} & 29.41 &{\small  [25.27-27.31]}\\
\bottomrule
\end{tabular}
\caption{ROUGE-1/2/L $F_1$ scores (\S\ref{sec:result:rouge}) of supervised (Sup.) and usupervised (Unsup.) methods over the test splits. In \textit{italics}, we report results reported from prior work, where we do not always have the outputs to compute the confidence intervals. \underline{Underlined} ROUGE scores of the \topedit{N} and \topedit{N}+Para are statistically significant ($p<0.05$) compared to the input \topn{N} ROUGE scores, N=\{5,6\}.}
\label{tab:results:rouge:full}
\end{table*}

% \bgroup
% \def\arraystretch{1.2}

\begin{table*}[t]
\centering
\fontsize{10}{10}\selectfont
\begin{tabular}{llrrrr}
\toprule
& \textbf{Method} &  {\textbf{\small Flesch $\nearrow$}} &  {\textbf{\small Flesch  CI}} & {\textbf{\small DC $\searrow$}}  & {\textbf{\small DC CI}}  \\
\midrule
\multicolumn{6}{c}{\textbf{LIAR-PLUS}}\\ 
\multirow{2}{*}{\textbf{Baselines}} & \leadn{4} & 50.89 & {\small [50.01-51.63]} & 8.75 & {\small  [8.71-8.80]} \\
& \leadn{6} &  53.01 &  {\small  [52.41-53.64]} & 8.43  & {\small  [8.39-8.47]} \\
\cdashline{1-6}
\multirow{3}{*}{\textbf{Supervised}} & \topn{6} (Supervised) & 57.77 &  {\small[57.15-58.38]} & 7.91 &  {\small[7.87-7.95]} \\
& \topn{6}+Par & 63.88 &  {\small[63.31-64.45]} & 7.55 & {\small [7.51-7.59]} \\
& \topn{6}+$\mathrm{Edits^{IE}}$ & 55.70 & {\small[55.03-56.36]} & 6.53 & {\small[6.50-6.56]}\\
& \topn{6}+$\mathrm{Edits^{IE}}$+$\mathrm{Edits^{Gram}}$ & 59.52 &  {\small [58.89-60.17]} & 7.78  &   {\small[7.73-7.83]} \\
& \topn{6}+$\mathrm{Edits^{IE}}$+$\mathrm{Edits^{Gram}}$+$\mathrm{Par}$ & 66.05 &  {\small [65.53-66.61]} & 7.46  & {\small [7.41-7.50]} \\
\cdashline{1-6}
\multirow{3}{*}{\textbf{Unsupervised}} & \topn{6} (Unsupervised) & 52.84  & {\small [52.27-53.36]} & 8.52 &  {\small [8.48-8.55]}  \\
& \topn{6}+Par & 50.92 &  {\small [50.18-51.58]} &  6.97 &  {\small[6.94-7.01]} \\
& \topn{6}+$\mathrm{Edits^{IE}}$ & 50.70 & {\small [50.13-51.27]} & 6.92 &  {\small[6.89-6.94]} \\
& \topn{6}+$\mathrm{Edits^{IE}}$+$\mathrm{Edits^{Gram}}$ & 54.76 &  {\small [54.15-55.34]} & 8.39 &  {\small [8.34-8.43]} \\
& \topn{6}+$\mathrm{Edits^{IE}}$+$\mathrm{Edits^{Gram}}$+$\mathrm{Par}$ & 61.80 &  {\small [61.17-62.42]} & 8.01 &  {\small [7.97-8.05]} \\
\cdashline{1-6}
 & $\mathrm{MTSum}^4$ \scriptsize \citeyearpar{atanasova-etal-2020-generating} & 58.08 &  {\small[57.33-58.83]} & 8.03 & {\small [7.97-8.08]}\\
& Justification & 58.90 & {\small [58.23-59.68]} & 8.26 & {\small [8.20-8.32]}\\
\midrule
\multicolumn{6}{c}{\textbf{LIAR-PLUS Test Split Ablation}} \\
\multirow{2}{*}{\textbf{Supervised}} & \topn{6}+Par & 64.45 &  [63.81-65.04] & 7.52 &  [7.48-7.56] \\
& \topn{6}+$\mathrm{Edits^{IE}}$ & 56.26 & {\small [55.37-57.04]} & 6.51 &  {\small [6.48-6.55]} \\
\cdashline{1-6}
\multirow{2}{*}{\textbf{Unsupervised}} & \topn{6}+Par & 59.83 & {\small [59.20-60.36]} & 8.21 & {\small [8.16-8.25]} \\
& \topn{6}+$\mathrm{Edits^{IE}}$ & 59.34 & {\small [58.72-59.91]} & 8.14 & {\small [8.10-8.18]} \\
\midrule
\multicolumn{6}{c}{\textbf{PubHealth}} \\
& \leadn{3} & 44.76 & {\small [43.49-45.87]} & 9.12 & {\small [9.05-9.20]} \\
& \leadn{5} & 46.00 & {\small [44.80-46.92]} & 8.88 & {\small [8.83-8.94]} \\
\cdashline{1-6}
\multirow{3}{*}{\textbf{Supervised}} & \topn{5} (Supervised) & 49.56 &  {\small[48.73-50.27]} & 8.63 & {\small [8.58-8.68]}  \\
& \topn{5}+Par & 47.38 & {\small [46.50-48.15]} & 7.07 & {\small [7.04-7.12]}\\
& \topn{5}+$\mathrm{Edits^{IE}}$ & 57.53 & {\small [56.85-58.19]} & 8.18 & {\small [8.13-8.24]}\\
& \topn{5}+$\mathrm{Edits^{IE}}$+$\mathrm{Edits^{Gram}}$ & 54.30 &  {\small [53.58-54.97]} & 8.33 & {\small [8.27-8.38]} \\
& \topn{5}+$\mathrm{Edits^{IE}}$+$\mathrm{Edits^{Gram}}$+$\mathrm{Par}$& 61.51 & {\small [60.89-62.19]} & 7.96 & {\small  [7.91-8.01]} \\
\cdashline{1-6}
\multirow{3}{*}{\textbf{Unsupervised}} & \topn{5} (Unsupervised) & 43.55 &  {\small[42.51-44.52]} & 9.26 &  {\small[9.19-9.32] }\\
& \topn{5}+Par & 42.70 &  {\small[41.60-43.59]} & 7.35 & {\small [7.31-7.40]} \\
& \topn{5}+$\mathrm{Edits^{IE}}$ & 56.33 & {\small [55.68-56.97]} & 8.35 & {\small [8.31-8.40]} \\
& \topn{5}+$\mathrm{Edits^{IE}}$+$\mathrm{Edits^{Gram}}$ & 50.46 & {\small [49.62-51.23]} & 8.65 &  {\small[8.59-8.70]} \\
& \topn{5}+$\mathrm{Edits^{IE}}$+$\mathrm{Edits^{Gram}}$+$\mathrm{Par}$& 60.25 & {\small [59.56-60.89]} & 8.13 &  {\small[8.08-8.19]} \\
\cdashline{1-6}
& $\mathrm{MTSum}^3$ \scriptsize \citeyearpar{atanasova-etal-2020-generating} & 49.69 & {\small [48.73-50.53]} & 8.81 & {\small [8.75-8.88]} \\
& Justification & 48.20 &  {\small[47.25-49.16]} & 9.22 &  {\small[9.15-9.32]} \\

\midrule
\multicolumn{6}{c}{\textbf{PubHealth - Test Split Ablation}} \\
\multirow{2}{*}{\textbf{Supervised}} & \topn{5}+Par & 46.23 & {\small [45.33-47.07]} & 7.11 & {\small [7.07-7.15]} \\
& \topn{5}+$\mathrm{Edits^{IE}}$ & 57.29 &  {\small[56.58-57.96]} & 8.21 &  {\small [8.16-8.26] } \\
\cdashline{1-6}
\multirow{2}{*}{\textbf{Unsupervised}} & \topn{5}+Par & 42.30 & {\small [41.34-43.28]} & 7.36 &  {\small [7.31-7.41]} \\
& \topn{5}+$\mathrm{Edits^{IE}}$ & 55.79 & {\small [55.16-56.44]} & 8.39 & {\small [8.34-8.43]} \\
\bottomrule
\end{tabular}
\caption{Readability measures (\S\ref{sec:result:readability}) over the validation splits. Readability measures include 95\% confidence intervals (\S\ref{sec:result:readability}, Metrics.). We report results from prior work -- $\mathrm{MTSum}$, where we have the outputs to compute readability.}
\label{tab:results:readability:val}
\end{table*}

\begin{table*}[t]
% \centering
\fontsize{9}{9}\selectfont
\begin{tabular}{llrrrrrr}
\toprule
& \textbf{Method} & \textbf{R-1}$\nearrow$ & \textbf{R-1 CI}  & \textbf{R-2}$\nearrow$ & \textbf{R-2 CI} &  \textbf{R-L}$\nearrow$ & \textbf{R-L CI}  \\
\midrule
\multicolumn{8}{c}{\textbf{LIAR-PLUS}}\\ 
\multirow{2}{*}{\textbf{Base.}} & \leadn{4} & 27.52 & {\small [26.99-28.00]} & 6.90 & {\small [6.54-7.30]} & 24.01 & {\small [23.53-24.49}] \\
& \leadn{6} & 28.93 & {\small [28.39-29.43]} & 8.32 & {\small [7.92-8.76]} & 25.67 & {\small [25.15-26.16]} \\
\cdashline{1-8}
\multirow{3}{*}{\textbf{Sup.}} & \topn{6} (Sup.) & 34.35 & {\small [33.71-34.94]} & 12.20 & {\small [11.72-12.70]} & 30.51 & {\small [29.97-31.09]}\\
& \topn{6}+Par & 34.51 & {\small[33.90-35.08]} & 11.49 & {\small[11.04-11.96]} & 30.68 & {\small[30.15-31.25]}\\
& \topn{6}+$\mathrm{Edits^{IE}}$ & 25.18 & {\small[24.63-25.72]} & 8.60 & {\small[8.23-8.98]} & 22.08 & {\small[21.58-22.55]}\\
& \topn{6}+$\mathrm{Edits^{IE}}$+$\mathrm{Edits^{Gram}}$ & 34.07 & {\small [33.45-34.71]} & 11.58 & {\small [11.14-12.05]} & 30.12 & {\small [29.55-30.70]} \\
& \topeditpara{6} & 34.20 & {\small [33.62-34.78]} & 11.05 & {\small [10.59-11.46]} & 30.26 & {\small [29.71-30.83]}\\
\cdashline{1-8}
\multirow{3}{*}{\textbf{Uns.}} & \topn{6} (Uns.) & 29.29 & {\small [28.79-29.78]} & 7.99 & {\small [7.64-8.37]} & 25.84 & {\small [25.36-26.30]} \\
& \topn{6}+Par & 22.74 & {\small[22.31-23.24]} & 5.56 & {\small[5.29-5.82]} & 19.50 & {\small[19.06-19.93]}\\
& \topn{6}+$\mathrm{Edits^{IE}}$ & 21.46 & {\small[20.98-21.93]} & 5.67 & {\small[5.43-5.93]} & 18.76 & {\small[18.34-19.20]}\\
& \topedit{6}+$\mathrm{Edits^{Gram}}$ & 29.02 & {\small [28.50-29.53]} & 7.47 & {\small [7.10-7.83]} & 25.52 & {\small [25.02-25.97]} \\
& \topeditpara{6} & 29.41 & {\small [28.89-29.90]} & 7.26 & {\small [6.90-7.60]} & 25.90 & {\small [25.43-26.37]}\\
\cdashline{1-8}
 & $\mathrm{MTSum}^4$ \scriptsize \citeyearpar{atanasova-etal-2020-generating} & 34.80 & {\small [34.13-35.39]} & 12.87 & {\small [12.29-13.46]} & 30.66 & {\small [30.08-31.26]} \\
\midrule
\multicolumn{8}{c}{\textbf{LIAR-PLUS Test Split Ablation}}\\ 
\multirow{2}{*}{\textbf{Sup.}} & \topn{6}+Para & 34.62 & {\small[34.02-35.23]} & 11.79 & {\small[11.33-12.21]} & 30.81 & {\small[30.26-31.35]}\\
& \topn{6}+$\mathrm{Edits^{IE}}$ & 25.48 & {\small[25.00-26.03]} & 8.75 & {\small[8.41-9.09]} & 22.29 & {\small[21.78-22.79]}\\
\cdashline{1-8}
\multirow{2}{*}{\textbf{Uns.}} & \topn{6}+Para & 34.61 & {\small[34.08-35.19]} & 11.77 & {\small[11.30-12.26]} & 30.81 & {\small[30.20-31.38]}\\
& \topn{6}+$\mathrm{Edits^{IE}}$ & 25.48 & {\small[24.87-26.01]} & 8.75 & {\small[8.41-9.13]} & 22.29 & {\small[21.81-22.76]}\\
\midrule

\multicolumn{8}{c}{\textbf{PubHealth}} \\
\multirow{2}{*}{\textbf{Base.}}& \leadn{3} & 23.18 & {\small [22.69-23.71]} & 5.55 & {\small [5.14-6.00]} & 18.74 & {\small [18.31-19.19]} \\
& \leadn{5} & 23.31 & {\small [22.71-23.90]} & 6.07 & {\small [5.70-6.46]} & 19.60 & {\small [19.04-20.14]}\\
\cdashline{1-8}
\multirow{3}{*}{\textbf{Sup.}} & \topn{5} (Sup.) & 30.37 & {\small [29.40-31.42]} & 12.64 & {\small [11.51-13.77]} & 26.46 & {\small [25.42-27.49]} \\
& \topn{5}+Par & 22.49 & {\small[21.62-23.39]} & 8.96 & {\small[8.22-9.75]} & 19.73 & {\small[18.82-20.64]}
\\
& \topn{5}+$\mathrm{Edits^{IE}}$ & 29.76 & {\small[28.81-30.70]} & 10.75 & {\small[9.91-11.63]} & 25.44 & {\small[24.56-26.32]}
\\
& \topedit{5}+$\mathrm{Edits^{Gram}}$ & 29.62 & {\small [28.69-30.49]} & 11.20 & {\small [10.31-12.19]} & 25.54 & {\small [24.62-26.46]} \\
& \topeditpara{5}& 28.81 & {\small [27.97-29.70]} & 9.67 & {\small [8.94-10.37]} & 24.47 & {\small [23.71-25.31]}\\
\cdashline{1-8}
\multirow{3}{*}{\textbf{Uns.}} & \topn{5} (Uns.) & 23.80 & {\small [23.27-24.31]} & 5.76 & {\small [5.36-6.13]} & 19.24 & {\small [18.74-19.70]} \\
& \topn{5}+Par & 18.28 & {\small[17.64-18.88]} & 4.50 & {\small[4.22-4.79]} & 15.50 & {\small[14.90-16.14]}
\\
& \topn{5}+$\mathrm{Edits^{IE}}$ & 24.44 & {\small[23.85-25.04]} & 5.97 & {\small[5.67-6.31]} & 20.51 & {\small[19.98-21.01]}
\\
& \topedit{5}+$\mathrm{Edits^{Gram}}$ & 23.74 & {\small [23.20-24.28]} & 5.72 & {\small [5.41-6.03]} & 19.76 & {\small [19.27-20.30]} \\
& \topeditpara{5}& 23.96 & {\small [23.49-24.49]} & 5.46 & {\small [5.17-5.78]} & 19.98 & {\small [19.52-20.45]} \\
\cdashline{1-8}
&$\mathrm{MTSum}^3$ \scriptsize \citeyearpar{atanasova-etal-2020-generating} & 31.05 & {\small [30.08-32.09]} & 12.66 & {\small [11.44-13.82]} & 26.45 & {\small [25.46-27.50]}\\

\midrule

\multicolumn{8}{c}{\textbf{PubHealth Test Split Ablation}}\\ 
\multirow{2}{*}{\textbf{Sup.}} & \topn{5}+Par & 22.09 & {\small[21.24-22.95]} & 8.75 & {\small[7.97-9.54]} & 19.48 & {\small[18.60-20.36]}
\\
& \topn{5}+$\mathrm{Edits^{IE}}$ & 29.46 & {\small[28.57-30.37]} & 10.71 & {\small[9.86-11.63]} & 25.53 & {\small[24.67-26.40]}
\\
\cdashline{1-8}
\multirow{2}{*}{\textbf{Uns.}} & \topn{5}+Par & 18.11 & {\small[17.44-18.77]} & 4.41 & {\small[4.16-4.70]} & 15.48 & {\small[14.92-16.10]}
\\
& \topn{5}+$\mathrm{Edits^{IE}}$ & 18.11 & {\small[17.44-18.77]} & 4.41 & {\small[4.16-4.70]} & 15.48 & {\small[14.92-16.10]}
\\

\bottomrule
\end{tabular}
\caption{ROUGE-1/2/L $F_1$ scores (\S\ref{sec:result:rouge}) of baselines (Base.), supervised (Sup.) and usupervised (Uns.) methods over the validation splits. In \textit{italics}, we report results reported from prior work, where we do not always have the outputs to compute the confidence intervals.}
\label{tab:results:rouge:val}
\end{table*}

\subsection{Case Studies}

\begin{table*}[t]
% \fontsize{8}{8}\selectfont
\scriptsize
\centering
\begin{tabular}{p{430pt}}%{p{206pt}}
\toprule
\textbf{\topn{5}:} Heavily-armed Muslims shouting “Allahu Akbar” open fire \myul[green_editing]{campers and hikers} in a park. A heavily armed group of Middle Eastern looking Muslim men was arrested \myul[green_editing]{outside Los Angeles} after opening fire upon hikers and campers in a large State Park \myul[orange]{in the area}. There was no evidence found that a crime had been committed by any of the subjects \myul[orange]{who were detained and they were released}. Also, the police report described the men only as “\myul[green_editing]{males},” not “Middle Eastern \myul[orange]{males}” or “Muslim \myul[orange]{males}.” The website that started this rumor was Superstation95, which is not a “superstation” at all but rather a repository of misinformation from Hal Turner, who in 2010 was sentenced to 33 months in prison for making death threats against three federal judges. No credible news reports made any mention of the “Allahu Akbar” claim, and no witnesses stated they had been “shot at” \myul[orange]{by the men while hiking or camping.}
 \\
\hline
\textbf{\topedit{5}:} \myul[blue_editing]{Heavily-armed Muslims} \textbf{\textcolor{green_editing}{\myul[blue_editing]{males}}} \myul[blue_editing]{shouting} “Allahu Akbar” open fire in a park. A heavily armed group of Middle Eastern looking Muslim men was arrested after opening fire upon hikers and campers in a large State Park \textbf{\textcolor{green_editing}{outside Los Angeles}}. There was no evidence found that a crime had been committed \myul[blue_editing]{by any of the subjects on}
\textbf{\textcolor{green_editing}{\myul[blue_editing]{campers and hikers}}}. \myul[orange]{Also, the police report described the men only as ``,'' not ``Middle Eastern'' or ``Muslim.''} The website that started this rumor was Superstation95, which is not a “superstation” at all but rather a repository of misinformation from Hal Turner, who in 2010 was sentenced to 33 months in prison \myul[blue_editing]{for making death threats against three federal judges}. \myul[blue_editing]{No credible news reports made any mention of the ``Allahu Akbar'' claim, and no witnesses stated they had been ``shot at''.}
 \\
\hline
\textbf{\topeditpara{5}} \textbf{\textcolor{blue_editing}{Muslims shout}} "Allahu Akbar" open fire in a park. A heavily armed group of Middle Eastern looking Muslim men was arrested after opening fire on hikers and campers in a large State Park outside Los Angeles. There was no evidence that a crime had been committed by \textbf{\textcolor{blue_editing}{any of the campers or hikers}}. The website that started this rumor was Superstation95, which is not a “superstation” at all but rather a repository of misinformation from Hal Turner, who in 2010 was sentenced to 33 months in prison. \textbf{\textcolor{blue_editing}{There were no credible news reports that mentioned the Allahu Akbar claim, and no witnesses that said they had been shot at.}}\\
\hline
\textbf{Original Explanation:} Secondary reporting claiming that Muslim men fired upon hikers (and that the media covered it up) appeared on a site that had previously inaccurately claimed Illinois had applied Sharia law to driver’s licenses, that Target introduced “Sharia-compliant” checkout lanes, and that Muslims successfully banned Halloween at a New Jersey school. \\
\hline
\textbf{Claim:} The media covered up an incident in San Bernardino during which several Muslim men fired upon a number of Californian hikers. \textbf{Label:} False\\
\midrule
\textbf{\topn{5}}: The article claims the CDC might have to stop calling COVID-19 an epidemic because the death rate is becoming so low that it wouldn't meet the CDC's definition of epidemic. The latest CDC statement \myul[orange]{made public when the Facebook post was made said deaths attributed to COVID-19 decreased} from the previous week, but \myul[orange]{remained at the epidemic threshold, and were likely to increase}. Moreover death rates \myul[orange]{alone} do not define an epidemic. Amid news headlines that the United States set a daily record for the number of new coronavirus cases,\myul[orange]{ an article widely shared on Facebook made a contrarian claim}. The CDC page says: ``Epidemic refers to an increase, often sudden, in the number of cases of a disease above what is normally expected \myul[green_editing]{in that population} in that area.'' \\
\hline
\textbf{\topedit{5}:} \myul[blue_editing]{The article claims} the CDC might have to stop calling COVID-19 an epidemic \textbf{\textcolor{green_editing}{\myul[blue_editing]{in that population}}} because the death rate is \myul[blue_editing]{becoming} so low that it wouldn't meet \myul[blue_editing]{the CDC's} definition of epidemic. \myul[blue_editing]{The latest CDC statement an article from the previous week said deaths decreased, but.} \myul[blue_editing]{Moreover,} death rates do not define an epidemic. \myul[orange]{Amid news headlines that the United States} \textbf{\textcolor{green_editing}{\myul[orange]{on Facebook}}} \myul[orange]{set a daily record for the number of new coronavirus cases,} The CDC page \myul[blue_editing]{ when the Facebook post was made} says: ``Epidemic refers to an increase, often sudden, in the number of cases of a disease attributed to COVID-19.'' \\
\hline
\textbf{\topeditpara{5}:}
\textbf{\textcolor{blue_editing}{According to the article}}, the CDC might have to stop calling COVID-19 an epidemic because the death rate is so low that it wouldn't meet \textbf{\textcolor{blue_editing}{their}} definition of an epidemic. \textbf{\textcolor{blue_editing}{An article from the previous week said deaths decreased, but that's what the latest CDC statement says.}} Death rates do not define an epidemic. The CDC's page on Facebook says Epidemic refers to an increase, often sudden, in the number of cases of a disease attributed to COVID-19. \\
\hline
\textbf{Original Explanation:}
Despite a dip in death rates, which are expected to rise again, the federal Centers for Disease Control and Prevention still considers COVID-19 an epidemic. Death rates alone don’t determine whether an outbreak is an epidemic. \\
\hline 
\textbf{Claim:} The CDC may have to stop calling COVID-19 an ‘epidemic’ due to a remarkably low death rate. \textbf{Label:} False \\
\bottomrule
\end{tabular}
\caption{Example explanations -- extracted \topn{5} RCs, the iterative editing, and the latter with paraphrasing on top, taken from the test split of PubHealth. Each color designates an edit operation -- \textbf{\textcolor{green_editing}{reordering}}, \textbf{\textcolor{orange}{deletion}}, and \textbf{\textcolor{blue_editing}{paraphrasing}}. The underlining designates the position in the text where the corresponding operation will be applied in the next step -- post-editing and paraphrasing.} 
%\textbf{\textcolor{purple}{insertion}}
\label{tab:qualexamples}
\end{table*}
Table~\ref{tab:qualexamples} presents a case study from the PubHealth dataset. Overall, the initial extracted RC sentences are transformed to be more concise, fluent and human-readable by applying the iterative post-editing algorithm followed by paraphrasing. We can also see that compared to the original explanation, the post-edited explanations contain words that do not change the semantics of the explanation, but would not be scored as correct according to the ROUGE scores. For example, in the second instance, ``Death rates do not define an epidemic'' in the post-edited explanation and ``Death rates alone don’t determine whether an outbreak is an epidemic'' from the original explanation express the same meaning, but contain both paraphrases and words without added meaning that would decrease the final ROUGE scores. Finally, compared to the original explanation, the post-edited explanations for both instances have preserved the information needed for fact checking. 
% \bgroup
% \def\arraystretch{1}

\chapter{Multi-Hop Fact Checking of Political Claims}
\label{chap:multihop}

\section{Introduction}
\begin{figure}[t]
    \centering
    \includegraphics[scale=1]{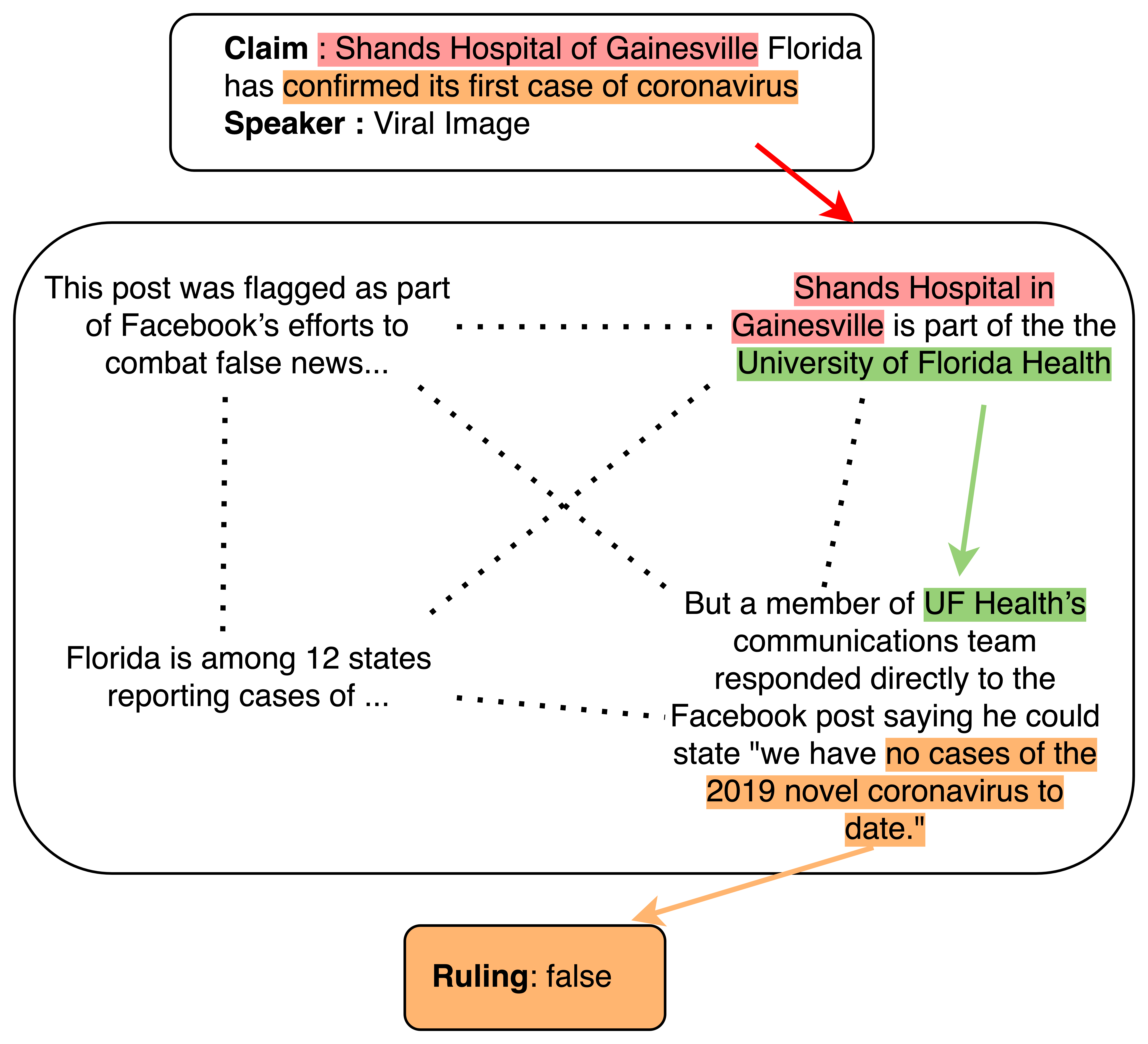}
    \caption{An illustration of multiple hops over an instance from \politihop. Each instance consists of a claim, a speaker% (author of the claim)%
    , a veracity label, and a PolitiFact article the annotated evidence sentences. The highlighted sentences represent the evidence sentences a model needs to connect to arrive at the correct veracity prediction.}
    \label{fig:example_4}
\end{figure}
\noindent Recent progress in machine learning has seen interest in automating complex reasoning, where a conclusion can be reached only after following logically connected arguments.
To this end, multi-hop datasets and models have been introduced, which learn to combine information from several sentences to arrive at an answer. While most of them concentrate on question answering, fact checking is another task that often requires a combination of multiple evidence pieces to predict a claim's veracity. 

Existing fact checking models usually optimize only the veracity prediction objective and assume that the task requires a single inference step. Such models ignore that often several linked evidence chunks have to be explicitly retrieved and combined to make the correct veracity prediction. Moreover, they do not provide explanations of their decision-making, which is an essential part of fact checking.

\citet{atanasova-etal-2020-generating-fact} note the importance of providing explanations for fact checking verdicts, and propose an extractive summarization model, which optimizes a ROUGE score metric w.r.t. a gold explanation. 
Gold explanations for this are obtained from the LIAR-PLUS~\cite{alhindi-etal-2018-evidence} dataset, which is constructed from PolitiFact\footnote{\url{https://www.politifact.com/}} articles written by professional fact checking journalists.
However, the dataset does not provide guidance on the several relevant evidence pieces that have to be linked and assume that the explanation requires a single reasoning step. FEVER~\cite{thorne-etal-2018-fever} is another fact checking dataset, which contains annotations of evidence sentences from Wikipedia pages. However, it consists of manually augmented claims, which require limited reasoning capabilities for verification as the evidence mostly consists of one or two sentences.

To provide guidance for the multi-hop reasoning process of a claim's verification and facilitate progress on explainable fact checking, we introduce \politihop, a dataset of 500 real-world claims with manual annotations of sets of interlinked evidence chunks from PolitiFact articles needed to predict the claims' labels. We provide insights from the annotation process, indicating that fact checking real-world claims is an elaborate process requiring multiple hops over evidence chunks, where multiple evidence sets are also possible.

To assess the difficulty of the task, we conduct experiments with lexical baselines, as well as a single-inference step model -- BERT~\cite{devlin-etal-2019-bert}, and a multi-hop model -- Transformer-XH~\cite{zhao2020transformer-xh}. Transformer-XH allows for the sharing of information between sentences located anywhere in the document by eXtra Hop attention and achieves the best performance.
% We show that a multi-hop model indeed performs better for multi-hop fact-checking with real claims compared to a single-step inference model. Pre-trained on the LIAR-PLUS dataset and fine-tuned on \politihop, Transformer-XH with three hop layers achieves a FEVER score of 24.5 \cite{fever}, 57.3 in Macro F1 label prediction and an F1 of 47.2 F1 for evidence retrieval, which still leaves plenty of room for improvement in future work.
We further study whether multi-hop reasoning learned with Transformer-XH can be transferred to \politihop. We find that the model cannot leverage any reasoning skills from training on FEVER, while training on LIAR-PLUS improves the performance on \politihop. We hypothesize that this is partly due to a domain discrepancy, as FEVER is constructed from Wikipedia and consists of claims requiring only one or two hops for verification. In contrast, LIAR-PLUS is based on PolitiFact, same as \politihop.

Finally, we perform a detailed error analysis to understand the models' shortcomings and recognize possible areas for improvement. We find that the models perform worse when the gold evidence sets are larger and that, surprisingly, named entity (NE) overlap between evidence and non-evidence sentences does not have a negative effect on either evidence retrieval or label prediction. The best results for Transformer-XH on the dev and test sets are for a different number of hops -- 2 and 6, indicating that having a fixed parameter for the number of hops is a downside of Transformer-XH; this should instead be learned for each claim. Overall, our experiments constitute a solid basis to be used for future developments.% during training.

%To address the aforementioned problems and to facilitate the progress on explainable, multi-hop reasoning fact-checking, we introduce MultiFC, a dataset for multi-hop fact-checking. The dataset contains 500 annotated PolitiFact articles. Each annotation consists of the claim veracity label (either `true', `half-true' or `false') and evidence sentences. Each sentence in the article is annotated as either evidence or not evidence. Furthermore, the evidence sentences are grouped into chains of reasoning, where a chain is a sequence of sentences which provides an argument for or against a claim and aim to provide enough context to form a solid explanation of the verdict. These chains require multi-hop reasoning to be successfully retrieved.

%In this paper we describe our annotation process and the main issues it created. Our inter-annotation agreement studies suggest that the task of multi-hop, explainable fact-checking is challenging even for human annotators. %Next, we present the main statistics of the dataset.

%We compare several baselines to prove that multi-hop models are favourable on our dataset. As the multi-hop baseline we use Transformer-XH \cite{trans-xh}. %We also present our ablation studies and error analysis. 
%Our results show that there is ample space for improvement.

To summarise, our \textbf{contributions} are as follows:
\begin{itemize}%[nosep]
\item We document the first study on multi-hop fact checking of political claims.
\item We create a dataset, \politihop, for the task.
\item We study whether reasoning skills learned with a multi-hop model on similar datasets can be transferred to \politihop.
\item We analyze to what degree existing multi-hop reasoning methods are suitable for the task.
\end{itemize}

\section{Multi-Hop Fact Checking}
%Frame the problem here formally and describe what's challenging about it and different from multi-hop reasoning.
%The formal definition should say, on a high level: how the input is structured, how the output is structured, and what high-level function the model takes.

%This works best by pointing to a figure with an example, which can already appear on Page 1.

A multi-hop fact checking model $f(X)$ receives as an input $X=\{(claim_i, document_i)\\| i\in[1,|X|]\}$, where $document_i = [sentence_{ij}|j \in [1, |document_i|]]$ is the corresponding PolitiFact article for $claim_i$ and consists of a list of sentences. During the training process, the model learns to (i) select which sentences from the input contain evidence needed for the veracity prediction $y_i^S = [y_{ij}^S \in \{0, 1\}|j \in [1, |document_i|]]$ (\textbf{sentence selection task}), where 1 indicates that the sentence is selected as an evidence sentence; and (ii) predict the veracity label of the claim $y_i^L \in \{True, False, Half-True\}$, based on the extracted evidence (\textbf{veracity prediction task}). The sentences selected by the model as evidence provide \textit{sufficient} explanation, which allows to verify the corresponding claim \textit{efficiently} instead of reading the whole article. Each evidence set consists of $k$ sentences, where $k \in [1, max_{i\in[1,|X|]}(|document_i|)]$ is a hyper-parameter of the model. Figure~\ref{fig:example_4} illustrates the process of multi-hop fact checking, where multiple evidence sentences provide different information, which needs to be connected in logical order to reach the final veracity verdict.

\begin{table}%[h!]
% \footnotesize
\centering
\begin{tabular}{p{8cm}rr}
\toprule
\bf Statistic & \bf Test & \bf Train\\
\midrule
\#Words per article & 569 (280.8) & 573 (269.1)\\
\#Sent. per article & 28 (12.8) & 28 (12.8)\\
\#Evidence sent. per article & 11.75 (5.56) & 6.33 (2.98)\\
\#Evidence sent. per set & 2.88 (1.43) & 2.59 (1.51)\\
% #Evidence sentences with an URL & 7.9 (3.89) & 7.7 (4.24)\\
% #sentences with matching text & 2.44 (2.26) & 1.9 (1.92)\\
\#Sets per article & 4.08 (1.83) & 2.44 (1.28)\\
\midrule
\multicolumn{3}{c}{\textbf{Label Distribution}}\\
False & 149 & 216\\
Half-true & 30 & 47\\
True & 21 & 37\\
\bottomrule
\end{tabular}

\caption{\politihop\ dataset statistics. Test set statistics are calculated for a union of two annotators; train instances are annotated by one annotator only, which makes some measures different across splits. We report the mean and standard deviation (in parentheses).} %over the instances in the corresponding split.}
\label{table:data}
\end{table}

%\begin{table}%[!ht]
%\footnotesize
%\centering
%    \begin{tabular}{lrrrrrr}
%      \toprule
%      & \bf 1 & \bf 2 & \bf 3 &  \bf 4 &  \bf 5 &  \bf 6+\\
%      \midrule
%    %   & \multicolumn{6}{c}{\textbf{PolitiHop}}\\
%      Test & 24.8 & 27.1 & 23.2 & 13.6 & 6.1
%      & 4.9\\
%      Train & 27.4 & 30.8 & 22.4 & 11.0 & 5.3
%      & 3.1\\
%      Dev & 29.5 & 32.9 & 19.5 & 10.1 & 3.4
%      & 4.7\\
    %   \midrule
    %   & \multicolumn{6}{c}{\textbf{FEVER}}\\
    %   & 1 & 2+ & & & &\\
    %   All & 83.2 & 16.8 & & & &\\
%      \bottomrule
%    \end{tabular}
%  \caption{Number of evidence sentences per chain in \politihop.}
%  \label{table:length}
%\end{table}

\subsection{Dataset}
%How was the dataset created and why? Show some dataset statistics here already. Non-essential things can go in the appendix.
%e.g. https://www.aclweb.org/anthology/N18-1023.pdf
%Maybe also compare to FEVER to show the differences? - percentage of single-hop, average number of hops
We present \politihop, the first dataset for multi-hop fact checking of real-world claims. It consists of 500 manually annotated claims in written English, split into a training (300 instances) and a test set (200 instances). For each claim, the corresponding PolitiFact article was retrieved, which consists of a discussion of each claim and its veracity, written by a professional fact checker. The annotators then selected sufficient sets of evidence sentences from said articles.
% We define an `evidence chain' as a list of sentences where each sentence follows from the previous one, which together constitute enough evidence to verify the corresponding claim and provide an explanation for it.
As sometimes more than one set can be found to describe a reason behind the veracity of a claim independently, we further take each set in the training split as a separate instance, resulting in 733 training examples.
Each training example is annotated by one annotator, whereas each test example is annotated by two. We split the training data into train and dev datasets, where the former has 592 examples and the latter -- 141. For veracity prediction, we arrived at Krippendorf's $\alpha$ and Fleiss' $\kappa$ agreement values of 0.638 and 0.637, respectively. By comparison, \citet{thorne-etal-2018-fever} reported Fleiss' $\kappa$ of 0.684 on FEVER. For the sentence prediction, we attain Krippendorf's $\alpha$ of 0.437. A more in-depth description of the annotation process can be found in the appendix. %, Section~\ref{ref:annotation}.

Table~\ref{table:data} presents statistics of the dataset. 
% We observe that 
The average number of evidence sentences per set is above 2, which already indicates that the task is more complex than the FEVER dataset. In FEVER, 83.2\% of the claims require one sentence, whereas in \politihop, only 24.8\% require one sentence. %The distribution of the number of evidence sentences in \politihop\ is presented in Table~\ref{table:length}.

% Similarly to label prediction, we measured F1, precision and recall scores for evidence retrieval. The results are: 52.4\%, 49.2\% and 56\%, respectively.

% In order to asses the dataset's difficulty and potential for automation, we performed inter-annotator agreement analysis. 
% We used Krippendorff's alpha and Fleiss' kappa to measure the agreement between annotators on both label prediction and evidence extraction tasks.

% Since each test example is annotated by exactly two annotators, we measured the F1 macro scores by treating one set of annotations as the gold label. This way, the annotators achieve 76.3\% F1 macro score on label prediction.

\section{Models}
We compare the performance of five different models to measure the difficulty of automating the task. 
% The models are random, TF-IDF, BERT, and Transformer-XH.

\textbf{Majority.}
Label prediction only. The majority baseline labels all claims as false.

\textbf{Random.}
% For the random baseline, 
We pick a random number $k \in [1, 10]$ and then randomly choose $k$ sentences from the document as evidence. For label prediction, we randomly pick one of the labels. %with equal probability

\textbf{TF-IDF.}
% For our TF-IDF baseline, 
For each instance $x_i$ we construct a vector $v_i^C=[v_{il}^C| l \in [0, |N^C|]]$ with TF-IDF scores $v_{il}^C$ for all n-grams $N^C$ found in all of the claims; and one vector $v_i^D=[v_{im}^D| m \in [0, |N^D|]]$ with TF-IDF scores $v_{im}^D$ for all n-grams $N^D$ found in all of the documents, where $n \in [2, 3]$. We then train a Naive Bayes model $g(V)$, where $V = \{v_i = (v_i^C \cdot v_i^D) | i \in [0, |X|]\}$ is the concatenation of the two feature vectors. 
% We also remove English stop words using the built-in list in the Scikit-learn library~\cite{scikit-learn}.

\textbf{BERT.}
We first train a Transformer model~\cite{vaswani2017attention}, which does not include a multi-hop mechanism, but applies a single inference step to both the evidence retrieval and the label prediction tasks. We employ BERT~\cite{devlin-etal-2019-bert} with the base pre-trained weights. Each sentence from a fact checking document is encoded separately, combined with the claim and the author of the claim. We refer to the encoded triple as \textit{node $\tau$}. The tokens of one node $x_{\tau} = \{x_{\tau, j} | j \in [1, |x_{\tau}|]\}$ are encoded with the BERT model into contextualized distributed representations: $h_{\tau} = \{h_{\tau,j} | j \in [1, |x_{\tau}|\}$. The encoded representations of all nodes are passed through two feed-forward layers:
\begin{gather}
\small
p(y^{L}|\tau) = softmax(Linear(h_{\tau,0})) \label{eq:lone}\\
p(y^{S}|\tau) = softmax(Linear(h_{\tau,0})) \label{eq:ltwo}\\
p(y^{L}|X) = \sum_{\tau}p(y^{L}|\tau)p(y^{S}|\tau) \label{eq:final}
\end{gather}
The first layer predicts the veracity of the claim given a particular node $\tau$ by using the contextual representation of the ``[CLS]'' token, located at the first position (Eq.~\ref{eq:lone}). The second feed-forward layer learns the importance of each node in the graph (Eq.~\ref{eq:ltwo}). The outputs of these two layers are combined for the final label prediction (Eq.~\ref{eq:final}).
For evidence prediction, we choose $k$ most important sentences, as ranked by the second linear layer. In our experiments, we set $k=6$ since this is the average number of evidence sentences selected by a single annotator.
The implementation of the feed-forward prediction layers is the same as in Transformer-XH, described below, and can be viewed as an ablation of Transformer-XH removing the eXtra Hop attention layers.

\textbf{Transformer-XH.}
Transformer-XH is a good candidate for a multi-hop model for our setup as it has previously achieved the best multi-hop evidence retrieval results on FEVER. It is also inspired by and improves over other multi-hop architectures~\cite{liu-etal-2020-fine,zhou-etal-2019-gear}, and we conjecture that the results should be generalisable for its predecessors as well. Not least, its architecture allows for ablation studies of the multi-hop mechanism. % and also for an ablation as a model without a multi-hop mechanism, which we described above.
Following previous work on applying Transformer-XH to FEVER \cite{zhao2020transformer-xh}, we encode node representations as with the BERT model and construct a fully connected graph with them. Transformer-XH uses eXtra hop attention layers to enable information sharing between the nodes.
An eXtra hop attention layer is a Graph Attention layer (GAT) \cite{gat}, which receives as input a graph $\{X, E\}$ of all evidence nodes $X$ and the edges between them $E$, where the edges encode the attention between two nodes in the graph. Each eXtra hop layer computes the attention between a node and its neighbors, which corresponds to one hop of reasoning across nodes.
Transformer-XH applies L eXtra hop layers to the BERT node encodings $H_0$, which results in new representations $H_L$ that encode the information shared between the nodes, unlike BERT, which encodes each input sentence separately.
We use three eXtra hop layers as in \cite{zhao2020transformer-xh}, which corresponds to three-hop reasoning, and we experiment with varying the number of hops.
The representations $H_L$ are passed to the final two linear layers for label and evidence prediction as in BERT. The final prediction of the veracity label $p(y^{L}|\{X, E\})$ now can also leverage information exchanged in multiple hops between the nodes through the edges $E$ between them.

\section{Experiments}
We address the following research questions:
\begin{itemize}%[nosep]
\item Can multi-hop architectures successfully reason over evidence sets on \politihop? 
%Does explicitly modelling relationships between evidence sentences using a multi-hop architecture (with Transformer-XH) positively impact results on \politihop? %outperform other baselines, i.e. does the Extra Hop attention give an advantage on
\item How do multi-hop vs. single inference architectures fare in an adversarial evaluation, where named entities (NE) in evidence and non-evidence sentences overlap? %Is Transformer-XH robust to named entity overlap between evidence and non-evidence sentences?
\item Does pre-training on related small in-domain or large out-of-domain datasets improve model performance? % (FEVER)  (LIAR-PLUS) 
\end{itemize}

We further perform ablation studies to investigate the influence of different factors on performance (see Section~\ref{sec:discussion_4}).
% : changing the loss function, varying the number of evidence sentences retrieved, adding sentence positions to sentence encodings, varying the number of hops in Transformer-XH, length of the chains, NE overlap, label prediction confidence, edge attention between evidence, and non-evidence sentences.

\subsection{Experimental Setup}
\textbf{Metrics.} We use macro $F_1$ score and accuracy for the veracity prediction task and $F_1$ and precision for the evidence retrieval task. To calculate the performance on both tasks jointly, we use the FEVER score \cite{thorne-etal-2018-fever}, where the model has to retrieve at least one full evidence set and predict the veracity label correctly for the label prediction to count as correct. We consider a single evidence set to be sufficient for correct label prediction. As each example from train and dev sets in \politihop, and every example from LIAR-PLUS, has one evidence set, all evidence sentences need to be retrieved for these. The employed measures for evidence retrieval allow for comparison to related work and for relaxing the requirements on the models. We consider the FEVER score to be the best for evaluating explainable fact checking.

\textbf{Dataset settings.}
We consider three settings: $\mathtt{adversarial}$, $\mathtt{full}$ article, and $\mathtt{even}$ $\mathtt{split}$.
For $\mathtt{full}$, the whole article for each claim is given as input. %, All sentences from each article are provided, 
% with the annotated evidence sentences labelled as 1 and all others labelled as 0
For $\mathtt{even}$ $\mathtt{split}$, we pick all sentences from the same article, but restrict the number of non-evidence sentences to be at most equal to the number of evidence sentences. Non-evidence sentences are picked randomly. %as many as there are evidence sentences. , if such number is available, which is usually the case.  If not, then we take all non-evidence sentences. 
This results in a roughly even split between evidence and non-evidence sentences for the test set. 
Since we divide train and dev datasets into one evidence set per example, but keep all non-evidence for each, the number of non-evidence sentences for instances in these splits is usually 2-3 times larger than the number of evidence sentences. To examine if the investigated multi-hop models overfit on named entity (NE) overlaps, we further construct an $\mathtt{adversarial}$ dataset from the even split dataset by changing each non-evidence sentence to a random sentence from any PolitiFact article, which contains at least one NE present in the original evidence sentences. While such sentences can share information about a relevant NE, they are irrelevant for the claim. We argue that this is a good testbed to understand if a fact checking model can successfully reason over evidence sets and identify non-evidence sentences, even if they contain relevant NEs, which are rather surface features not indicating whether the sentence is relevant to the claim.

\textbf{Training settings.} We perform transfer learning, training on in-domain data (LIAR-PLUS, \politihop), out-of-domain data (FEVER), or a combination thereof. See the appendix for details on training regimes and hyper-parameters.

Note that the measures do not consider the order of the sentences in the evidence set, and the systems do not predict that as well. We believe other measures and models that take that into account should be explored in future work. Here we consider them to appear in the same order as in the document. This also corresponds to the way they were annotated.

\begin{table*}[t]
\small
\centering
    \begin{tabular}{l@{\hskip 0.05in \vline \hskip 0.05in}rr@{\hskip 0.05in \vline \hskip 0.05in}rr@{\hskip 0.05in \vline \hskip 0.05in}r@{\hskip 0.05in \vline \hskip 0.05in}rr@{\hskip 0.05in \vline \hskip 0.05in}rr@{\hskip 0.05in \vline \hskip 0.05in}rr@{\hskip 0.05in \vline \hskip 0.05in}rr@{\hskip 0.05in \vline \hskip 0.05in}r}
      \toprule
      & \multicolumn{5}{c@{\hskip 0.05in \vline \hskip 0.05in}}{\textbf{Dev}} & \multicolumn{5}{c}{\textbf{Test}}\\
      & \bf L-$\mathbf{F_1}$ & \bf L-Acc & \bf E-$\mathbf{F_1}$ & \bf E-Prec & \bf FEVER &
      \bf 
      L-$\mathbf{F_1}$ & \bf L-Acc & \bf E-$\mathbf{F_1}$ & \bf E-Prec & \bf FEVER\\
      \midrule
       Random & 34.1 & 38.5 & 22.9 & 30.2 & 4.5
      & 24.2 & 27.7 & 14.7 & 12.2 & 0.7\\
      Majority & 27.3 & 69.5 & - & - & - & 28.6 & 75.0 & - & - & -\\
      Annotator & - & - & - & - & - & 76.3 & - & 52.4 & 49.2 & -\\
    %   TF-IDF (concat) & 27.0 & 68.0 & - & - & - & 28.6 & 76.0 & - & - & -\\
      TF-IDF & 34.4 & 69.5 & - & - & - & 34.0 & 76.0 & - & - & -\\
      \midrule
      \multicolumn{11}{c}{\textbf{LIAR-PLUS full articles dataset}} \\
      BERT & 45.4 & 70.9 & \bf 18.4 & \bf 13.7 & \bf 14.9
      & \bf 57.0 & 76.0 & \bf 32.9 & \bf 38.9 & \bf 13.0\\
      Transformer-XH & \bf 56.2 & \bf 74.5 & 17.1 & 12.8 & 14.2
      & 56.3 & \bf 79.5 & 30.3 & 35.8 & 12.0\\
     \midrule
     \multicolumn{11}{c}{\textbf{PolitiHop full articles dataset}} \\
     BERT & 54.7 & 69.5 & \bf 32.0 & \bf 23.6 & 31.9
      & \bf 44.8 & \bf 76.0 & \bf 47.0 & \bf 54.2 & \bf \underline{24.5}\\
      Transformer-XH & \bf 61.1 & \bf 76.6 & 30.4 & 22.3 & \bf 34.8
      & 43.3 & 75.5 & 44.7 & 51.7 & 23.5\\
      \midrule
    \multicolumn{11}{c}{\textbf{LIAR-PLUS and PolitiHop full articles}} \\
      BERT & 64.4 & 75.9 & 29.6 & 21.7 & 28.4
      & \bf \underline{57.8} & 79.5 & 45.1 & 52.2 & 23.5\\
      Transformer-XH & \bf \underline{64.6} & \bf \underline{78.7} & \bf \underline{32.4} & \bf \underline{23.8} & \bf \underline{38.3}
      & 57.3 & \bf \underline{80.5} & \bf \underline{47.2} & \bf \underline{54.5} & \bf \underline{24.5}\\
      \bottomrule
    \end{tabular}
  
  \caption{\politihop\ results for label (L), evidence (E) and joint (FEVER) performance in the $\mathtt{full}$ setting. Best results with a particular training dataset (LIAR-PLUS/PolitiHop/LIAR-PLUS and PolitiHop) are emboldened and the best results across all set-ups are underlined.} %Random baseline and annotator performances added for reference.}
  \label{table:baseline}
\end{table*}

\section{Results}
\label{sec:results}
%Can be structured by model type etc. Should include some error analysis. Results and Discussion can be split into two sections if that's easier.
%Baseline Comparison in the Full Article Setting

\textbf{Full article setting.} From the results
% for the $\mathtt{full}$ training setting with models trained on different datasets, then evaluted with the test split of \politihop, are
in Table~\ref{table:baseline}, we can observe that both BERT and Transformer-XH greatly outperform the Random and TF-IDF baselines. 
Out of BERT and Transformer-XH, neither model clearly outperforms the other on our dataset. This is surprising as Transformer-XH outperforms the BERT baselines by a significant margin on both FEVER and the multi-hop dataset HotpotQA \cite{zhao2020transformer-xh}. 
However, we observe that the best performance is achieved with Transformer-XH trained on LIAR-PLUS, then fine-tuned on \politihop. It also achieves the highest FEVER scores on \politihop\ in that setting. %As Table~\ref{table:liar} indicates, it also outperforms BERT on the LIAR-PLUS dataset.
Further, very low FEVER scores of both Transformer-XH and BERT indicate how challenging it is to retrieve the whole evidence set.
%Very low FEVER scores of both models indicate how challenging it is to retrieve the whole chain of evidence. %when choosing only the 6 top evidence sentences. This is especially difficult with LIAR-PLUS dataset, where the whole evidence was treated as one chain.\\

\textbf{Adversarial setting.} %To see if Transformer-XH is robust to named entities, We constructed an adversarial dataset by taking the even splits dataset and changing each non-evidence sentence to to a random sentence, taken from this or other PolitiFact article, as long as it contained a NE from one of the evidence sentences.
%We took Transformer-XH models trained in three different dataset settings: adversarial, full article and even split. We evaluated them on the adversarial dataset. 
We train the Transformer-XH models on the $\mathtt{even}$ $\mathtt{split}$ setting, then evaluate it on both  $\mathtt{adversarial}$ and $\mathtt{even}$ $\mathtt{split}$ datasets (see Table~\ref{table:adv}). 
The model performs similarly in both settings. When compared on test sets, it achieves a higher FEVER score on the $\mathtt{adversarial}$, but the dev set's FEVER score is higher on the $\mathtt{even}$ $\mathtt{split}$ setting. Overall, the results show Transformer-XH is robust towards NE overlap.

\begin{table*}[t]
\small
\centering
    \begin{tabular}{l@{\hskip 0.05in \vline \hskip 0.05in}rr@{\hskip 0.05in \vline \hskip 0.05in}rr@{\hskip 0.05in \vline \hskip 0.05in}r@{\hskip 0.05in \vline \hskip 0.05in}rr@{\hskip 0.05in \vline \hskip 0.05in}rr@{\hskip 0.05in \vline \hskip 0.05in}rr@{\hskip 0.05in \vline \hskip 0.05in}rr@{\hskip 0.05in \vline \hskip 0.05in}r}
      \toprule
      & \multicolumn{5}{c@{\hskip 0.05in \vline \hskip 0.05in}}{\textbf{Dev}} & \multicolumn{5}{c}{\textbf{Test}}\\
      & \bf L-$\mathbf{F_1}$ & \bf L-Acc & \bf E-$\mathbf{F_1}$ & \bf E-Prec & \bf FEVER &
      \bf L-$\mathbf{F_1}$ & \bf L-Acc & \bf E-$\mathbf{F_1}$ & \bf E-Prec & \bf FEVER\\
      \midrule
      even split & \bf 58.1 & \bf 71.6 & 47.7 & 35.5 & \bf 52.5
      & \bf 62.9 & \bf 82.0 & 58.2 & 66.7 & 31.0\\
      adversarial & 56.5 & 70.9 & \bf 49.9 & \bf 38.7 & 46.8
      & 56.4 & 77.0 & \bf 63.6 & \bf 76.0 & \bf 33.5\\
      \bottomrule
    \end{tabular}
  
  \caption{\politihop\ $\mathtt{adversarial}$ vs $\mathtt{even}$ $\mathtt{split}$ dataset results for label (L), evidence (E) and joint (FEVER) performance for Transformer-XH trained on LIAR-PLUS and \politihop\ on the $\mathtt{even}$ $\mathtt{split}$ setting. Best result emboldened.}
  \label{table:adv}
\end{table*}

\textbf{Out-of-domain pre-training on FEVER.}
In this experiment, we examine whether pre-training Transformer-XH on the large, but out-of-domain dataset FEVER, followed by fine-tuning on LIAR-PLUS, then on \politihop\, improves results on \politihop. 
As can be seen from Table~\ref{table:fever-pre}, it does not have a positive effect on performance in the $\mathtt{full}$ setting, unlike pre-training on LIAR-PLUS. We hypothesize that the benefits of using a larger dataset are outweighed by the downsides of it being out-of-domain.
We further quantify the domain differences between datasets. We use Jensen-Shannon divergence \cite{jlin}, commonly employed for this purpose \cite{ruder-plank-2017-learning}. The divergence between FEVER and \politihop\ is 0.278, while between LIAR-PLUS and \politihop\ is 0.063, which further corroborates our hypothesis.
Another reason might be that \politihop\ has several times more input sentences compared to FEVER.
Labelling difference might matter as well: FEVER uses `true', `false' and `not enough info', while \politihop\ uses `true', `false' and `half-true'.

\begin{table*}[t]
\footnotesize
\centering
    \begin{tabular}{l@{\hskip 0.05in \vline \hskip 0.05in}rr@{\hskip 0.05in \vline \hskip 0.05in}rr@{\hskip 0.05in \vline \hskip 0.05in}r@{\hskip 0.05in \vline \hskip 0.05in}rr@{\hskip 0.05in \vline \hskip 0.05in}rr@{\hskip 0.05in \vline \hskip 0.05in}rr@{\hskip 0.05in \vline \hskip 0.05in}rr@{\hskip 0.05in \vline \hskip 0.05in}r}
      \toprule
      & \multicolumn{5}{c@{\hskip 0.05in \vline \hskip 0.05in}}{\textbf{Dev}} & \multicolumn{5}{c}{\textbf{Test}}\\
      & \bf L-$\mathbf{F_1}$ & \bf L-Acc & \bf E-$\mathbf{F_1}$ & \bf E-Prec & \bf FEVER &
      \bf L-$\mathbf{F_1}$ & \bf L-Acc & \bf E-$\mathbf{F_1}$ & \bf E-Prec & \bf FEVER\\
      \midrule
      \specialcell{FEVER+LIAR-PLUS\\+\politihop}& 48.6 & 70.2 & 30.5 & 22.2 & 32.6
      & \bf 59.9 & \bf 83.0 & 45.1 & 52.7 & 21.5\\
      \specialcell{LIAR-PLUS\\+\politihop}& \bf 64.6 & \bf 78.7 & \bf 32.4 & \bf 23.8 & \bf 38.3
      & 57.3 & 80.5 & \bf 47.2 & \bf 54.5 & \bf 24.5\\
      \bottomrule
    \end{tabular}
  
  \caption{\politihop\ $\mathtt{full}$ results for label (L), evidence (E) and joint (FEVER) performance for Transformer-XH trained on different datasets. Best model emboldened.} %trained on LIAR-PLUS and \politihop\ vs. same, but pre-trained on FEVER. Best model emboldened.}
  \label{table:fever-pre}
\end{table*}

\section{Analysis and Discussion}
\label{sec:discussion_4}
In Section~\ref{sec:results}, we documented experimental results on multi-hop fact checking of political claims. Overall, we found that multi-hop training on Transformer-XH gives small improvements over BERT, that pre-training on in-domain data helps, and that Transformer-XH deals well with an adversarial test setting.
Below, we aim to further understand the impact of modeling multi-hop reasoning explicitly with a number of ablation studies:
\begin{itemize}%[nosep]
\item How the evidence set size affects performance;
\item Varying the hops' number in Transformer-XH;
\item The impact of evidence set size on performance;
\item How NE overlap affects performance;
\item To what extent Transformer-XH pays attention to relevant evidence sentences.
\end{itemize}

Further ablation studies can be found in the appendix, namely on: impact of varying the number of evidence sentences on evidence retrieval; how to weigh the different loss functions (for label vs. evidence prediction); if providing supervision for evidence sentence positions impacts performance; and to what degree high label confidence is an indication of high performance.

\textbf{Varying the number of hops in Transformer-XH.}
We train Transformer-XH with a varying number of hops to see if there is any pattern in how many hops result in the best performance. \citet{zhao2020transformer-xh} perform a similar experiment and find that 3 hops are best, similar for 2-5 hops, while the decrease in performance is noticeable for 1 and 6 hops. 
% Still, the authors only observed small differences in results of up to 2 points in F1 score.
We experiment with hops between 1 and 7 (see Table~\ref{table:hops}). Evidence retrieval performance is quite similar in each case. There are some differences for the label prediction task: 1 and 2 hops have slightly worse performance, the 4-hop model has the highest test score and the lowest dev score, while the exact opposite holds for the 5-hop model. Therefore, no clear pattern can be found. One reason for this could be the high variance of the annotated evidence sentences in \politihop. 
% In other words -- some claims might require more and some fewer hops, which may even out the performance between the options. 
%We investigate this issue in more detail in next. %the error analysis.

\begin{table*}[!ht]
% \small
\centering
    \begin{tabular}{lrr@{\hskip 0.05in \vline \hskip 0.05in}rr@{\hskip 0.05in \vline \hskip 0.05in}r@{\hskip 0.05in \vline \hskip 0.05in}rr@{\hskip 0.05in \vline \hskip 0.05in}rr@{\hskip 0.05in \vline \hskip 0.05in}rr@{\hskip 0.05in \vline \hskip 0.05in}rr@{\hskip 0.05in \vline \hskip 0.05in}r}
      \toprule
      & \multicolumn{5}{c@{\hskip 0.05in \vline \hskip 0.05in}}{\textbf{Dev}} & \multicolumn{5}{c}{\textbf{Test}}\\
      & \bf L-$\mathbf{F_1}$ & \bf L-Acc & \bf E-$\mathbf{F_1}$ & \bf E-Prec & \bf FEVER &
      \bf L-$\mathbf{F_1}$ & \bf L-Acc & \bf E-$\mathbf{F_1}$ & \bf E-Prec & \bf FEVER\\
      \midrule
      1 & 54.0 & \bf 75.2 & 47.1 & 35.2 & 52.5
      & 58.9 & 79.5 & 58.7 & \bf 67.6 & 33.0\\
      2 & 56.1 & 73.0 & 47.5 & 35.4 & \bf 53.9
      & 59.8 & 78.5 & 58.1 & 66.7 & 32.5\\
      3 & 58.1 & 71.6 & \bf 47.7 & \bf 35.5 & 52.5
      & 62.9 & 82.0 & 58.2 & 66.7 & 31.0\\
      4 & 53.3 & 70.9 & 47.0 & 34.9 & 50.4
      & \bf 65.0 & \bf 82.0 & \bf 58.9 & \bf 67.6 & 33.0\\
      5 & \bf 59.6 & 73.0 & \bf 47.7 & \bf 35.5 & 51.8
      & 55.3 & 76.5 & 58.7 & 67.3 & 32.0\\
      6 & 56.5 & 73.0 & 45.9 & 34.2 & 50.4
      & 64.9 & 81.5 & 57.5 & 66.0 & \bf 35.0\\
      7 & 56.3 & 71.6 & 46.4 & 34.6 & 50.4
      & 62.8 & 81.5 & 57.9 & 66.4 & 33.0\\
      \bottomrule
    \end{tabular}
  \caption{\politihop\ Transformer-XH results for label (L), evidence (E) and joint (FEVER) performance for training on the LIAR-PLUS + \politihop\ $\mathtt{even}$ $\mathtt{split}$ datasets with a varying number of hop layers. Best sentence number emboldened.}
  \label{table:hops}
\end{table*}

\textbf{Evidence set size vs. performance.}
% Chain length is equal to the number of sentences that constitute the evidence chain. 
Not surprisingly, larger number of evidence sentences leads to higher precision and lower recall, resulting in a lower FEVER score. This is true for both models, as Table~\ref{table:chain-len} (top) indicates. We also notice that the smaller the number, the smaller the ratio of evidence to non-evidence sentences.
% , because the number of non-evidence sentences is equal to the sum of lengths of chains for the given claim. 
For instance, if a claim has two sets of evidence, one of size 1 and the other of size 3, then after splitting into one example per set, there are 4 non-evidence sentences in each of the two examples, but the one with set of size 1 has only one evidence sentence -- which decreases the evidence to non-evidence ratio and makes it more difficult to achieve high precision.

\begin{table*}[!ht]
\small
\centering
    \begin{tabular}{l@{\hskip 0.05in \vline \hskip 0.05in}rr@{\hskip 0.05in \vline \hskip 0.05in}rr@{\hskip 0.05in \vline \hskip 0.05in}r@{\hskip 0.05in \vline \hskip 0.05in}rr@{\hskip 0.05in \vline \hskip 0.05in}rr@{\hskip 0.05in \vline \hskip 0.05in}rr@{\hskip 0.05in \vline \hskip 0.05in}rr@{\hskip 0.05in \vline \hskip 0.05in}r}
      \toprule
      & \multicolumn{5}{c@{\hskip 0.05in \vline \hskip 0.05in}}{\textbf{Transformer-XH}} & \multicolumn{5}{c}{\textbf{BERT}}\\
      & \textbf{L-$\mathbf{F_1}$} & \textbf{L-Acc} & \textbf{E-$\mathbf{F_1}$} & \textbf{E-Prec} & \textbf{FEVER} &
      \textbf{L-$\mathbf{F_1}$} & \textbf{L-Acc} & \textbf{E-$\mathbf{F_1}$} & \textbf{E-Prec} & \textbf{FEVER}\\
      \midrule
      \specialcell{1 or 2 evidence \\ sentences} & \bf 63.9 & \bf 76.8 & \bf 43.7 & \bf 29.2 & \bf 74.4
      & 53.5 & 72.0 & 41.3 & 27.5 & 62.2\\
      \specialcell{3+ evidence \\ sentences} & \bf 60.9 & 66.1 & \bf 67.5 & \bf 56.2 & \bf 42.4
      & 57.8 & 66.1 & 65.8 & 54.8 & 40.7\\
      \midrule
      $<$  40\% NE overlap & \bf 62.5 & \bf 77.0 & \bf 59.1 & \bf 46.2 & \bf 62.3
      & 62.0 & 75.4 & 57.7 & 45.1 & 59.0\\
      $\geq$ 40\% NE overlap & \bf 63.6 & \bf 71.0 & \bf 48.5 & \bf 35.5 & \bf 60.9
      & 47.5 & 66.7 & 46.2 & 33.8 & 49.3\\
      \bottomrule
    \end{tabular}
  \caption{\politihop\ $\mathtt{adversarial}$ dev set performance vs. (top) evidence set size and (bottom) NE overlap between evidence and non-evidence sentences for label (L), evidence (E) and joint (FEVER) performance. Better model emboldened.}
  \label{table:chain-len}
\end{table*}

\textbf{Named entity overlap vs. performance.}
To measure the effect of having the same NEs in evidence and non-evidence sentences, we computed NE overlap -- a measure of the degree to which evidence and non evidence sentences share NEs. We compute the overlap as $|E \cap N|/|E \cup N|$; E and N are sets of NEs in evidence and non-evidence sentences, respectively.
Table~\ref{table:chain-len} (bottom) shows that a higher NE overlap results in more confusion when retrieving evidence sentences, but it does not have a significant influence on label prediction in the case of Transformer-XH. For BERT, higher NE overlap leads to a bigger, negative effect on both tasks. This suggests Transformer-XH is more robust to NE overlaps.

\begin{table}[t]
\footnotesize
 \centering
 \setlength{\tabcolsep}{4pt}
    \begin{tabular}{cccc}
      \toprule
      \textbf{ev} $\rightarrow$ \textbf{non-ev} & \textbf{ev} $\rightarrow$ \textbf{ev} & \textbf{non-ev} $\rightarrow$ \textbf{non-ev} & \textbf{non-ev} $\rightarrow$ \textbf{ev}\\
      \midrule
      \bf 1.085 & 1.076 & 0.966 & 0.964\\
      \bottomrule
    \end{tabular}
  \caption{Attention weights in the last eXtra hop layer of Transformer-XH. The numbers are the average ratios of the actual attention weights to average attention weight of the given graph.}
  \label{table:attention}
\end{table}

%\textbf{Edge Attention between evidence and non-evidence and within each group}.
\textbf{Attention over evidence sentences.}
We investigate what attention patterns Transformer-XH learns. %to what extent Transformer-XH learns to pay more attention to evidence sentences when making predictions. 
Ideally, attention flowing from evidence sentences should be higher than from non-evidence ones since this determines how much they contribute to the final representations of each sentence. To do this, we inspect the weights in the final eXtra hop layer. We normalize results by measuring the ratio of the given attention to the average attention for the given graph: 1 means average, over/under 1 means more/less than average.
Table~\ref{table:attention} shows average ratios for evidence vs. non-evidence sentences. One notable finding is that attention weights from evidence sentences are higher than average, and attention from non-evidence sentences is lower. The Welch t-test indicates that the difference is significant with a $p$-value lower than 10$^{-30}$. So, attention weights get more importance on average, but the magnitude of this effect is quite limited. This shows the limitations of using Transformer-XH for this task.

\section{Related Work}
% NLP often deals with structured corpora of text, e.g. Wikipedia is a set of articles, which contain hyperlinks to other articles. Such structures can often be represented as graphs. Some NLP tasks can be formulated as a reasoning task on these structures.

% Multi-Hop reasoning is the type of reasoning where information from several sentences has to be combined in order to arrive at an answer. It has been mostly studied in the context of question answering. Some of the most notable multi-hop QA dataset include HotPotQA \cite{hotpot} and WikiHop \cite{fever}.

% Another related field where multi-hop reasoning has been studied is fact verification. FEVER dataset \cite{fever} contains many examples where the evidence is distributed over more than one example and some sort of multi-hop reasoning is required to retrieve it.

%There have been numerous efforts to automate different stages in the fact-checking process: finding check-worthy claims, finding evidence and predicting the veracity of a claim are all related but essentially different tasks. 
%: true, false and perhaps some moderate labels, e.g. half-true. 
%There are two main drawbacks of these models: 
% One is that the models treat fact-checking as a one-step problem, ignoring the fact that very often evidence from multiple sources needs to be retrieved and composed to make the correct veracity prediction. The other drawback is the lack of explanation of their decision-making.
\textbf{Fact checking.}
% Summarise what's been done on fact checking, end in the conclusion that there's practically no work on multi-hop FC
Several datasets have been released 
% in the past few years 
to assist in automating fact checking. 
\citet{vlachos-riedel-2014-fact} present a dataset with 106 political claim-verdict pairs. 
% collected from PolitiFact and the fact checking blog of Channel 4\footnote{ http://blogs.channel4.com/factcheck/}.
The FakeNewsChallenge~\footnote{\url{http://www.fakenewschallenge.org/}}, provides 50K headline-article pairs and formulates the task of fact checking as stance detection between the headline and the body of the article. The relationship between these two tasks is further explored in \citet{hardalov2021survey}.
\citet{wang-2017-liar} extract 12.8K claims from PolitiFact constituting the LIAR dataset.
\citet{alhindi-etal-2018-evidence} introduce the LIAR-PLUS dataset extending the latter with automatically extracted summaries from PolitiFact articles. These are, however, high-level explanations that omit evidence details. LIAR-PLUS also does not provide annotation of particular evidence sentences from the article leading to the final verdict and the possible different evidence sets. 
\citet{augenstein-etal-2019-multifc} present a real-world dataset constructed from 26 fact checking portals, including PolitiFact, consisting of 35k claims paired with crawled evidence documents.
\citet{thorne-etal-2018-fever} present the FEVER dataset, consisting of 185K claims produced by manually re-writing Wikipedia sentences. 
% However, as highlighted in the original paper, only 12.15\% of the claims were generated from multiple pages. 
Furthermore, \citet{niewinski-etal-2019-gem} from the FEVER'2019 shared task~\cite{thorne-etal-2019-fever2} and \citet{hidey-etal-2020-deseption} use adversarial attacks to show the vulnerability of models trained on the FEVER dataset to claims that require more than one inference step.
Unlike prior work, we construct a dataset with annotations of the different reasoning sets and the multiple hops that constitute them.

\textbf{Multi-hop datasets.}
% Summarise what's been done in multi-hop reasoning, conclude by saying that work here has been in the space of QA, not FC.
Multi-hop reasoning has been mostly studied in the context of Question Answering (QA). 
\citet{yang-etal-2018-hotpotqa} introduce HotpotQA with Wikipedia-based question-answer pairs requiring reasoning over multiple documents and provide gold labels for sentences supporting the answer. 
% The supporting evidence provides for strong supervision over the reasoning of the networks and helps interpretability. 
% The question-answer pairs are generated by showing crowd workers a set of related documents and asking them to come up with questions in order to increase the diversity of questions and keep them free of a strict structure.
\citet{welbl-etal-2018-constructing} introduce MedHop and WikiHop datasets for reasoning over multiple documents. These are constructed using Wikipedia and DrugBank as Knowledge Bases (KB), and are limited to entities and relations existing in the KB. This, in turn, limits the type of questions that can be generated. TriviaQA \cite{joshi-etal-2017-triviaqa} and SearchQA \cite{searchqa} contain multiple documents for question-answer pairs but have few examples where reasoning over multiple paragraphs from different documents is necessary.

\textbf{Multi-hop models.} \citet{chen-durrett-2019-understanding} observe that models without multi-hop reasoning are still able to perform well on a large portion of the test dataset. \citet{hidey-etal-2020-deseption} employ a pointer-based architecture, which re-ranks documents related to a claim and jointly predicts the sequence of evidence sentences and their stance to the claim. \citet{asai2019learning} 
% build a recurrent neural network that
sequentially extract paragraphs from the reasoning path conditioning on the documents extracted on the previous step. CogQA \cite{ding-etal-2019-cognitive}
% uses multiple BERT models for 
detect spans and entities of interest and then run a BERT-based Graph Convolutional Network for ranking. \citet{nie-etal-2019-revealing} 
% use BERT to 
perform semantic retrieval of relevant paragraphs followed by span prediction in the case of QA and 3-way classification for fact checking. \citet{zhou-etal-2019-gear}, \citet{liu-etal-2020-fine}, and \citet{zhao2020transformer-xh} model documents as a graph and apply attention networks across the nodes of the graph. We use \citet{zhao2020transformer-xh}'s model due to its strong performance in multi-hop QA on the HotpotQA dataset, in evidence-based fact checking on FEVER, and to evaluate its performance on real-world claim evidence reasoning.

\section{Conclusions}
In this paper, we studied the novel task of multi-hop reasoning for fact checking of real-world political claims, which encompasses both evidence retrieval and claim veracity prediction. 
We presented \politihop, the first political fact checking dataset with annotated evidence sentences.
We compared several models on \politihop\ and found that the multi-hop architecture Transformer-XH slightly outperforms BERT in most of the settings, especially in terms of evidence retrieval, where BERT is easily fooled by named entity overlaps between the claim and evidence sentences. 
The performance of Transformer-XH is further improved when retrieving more than two evidence sentences and the number of hops larger than one, which corroborates the assumption of the multi-hop nature of the task. %Further, Transformer-XH performs best in an in-domain transfer learning setting. %is significantly improved by pre-training on the LIAR-PLUS dataset, which is in the same domain as \politihop\. 
\section{Appendices}
\subsection{Annotation Process}
\label{ref:annotation}

\subsubsection{Annotation Pipeline}
We used the PolitiFact API to retrieve the articles, along with the source pages used in the article, the claim and the author of the claim. For each article we performed the annotation process as follows:

\begin{enumerate}%[nosep]
\item{Reading the claim and the article.}
\item{Picking the evidence sentences from the corresponding PolitiFact article. These sentences should sum up the whole article while providing as much evidence as possible.}
\item{Deciding on the veracity label.}
\item{Going to each relevant url and checking whether it contains the equivalent textual evidence.}
\end{enumerate}

In Step 2, we retrieved evidence sentences sentences, where each sentence follows from the previous one and together they constitute enough evidence to verify the claim and provide an explanation for it.

In Step 4, we wanted to examine how often the evidence can be retrieved from external sources, i.e. not relying on PolitiFact articles. However, we have not gathered enough data to carry out a reliable evaluation of this and thus left the idea for future work.

Originally, following \cite{thorne-etal-2018-fever}, we wanted to have a `not enough evidence' label, but due to a small frequency of this label in the annotations, as well as due to a significant disagreement between annotators on that label, we decided to discard it and re-label it with one of the remaining labels (false, half-true or true).
In case of conflicting label annotations, a third annotator was asked to resolve the conflict.

\subsubsection{Inter-Annotator Agreement}
\begin{figure}%[!h]
\center
\includegraphics[scale=0.7]{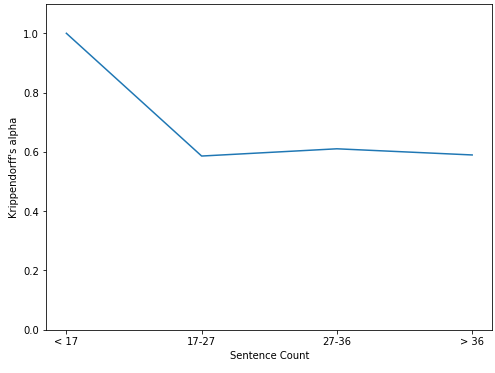} 
\caption{Article length vs. inter-annotator agreement.}
\label{figure:iaa}
\end{figure}

We report Inter-Annotator Agreement (IAA) agreement on the test set, where we had two annotators annotating each instance. For the veracity prediction task, annotators’ Krippendorf's $\alpha$ and Fleiss' $\kappa$ are equal to 0.638 and 0.637 respectively. By comparison, \citet{thorne-etal-2018-fever} reported Fleiss' $\kappa$ of 0.684 on the veracity label prediction, which is the another indication of the increased complexity when predicting veracity of claims occurring naturally. For the sentence prediction task, when treating each ruling article as a separate dataset and averaging over all articles, annotators achieve 0.437 Fleiss' $\kappa$ and 0.437 Krippendorff's $\alpha$ (we also compute IAA when treating all sentences from all articles as one dataset, where both IAA measures drop to 0.400). Figure~\ref{figure:iaa} confirms the intuition that annotators tend to agree more on the shorter articles, which are easier to annotate as they contain fewer sets and fewer hops per set.

\subsection{Training Details}
We used LIAR-PLUS and \politihop\ for training in three different settings:\\
1. Training on LIAR-PLUS only.\\
2. Training on \politihop\ only.\\
3. Pre-training on LIAR-PLUS and fine-tuning on \politihop. \\
4. Pre-training on FEVER, then fine-tuning on LIAR-PLUS and \politihop.

In the first setting, the models are trained for 4 epochs on LIAR-PLUS. In the second setting, the models are trained for 8 epochs on \politihop. In the third setting, models are trained for 4 epochs on LIAR-PLUS, followed by 4 epochs on \politihop. In every setting, models are evaluated on the dev set and the model with the best label prediction macro $F_1$ score is saved, which enables early stopping.
For the fourth setting, we pre-train the model for 2 epochs on the FEVER dataset, followed by 4 epochs on LIAR-PLUS, the fine-tune on \politihop\ for 4 epochs.

The models have been trained and evaluated using one NVIDIA TITAN RTX.
We report the results based on a single run with a random seed fixed to 42.

Both BERT and Transformer-XH are trained with the same hyperparameters as in \cite{zhao2020transformer-xh}: BERT 12 layers' with the hidden size of 768, 3 GAT layers with the hidden size of 64. Optimized with Adam with the learning rate of 1e-5. For the TF-IDF baseline, we also remove English stop words using the built-in list in the Scikit-learn library~\cite{scikit-learn}.

Some of the experiments are ablation studies of the number of GAT layers and the number of retrieved evidence sentences. The former varies between 1 and 7 while the latter varies between 1 and 10.

\subsection{Additional Results}

\textbf{Number of evidence sentences vs. evidence retrieval performance}. One of the challenges of generating fact checking explanations is deciding on the length of the explanation. By design, the explanations should be short, ideally just a few sentences. On the other hand, they have to provide a comprehensive motivation of the fact checking verdict. Transformer-XH handles evidence retrieval by ranking the importance of each input sentence. %It does not learn to classify sentences; this needs to be done by another method.  
We, therefore, pick the most highly ranked sentences, according to the model. By default for all experiments in Section~\ref{sec:results}, the top 6 sentences are used, as it is the average length of an annotation in the \politihop\ test set.

\begin{table}[t]
% \footnotesize
\centering
\setlength{\tabcolsep}{4pt}
    \begin{tabular}{lrrr@{\hskip 0.05in \vline \hskip 0.05in}rrr}
      \toprule
      & \multicolumn{3}{c@{\hskip 0.05in \vline \hskip 0.05in}}{\textbf{Test}} & \multicolumn{3}{c}{\textbf{Dev}}\\
    \bf  \#S. &\bf $\mathbf{F_1}$ &\bf Recall &\bf Precision &\bf $\mathbf{F_1}$ & \bf Recall & \bf Precision \\
      \midrule
      1 & 15.9 & 9.7 & \bf 62.0 & 17.3 & 13.5
      & 30.5 \\
      2 & 27.6 & 19.4 & 61.0 & 27.8 & 28.9
      & \bf 31.2 \\
      3 & 36.2 & 28.5 & 61.2 & 31.8 & 39.4
      & 30.0  \\
      4 & 41.3 & 35.5 & 59.1 & 32.3 & 47.0
      & 27.1 \\
      5 & 44.0 & 41.0 & 55.8 & 31.6 & 52.9
      & 24.5 \\
      6 & 47.2 & 47.2 & 54.5 & \bf 32.4 & 60.9
      & 23.8 \\
      7 & 49.2 & 52.5 & 52.9 & 31.8 & 65.8
      & 22.4 \\
      8 & 50.3 & 56.9 & 51.0 & 31.2 & 70.5
      & 21.2 \\
      9 & 50.8 & 60.7 & 49.0 & 30.5 & 74.7
      & 20.2 \\
      10 & \bf 51.0 & \bf 64.1 & 47.3 & 29.1 & \bf 76.4
      & 18.9 \\
      \bottomrule
    \end{tabular}

  \caption{\politihop\ evidence retrieval results for a model trained on LIAR-PLUS $\mathtt{full}$, then fine-tuned on \politihop\ $\mathtt{full}$, with a varying number of top sentences retrieved as evidence. Best number of sentences emboldened.}
  \label{table:evi-num}
\end{table}

Table~\ref{table:evi-num} shows how recall trades off against precision and improves as an increasing number of sentences is selected. 
% As expected, recall improves, and precision deteriorates. %as the number of sentences grows. 
Test set $F_1$ grows as the number of sentences grows, while the best train set $F_1$ is the highest for 3 sentences and it gets worse as the number of sentences increases. $F_1$ on dev set is much more even for different numbers, but it peaks at 6 sentences. Generally, 6 sentences gives the best trade-off between the performance on test and dev sets, while being short enough to be considered as short summary of the whole article.

\begin{table}%[!ht]
% \small
  \begin{center}
    \begin{tabular}{l@{\hskip 0.05in \vline \hskip 0.05in}rrr@{\hskip 0.05in \vline \hskip 0.05in}rrrrr}
      \toprule
      & \multicolumn{3}{c@{\hskip 0.05in \vline \hskip 0.05in}}{\textbf{Dev}} & \multicolumn{3}{c}{\textbf{Test}}\\
      & \bf Lab & \bf Evi & \bf Joint &
      \bf Lab & \bf Evi & \bf Joint\\
    %   & \bf $F_1$ & \bf $F_1$ & \bf FEVER &
    %   \bf $F_1$ & \bf $F_1$ & \bf FEVER\\
      \midrule
      BERT & 72.2 & 43.7 & 11.3
      & 72.0 & 43.5 & \bf 14.5\\
      TXH & \bf 73.7 & \bf 44.4 & \bf 12.1
      & \bf 72.6 & \bf 44.6 & 13.4\\
      \bottomrule
    \end{tabular}
  \end{center}
  \caption{LIAR-PLUS results when trained on the LIAR-PLUS full articles dataset. Best model emboldened. The Lab(el) and Evi(dence) results are $F_1$ scores, and Joint is measured with FEVER score.}
  \label{table:liar}
\end{table}

\textbf{LIAR-PLUS}. Here, we investigate training then testing on LIAR-PLUS. As Table~\ref{table:liar} shows, Transformer-XH outperforms BERT by a small margin. This confirms the results in \cite{zhao2020transformer-xh} that Transformer-XH generally performs well in multi-hop settings.

\textbf{Loss function comparison}. In this experiment we compare BERT and Transformer-XH performance on the \politihop\ full article setting when trained with three different loss functions. The default loss function is the sum of evidence prediction loss and label prediction loss. The EVI setting uses evidence loss only and saves the model with the highest validation set evidence $F_1$ score. The LAB setting uses label prediction loss only and saves the model with the highest macro $F_1$ score on validation data label prediction, just like in the default setting.

Table~\ref{table:loss} shows the best performance with the joint loss for BERT. EVI setting hurts the label prediction while LAB setting hurts the evidence prediction, without providing a clear boost in the second metric over the joint model. Transformer-XH performs much worse on evidence prediction when trained using label prediction loss only. Interestingly, there is no clear performance difference between EVI and default settings.

\begin{table*}[t]
  \begin{center}
    \begin{tabular}{l@{\hskip 0.05in \vline \hskip 0.05in}rr@{\hskip 0.05in \vline \hskip 0.05in}rr@{\hskip 0.05in \vline \hskip 0.05in}r@{\hskip 0.05in \vline \hskip 0.05in}rr@{\hskip 0.05in \vline \hskip 0.05in}rr@{\hskip 0.05in \vline \hskip 0.05in}rr@{\hskip 0.05in \vline \hskip 0.05in}rr@{\hskip 0.05in \vline \hskip 0.05in}r}
      \toprule
      & \multicolumn{5}{c@{\hskip 0.05in \vline \hskip 0.05in}}{\textbf{Dev}} & \multicolumn{5}{c}{\textbf{Test}}\\
      & \multicolumn{2}{c@{\hskip 0.05in \vline \hskip 0.05in}}{\bf Label} & \multicolumn{2}{c@{\hskip 0.05in \vline \hskip 0.05in}}{\bf Evidence} & \bf Joint &
      \multicolumn{2}{c@{\hskip 0.05in \vline \hskip 0.05in}}{\bf Label} & \multicolumn{2}{c@{\hskip 0.05in \vline \hskip 0.05in}}{\bf Evidence} & \bf Joint\\
      & \bf $\mathbf{F_1}$ & \bf Acc & \bf $\mathbf{F_1}$ & \bf Prec & \bf FEVER &
      \bf $\mathbf{F_1}$ & \bf Acc & \bf $\mathbf{F_1}$ & \bf Prec & \bf FEVER\\
      \midrule
      BERT & 64.4 & 75.9 & 29.6 & 21.7 & 28.4
      & 57.8 & 79.5 & 45.1 & 52.2 & 23.5\\
      Transformer-XH & \bf 64.6 & \bf 78.7 & 32.4 & 23.8 & \bf 38.3
      & 57.3 & 80.5 & \bf 47.2 & \bf 54.5 & \bf 24.5\\
      BERT-EVI & 34.0 & 51.1 & 31.8 & 23.4 & 26.2
      & 40.7 & 63.5 & 45.7 & 52.8 & 21.5\\
      Trans-XH-EVI & 56.0 & 68.1 & \bf 34.4 & \bf 25.2 & 33.3
      & 59.9 & 75.5 & 46.7 & 54.2 & 21.5\\
      BERT-LAB & 59.4 & 72.3 & 19.9 & 14.8 & 14.9
      & 60.3 & 77.5 & 33.8 & 40.1 & 13.5\\
      Trans-XH-LAB & 62.0 & 75.2 & 18.2 & 13.5 & 14.9
      & \bf 60.6 & \bf 81.5 & 32.4 & 38.8 & 13.0\\
      Random & 24.2 & 27.7 & 14.7 & 12.2 & 0.7
      & 34.1 & 38.5 & 22.9 & 30.2 & 4.5\\
      \bottomrule
    \end{tabular}
  \end{center}
  \caption{\politihop\ results trained on LIAR-PLUS + \politihop\ full articles datasets. EVI means the model was trained with loss on the evidence prediction task only. LAB means the loss on the label prediction only. The default loss was the sum of both. Best model emboldened.}
  \label{table:loss}
\end{table*}

\textbf{Adding sentence IDs to sentence encodings}.
The main goal of this experiment was to see whether providing the information about the positions of the sentences in articles can be leveraged to improve the performance of BERT and Transformer-XH models.

We took the models pre-trained on LIAR-PLUS without sentence positions and fine-tuned the model on \politihop\ with sentence positions, by prepending each sentence's encoding with the token [unusedN], where $N=$\ sentence position.
Table~\ref{table:sent-ids} shows a significant performance boost for Transformer-XH label prediction, but not for evidence retrieval. BERT does not exhibit any improvement, which is to be expected as it considers each sentence in isolation, it doesn't learn any interactions between sentences.

\begin{table*}[t]
  \begin{center}
    \begin{tabular}{l@{\hskip 0.05in \vline \hskip 0.05in}rr@{\hskip 0.05in \vline \hskip 0.05in}rr@{\hskip 0.05in \vline \hskip 0.05in}r@{\hskip 0.05in \vline \hskip 0.05in}rr@{\hskip 0.05in \vline \hskip 0.05in}rr@{\hskip 0.05in \vline \hskip 0.05in}rr@{\hskip 0.05in \vline \hskip 0.05in}rr@{\hskip 0.05in \vline \hskip 0.05in}r}
      \toprule
      & \multicolumn{5}{c@{\hskip 0.05in \vline \hskip 0.05in}}{\textbf{Dev}} & \multicolumn{5}{c}{\textbf{Test}}\\
      & \multicolumn{2}{c@{\hskip 0.05in \vline \hskip 0.05in}}{\bf Label} & \multicolumn{2}{c@{\hskip 0.05in \vline \hskip 0.05in}}{\bf Evidence} & \bf Joint &
      \multicolumn{2}{c@{\hskip 0.05in \vline \hskip 0.05in}}{\bf Label} & \multicolumn{2}{c@{\hskip 0.05in \vline \hskip 0.05in}}{\bf Evidence} & \bf Joint\\
      & \bf $\mathbf{F_1}$ & \bf Acc & \bf $\mathbf{F_1}$ & \bf Prec & \bf FEVER &
      \bf $\mathbf{F_1}$ & \bf Acc & \bf $\mathbf{F_1}$ & \bf Prec & \bf FEVER\\
      \midrule
      BERT & 62.4 & 75.2 & 27.7 & 20.3 & 24.1
      & 57.4 & 79.0 & 42.9 & 49.2 & 21.5\\
      Transformer-XH & \bf 65.2 & \bf 78.0 & \bf 32.2 & \bf 23.6 & \bf 38.3
      & \bf 66.5 & \bf 84.0 & \bf 47.2 & \bf 54.9 & \bf 26.5\\
      \bottomrule
    \end{tabular}
  \end{center}
  \caption{\politihop\ results for training on LIAR-PLUS $\mathtt{full}$, then fine-tuning on \politihop\ $\mathtt{full}$ with sentence ID encodings. Best model emboldened.}
  \label{table:sent-ids}
\end{table*}

\textbf{Label confidence vs. performance}.
The goal here was to measure how confident the models are in their label predictions and to see if higher confidence means better performance. The results are presented in Table~\ref{table:lab-conf}.

Transformer-XH is usually sure of its predictions, so not much can be observed based on that - it does indeed have higher $F_1$ score when it is more confident, but there are too few instances where it is not confident to make any conclusions - apart from the one that it's often sure but makes a mistake anyway. Besides, NE overlap was not particularly high for the instances where the model got confused.

The effect is even stronger with BERT to the point where having less than 95\% confidence usually results in a bad prediction.

\begin{table*}[t]
  \begin{center}
    \begin{tabular}{l@{\hskip 0.05in \vline \hskip 0.05in}rr@{\hskip 0.05in \vline \hskip 0.05in}rr@{\hskip 0.05in \vline \hskip 0.05in}r@{\hskip 0.05in \vline \hskip 0.05in}rr@{\hskip 0.05in \vline \hskip 0.05in}rr@{\hskip 0.05in \vline \hskip 0.05in}rr@{\hskip 0.05in \vline \hskip 0.05in}rr@{\hskip 0.05in \vline \hskip 0.05in}r}
      \toprule
      & \multicolumn{5}{c@{\hskip 0.05in \vline \hskip 0.05in}}{\textbf{Transformer-XH}} & \multicolumn{5}{c}{\textbf{BERT}}\\
      & \multicolumn{2}{c@{\hskip 0.05in \vline \hskip 0.05in}}{\bf Label} & \multicolumn{2}{c@{\hskip 0.05in \vline \hskip 0.05in}}{\bf Evidence} & \bf Joint &
      \multicolumn{2}{c@{\hskip 0.05in \vline \hskip 0.05in}}{\bf Label} & \multicolumn{2}{c@{\hskip 0.05in \vline \hskip 0.05in}}{\bf Evidence} & \bf Joint\\
      &\bf $\mathbf{F_1}$ &\bf Acc &\bf $\mathbf{F_1}$ &\bf Prec &\bf FEVER &\bf
      $\mathbf{F_1}$ & \bf Acc &\bf $\mathbf{F_1}$ &\bf Prec &\bf FEVER\\
      \midrule
      $<$ 90\% & \bf 45.5 & \bf 46.2 & \bf 51.6 & \bf 39.7 & \bf 34.6
      & 30.9 & 32.6 & 52.9 & 41.6 & 20.9\\
      $\geq$ 90\% & 67.2 & 78.3 & \bf 54.1 & \bf 40.7 & 67.0
      & \bf 72.5 & \bf 85.7 & 50.9 & 37.8 & \bf 67.3\\
      \bottomrule
    \end{tabular}
  \end{center}
  \caption{\politihop\ adversarial dev set performance vs. label confidence.}
  \label{table:lab-conf}
\end{table*}

\subsection{PolitiHop Example}
Table \ref{tab:Example_4} shows an example from the proposed \politihop{} dataset.
\begin{table}[t]
% \fontsize{8.4}{8.4}\selectfont
\begin{center}
\begin{tabular}{p{15cm}}
\toprule
\textbf{Claim}: Claim: Says 20 million Chinese converted to Islam after it’s proven that the coronavirus doesn’t affect Muslims.\\ 
\textbf{Speaker}: Viral image\\
\textbf{Label}: false \\
\midrule
\textbf{Ruling Comments}: 
% \hspace*{3mm} 
[1] Amid fears about the coronavirus disease , a YouTube video offers a novel way to inoculate yourself: convert to Islam. [2] ``20m Chinese gets converted to Islam after it is proven that corona virus did not affect the Muslims,'' reads the title of a video posted online Feb. 18 (...) \textbf{[5] That’s because the footage is from at least as far back as May 26, 2019, when it was posted on Facebook with this caption: 
``Alhamdulillah welcome to our brothers in faith.'' [6] On Nov. 7, 2019, it was posted on YouTube with this title: ``MashaaAllah hundreds converted to Islam in Philippines.''} [7] Both posts appeared online before the current outbreak of the new coronavirus, COVID-19, was first reported in Wuhan, China, on Dec. 31, 2019. \textit{[8] But even if the footage followed the outbreak, Muslims are not immune to COVID-19, as the Facebook post claims. [9] After China, Iran has emerged as the second focal point for the spread of COVID-19, the New York Times reported on Feb. 24 .} [10] 
``The Middle East is in many ways the perfect place to spawn a pandemic, experts say, with the constant circulation of both Muslim pilgrims and itinerant workers who might carry the virus.'' [11] On Feb. 18, Newsweek reported that coronavirus ``poses a serious risk to millions of inmates in China’s Muslim prison camps.'' \\
\bottomrule
\end{tabular}
\end{center}
\caption{An example from the \politihop\ dataset. Each example consists of a claim, a speaker (author of the claim), a veracity label and a PolitiFact article with the annotated evidence sentences. One of the evidence sets is in bold, and the other in italics.}
\label{tab:Example_4}
\end{table}

\part{Diagnostic Explainability Methods}
\chapter{A Diagnostic Study of Explainability Techniques for Text Classification}
\label{chap:diagnostic_study_saliency}

\section{Introduction}
\begin{figure}
\centering
\includegraphics[width=300pt]{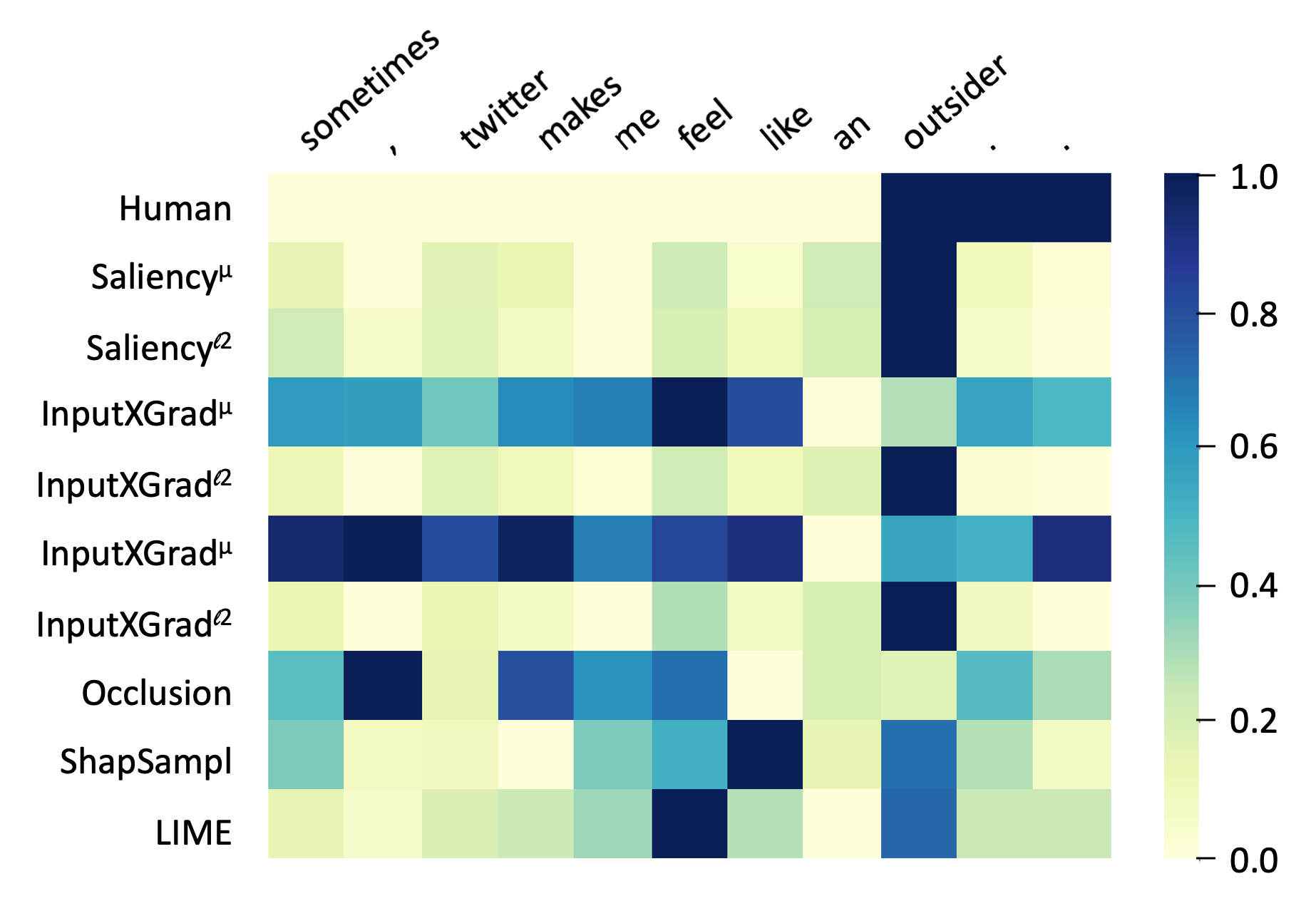}
\caption{Example of the saliency scores for the words (columns) of an instance from the Twitter Sentiment Extraction dataset. They are produced by the explainability techniques (rows) given a \trans{} model. The first row is the human annotation of the salient words. The scores are normalized in the range $[0, 1]$.}
\label{fig:example}
\end{figure}
% Advancing the state of machine learning with models nearing human performance has set the ground for the broad adoption of automated decision making in a multitude of areas. 
Understanding the rationales behind models' decisions is becoming a topic of pivotal importance, as both the architectural complexity of machine learning models and the number of their application domains increases. Having greater insight into the models' reasons for making a particular prediction has already proven to be essential for discovering potential flaws or biases in medical diagnosis \cite{caruana2015intelligible} and judicial sentencing \cite{rich2016machine}. In addition, European law has mandated ``the right $\dots$ to obtain an explanation of the decision reached'' ~\cite{regulation2016regulation}.

% Models' reasoning processes can be elucidated with either explanation or interpretation methods~\cite{10.1145/3236386.3241340}. \textit{Interpretability} techniques attempt to compile a summary of possible decision paths that apply to most input data points. Interpretability techniques that provide information on the neuron level end up producing a large number of outputs and are thus too fine-grained to be of benefit for most end users. 
\textit{Explainability methods} attempt to reveal the reasons behind a model's prediction for a single data point, as shown in Figure~\ref{fig:example}. They can be produced post-hoc, i.e., with already trained models. Such post-hoc explanation techniques can be applicable to one specific model~\cite{barakat2007rule, wagner2019interpretable} or to a broader range thereof~\cite{ribeiromodel, lundberg2017unified}. 
They can further be categorised as: employing model gradients \cite{sundararajan2017axiomatic, Simonyan2013DeepIC}, being perturbation based \cite{shapley1953value, zeiler2014visualizing} or providing explanations through model simplifications \cite{ribeiromodel, johansson2004truth}. There also exist explainability methods that generate textual explanations~\cite{NIPS2018_8163} and are trained post-hoc or jointly with the model at hand. 

While there is a growing amount of explainability methods, we find that they can produce varying, sometimes contradicting explanations, as illustrated in Figure~\ref{fig:example}.
% The recent interest in explaining machine learning predictions has created an abundance of explainability methods that can produce varying explanations as shown in Figure~\ref{fig:example}. 
Hence, it is important to \emph{assess existing techniques} and to \emph{provide a generally applicable and automated methodology} for choosing one that is suitable for a particular model architecture and application task~\cite{jacovi-goldberg-2020-towards}. 
% The latter has also been put forward as an important open challenge in~\cite{jacovi2020towards}. 
\citet{robnik2018perturbation} compiles a list of property definitions for explainability techniques, but it remains a challenge to evaluate them in practice. Several other studies have independently proposed different setups for probing varied aspects of explainability techniques~\cite{deyoung-etal-2020-eraser, sundararajan2017axiomatic}.
% as Comprehensiveness and Sufficiency~\cite{deyoung-etal-2020-eraser} or Sensitivity and Implementation Invariance~\cite{sundararajan2017axiomatic}. 
However, existing studies evaluating explainability methods are discordant and do not compare to properties from previous studies. In our work, we consider properties from related work and extend them to be applicable to a broader range of downstream tasks.

Furthermore, to create a thorough setup for evaluating explainability methods, one should include at least: (i) different groups of explainability methods (explanation by simplification, gradient-based, etc.), (ii) different downstream tasks, and (iii) different model architectures. However, existing studies usually consider at most two of these aspects, thus providing insights tied to a specific setup.  
% Finally, only a few of these studies address explainability for NLP tasks \cite{deyoung-etal-2020-eraser}.
%There also exist studies that evaluate explainability techniques solely using human feedback and disregarding the functional characteristics of explanations. 
% The main techniques for producing explanations for machine learning models, applicable also for Natural Language Processing (NLP) systems

We propose a number of \propertyplural{} for explainability methods and evaluate them in a comparative study. We consider explainability methods from different groups, all widely applicable to most ML models and application tasks. We conduct an evaluation on three text classification tasks, which contain human annotations of salient tokens. Such annotations are available for Natural Language Processing (NLP) tasks, as they are relatively easy to obtain. This is in contrast to ML sub-fields such as image analysis, for which we only found one relevant dataset -- 536 manually annotated object bounding boxes for Visual Question Answering~\cite{subramanian-etal-2020-obtaining}. 

We further compare explainability methods across three of the most widely used model architectures -- \cnn, \lstm, and \trans{}. The \trans{} model achieves state-of-the-art performance on many text classification tasks but has a complex architecture, hence methods to explain its predictions are strongly desirable. The proposed properties can also be directly applied to Machine Learning (ML) subfields other than NLP. The code for the paper is publicly available.\footnote{\url{https://github.com/copenlu/xai-benchmark}} \\In summary, the \textbf{contributions} of this work are:
\begin{itemize}%[noitemsep]
\item We compile a comprehensive list of \propertyplural{} for explainability and automatic measurement of them, allowing for their effective assessment in practice.
% \item We devise an inclusive benchmark for explainability techniques that covers different groups of explainability tools, three NLP datasets, and diverse model architectures.
\item We study and compare the characteristics of different groups of explainability techniques in three different application tasks and three different model architectures.
% for the above-mentioned setups and provide insights for guiding their selection.
\item We study the attributions of the explainability techniques and human annotations of salient regions to compare and contrast the rationales of humans and machine learning models. 
\end{itemize}
% TODO a picture of explainability over different techniques, datasets, models

\section{Related Work}
% The development of increasingly complex model architectures has triggered new studies in producing model explanations that can help to make models' predictions more transparent. 
Explainability methods can be divided into explanations by simplification, e.g., LIME \cite{ribeiromodel}; gradient-based explanations \cite{sundararajan2017axiomatic}; perturbation-based explanations \cite{shapley1953value, zeiler2014visualizing}. Some studies propose the generation of text serving as an explanation, e.g., \cite{NIPS2018_8163,lei-etal-2016-rationalizing,atanasova-etal-2020-generating-fact}. For extensive overviews of existing explainability approaches, see \citet{BARREDOARRIETA202082}.

Explainability methods provide explanations of different qualities, so assessing them systematically is pivotal. A common attempt to reveal shortcomings in explainability techniques is to reveal a model's reasoning process with counter-examples \cite{alvarez2018robustness, kindermans2019reliability,atanasova-etal-2020-generating}, finding different explanations for the same output. However, single counter-examples do not provide a measure to evaluate explainability techniques~\cite{jacovi-goldberg-2020-towards}.%, and it would be possible to find such in many cases~\cite{jacovi2020towards}. 

Another group of studies performs human evaluation of the outputs of explainability methods \cite{lertvittayakumjorn-toni-2019-human, narayanan2018humans}. Such studies exhibit low inter-annotator agreement and reflect mostly what appears to be reasonable and appealing to the annotators, not the actual properties of the method.

The most related studies to our work design measures and properties of explainability techniques. \citet{robnik2018perturbation} propose an extensive list of properties. The \textit{Consistency} property captures the difference between explanations of different models that produce the same prediction; and the \textit{Stability} property measures the difference between the explanations of similar instances given a single model. We note that similar predictions can still stem from different reasoning paths. Instead, we propose to explore instance activations, which reveal more of the model's reasoning process than just the final prediction. The authors propose other properties as well, which we find challenging to apply in practice. We construct a comprehensive list of \propertyplural{} tied with measures that assess the degree of each characteristic.

Another common approach to evaluate explainability methods is to measure the sufficiency of the most salient tokens for predicting the target label~\cite{deyoung-etal-2020-eraser}. We also include a sufficiency estimate, but instead of fixing a threshold for the tokens to be removed, we measure the decrease of a model's performance, varying the proportion of excluded tokens. Other perturbation-based evaluation studies and measures exist~\cite{sundararajan2017axiomatic, Adebayo:2018:SCS:3327546.3327621}, but we consider the above, as it is the most widely applied.

Another direction of explainability evaluation is to compare the agreement of salient words annotated by humans to the saliency scores assigned by explanation techniques \cite{deyoung-etal-2020-eraser}. We also consider the latter and further study the agreement across model architectures, downstream tasks, and explainability methods. 
While we consider human annotations at the word level \cite{NIPS2018_8163, lei-etal-2016-rationalizing}, there are also datasets \cite{clark-etal-2019-boolq,khashabi-etal-2018-looking} with annotations at the sentence-level, which would require other model architectures, so we leave this for future work.

Existing studies for evaluating explainability heavily differ in their scope. Some concentrate on a \textbf{single model architecture} - BERT-LSTM \cite{deyoung-etal-2020-eraser}, RNN \cite{arras-etal-2019-evaluating}, CNN \cite{lertvittayakumjorn-toni-2019-human}, whereas a few consider \textbf{more than one} model \cite{guan2019towards, poerner-etal-2018-evaluating}. Some studies concentrate on one \textbf{particular dataset} \cite{guan2019towards,arras-etal-2019-evaluating}, while only a few generalize their findings over \textbf{downstream tasks} \cite{deyoung-etal-2020-eraser, vashishth2019attention}. Finally, existing studies focus on one \cite{vashishth2019attention} or a single group of explainability methods \cite{deyoung-etal-2020-eraser, Adebayo:2018:SCS:3327546.3327621}. Our study is the first to propose a unified comparison of different groups of explainability techniques across three text classification tasks and three model architectures.

% A recent study by \citet{hooker2019benchmark} for image analysis shows, however, that random perturbations of the model's input are exploiting the model's sensitivity to out-of-distribution data and observes that test set accuracy quickly erodes even if we start by replacing the most redundant features and we end up by evaluating an incorrect feature ranking. This poses the question of whether the previous perturbation-based approaches of evaluation can reveal the actual properties of the explanation approaches. We are the first to consider a fair comparison of explainability techniques for NLP.

\section{Evaluating Attribution Maps}
We now define a set of \propertyplural{} of explainability techniques, and propose how to quantify them. Similar notions can be found in related work~\cite{robnik2018perturbation, deyoung-etal-2020-eraser}, and we extend them to be generally applicable to downstream tasks. We first introduce the prerequisite notation. 
Let $X = \{(x_i, y_i, w_i)|i \in [1,N]\}$ be the test dataset, where each instance consists of a 
list of \emph{tokens} $x_i$ = $\{x_{i,j}| j \in [1, |x_i|]\}$, a \emph{gold label} $y_i$, and a 
\emph{gold saliency score} for each of the tokens in $x_i$: $w_i = \{w_{i,j} | j \in [1, |x_i|]\}$ 
with each $w_{i,j} \in \{0, 1\}$. Let $\omega$ be an explanation technique that, given a model $M$,  a class $c$, and a single instance $x_i$, computes saliency scores for each token in the input: 
\salscores $ = \{\omega_{(i,j),c}^{M} |j \in [1, |x_i|]\}$. Finally, let $M = M_1, \dots M_K$ be models with the same architecture, each trained from a randomly chosen seed, and let $M' = M_1', \dots M_K'$ be models of the same architecture, but with randomly initialized weights.

\textbf{Agreement with human rationales (HA)}. This \property{} measures the degree of overlap between saliency scores provided by human annotators, specific to the particular task, and the word saliency scores computed by an explainability technique on each instance. The property is a simple way of approximating the quality of the produced feature attributions. While it does not necessarily mean that the saliency scores explain the predictions of a model, we assume that explanations with high agreement scores would be more comprehensible for the end-user as they would adhere more to human reasoning. With this \property{}, we can also compare how the type and the performance of a model and/or dataset affect the agreement with human rationales when observing one type of explainability technique. 
% Moreover, by comparing the agreement of the saliency scores of a trained as opposed to a randomly initialized model, we observe the degree to which a saliency method remains \textit{invariant} given different model weights.

During evaluation, we provide an estimate of the average agreement of the explainability technique across the dataset. To this end, we start at the instance level and compute the Average Precision (AP) of produced saliency scores \salscores{} by comparing them to the gold saliency annotations $w_i$. Here, the label for computing the saliency scores is the gold label: $c=y_i$.
%, as it is also the one for the gold saliency map. 
Then, we compute the average across all instances, arriving at Mean AP (MAP):
\begin{equation}
%\small
\textrm{MAP}(\omega, M, X) = \frac{1}{N}\sum \limits_{i \in [1, N]} AP(w_{i}, \omega_{x_i, y_i}^M)   
\end{equation}
\textbf{Confidence Indication (CI)}. A token from a single instance can receive several saliency scores, indicating its contribution to the prediction of each of the classes. Thus, when a model recognizes a highly indicative pattern of the predicted class $k$, the tokens involved in the pattern would have highly positive saliency scores for this class and highly negative saliency scores for the remaining classes. On the other hand, when the model is not highly confident, we can assume that it is unable to recognize a strong indication of any class, and the tokens accordingly do not have high saliency scores for any class. Thus, the computed explanation of an instance $i$ should indicate the confidence $p_{i,k}$ of the model in its prediction.

We propose to measure the predictive power of the produced explanations for the confidence of the model. We start by computing the Saliency Distance (SD) between the saliency scores for the predicted class $k$ to the saliency scores of the other classes $K/k$ (Eq.~\ref{eq:conf1}). Given the distance between the saliency scores, we predict the confidence of the class with logistic regression (LR) and finally compute the Mean Absolute Error -- MAE (Eq.~\ref{eq:conf2}), of the predicted confidence to the actual one.
\begin{gather}
\textrm{SD} = \sum \limits_{j \in [0,|x|]}D(\omega_{x_{i,j}, k}^{M}, \omega_{x_{i,j}, K/k}^{M}) \label{eq:conf1}\\
\textrm{MAE}(\omega, M, X) = \sum \limits_{\substack{i \in [1, N]}} |p_{i,k} - \textrm{LR}(\salmap)|~\label{eq:conf2}
\end{gather}
For tasks with two classes, D is the subtraction of the saliency value for class k and the other class. For more than two classes, D is the concatenation of the max, min, and average across the differences of the saliency value for class k and the other classes. Low MAE indicates that model's confidence can be easily identified by looking at the produced explanations.

\textbf{Faithfulness (F)}. Since explanation techniques are employed to explain model predictions for a single instance, an essential property is that they are faithful to the model's inner workings and not based on arbitrary choices. A well-established way of measuring this property is by replacing a number of the most-salient words with a mask token~\cite{deyoung-etal-2020-eraser} and observing the drop in the model's performance. To avoid choosing an unjustified percentage of words to be perturbed, we produce several dataset perturbations by masking 0, 10, 20, \dots, 100\% of the tokens in order of decreasing saliency, thus arriving at $X^{\omega^0}$, $X^{\omega^{10}}$, \dots, $X^{\omega^{100}}$. Finally, to produce a single number to measure faithfulness, we compute the area under the threshold-performance curve (AUC-TP):
\begin{equation}
\begin{aligned}
\textrm{AUC-TP}(\omega, M, X) = \\ 
\textrm{AUC}([(i, P(M(X^{\omega^0}))-M(X^{\omega^i}))]) 
\end{aligned}
\end{equation}
where P is a task specific performance measure and $i \in [0, 10, \dots, 100]$. 
We also compare the AUC-TP of the saliency methods to a random saliency map to find whether there are explanation techniques producing saliency scores without any contribution over a random score. 

Using AUC-TP, we perform an ablation analysis
% by observing performance changes when masking the most salient words according to an explainability technique. 
% On the one hand, the approach 
which is a good approximation of whether the most salient words are also the most important ones for a model's prediction. 
However, some prior studies~\cite{feng-etal-2018-pathologies} find that models remain confident about their prediction even after stripping most input tokens, leaving a few that might appear nonsensical to humans. The \propertyplural\ that follow aim to facilitate a more in-depth analysis of the alignment between the inner workings of a model and produced saliency maps.

\textbf{Rationale Consistency (RC)}.
A desirable property of an explainability technique is to be consistent with the similarities in the reasoning paths of several models on a single instance. Thus, when two reasoning paths are similar, the scores provided by an explainability technique $\omega$ should also be similar, and vice versa. Note that we are interested in similar reasoning paths as opposed to similar predictions, as the latter does not guarantee analogous model rationales. For models with diverse architectures, we expect rationales to be diverse as well and to cause low consistency. Therefore, we focus on a set of models with the same architecture, trained from different random seeds as well as the same architecture, but with randomly initialized weights. The latter would ensure that we can have model pairs $(M_s, M_p)$ with similar and distant rationales. We further claim that the similarity in the reasoning paths could be measured effectively with the distance between the activation maps (averaged across layers and neural nodes) produced by two distinct models (Eq.~\ref{eq:consist1}). The distance between the explanation scores is computed simply by subtracting the two (Eq.~\ref{eq:consist2}). Finally, we compute  Spearman's $\rho$ between the similarity of the explanation scores and the similarity of the attribution maps (Eq.~\ref{eq:consist3}).

\begin{gather}
D(M_s, M_p, x_i) = D(M_s(x_i), M_p(x_i)) \label{eq:consist1}\\
D(M_s, M_p, x_i, \omega) = D(\omega_{x_i, y_i}^{M_s}, \omega_{x_i, y_i}^{M_p}) \label{eq:consist2} \\
\begin{split}
\rho(M_s, M_p, X, \omega) = \rho(D(M_s, M_p, x_i), \\
D(M_s, M_p, x_i, \omega)| i \in [1, N] ) \label{eq:consist3}
\end{split}
\end{gather}
The higher the positive correlation is, the more consistent the attribution method would be. We choose Spearman's $\rho$ as it measures the monotonic correlation between the two variables. On the other hand, Pearson's $\rho$ measures only the linear correlation, and we can have a non-linear correlation between the activation difference and the saliency score differences.
When subtracting saliency scores and layer activations, we also take the absolute value of the vector difference as the property should be invariant to order of subtraction.
An additional benefit of the property is that low correlation scores would also help to identify explainability techniques that are not faithful to a model's rationales.

\textbf{Dataset Consistency (DC)}. 
The next \property{} is similar to the above notion of rationale consistency but focuses on consistency across instances of a dataset as opposed to consistency across different models of the same architecture. In this case, we test whether instances with similar rationales also receive similar explanations. While Rationale Consistency compares instance explanations of the same instance for different model rationales, Dataset Consistency compares explanations for pairs of instances on the same model. We again measure the similarity between instances $x_i$ and $x_j$ by comparing their activation maps, as in Eq.~\ref{eq:consistdata1}. The next step is to measure the similarity of the explanations produced by an explainability technique $\omega$, which is the difference between the saliency scores as in Eq.~\ref{eq:consistdata2}. Finally, we measure Spearman's $\rho$  between the similarity in the activations and the saliency scores as in Eq.~\ref{eq:consistdata3}. We again take the absolute value of the difference.
\begin{gather}
D(M, x_i, x_j) = D(M(x_i), M(x_j)) \label{eq:consistdata1} \\
D(M, x_i, x_j, \omega) = D(\omega_{x_i, y_i}^{M}, \omega_{x_j, y_i}^{M})
\label{eq:consistdata2} \\
\begin{split}
\rho(M, X, \omega) = \rho(D(M, x_i, x_j), \\ D(M, x_i, x_j, \omega)| i, j \in [1, N])
\end{split} 
\label{eq:consistdata3}
\end{gather}
% Low correlation scores would again indicate attribution maps that are not faithful to model's rationales.

\section{Experiments}
\subsection{Datasets}
\begin{table}[t]
\centering
% \scriptsize
\begin{tabular}{p{34mm}p{65mm}p{28mm}p{20mm}}
\toprule
\textbf{Dataset} & \textbf{Example} & \textbf{Size} & \textbf{Length} \\ \midrule
e-SNLI \newline \cite{NIPS2018_8163} & \textit{Premise:} An adult dressed in black \textbf{holds a stick.} \newline \textit{Hypothesis:} An adult is walking away, \textbf{empty-handed}. \newline \textit{Label}: contradiction & 549 367 Train \newline 9 842 Dev \newline 9 824 Test & 27.4 inst. \newline 5.3 expl. \\ \midrule
Movie \newline Reviews \newline \cite{zaidan-etal-2007-using} \newline & \textit{Review:} he is one of \textbf{the most exciting martial artists on the big screen}, continuing to perform his own stunts and \textbf{dazzling audiences} with his flashy kicks and punches.\newline \textit{Class:} Positive & 1 399 Train \newline 199 Dev \newline 199 Test & 834.9 inst. \newline 56.18 expl. \\ \midrule
Tweet \newline Sentiment \newline Extraction \newline (TSE)~\footnotemark & 
\textit{Tweet:} im soo \textbf{bored}...im deffo missing my music channels \newline
\textit{Class:} Negative & 
21 983 Train \newline 2 747 Dev \newline 2 748 Test & 20.5 inst. \newline 9.99 expl. \\
\bottomrule
\end{tabular}
\caption{Datasets with human-annotated saliency explanations. The \textit{Size} column presents the dataset split sizes we use in our experiments. The \textit{Length} column presents the average number of instance tokens in the test set \textit{(inst.)} and the average number of human annotated explanation tokens \textit{(expl.)}.}
\label{tab:datasets6}
\end{table}
\footnotetext{\url{https://www.kaggle.com/c/tweet-sentiment-extraction}}

% We use datasets with human annotations of salient tokens as the post-hoc explainability methods generate saliency scores at the token level. 
% \footnote{The dataset CoS-E \cite{rajani2019explain} also provides word-level rationales. However, it requires complex (usually Transformer) models employing external common sense resources, while our goal is to train models with diverse architectures only on the input of the dataset. Hence, we do not use the dataset in our experimental setup.}
Table~\ref{tab:datasets6} provides an overview of the used datasets.
For e-SNLI, models predict inference -- contradiction, neutral, or entailment -- between sentence tuples. For the Movie Reviews dataset, models predict the sentiment -- positive, negative, or neutral -- of reviews with multiple sentences. Finally, for the TSE dataset, models predict tweets' sentiment -- positive, negative, or neutral.
The e-SNLI dataset provides three dataset splits with human-annotated rationales, which we use as training, dev, and test sets, respectively. The Movie Reviews dataset provides rationale annotations for nine out of ten splits. Hence, we use the ninth split as a test and the eighth split as a dev set, while the rest are used for training. Finally, the TSE dataset only provides rationale annotations for the training dataset, and we therefore  randomly split it into 80/10/10\% chunks for training, development and testing.

\subsection{Models}  
\begin{table}[t]
\centering
\small
\begin{tabular}{lrr}
\toprule
\textbf{Model} & \textbf{Val} & \textbf{Test} \\ \midrule
\multicolumn{3}{c}{\textbf{e-SNLI}} \\
\trans & 0.897 ($\pm$0.002) & 0.892 ($\pm 0.002$) \\
\transrand & 0.167 ($\pm$0.003) & 0.167 ($\pm 0.003$)\\
\cnn & 0.773 ($\pm$0.003) & 0.768 ($\pm 0.002$)\\
\cnnrand & 0.195 ($\pm 0.038$) & 0.194 ($\pm 0.037$) \\
\lstm & 0.794 ($\pm$0.005) & 0.793 ($\pm 0.009$)\\
\lstmrand & 0.176 ($\pm 0.013$) & 0.176 ($\pm 0.000$) \\ \midrule

\multicolumn{3}{c}{\textbf{Movie Reviews}}\\
\trans & 0.859 ($\pm$0.044) & 0.856 ($\pm$0.018) \\
\transrand & 0.335 ($\pm$0.003)& 0.333 ($\pm 0.000$)\\
\cnn & 0.831 ($\pm$0.014) & 0.773 ($\pm$0.005)\\
\cnnrand & 0.343 ($\pm$0.020) & 0.333 ($\pm 0.001$) \\
\lstm & 0.614 ($\pm$0.017) & 0.567 ($\pm 0.019$)\\
\lstmrand & 0.362 ($\pm$0.030) & 0.363 ($\pm 0.041$) \\ \midrule

\multicolumn{3}{c}{\textbf{TSE}} \\
\trans & 0.772 ($\pm$0.005) & 0.781 ($\pm 0.009$) \\
\transrand &0.165 ($\pm$0.025) & 0.171 ($\pm 0.022$)\\
\cnn & 0.708 ($\pm$0.007) &  0.730 ($\pm 0.007$)\\
\cnnrand & 0.221 ($\pm$0.060) & 0.226 ($\pm 0.055$) \\
\lstm & 0.701 ($\pm$0.005) & 0.727 ($\pm 0.004$)\\
\lstmrand & 0.196 ($\pm$0.070) & 0.204 ($\pm 0.070$) \\
\bottomrule
\end{tabular}
\caption{Models' $F_1$ score on the test and the validation datasets. The results present the average and the standard deviation of the Performance measure over five models trained from different seeds. The random versions of the models are again five models, but only randomly initialized, without training.}
\label{tab:modeleval}
\end{table}

We experiment with different commonly used base models, namely \cnn{}~\cite{fukushima1980neocognitron}, \lstm{} ~\cite{hochreiter1997long}, and the \trans{} ~\cite{vaswani2017attention} architecture BERT \cite{devlin-etal-2019-bert}. The selected models allow for a comparison of the explainability techniques on diverse model architectures.  Table~\ref{tab:modeleval} presents the performance of the separate models on the datasets.
% The \cnn{} model has access to a limited context window and exhibits an information bottleneck due to the compression of the windows. The \lstm{} model builds a continuous representation of the input instance in a sequential manner, where at each step, it decides which information to update, reset, and output to the next step. The \lstm{} model can account for long-distance relations, but its loss of information along the steps can become severe. Finally, in the \trans{} model, the representation of each token is constructed by attending to all tokens in the input instance with several attention heads. While the Transformer architecture does not encode the positional information in the same way as recurrent models do, the full sequence order is preserved using positional embeddings.

For the \cnn{} model, we use an embedding, a convolutional, a max-pooling, and a linear layer. The embedding layer is initialized with GloVe~\cite{pennington-etal-2014-glove} embeddings and is followed by a dropout layer. The convolutional layer computes convolutions with several window sizes and multiple-output channels with ReLU~\cite{hahnloser2000digital} as an activation function. The result is compressed down with a max-pooling layer, passed through a dropout layer, and into a fine linear layer responsible for the prediction. The final layer has a size equal to the number of classes in the dataset.

The \lstm{} model again contains an embedding layer initialized with the GloVe embeddings. The embeddings are passed through several bidirectional LSTM layers. The final output of the recurrent layers is passed through three linear layers and a final dropout layer.

For the \trans{} model, we fine-tune the pre-trained basic, uncased language model (LM)~\cite{Wolf2019HuggingFacesTS}. The fine-tuning is performed with a linear layer on top of the LM with a size equal to the number of classes in the corresponding task. Further implementation details for all of the models, as well as their $F_1$ scores, are presented in~\ref{appendix:A}.

\subsection{Explainability Techniques}
We select the explainability techniques to be representative of different groups -- gradient~\cite{sundararajan2017axiomatic, Simonyan2013DeepIC}, perturbation ~\cite{shapley1953value, zeiler2014visualizing} and simplification based~\cite{ribeiromodel, johansson2004truth}. 

Starting with the \textbf{gradient-based} approaches, we select \textit{Saliency}~\cite{Simonyan2013DeepIC} as many other gradient-based explainability methods build on it. It computes the gradient of the output w.r.t. the input. We also select two widely used improvements of the \textit{Saliency} technique, namely \textit{InputXGradient}~\cite{Kindermans2016InvestigatingTI}, and \textit{Guided Backpropagation}~\cite{springenberg2014striving}. InputXGradient additionally multiplies the gradient with the input and \textit{Guided Backpropagation} overwrites the gradients of ReLU functions so that only non-negative gradients are backpropagated.

From the \textbf{perturbation-based} approaches, we employ \textit{Occlusion}~\cite{zeiler2014visualizing}, which replaces each token with a baseline token (as per standard, we use the value zero) and measures the change in the output. Another popular perturbation-based technique is the \textit{Shapley Value Sampling}~\cite{castro2009polynomial}. It is based on the Shapley Values approach that computes the average marginal contribution of each word across all possible word perturbations. The Sampling variant allows for a faster approximation of Shapley Values by considering only a fixed number of random perturbations as opposed to all possible perturbations.

Finally, we select the \textbf{simplification-based} explanation technique LIME~\cite{ribeiromodel}. For each instance in the dataset, LIME trains a linear model to approximate the local decision boundary for that instance.
% which trains local linear approximations of the complex black-box function for each instance in the dataset.

\textbf{Generating explanations.} 
The saliency scores from each of the explainability methods are generated for each of the classes in the dataset. As all of the gradient approaches provide saliency scores for the embedding layer (the last layer that we can compute the gradient for), we have to aggregate them to arrive at one saliency score per input token. As we found different aggregation approaches in related studies~\cite{bansal2016ask, deyoung-etal-2020-eraser}, we employ the two most common methods -- 
L2 norm and averaging (denoted as $\mu$ and $\ell2$ in the explainability method names). 
% For completeness, we present results with both aggregation methods, denoted as a superscript next to the names of an explanation method.
\section{Results and Discussion}
We report the measures of each \property{} as well as FLOPs as a measure of the computing time used by the particular method. For all \propertyplural, we also include the randomly assigned saliency as a baseline.

\subsection{Results}
\begin{table}[t]
\centering
% \small
\fontsize{10}{10}\selectfont
\begin{tabular}{llrrr}
\toprule
\textbf{Model} & \textbf{Saliency} & \textbf{e-SNLI} & \textbf{IMDB} & \textbf{TSE} \\
\midrule
% \\
\multirow{10}{*}{\trans} & \rand & 0.201 & 0.517 & 0.185  \\
&\shapsamp & 0.479 & 0.481 & 0.667  \\
&\lime & \textbf{0.809} & 0.604 & 0.553  \\
&\occlusion & 0.523 & 0.323 & 0.556  \\
&\salmean & 0.772 & 0.671 & \underline{0.707}  \\
&\salnorm & 0.781 & \textbf{0.687} & 0.696   \\
&\inputxmean & 0.364 & 0.432 & 0.307  \\
&\inputxnorm & \underline{0.796} & \underline{0.676} & \textbf{0.754}  \\
&\guidedmean & 0.468 & 0.236 & 0.287  \\
&\guidednorm & 0.782 & \underline{0.676} & 0.685  \\
\midrule
% \multicolumn{4}{c}{\cnn}\\
\multirow{10}{*}{\cnn} &\rand & 0.209 & 0.468 & 0.384  \\
&\shapsamp & 0.460 & 0.648 & 0.630  \\
&\lime & 0.571 & 0.572 & \textbf{0.681} \\
&\occlusion & 0.554 & 0.411 & 0.594  \\
&\salmean & 0.853 & 0.712 & 0.595  \\
&\salnorm & \underline{0.875} & \textbf{0.796} & 0.631  \\
&\inputxmean & 0.576 & 0.662 & 0.613  \\
&\inputxnorm & \textbf{0.881} & 0.759 & \underline{0.636}  \\
&\guidedmean & 0.403 & 0.346 & 0.438  \\
&\guidednorm & \underline{0.875} & \underline{0.788} & 0.628  \\
\midrule
% \multicolumn{4}{c}{\lstm}\\
\multirow{10}{*}{\lstm}&\rand & 0.166 & 0.343 & 0.225  \\
&\shapsamp & 0.606 & 0.605 & 0.526  \\
&\lime & 0.759 & 0.233 & 0.630 \\
&\occlusion & 0.609 & 0.589 & 0.681  \\
&\salmean & 0.795 & 0.568 & 0.702  \\
&\salnorm & 0.800 & 0.583 & \textbf{0.704}  \\
&\inputxmean & 0.432 & 0.481 & 0.441  \\
&\inputxnorm & \textbf{0.820} & \textbf{0.685} & 0.693  \\
&\guidedmean & 0.492 & 0.553 & 0.410  \\
&\guidednorm & \underline{0.805} & \underline{0.660} & \textbf{0.720} \\
\bottomrule
\end{tabular}
\caption{Mean of the \property{} measures for all tasks and models. The best result for the particular model architecture and downstream task is in bold and the second-best is underlined.}
\label{tab:meanprop}
\end{table}
Of the three model architectures, unsurprisingly, the \trans\ model performs best, while the \cnn\ and the \lstm\ models are close in performance. It is only for the IMDB dataset that the \lstm\ model performs considerably worse than the \cnn, which we attribute to the fact that the instances contain a large number of tokens, as shown in Table~\ref{tab:datasets6}. As this is not the core focus of this paper, detailed results can be found in the supplementary material.
% .~\ref{appendix:A}.

\textbf{Overall results.} Table~\ref{tab:meanprop} presents the mean of all properties across tasks and models with all property measures normalized to be in the range [0,1]. 
% For more detailed results on each \property, we refer the reader to Appendix~\ref{appendix:B}. 
We see that gradient-based explainability techniques always have the best or the second-best performance for the \propertyplural{} across all three model architectures and all three downstream tasks. Note that, \inputxmean{} and \guidedmean{}, which are computed with a mean aggregation of the scores, have some of the worst results. We conjecture that this is due to the large number of values that are averaged -- the mean smooths out any differences in the values. In contrast, the L2 norm aggregation amplifies the presence of large and small values in the vector. From the non-gradient based explainability methods, \lime{} has the best performance, where in two out of nine cases it has the best performance. It is followed by \shapsamp{} and \occlusion{}. We can conclude that the occlusion based methods overall have the worst performance according to the \propertyplural{}.

Furthermore, we see that the explainability methods achieve better performance for the e-SNLI and the TSE datasets with the \trans{} and \lstm{} architectures, whereas the results for the IMDB dataset are the worst. We hypothesize that this is due to the longer text of the input instances in the IMDB dataset. The scores also indicate that the explainability techniques have the highest \property{} measures for the \cnn{} model with the e-SNLI and the IMDB datasets, followed by the \lstm{}, and the \trans{} model. We suggest that the performance of the explainability tools can be worse for large complex architectures with a huge number of neural nodes, like the \trans{} one, and perform better for small, linear architectures like the \cnn{}.
% However, we saw that due to the different nature of the properties and the characteristics of the models and the dataset, there are techniques that are better suited for a single diagnostic property and particular cases.
\begin{figure}
\centering
\includegraphics[width=335pt]{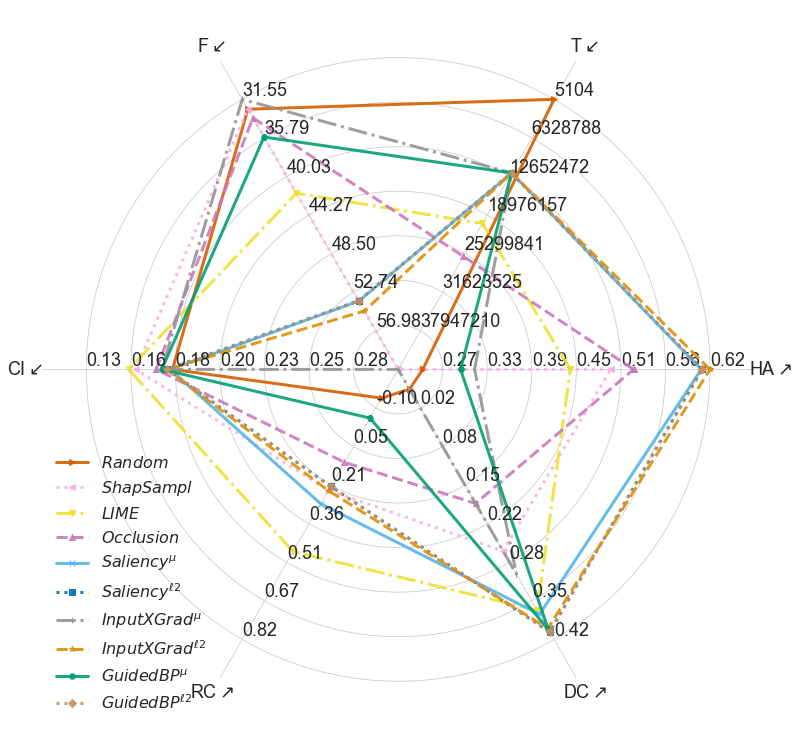}
\caption{Diagnostic property evaluation for all explainability techniques, on the e-SNLI dataset, \trans\ model. 
The $\nearrow$ and $\swarrow$ signs indicate that higher, correpspondingly lower, values of the property measure are better.}
\label{fig:spider11}
\end{figure}

\begin{figure}
\centering
\includegraphics[width=335pt]{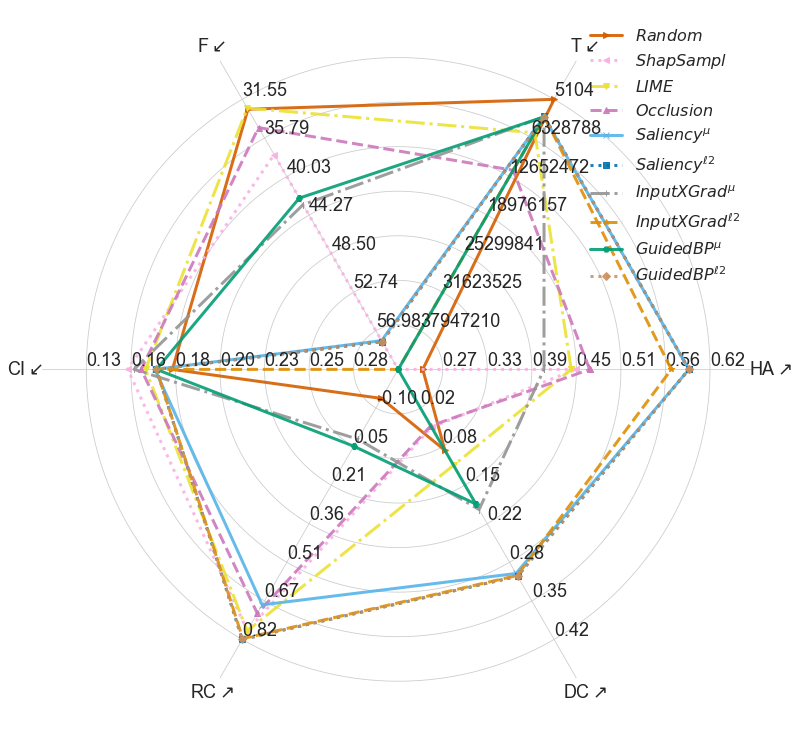}
\caption{Diagnostic property evaluation for 
all explainability techniques, on the e-SNLI dataset, \cnn\ model. 
The $\nearrow$ and $\swarrow$ signs indicate that higher, correpspondingly lower, values of the property measure are better.}
\label{fig:spider12}
\end{figure}

\begin{figure}
\centering
\includegraphics[width=335pt]{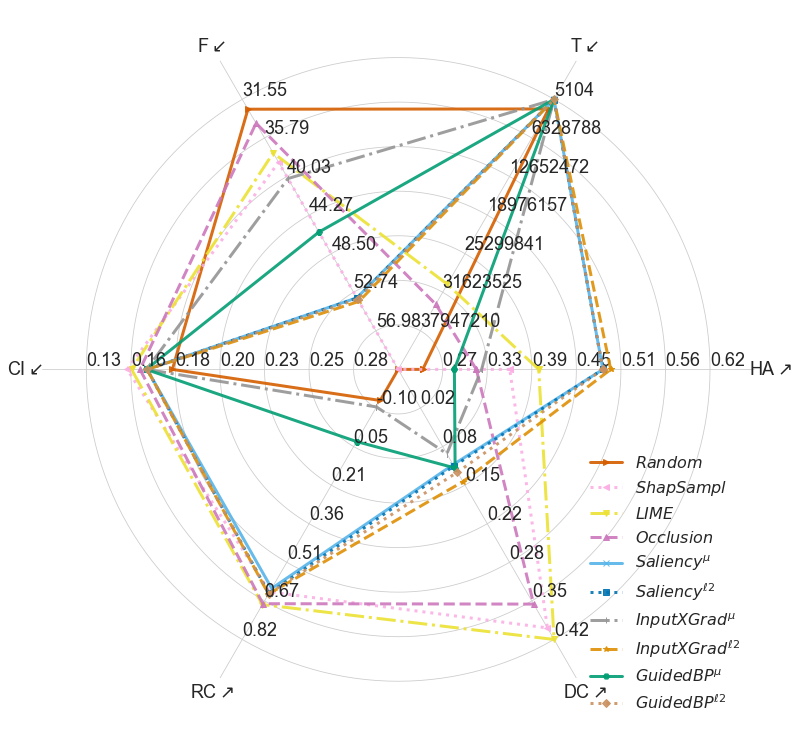}
\caption{Diagnostic property evaluation for 
all explainability techniques, on the e-SNLI dataset, \lstm\ model. 
The $\nearrow$ and $\swarrow$ signs indicate that higher, correpspondingly lower, values of the property measure are better.}
\label{fig:spider13}
\end{figure}

\textbf{Diagnostic property performance.} Figures~\ref{fig:spider11}, \ref{fig:spider12}, \ref{fig:spider13} show the performance of each explainability technique for all \propertyplural\ on the e-SNLI dataset. The TSE and IMDB datasets show similar tendencies and corresponding figures can be found in the supplementary material.

\textbf{Agreement with human rationales.}
We observe that the best performing explainability technique for the \trans{} model is \inputxnorm{} followed by the gradient-based ones with L2 norm aggregation.
% , which also holds true across the three tasks. 
While for the \cnn{} and the \lstm{} models, we observe similar trends, their MAP scores are always lower than for the \trans, which indicates a correlation between the performance of a model and its agreement with human rationales. 
% To further investigate this, we have estimated the MAP difference between the correctly and wrongly classified instances. We found that the set of correctly classified instances always has higher MAP with an average distance of:  0,045 for e-SNLI,  0,015 for IMDB,  0,005 for TSE datasets, and  0,024 for \trans, 0,043 for \cnn, and  0,002 for the \lstm{} models. 
% Finally, we note that the IMDB dataset has very low MAP scores, which is mainly dues to the large number of tokens per instance. 
Furthermore, the MAP scores of the \cnn{} model are higher than for the \lstm{} model, even though the latter achieves higher $F_1$ scores on the e-SNLI dataset.
% and almost the same performance on the TSE dataset. 
This might indicate that the representations of the \lstm{} model are less in line with human rationales.
Finally, we note that the mean aggregations of the gradient-based explainability techniques have MAP scores close to or even worse than those from the randomly initialized models.
% This is in line with studies that train models with human-annotated saliency to improve its performance~\cite{camburu2018snli}. 

\textbf{Faithfulness.} 
% Across all models and datasets,
We find that gradient-based techniques have the best performance for the Faithfulness \property. On the e-SNLI dataset, it is particularly \inputxnorm{}, which performs well across all model architectures. We further find that the \cnn{} exhibits the highest Faithfulness scores for seven out of nine explainability methods. We hypothesize that this is due to the simple architecture with relatively few neural nodes compared to the recurrent nature of the \lstm{} model and the large number of neural nodes in the \trans{} architecture. Finally, models with high Faithfulness scores do not necessarily have high Human agreement scores and vice versa. This suggests that these two are indeed separate \propertyplural{}, and the first should not be confused with estimating the faithfulness of the techniques.

% The only exception is for the IMDB dataset, where the \lstm{} model is the one that the explainability techniques are more faithful to. The latter comes as a surprise as it has the lowest performance. We assume that the length of the input is an important factor for the choice of a model architecture that can be easier interpreted. We also find that for IMDB, the explainability tools on the \cnn{} and \trans{} models perform worse than randomly sampled saliency scores, which further supports the assumption that models for longer text sequences are harder for interpreting.
% The explainability tools that select the most salient words for the model first are \salmean, \salnorm, \guidednorm, and \inputxnorm. On the contrary, the mean aggregations \guidedmean{} and \inputxmean{} perform similarly, and sometimes even worse, than a random assignment of saliency scores. While for the \trans{} model, only the \inputxmean{} attribution is worse than the random one, for the \cnn{} and the \lstm{} model, \shapsamp, \lime{} and \occlusion{} perform considerably worse than a random attribution. 
\textbf{Confidence Indication.} 
We find that the Confidence Indication of all models is predicted most accurately by the \shapsamp{}, \lime{}, and \occlusion{} explainability methods. This result is expected, as they compute the saliency of words based on differences in the model's confidence using different instance perturbations. We further find that the \cnn{} model's confidence 
% across all datasets 
is better predicted with \inputxmean{}. The lowest MAE with the balanced dataset is for the \cnn{} and \lstm{} models. We hypothesize that this could be due to these models' overconfidence, which makes it challenging to detect when the model is not confident of its prediction.

% We also note that for the IMDB dataset, the explainability tools apart from the best one, cannot be used as a good indication for the confidence of the model as they perform worse than randomly assigned saliency scores.

\textbf{Rationale Consistency.} 
There is no single universal explainability technique that achieves the highest score for Rationale Consistency property. 
% This is also true across all of the datasets. 
We see that \lime{} can be good at achieving a high performance, which is expected, as it is trained to approximate the model's performance. The latter is beneficial, especially for models with complex architectures like the \trans. The gradient-based approaches also have high Rationale Consistency scores. We find that the \occlusion{} technique is the best performing for the \lstm{} across all tasks, as it is the simplest of the explored explainability techniques, and does not inspect the model's internals or try to approximate them. This might serve as an indication that \lstm{} models, due to their recurrent nature, can be best explained with simple perturbation based methods that do not examine a model's reasoning process. 
% Finally, for the IMDB dataset, the Rationale Consistency scores for the \lstm{} model are mostly worse than random ones.

\textbf{Dataset Consistency.} Finally, the results for the Dataset Consistency property show low to moderate correlations of the explainability techniques with similarities across instances in the dataset. The correlation is present for LIME and the gradient-based techniques, again with higher scores for the L2 aggregated gradient-based methods. 

\textbf{Overall.} To summarise, the proposed list of \propertyplural{} allows for assessing existing explainability techniques from different perspectives and supports the choice of the best performing one. Individual property results indicate that gradient-based methods have the best performance.
The only strong exception to the above is the better performance of \shapsamp{} and \lime{} for the Confidence Indication \property. However, \shapsamp{}, \lime{} and \occlusion{} take considerably more time to compute and have worse performance for all other \propertyplural. 

\section{Conclusion}
We proposed a comprehensive list of \propertyplural{} for the evaluation of explainability techniques from different perspectives. We further used them to compare and contrast different groups of explainability techniques on three downstream tasks and three diverse architectures. We found that gradient-based explanations are the best for all of the three models and all of the three downstream text classification tasks that we consider in this work. Other explainability techniques, such as \shapsamp{}, \lime{} and \occlusion{} take more time to compute, and are in addition considerably less faithful to the models and less consistent with the rationales of the models and similarities in the datasets. 
% We also find that the explainability methods can have different property measures depending on the task and model architecture at hand. They achieve lower property measures for the IMDB dataset, where the input text is longer.
% In future work, we plan to employ the proposed \propertyplural{} to develop explainability techniques that optimise them. We further plan to extend our study to tasks with sentence-level saliency or to text generated explanations. Finally, we would like to compare the automated property measures with human evaluations of the same properties.

\section*{Acknowledgements}
$\begin{array}{l}\includegraphics[width=1cm]{euflag2.png} \end{array}$ This project has received funding from the European Union's Horizon 2020 research and innovation programme under the Marie Sk\l{}odowska-Curie grant agreement No 801199.

%\section*{Acknowledgments}
%\begin{wrapfigure}{L}{0.10\columnwidth}
%\vspace{-13pt}
%\includegraphics[width=0.17\columnwidth]{euflag2.png}
%\vspace{-25pt}
%\end{wrapfigure}
%This project has received funding from the European Union’s Horizon 2020 research and innovation programme under the Marie Skłodowska-Curie grant agreement No 801199.

% \section*{Acknowledgments}

% The acknowledgments should go immediately before the references.  Do
% not number the acknowledgments section. Do not include this section
% when submitting your paper for review. \\

% \noindent \textbf{Preparing References:} \\
% Include your own bib file like this:
% \verb|\bibliographystyle{acl_natbib}|
% \verb|\bibliography{acl2019}| 

% \clearpage
% \appendix
\section{Appendices}
\subsection{Experimental Setup}~\label{appendix:A}
\begin{table}[t]
\centering
% \small
\begin{tabular}{lrr}
\toprule
\textbf{Model} & \textbf{Time} & \textbf{Score} \\ \midrule
\multicolumn{3}{c}{\textbf{e-SNLI}} \\
\trans & 244.763 ($\pm$62.022) & 0.523 ($\pm$0.356) \\
\cnn & 195.041 ($\pm$53.994) & 0.756 ($\pm$0.028) \\
\lstm & 377.180 ($\pm$232.918) & 0.708 ($\pm$0.205)\\

\multicolumn{3}{c}{\textbf{Movie Reviews}}\\
\trans & 3.603 ($\pm$0.031) & 0.785 ($\pm$0.226) \\
\cnn & 4.777 ($\pm$1.953) & 0.756 ($\pm$0.058)\\
\lstm & 5.344 ($\pm$1.593) & 0.584 ($\pm$0.061) \\

\multicolumn{3}{c}{\textbf{TSE}} \\
\trans & 9.393 ($\pm$1.841) & 0.783 ($\pm$0.006) \\
\cnn & 2.240 ($\pm$0.544) & 0.730 ($\pm$0.035) \\
\lstm & 3.781 ($\pm$1.196) & 0.713 ($\pm$0.076) \\
\bottomrule
\end{tabular}
\caption{Hyper-parameter tuning details. \textit{Time} is the average time (mean and standard deviation in brackets) measured in minutes required for a particular model with all hyper-parameter combinations. \textit{Score} is the mean and standard deviation of the performance on the validation set as a function of the number of the different hyper-parameter searches.}
\label{tab:modeleval_app}
\end{table}

\textbf{Machine Learning Models}. The models used in our experiments are trained on the training splits, and the parameters are selected according to the development split. We conducted fine-tuning in a grid-search manner with the ranges and parameters we describe next. We use superscripts to indicate when a parameter value was selected for one of the datasets e-SNLI -- 1, Movie Review -- 2, and TSE -- 3. For the \cnn{} model, we experimented with the following parameters: embedding dimension $\in$ \{50, 100, 200, 300$^{1, 2, 3}$\}, batch size $\in$ \{16$^{2}$, 32, 64$^{3}$, 128, 256$^{1}$\}, dropout rate $\in$ \{0.05$^{1,2,3}$, 0.1, 0.15, 0.2\}, learning rate for an Adam optimizer $\in$ \{0.01, 0.03, 0.001$^{2, 3}$, 0.003, 0.0001$^{1}$, 0.0003\}, window sizes $\in$ \{[2, 3, 4]$^{2}$, [2, 3, 4, 5], [3, 4, 5]$^{3}$, [3, 4, 5, 6],$ $[4, 5, 6], [4, 5, 6, 7]$^{1}$\}, and number of output channels $\in$ \{50$^{2, 3}$, 100, 200, 300$^{1}$\}. We leave the stride and the padding parameters to their default values -- one and zero. 

For the \lstm{} model we fine-tuned over the following grid of parameters: embedding dimension $\in$ \{50, 100$^{1, 2}$, 200$^{3}$, 300\}, batch size $\in$ \{16$^{2,3}$, 32, 64, 128, 256$^{1}$\}, dropout rate $\in$ \{0.05$^{3}$, 0.1$^{1, 2}$, 0.15, 0.2\}, learning rate for an Adam optimizer $\in$ \{0.01$^{1}$, 0.03$^{2}$, 0.001$^{2, 3}$, 0.003, 0.0001, 0.0003\}, number of LSTM layers $\in$ \{1$^{2, 3}$, 2, 3, 4$^{1}$\}, LSTM hidden layer size $\in$ \{50, 100$^{1, 2, 3}$, 200, 300\}, and size of the two linear layers $\in$ \{[50, 25]$^{2}$, [100, 50]$^{1}$, [200, 100]$^{3}$\}. We also experimented with other numbers of linear layers after the recurrent ones, but having three of them, where the final was the prediction layer, yielded the best results. 

The \cnn{} and \lstm{} models are trained with an early stopping over the validation accuracy with a patience of five and a maximum number of training epochs of 100. We also experimented with other optimizers, but none yielded improvements.

Finally, for the \trans{} model we fine-tuned the pre-trained basic, uncased LM~\cite{Wolf2019HuggingFacesTS}(110M parameters) where the maximum input size is 512, and the hidden size of each layer of the 12 layers is 768. We performed a grid-search over learning rate of $\in \{1e-5, 2e-5^{1, 2}, 3e-5^{3}, 4e-5, 5e-5\}$. The models were trained with a warm-up period where the learning rate increases linearly between 0 and 1 for 0.05\% of the steps found with a grid-search. We train the models for five epochs with an early stopping with patience of one as the Transformer models are easily fine-tuned for a small number of epochs.

All experiments were run on a single NVIDIA TitanX GPU with 8GB, and 4GB of RAM and 4 Intel Xeon Silver 4110 CPUs.

The models were evaluated with macro $F_1$ score, which can be found here \url{https://scikit-learn.org/stable/modules/generated/sklearn.metrics} and is defined as follows:
\begin{equation*}
   Precision (P) = \frac{\mathrm{TP}}{\mathrm{TP} + \mathrm{FP}} 
\end{equation*}
\begin{equation*}
   Recall (R) = \frac{\mathrm{TP}}{\mathrm{TP} + \mathrm{FN}} 
\end{equation*}
\begin{equation*}
   F_1 = \frac{2*\mathrm{P}*\mathrm{R}}{\mathrm{P}+\mathrm{R}} 
\end{equation*}
where TP is the number of true positives, FP is the number of false positives, and FN is the number of false negatives.

\textbf{Explainability generation}. When evaluating the Confidence Indication property of the explainability measures, we train a logistic regression for 5 splits and provide the MAE over the five test splits. As for some of the models, e.g. \trans{}, the confidence is always very high, the LR starts to predict only the average confidence. To avoid this, we additionally randomly up-sample the training instances with a smaller confidence, making the number of instances in each confidence interval [0.0-0.1],\dots[0.9-1.0]) to be the same as the maximum number of instances found in one of the separate intervals.

For both Rationale and Dataset Consistency properties, we consider Spearman's $\rho$. While 
Pearson's $\rho$ measures only the linear correlation between two variables (a change in one variable should be proportional to the change in the other variable), Spearman's $\rho$ measures the monotonic correlation (when one variable increases, the other increases, too). In our experiments, we are interested in the monotonic correlation as all activation differences don't have to be linearly proportional to the differences of the explanations and therefore measure Spearman's $\rho$. 

The Dataset Consistency property is estimated over instance pairs from the test dataset. As computing it for all possible pairs in the dataset is computationally expensive, we select 2 000 pairs from each dataset in order of their decreasing word overlap and sample 2 000 from the remaining instance pairs. This ensures that we compute the \property\ on a set containing tuples of similar and different instances. 

Both the Dataset Consistency property and the Rationale Consistency property estimate the difference between the instances based on their activations. For the \lstm\ model, the activations of the LSTM layers are limited to the output activation also used for prediction as it isn't possible to compare activations with different lengths due to the different token lengths of the different instances. We also use min-max scaling of the differences in the activations and the saliencies as the saliency scores assigned by some explainability techniques are very small. 

% For the spider figures and the mean of the diagnostic property measures for each explainability technique, as presented in Table \ref{tab:meanprop}, we normalise all measures in the range $[0, 1]$ across all scores for the particular dataset and model. For the properties, where lower scores are better, we take 1 - score of the property. Finally, for the Consistency property, we use the MAE-up score from Table \ref{tab:confidence}.

\subsection{Spider Figure for the IMDB dataset}~\label{appendix:C}
\begin{figure}
\centering
\includegraphics[width=335pt]{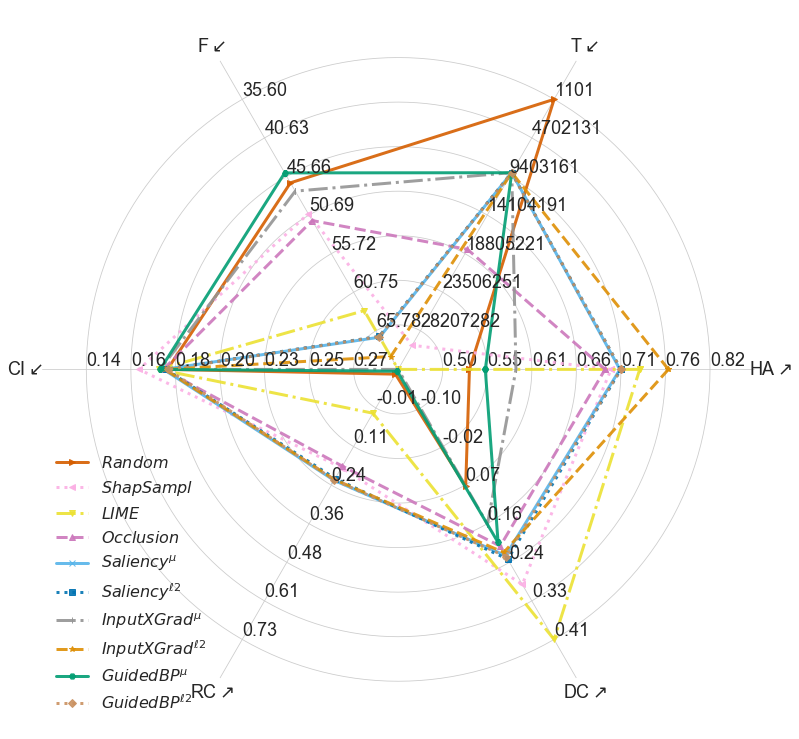}
\caption{Diagnostic property evaluation for 
all explainability techniques, on the TSE dataset, \trans\ model. 
The $\nearrow$ and $\swarrow$ signs indicate that higher, correspondingly lower, values of the property measure are better.}
\label{fig:spider31}
\end{figure}

\begin{figure}
\centering
\includegraphics[width=335pt]{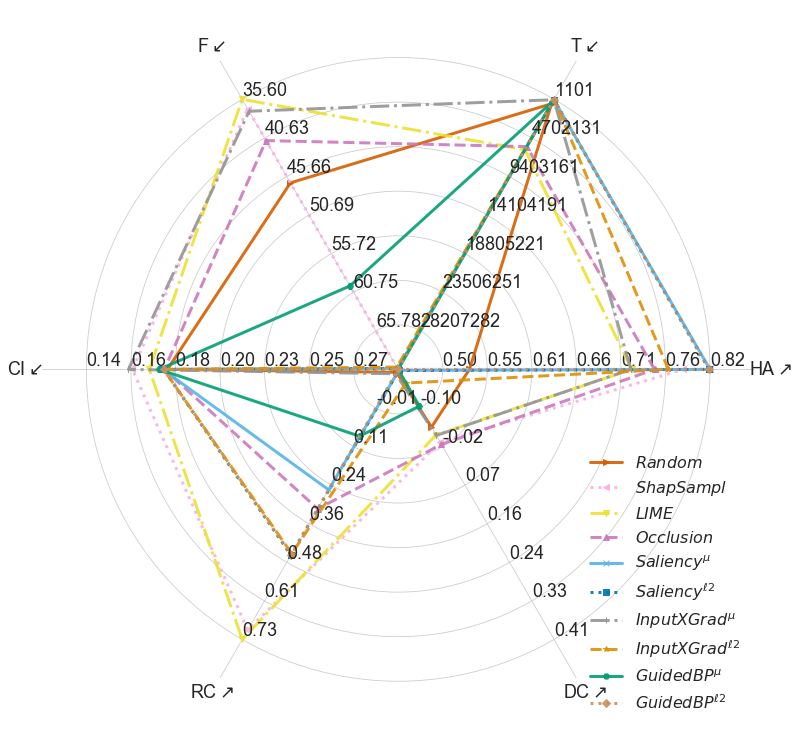}
\caption{Diagnostic property evaluation for 
all explainability techniques, on the TSE dataset, \cnn\ model. 
The $\nearrow$ and $\swarrow$ signs indicate that higher, correspondingly lower, values of the property measure are better.}
\label{fig:spider32}
\end{figure}

\begin{figure}
\centering
\includegraphics[width=335pt]{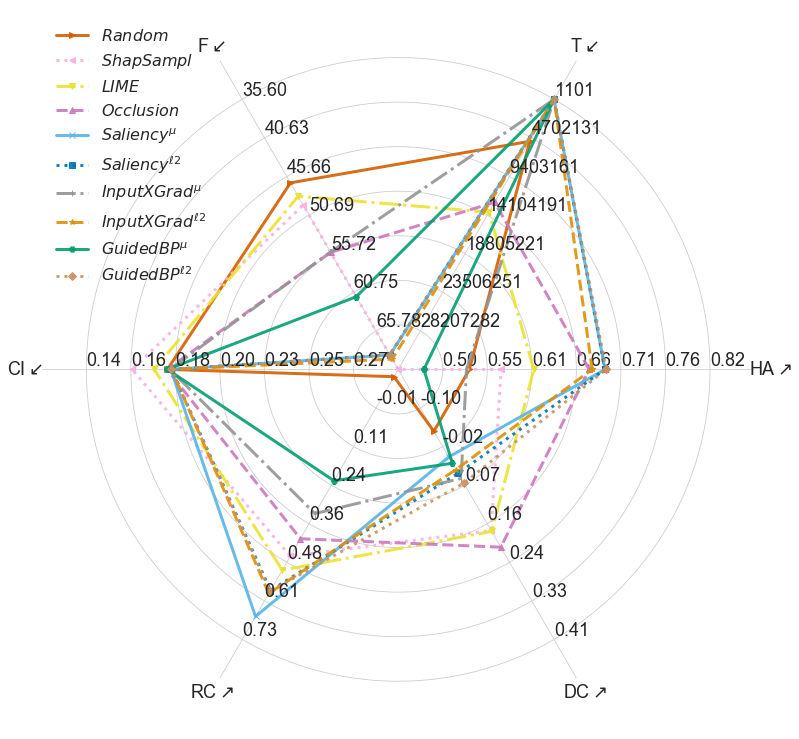}
\caption{Diagnostic property evaluation for 
all explainability techniques, on the TSE dataset, \lstm\ model. 
The $\nearrow$ and $\swarrow$ signs indicate that higher, correspondingly lower, values of the property measure are better.}
\label{fig:spider33}
\end{figure}

\begin{figure}
\centering
\includegraphics[width=335pt]{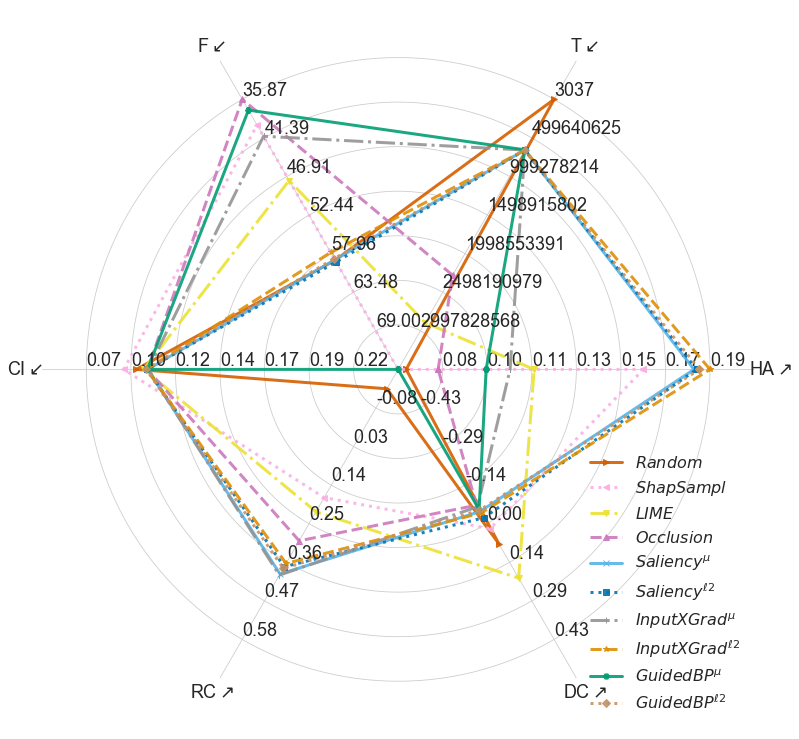}
\caption{Diagnostic property evaluation for 
all explainability techniques, on the IMDB dataset, \trans\ model. 
The $\nearrow$ and $\swarrow$ signs following the names of each explainability method indicate that higher, correspondingly lower, values of the property measure are better.}
\label{fig:spider21}
\end{figure}

\begin{figure}
\centering
\includegraphics[width=335pt]{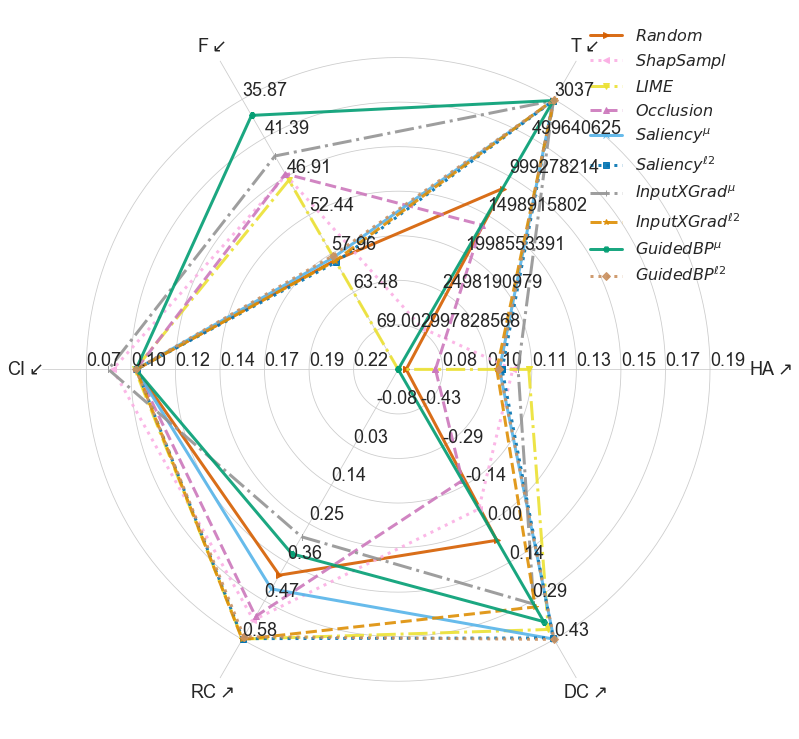}
\caption{Diagnostic property evaluation for 
all explainability techniques, on the IMDB dataset, \cnn\ model. 
The $\nearrow$ and $\swarrow$ signs following the names of each explainability method indicate that higher, correspondingly lower, values of the property measure are better.}
\label{fig:spider22}
\end{figure}

\begin{figure}
\includegraphics[width=335pt]{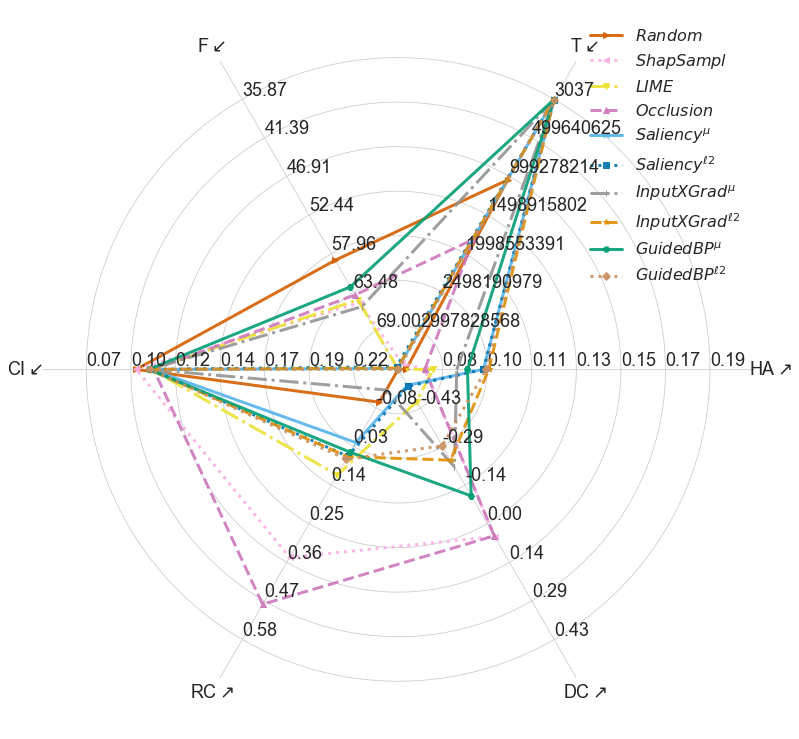}
\caption{Diagnostic property evaluation for 
all explainability techniques, on the IMDB dataset, \lstm\ model. 
The $\nearrow$ and $\swarrow$ signs following the names of each explainability method indicate that higher, correspondingly lower, values of the property measure are better.}
\label{fig:spider23}
\end{figure}

Figures \ref{fig:spider31}, \ref{fig:spider32}, \ref{fig:spider33} present the diagnostic property evaluation for the TSE dataset for the \trans{}, \cnn{}, \lstm{} models correspondingly. 
Figures \ref{fig:spider21}, \ref{fig:spider22}, \ref{fig:spider23} present the diagnostic property evaluation for the IMDB dataset for the \trans{}, \cnn{}, \lstm{} models correspondingly. 

\subsection{Detailed explainability techniques evaluation results.}
~\label{appendix:B}

Tables \ref{tab:human}, \ref{tab:faith}, \ref{tab:confidence}, \ref{tab:consistency:rat}, and \ref{tab:consistency:data} present detailed diagnostic property results. 

\begin{landscape}
\begin{table}
\centering
\tiny
\begin{tabular}{l@{\hspace{0.7\tabcolsep}}@{\hskip 0.05in \vline \hskip 0.05in}r@{\hspace{0.7\tabcolsep}}r@{\hspace{0.7\tabcolsep}}l@{\hspace{0.5\tabcolsep}}@{\hskip 0.05in \vline \hskip 0.05in}r@{\hspace{0.7\tabcolsep}}r@{\hspace{0.7\tabcolsep}}l@{\hspace{0.5\tabcolsep}}@{\hskip 0.05in \vline \hskip 0.05in}r@{\hspace{0.7\tabcolsep}}r@{\hspace{0.7\tabcolsep}}l@{\hspace{0.5\tabcolsep}}}
\toprule
\textbf{Explain.}&\multicolumn{3}{c}{\textbf{e-SNLI}}&\multicolumn{3}{c}{\textbf{IMDB}}&\multicolumn{3}{c}{\textbf{TSE}}\\
&\textbf{MAP}&\textbf{MAP RI}&\textbf{FLOPs}&\textbf{MAP}&\textbf{MAP RI}&\textbf{FLOPs}&\textbf{MAP}&\textbf{MAP RI}&\textbf{FLOPs}\\
\midrule
\rand&.297 ($\pm$.001)&--&6.12e+3 ($\pm$4.6e+1)&.079 ($\pm$.001)&--&9.41e+4 ($\pm$1.8e+2)&.573 ($\pm$.001)&--&4.62e+3 ($\pm$2.2e+1)\\
\midrule
\multicolumn{10}{c}{\trans}\\
\shapsamp&.511 ($\pm$.004)&.292 ($\pm$.011)&1.78e+7 ($\pm$5.5e+5)&.168 ($\pm$.003)&.084 ($\pm$.001)&3.00e+9 ($\pm$1.3e+8)&.716 ($\pm$.003)&.575 ($\pm$.027)&1.29e+7 ($\pm$2.0e+6)\\
\lime&.465 ($\pm$.008)&.264 ($\pm$.004)&2.39e+5 ($\pm$1.5e+4)&.127 ($\pm$.004)&.075 ($\pm$.004)&4.98e+8 ($\pm$1.4e+8)&.745 ($\pm$.003)&.570 ($\pm$.028)&2.82e+7 ($\pm$1.6e+6)\\
\occlusion&.537 ($\pm$.014)&.292 ($\pm$.009)&6.33e+5 ($\pm$1.0e+3)&.091 ($\pm$.001)&.084 ($\pm$.001)&8.05e+7 ($\pm$4.5e+5)&.710 ($\pm$.008)&.577 ($\pm$.012)&5.86e+5 ($\pm$1.6e+2)\\
\salmean&.614 ($\pm$.003)&.255 ($\pm$.008)&5.38e+4 ($\pm$1.8e+2)&.187 ($\pm$.005)&.079 ($\pm$.001)&6.59e+5 ($\pm$1.8e+3)&.725 ($\pm$.011)&.499 ($\pm$.002)&4.93e+4 ($\pm$2.1e+2)\\
\salnorm&.615 ($\pm$.003)&.255 ($\pm$.009)&5.39e+4 ($\pm$1.3e+2)&.188 ($\pm$.006)&.078 ($\pm$.001)&6.62e+5 ($\pm$8.4e+2)&.726 ($\pm$.014)&.498 ($\pm$.001)&4.93e+4 ($\pm$1.4e+2)\\
\inputxmean&.356 ($\pm$.005)&.280 ($\pm$.016)&5.38e+4 ($\pm$1.8e+2)&.118 ($\pm$.003)&.083 ($\pm$.001)&6.60e+5 ($\pm$4.5e+3)&.620 ($\pm$.008)&.558 ($\pm$.011)&4.92e+4 ($\pm$1.4e+2)\\
\inputxnorm&\underline{\textbf{.624 ($\boldsymbol \pm$.004)}}&.254 ($\pm$.013)&5.39e+4 ($\pm$1.5e+2)&\underline{\textbf{.193 ($\boldsymbol \pm$.005)}}&.079 ($\pm$.001)&6.62e+5 ($\pm$2.1e+3)&\textbf{.774 ($\boldsymbol \pm$.009)}&.499 ($\pm$.005)&4.92e+4 ($\pm$8.0e+1)\\
\guidedmean&.340 ($\pm$.012)&.281 ($\pm$.025)&5.39e+4 ($\pm$1.8e+2)&.109 ($\pm$.003)&.086 ($\pm$.005)&6.54e+5 ($\pm$7.5e+3)&.589 ($\pm$.006)&.567 ($\pm$.008)&4.94e+4 ($\pm$4.1e+2)\\
\guidednorm&.615 ($\pm$.003)&.255 ($\pm$.009)&5.38e+4 ($\pm$1.1e+2)&.189 ($\pm$.005)&.079 ($\pm$.001)&6.59e+5 ($\pm$2.8e+3)&.726 ($\pm$.012)&.498 ($\pm$.001)&4.97e+4 ($\pm$4.2e+2)\\
\midrule
\multicolumn{10}{c}{\cnn}\\
\shapsamp&.471 ($\pm$.003)&.298 ($\pm$.008)&3.79e+7 ($\pm$3.1e+3)&.119 ($\pm$.004)&.084 ($\pm$.001)&1.26e+7 ($\pm$1.6e+5)&.789 ($\pm$.004)&.586 ($\pm$.017)&4.53e+6 ($\pm$2.1e+4)\\
\lime&.466 ($\pm$.002)&.300 ($\pm$.017)&1.81e+4 ($\pm$1.2e+3)&\textbf{.125 ($\boldsymbol \pm$.005)}&.079 ($\pm$.004)&5.39e+7 ($\pm$1.9e+4)&.737 ($\pm$.002)&.581 ($\pm$.021)&1.52e+4 ($\pm$7.1e+1)\\
\occlusion&.487 ($\pm$.003)&.298 ($\pm$.006)&6.06e+4 ($\pm$2.9e+2)&.090 ($\pm$.001)&.084 ($\pm$.001)&3.36e+5 ($\pm$2.6e+3)&.760 ($\pm$.004)&.580 ($\pm$.006)&1.40e+4 ($\pm$3.6e+1)\\
\salmean&\textbf{.600 ($\boldsymbol \pm$.002)}&.339 ($\pm$.007)&1.08e+4 ($\pm$5.6e+1)&.114 ($\pm$.005)&.091 ($\pm$.001)&4.28e+3 ($\pm$2.3e+2)&\underline{\textbf{.816 ($\pm$.003)}}&.593 ($\pm$.008)&4.16e+3 ($\pm$1.9e+1)\\
\salnorm&\textbf{.600 ($\boldsymbol \pm$.002)}&.339 ($\pm$.007)&1.06e+4 ($\pm$5.6e+1)&.115 ($\pm$.005)&.090 ($\pm$.001)&4.29e+3 ($\pm$9.9e+1)&.815 ($\pm$.003)&.596 ($\pm$.009)&4.16e+3 ($\pm$1.2e+1)\\
\inputxmean&.435 ($\pm$.001)&.294 ($\pm$.014)&1.07e+4 ($\pm$2.3e+1)&.121 ($\pm$.003)&.086 ($\pm$.002)&4.27e+3 ($\pm$1.8e+2)&.736 ($\pm$.002)&.572 ($\pm$.011)&4.16e+3 ($\pm$1.2e+1)\\
\inputxnorm&.580 ($\pm$.001)&.280 ($\pm$.003)&1.06e+4 ($\pm$6.5e+1)&.113 ($\pm$.004)&.093 ($\pm$.002)&4.09e+3 ($\pm$1.8e+2)&.774 ($\pm$.003)&.501 ($\pm$.006)&4.12e+3 ($\pm$2.7e+1)\\
\guidedmean&\textcolor{bad_res}{.269 ($\pm$.001)}&.299 ($\pm$.017)&1.08e+4 ($\pm$1.7e+2)&\textcolor{bad_res}{.076 ($\pm$.002)}&.086 ($\pm$.002)&4.27e+3 ($\pm$2.2e+2)&\textcolor{bad_res}{.501 ($\pm$.006)}&.573 ($\pm$.013)&4.32e+3 ($\pm$4.0e+2)\\
\guidednorm&\textbf{.600 ($\boldsymbol \pm$.002)}&.339 ($\pm$.007)&1.07e+4 ($\pm$3.4e+1)&.114 ($\pm$.005)&.091 ($\pm$.002)&4.21e+3 ($\pm$2.2e+2)&.815 ($\pm$.003)&.594 ($\pm$.009)&4.14e+3 ($\pm$1.7e+1)\\
\midrule
\multicolumn{10}{c}{\lstm}\\
\shapsamp&.396 ($\pm$.012)&.291 ($\pm$.008)&8.42e+5 ($\pm$1.2e+4)&.086 ($\pm$.001)&.084 ($\pm$.000)&2.30e+8 ($\pm$2.5e+5)&.605 ($\pm$.034)&.588 ($\pm$.020)&1.12e+7 ($\pm$2.1e+6)\\
\lime&.429 ($\pm$.012)&.309 ($\pm$.018)&1.68e+5 ($\pm$2.1e+5)&.089 ($\pm$.001)&.081 ($\pm$.002)&3.00e+8 ($\pm$1.8e+5)&.638 ($\pm$.025)&.588 ($\pm$.021)&5.20e+4 ($\pm$4.1e+3)\\
\occlusion&.358 ($\pm$.003)&.281 ($\pm$.007)&2.46e+5 ($\pm$5.7e+0)&.086 ($\pm$.002)&.083 ($\pm$.002)&1.18e+6 ($\pm$1.1e+3)&.694 ($\pm$.011)&.578 ($\pm$.016)&3.71e+4 ($\pm$2.7e+0)\\
\salmean&.502 ($\pm$.008)&.411 ($\pm$.011)&5.11e+3 ($\pm$6.8e+0)&.108 ($\pm$.001)&.106 ($\pm$.000)&3.04e+3 ($\pm$7.7e+1)&.710 ($\pm$.009)&.546 ($\pm$.000)&1.11e+3 ($\pm$2.8e+0)\\
\salnorm&.502 ($\pm$.008)&.410 ($\pm$.010)&5.12e+3 ($\pm$4.6e+0)&.108 ($\pm$.002)&.106 ($\pm$.002)&3.07e+3 ($\pm$3.9e+1)&.710 ($\pm$.010)&.546 ($\pm$.001)&1.10e+3 ($\pm$1.4e+0)\\
\inputxmean&.364 ($\pm$.004)&.349 ($\pm$.027)&5.12e+3 ($\pm$7.2e+0)&.098 ($\pm$.002)&.096 ($\pm$.002)&3.06e+3 ($\pm$7.0e+1)&\textcolor{bad_res}{.570 ($\pm$.010)}&.601 ($\pm$.017)&1.11e+3 ($\pm$2.2e+0)\\
\inputxnorm&\textbf{.511 ($\pm$.007)}&.389 ($\pm$.004)&5.12e+3 ($\pm$4.2e+0)&\textbf{.110 ($\boldsymbol \pm$.001)}&.107 ($\pm$.000)&3.05e+3 ($\pm$9.9e+1)&.697 ($\pm$.007)&.544 ($\pm$.001)&1.10e+3 ($\pm$1.6e+0)\\
\guidedmean&.333 ($\pm$.009)&.382 ($\pm$.033)&5.11e+3 ($\pm$4.4e+0)&.102 ($\pm$.005)&.098 ($\pm$.003)&3.06e+3 ($\pm$1.0e+2)&\textcolor{bad_res}{.527 ($\pm$.005)}&.570 ($\pm$.031)&1.10e+3 ($\pm$2.2e+0)\\
\guidednorm&.502 ($\pm$.009)&.410 ($\pm$.009)&5.10e+3 ($\pm$2.5e+1)&.109 ($\pm$.001)&.107 ($\pm$.001)&3.08e+3 ($\pm$9.2e+1)&\textbf{.711 ($\boldsymbol \pm$.009)}&.547 ($\pm$.001)&1.10e+3 ($\pm$2.4e+0)\\
\bottomrule
\end{tabular}
\caption{Evaluation of the explainability techniques with Human Agreement (HA) and time for computation. HA is measured with Mean Average Precision (MAP) with the gold human annotations, MAP of a Randomly initialized model (MAP RI). The time is computed with FLOPs. The presented numbers are averaged over five different models and the standard deviation of the scores is presented in brackets. Explainability methods with the best MAP for a particular dataset and model are in bold, while the best MAP across all models for a dataset is underlined as well. Methods that have MAP worse than the randomly generated saliency are in \textcolor{bad_res}{red}.}
\label{tab:human}
\end{table}
\end{landscape}

\begin{table*}[t]
\centering
\begin{tabular}{lrrr}
\toprule
\textbf{Explain.}&\textbf{e-SNLI}&\textbf{IMDB}&\textbf{TSE}\\
\midrule
\rand&56.05 ($\pm$0.71)&49.26 ($\pm$1.94)&56.45 ($\pm$2.37)\\
\midrule
\multicolumn{4}{c}{\textbf{\trans}}\\
\shapsamp&56.05 ($\pm$0.71)&\textcolor{bad_res}{65.84 ($\pm$11.8)}&52.99 ($\pm$4.24)\\
\lime&48.14 ($\pm$10.8)&\textcolor{bad_res}{59.04 ($\pm$13.7)}& 42.17 ($\pm$7.89)\\
\occlusion&55.24 ($\pm$3.77)&\textcolor{bad_res}{69.00 ($\pm$6.22)}&52.23 ($\pm$4.29)\\
\salmean&37.98 ($\pm$2.18)&\textcolor{bad_res}{49.32 ($\pm$9.01)}&\textbf{39.20 ($\boldsymbol \pm$3.06)}\\
\salnorm&38.01 ($\pm$2.19)&\textbf{49.05 ($\boldsymbol \pm$9.16)}&39.29 ($\pm$3.14)\\
\inputxmean&\textcolor{bad_res}{56.98 ($\pm$1.89)}&\textcolor{bad_res}{64.47 ($\pm$8.70)}&55.52 ($\pm$2.59)\\
\inputxnorm&\textbf{37.05 ($\boldsymbol \pm$2.29)}&\textcolor{bad_res}{50.22 ($\pm$8.85)}&37.04 ($\pm$2.69)\\
\guidedmean&53.43 ($\pm$1.00)&\textcolor{bad_res}{67.68 ($\pm$6.94)}&\textcolor{bad_res}{57.56 ($\pm$2.60)}\\
\guidednorm&38.01 ($\pm$2.19)&\textcolor{bad_res}{49.47 ($\pm$8.89)}&39.26 ($\pm$3.18)\\
\midrule
\multicolumn{4}{c}{\cnn}\\
\shapsamp&51.78 ($\pm$2.24)&\textcolor{bad_res}{59.69 ($\pm$8.37)}&\textcolor{bad_res}{64.72 ($\pm$1.75)}\\
\lime&\textcolor{bad_res}{56.16 ($\pm$1.67)}&\textcolor{bad_res}{59.09 ($\pm$8.48)}&\textcolor{bad_res}{65.78 ($\pm$1.59)}\\
\occlusion&54.32 ($\pm$0.94)&\textcolor{bad_res}{59.86 ($\pm$7.78)}&\textcolor{bad_res}{61.17 ($\pm$1.48)}\\
\salmean&34.26 ($\pm$1.78)&\textcolor{bad_res}{49.61 ($\pm$5.26)}&35.70 ($\pm$2.94)\\
\salnorm&34.16 ($\pm$1.81)&\textbf{49.04 ($\boldsymbol \pm$5.60)}&35.67 ($\pm$2.91)\\
\inputxmean&47.06 ($\pm$3.82)&\textcolor{bad_res}{62.05 ($\pm$7.54)}&\textcolor{bad_res}{64.45 ($\pm$2.99)}\\
\inputxnorm&\underline{\textbf{31.55 ($\boldsymbol \pm$2.83)}}&49.20 ($\pm$5.96)&35.86 ($\pm$3.22)\\
\guidedmean&47.68 ($\pm$2.65)&\textcolor{bad_res}{67.03 ($\pm$4.36)}&44.93 ($\pm$1.57)\\
\guidednorm&34.16 ($\pm$1.81)&\textcolor{bad_res}{49.80 ($\pm$5.99)}&\underline{\textbf{35.60 ($\boldsymbol \pm$2.91)}}\\
\midrule
\multicolumn{4}{c}{\lstm}\\
\shapsamp&51.05 ($\pm$4.47)&44.05 ($\pm$3.06)&53.97 ($\pm$6.00)\\
\lime&51.93 ($\pm$7.73)&\textcolor{bad_res}{44.41 ($\pm$3.04)}&54.95 ($\pm$3.19)\\
\occlusion&54.73 ($\pm$3.12)&45.01 ($\pm$3.84)&48.68 ($\pm$2.28)\\
\salmean&38.29 ($\pm$1.77)&35.98 ($\pm$2.11)&\textbf{37.20 ($\boldsymbol \pm$3.48)}\\
\salnorm&38.26 ($\pm$1.84)&36.22 ($\pm$2.04)&37.23 ($\pm$3.50)\\
\inputxmean&49.52 ($\pm$1.81)&43.57 ($\pm$4.98)&48.71 ($\pm$3.23)\\
\inputxnorm&\textbf{37.95 ($\boldsymbol \pm$2.06)}&36.03 ($\pm$1.97)&36.75 ($\pm$3.35)\\
\guidedmean&44.48 ($\pm$2.12)&46.00 ($\pm$3.20)&43.72 ($\pm$5.69)\\
\guidednorm&38.17 ($\pm$1.80)&\underline{\textbf{35.87 ($\boldsymbol \pm$1.99)}}&37.21 ($\pm$3.48)\\
\bottomrule
\end{tabular}

\caption{Faithfulness-AUC for thresholds $\in$ [0, 10, 20, \dots, 100]. \textit{Lower scores} indicate the ability of the saliency approach to assign higher scores to words more responsible for the final prediction. The scores are mean of different random initializations; the standard deviation is shown in brackets. The smallest AUC for a particular dataset and model are in bold; the smallest AUC across all models for a dataset is underlined. AUC worse than the randomly generated saliency are in \textcolor{bad_res}{red}.}
\label{tab:faith}
\end{table*}

\begin{landscape}
\begin{table}[p]
\centering
\tiny
\begin{tabular}{l@{\hspace{0.2\tabcolsep}}@{\hskip 0.05in \vline \hskip 0.05in}r@{\hspace{0.7\tabcolsep}}r@{\hspace{0.7\tabcolsep}}r@{\hspace{0.7\tabcolsep}}r@{\hspace{0.5\tabcolsep}}@{\hskip 0.05in \vline \hskip 0.05in}r@{\hspace{0.7\tabcolsep}}r@{\hspace{0.7\tabcolsep}}r@{\hspace{0.7\tabcolsep}}r@{\hspace{0.5\tabcolsep}}@{\hskip 0.05in \vline \hskip 0.05in}r@{\hspace{0.7\tabcolsep}}r@{\hspace{0.7\tabcolsep}}r@{\hspace{0.7\tabcolsep}}r}
\toprule
& \multicolumn{4}{c}{\textbf{e-SNLI}}&\multicolumn{4}{c}{\textbf{IMDB}}&\multicolumn{4}{c}{\textbf{TSE}} \\
\textbf{Explain.} & \textbf{MAE} & \textbf{MAX} &\textbf{MAE-up} & \textbf{MAX-up} & \textbf{MAE} & \textbf{MAX} & \textbf{MAE-up} & \textbf{MAX-up} & \textbf{MAE} & \textbf{MAX} & \textbf{MAE-up} & \textbf{MAX-up} \\ \midrule
\rand&.087 ($\pm$.004)&.527 ($\pm$.007)&.276 ($\pm$.005)&.377 ($\pm$.002)&.130 ($\pm$.007)&.286 ($\pm$.014)&.160 ($\pm$.003)&.251 ($\pm$.008)&.092 ($\pm$.009)&.466 ($\pm$.021)&.260 ($\pm$.017)&.428 ($\pm$.064) \\
\midrule
\multicolumn{13}{c}{\textbf{\trans}} \\
\shapsamp&.071 ($\pm$.005)&.456 ($\pm$.037)&.158 ($\pm$.029)&.437 ($\pm$.046)&\textbf{.071 ($\boldsymbol \pm$.008)}&\textbf{.238 ($\boldsymbol \pm$.036)}&\textbf{.120 ($\boldsymbol \pm$.033)}&\textbf{.213 ($\boldsymbol \pm$.035)}&\underline{\textbf{.073 ($\boldsymbol \pm$.012)}}&\textbf{.408 ($\boldsymbol \pm$.043)}&\textbf{.169 ($\boldsymbol \pm$.052)}&\textbf{.415 ($\boldsymbol \pm$.030)} \\
\lime&\underline{\textbf{.068 ($\boldsymbol \pm$.002)}}& \underline{\textbf{.368 ($\boldsymbol \pm$.151)}}&\textbf{.136 ($\boldsymbol \pm$.028)}&\textbf{.395 ($\boldsymbol \pm$.128)}&.077 ($\pm$.008)&.288 ($\pm$.024)&.184 ($\pm$.018)&.260 ($\pm$.021)&.084 ($\pm$.009)&.521 ($\pm$.072)&.232 ($\pm$.013)&.661 ($\pm$.225) \\
\occlusion&.074 ($\pm$.004)&.499 ($\pm$.020)&.224 ($\pm$.006)&.518 ($\pm$.048)&.085 ($\pm$.011)&.306 ($\pm$.015)&.196 ($\pm$.015)&.252 ($\pm$.011)&.085 ($\pm$.011)&.463 ($\pm$.035)&.247 ($\pm$.015)&.482 ($\pm$.091) \\
\salmean&.078 ($\pm$.005)&.544 ($\pm$.014)&.269 ($\pm$.004)&.416 ($\pm$.043)&.083 ($\pm$.009)&.303 ($\pm$.008)&.197 ($\pm$.017)&.269 ($\pm$.023)&.085 ($\pm$.012)&.474 ($\pm$.021)&.248 ($\pm$.017)&.467 ($\pm$.091) \\ 
\salnorm&.078 ($\pm$.005)&.565 ($\pm$.051)&.259 ($\pm$.007)&.571 ($\pm$.095)&.083 ($\pm$.009)&.306 ($\pm$.017)&.195 ($\pm$.021)&.245($\pm$.004)&.085 ($\pm$.012)&.465 ($\pm$.021)&.255 ($\pm$.012)&.479 ($\pm$.074) \\ 
\inputxmean&.079 ($\pm$.005)&.502 ($\pm$.015)&.242 ($\pm$.006)&.518 ($\pm$.031)&.084 ($\pm$.011)&.310 ($\pm$.011)&.198 ($\pm$.013)&.246 ($\pm$.008)&.085 ($\pm$.011)&.463 ($\pm$.015)&.237 ($\pm$.010)&.480 ($\pm$.071) \\
\inputxnorm&.078 ($\pm$.005)&.568 ($\pm$.057)&.258 ($\pm$.007)&.581 ($\pm$.096)&.083 ($\pm$.011)&.301 ($\pm$.014)&.193 ($\pm$.023)&.249 ($\pm$.016)&.086 ($\pm$.013)&.469 ($\pm$.022)&.252 ($\pm$.016)&.480 ($\pm$.087) \\
\guidedmean&.080 ($\pm$.005)&.505 ($\pm$.016)&.242 ($\pm$.008)&.519 ($\pm$.037)&.084 ($\pm$.011)&.308 ($\pm$.009)&.196 ($\pm$.014)&.245 ($\pm$.014)&.085 ($\pm$.011)&.456 ($\pm$.014)&.237 ($\pm$.013)&.494 ($\pm$.069) \\
\guidednorm&.078 ($\pm$.005)&.565 ($\pm$.051)&.258 ($\pm$.007)&.573 ($\pm$.095)&.080 ($\pm$.012)&.306 ($\pm$.009)&.192 ($\pm$.018)&.244 ($\pm$.008)&.086 ($\pm$.012)&.503 ($\pm$.053)&.261 ($\pm$.017)&.450 ($\pm$.081) \\
\midrule
\multicolumn{13}{c}{\textbf{\cnn}} \\
\shapsamp&\textbf{.103 ($\boldsymbol \pm$.001)}&.439 ($\pm$.020)&\textbf{.133 ($\boldsymbol \pm$.003)}&.643 ($\pm$.032)&.077 ($\pm$.018)&.210 ($\pm$.041)&.085 ($\pm$.023)&.196 ($\pm$.026)&.093 ($\pm$.002)&\underline{\textbf{.372 ($\boldsymbol \pm$.011)}}&.148 ($\pm$.004)&.479 ($\pm$.030) \\
\lime&.125 ($\pm$.003)&.498 ($\pm$.018)&.190 ($\pm$.006)&.494 ($\pm$.028)&.128 ($\pm$.006)&.289 ($\pm$.019)&.156 ($\pm$.003)&.260 ($\pm$.011)&.103 ($\pm$.001)&.469 ($\pm$.027)&.202 ($\pm$.014)&.633 ($\pm$.090) \\
\occlusion&.119 ($\pm$.004)&.492 ($\pm$.018)&.176 ($\pm$.007)&.507 ($\pm$.037)&.130 ($\pm$.007)&.289 ($\pm$.018)&.160 ($\pm$.006)&.254 ($\pm$.005)&.114 ($\pm$.002)&.463 ($\pm$.018)&.250 ($\pm$.007)&.418 ($\pm$.035) \\
\salmean&.137 ($\pm$.002)&.496 ($\pm$.011)&.220 ($\pm$.006)&.399 ($\pm$.010)&.129 ($\pm$.007)&.288 ($\pm$.021)&.159 ($\pm$.003)&.253 ($\pm$.013)&.115 ($\pm$.002)&.467 ($\pm$.014)&.245 ($\pm$.007)&.425 ($\pm$.028) \\
\salnorm&.140 ($\pm$.003)&.492 ($\pm$.009)&.225 ($\pm$.005)&.354 ($\boldsymbol \pm$.009)&.130 ($\pm$.006)&.286 ($\pm$.019)&.161 ($\pm$.004)&.250 ($\pm$.005)&.114 ($\pm$.002)&.475 ($\pm$.016)&.248 ($\pm$.006)&.405 ($\pm$.031) \\
\inputxmean&.110 ($\pm$.001)&\textbf{.436 ($\boldsymbol \pm$.014)}&.153 ($\pm$.007)&.460 ($\pm$.009)&\textbf{.071 ($\boldsymbol \pm$.004)}&\underline{\textbf{.191 ($\boldsymbol \pm$.010)}}&\underline{\textbf{.071 ($\boldsymbol \pm$.005)}}&\underline{\textbf{.190 ($\boldsymbol \pm$.010)}}&\textbf{.090 ($\boldsymbol \pm$.002)}&.379 ($\pm$.012)&\underline{\textbf{.135 ($\boldsymbol \pm$.004)}}&.477 ($\boldsymbol \pm$.025) \\ 
\inputxnorm&.140 ($\pm$.003)&.492 ($\pm$.009)&.225 ($\pm$.005)&.355 ($\pm$.007)&.130 ($\pm$.007)&.285 ($\pm$.019)&.160 ($\pm$.004)&.251 ($\pm$.011)&.114 ($\pm$.002)&.475 ($\pm$.014)&.248 ($\pm$.006)&.416 ($\pm$.033) \\
\guidedmean&.140 ($\pm$.003)&.485 ($\pm$.011)&.225 ($\pm$.005)&.367 ($\pm$.023)&.129 ($\pm$.006)&.286 ($\pm$.019)&.159 ($\pm$.003)&.253 ($\pm$.011)&.114 ($\pm$.002)&.462 ($\pm$.013)&.234 ($\pm$.011)&.441 ($\pm$.036) \\
\guidednorm&.140 ($\pm$.003)&.492 ($\pm$.009)&.225 ($\pm$.005)&\underline{\textbf{.353 ($\boldsymbol \pm$.008)}}&.130 ($\pm$.007)&.289 ($\pm$.018)&.159 ($\pm$.004)&.252 ($\pm$.011)&.114 ($\pm$.002)&.473 ($\pm$.015)&.249 ($\pm$.006)&\textbf{.404 ($\boldsymbol \pm$.029)} \\
\midrule
\multicolumn{13}{c}{\textbf{\lstm}} \\
\shapsamp&\textbf{.118 ($\boldsymbol \pm$.003)}&.622 ($\boldsymbol \pm$.035)&\underline{\textbf{.131 ($\boldsymbol \pm$.005)}}&.648 ($\pm$.054)&\underline{\textbf{.060 ($\boldsymbol \pm$.018)}}&\textbf{.279 ($\boldsymbol \pm$.065)}&\textbf{.160 ($\boldsymbol \pm$.014)}&.277 ($\pm$.038)&\textbf{.087 ($\boldsymbol \pm$.007)}&\textbf{.433 ($\boldsymbol \pm$.053)}&\textbf{.147 ($\boldsymbol \pm$.015)}&\underline{\textbf{.393 ($\boldsymbol \pm$.029)}}\\
\lime&.127 ($\pm$.004)&.512 ($\pm$.052)&.145 ($\pm$.009)&.490 ($\pm$.040)&.069 ($\pm$.018)&.300 ($\pm$.051)&.209 ($\pm$.024)&.267 ($\pm$.031)&.090 ($\pm$.007)&.667 ($\pm$.150)&.218 ($\pm$.010)&.864 ($\pm$.362) \\
\occlusion&.147 ($\pm$.003)&.579 ($\pm$.065)&.172 ($\pm$.007)&.593 ($\pm$.083)&.069 ($\pm$.017)&.304 ($\pm$.055)&.216 ($\pm$.014)&.324 ($\pm$.032)&.099 ($\pm$.006)&.509 ($\pm$.015)&.259 ($\pm$.012)&.723 ($\pm$.063) \\
\salmean&.163 ($\pm$.002)&.450 ($\pm$.008)&.195 ($\pm$.008)&.398 ($\pm$.031)&.069 ($\pm$.018)&.301 ($\pm$.051)&.208 ($\pm$.026)&\textbf{.259 ($\boldsymbol \pm$.022)}&.101 ($\pm$.007)&.518 ($\pm$.013)&.271 ($\pm$.008)&.469 ($\pm$.071) \\
\salnorm&.163 ($\pm$.002)&.448 ($\pm$.011)&.195 ($\pm$.008)&.399 ($\pm$.034)&.070 ($\pm$.018)&.299 ($\pm$.051)&.206 ($\pm$.024)&.263 ($\pm$.027)&.101 ($\pm$.007)&.523 ($\pm$.011)&.273 ($\pm$.008)&.441 ($\pm$.051) \\
\inputxmean&.161 ($\pm$.002)&.454 ($\pm$.018)&.193 ($\pm$.007)&.502 ($\pm$.033)&.066 ($\pm$.018)&.295 ($\pm$.059)&.201 ($\pm$.033)&\textbf{.262 ($\boldsymbol \pm$.014)}&.098 ($\pm$.007)&.527 ($\pm$.005)&.268 ($\pm$.008)&.425 ($\pm$.035) \\
\inputxnorm&.163 ($\pm$.002)&\textbf{.445 ($\boldsymbol \pm$.011)}&.195 ($\pm$.007)&\textbf{.394 ($\boldsymbol \pm$.029)}&.068 ($\pm$.018)&.303 ($\pm$.050)&.201 ($\pm$.031)&.277 ($\pm$.024)&.101 ($\pm$.007)&.523 ($\pm$.008)&.273 ($\pm$.007)&.445 ($\pm$.038) \\
\guidedmean&.161 ($\pm$.001)&.453 ($\pm$.014)&.192 ($\pm$.007)&.516 ($\pm$.058)&.068 ($\pm$.019)&.298 ($\pm$.055)&.200 ($\pm$.024)&.287 ($\pm$.045)&.097 ($\pm$.006)&.523 ($\pm$.017)&.260 ($\pm$.016)&.460 ($\pm$.045) \\
\guidednorm&.163 ($\pm$.002)&.446 ($\pm$.010)&.195 ($\pm$.007)&.396 ($\pm$.042)&.069 ($\pm$.017)&.300 ($\pm$.050)&.204 ($\pm$.024)&.279 ($\pm$.025)&.101 ($\pm$.007)&.525 ($\pm$.010)&.273 ($\pm$.007)&.474 ($\pm$.051) \\
\bottomrule
\end{tabular}
\caption{Confidence Indication experiments are measured with the Mean Absolute Error (MAE) of the generated saliency scores when used to predict the confidence of the class predicted by the model and the Maximum Error (MAX). We present the result with and without up-sampling(MAE-up, MAX-up) of the model confidence.
The presented measures are an average over the set of models trained from from different random seeds. The standard deviation of the scores is presented in brackets.  AVG Conf. is the average confidence of the model for the predicted class. The best results for a particular dataset and model are in bold and the best results across a dataset are also underlined. Lower results are better.}
~\label{tab:confidence}
\end{table}
\end{landscape}

\begin{table*}[t]
\centering
\begin{tabular}{lrrr}
\toprule
\textbf{Explain.} & \textbf{e-SNLI} & \textbf{IMDB} & \textbf{TSE} \\
\midrule
\multicolumn{4}{c}{\textbf{\trans}} \\
\rand & -0.004 (2.6e-01)    & -0.035 (1.4e-01)  & 0.003 (6.1e-01) \\
\shapsamp & 0.310 (0.0e+00) & 0.234 (3.6e-12)   & 0.259 (0.0e+00) \\
\lime & \textbf{0.519 (0.0e+00)} & 0.269 (3.0e-31) & 0.110 (2.0e-29) \\
\occlusion & 0.215 (0.0e+00) & 0.341 (2.6e-50) & 0.255 (0.0e+00) \\
\salmean & 0.356 (0.0e+00) & 0.423 (3.9e-79) & \textbf{0.294 (0.0e+00)} \\
\salnorm & 0.297 (0.0e+00) & 0.405 (6.9e-72) & 0.289 (0.0e+00) \\
\inputxmean & \textcolor{bad_res}{-0.102 (2.0e-202)} & \textbf{0.426 (2.5e-80)} & \textcolor{bad_res}{-0.010 (1.3e-01)} \\
\inputxnorm & 0.311 (0.0e+00) & 0.397 (3.8e-69) & 0.292 (0.0e+00) \\
\guidedmean & 0.064 (1.0e-79) & \textcolor{bad_res}{-0.083 (4.2e-04)} & \textcolor{bad_res}{-0.005 (4.9e-01)} \\
\guidednorm & 0.297 (0.0e+00) & 0.409 (1.2e-73) & 0.293 (0.0e+00) \\
\midrule
\multicolumn{4}{c}{\textbf{\cnn}} \\
\rand & -0.003 (4.0e-01) & 0.426 (2.6e-106) & -0.002 (7.4e-01) \\
\shapsamp & 0.789 (0.0e+00) & 0.537 (1.4e-179) & 0.704 (0.0e+00) \\
\lime & 0.790 (0.0e+00) & 0.584 (1.9e-219) & \underline{\textbf{0.730 (0.0e+00)}} \\
\occlusion & 0.730 (0.0e+00) & 0.528 (2.4e-172) & 0.372 (0.0e+00) \\
\salmean & 0.701 (0.0e+00) & 0.460 (4.5e-126) & 0.320 (0.0e+00) \\
\salnorm & \underline{\textbf{0.819 (0.0e+00)}} & 0.583 (4.0e-218) & 0.499 (0.0e+00) \\
\inputxmean & 0.136 (0.0e+00) & \textcolor{bad_res}{0.331 (1.2e-62)} & 0.002 (7.5e-01) \\
\inputxnorm & 0.816 (0.0e+00) & \underline{\textbf{0.585 (8.6e-221)}} & 0.495 (0.0e+00) \\
\guidedmean & 0.160 (0.0e+00) &\textcolor{bad_res}{ 0.373 (5.5e-80)} & 0.173 (6.3e-121) \\
\guidednorm & \underline{\textbf{0.819 (0.0e+00)}} & 0.578 (2.4e-214) & 0.498 (0.0e+00) \\
\midrule
\multicolumn{4}{c}{\textbf{\lstm}} \\ 
\rand & 0.004 (1.8e-01) & 0.002 (9.2e-01) & 0.010 (1.8e-01) \\
\shapsamp & 0.657 (0.0e+00) & 0.382 (1.7e-63) & 0.502 (0.0e-00) \\
\lime & \textbf{0.700 (0.0e+00)} & 0.178 (3.3e-14) & 0.540 (0.0e-00) \\
\occlusion & 0.697 (0.0e+00) & \textbf{0.498 (1.7e-113)} & 0.454 (0.0e-00) \\
\salmean & 0.645 (0.0e+00) & 0.098 (3.1e-05) & \textbf{0.667 (0.0e-00)} \\
\salnorm & 0.662 (0.0e+00) & 0.132 (1.8e-08) & 0.596 (0.0e-00) \\
\inputxmean & 0.026 (1.9e-14) & \textcolor{bad_res}{-0.032 (1.7e-01)} & 0.385 (0.0e-00) \\
\inputxnorm & 0.664 (0.0e+00) & 0.133 (1.5e-08) & 0.604 (0.0e-00) \\
\guidedmean & 0.144 (0.0e+00) & 0.122 (2.0e-07) & 0.295 (0.0e-00) \\
\guidednorm & 0.663 (0.0e+00) & 0.139 (3.1e-09) & 0.598 (0.0e-00) \\
\bottomrule
\end{tabular}
\caption{Rationale Consistency Spearman's $\rho$ correlation; p-value is in brackets. The best results for a dataset and model are in bold and across a dataset are underlined. Correlation lower that the randomly sampled saliency scores is in \textcolor{bad_res}{red}.}
\label{tab:consistency:rat}
\end{table*}

\begin{table*}[t]
\centering
\begin{tabular}{llll}
\toprule
\textbf{Explain.} & \textbf{e-SNLI} & \textbf{IMDB} & \textbf{TSE} \\
\midrule
\multicolumn{4}{c}{\textbf{\trans}}  \\
\rand & 0.047 (2.7e-04) & 0.127 (6.6e-07)/ & 0.121 (2.5e-01) \\
\shapsamp & 0.285 (1.8e-02) & \textcolor{bad_res}{0.078 (5.8e-04)} & 0.308 (3.4e-36) \\
\lime & 0.372 (3.1e-90) & \textbf{0.236 (4.6e-07)} & \underline{\textbf{0.413 (3.4e-120)}} \\
\occlusion & 0.215 (9.6e-02) & \textcolor{bad_res}{0.003 (2.0e-04)} & 0.235 (7.3e-05) \\
\salmean & 0.378 (4.3e-57) & \textcolor{bad_res}{0.023 (4.3e-02)} & 0.253 (1.4e-20) \\
\salnorm & 0.027 (3.0e-05) & \textcolor{bad_res}{-0.043 (5.6e-02)} & 0.260 (6.8e-21) \\
\inputxmean & 0.319 (3.0e-03) & \textcolor{bad_res}{0.008 (1.2e-01)} & 0.193 (7.5e-05) \\
\inputxnorm & 0.399 (1.9e-78) & \textcolor{bad_res}{0.028 (2.3e-03)} & 0.247 (4.9e-17) \\
\guidedmean & 0.400 (6.7e-31) & \textcolor{bad_res}{0.017 (1.9e-01)} & 0.228 (5.2e-09) \\
\guidednorm & \underline{\textbf{0.404 (1.4e-84)}} & \textcolor{bad_res}{0.019 (4.3e-04)} & 0.255 (3.1e-20) \\
\midrule
\multicolumn{4}{c}{\textbf{\cnn}} \\
\rand & 0.018 (2.4e-01) & 0.115 (1.8e-04) & 0.008 (2.0e-01) \\
\shapsamp & \textcolor{bad_res}{0.015 (1.8e-01)} & \textcolor{bad_res}{-0.428 (5.3e-153)} & 0.037 (1.4e-01) \\
\lime & \textcolor{bad_res}{0.000 (4.4e-02)} & 0.400 (1.4e-126) & 0.023 (4.0e-01) \\
\occlusion & \textcolor{bad_res}{-0.076 (6.5e-02)} & \textcolor{bad_res}{-0.357 (1.9e-85)} & \textbf{0.041 (1.7e-01)} \\
\salmean & 0.381 (6.9e-91) & 0.431 (1.1e-146) & \textcolor{bad_res}{-0.100 (3.9e-06)} \\
\salnorm & 0.391 (1.7e-98) & 0.427 (3.5e-135) & \textcolor{bad_res}{-0.100 (3.7e-06)} \\
\inputxmean & 0.171 (5.1e-04) & 0.319 (1.4e-69) & 0.024 (3.5e-01) \\
\inputxnorm & \textbf{0.399 (1.0e-93)} & 0.428 (1.4e-132) & \textcolor{bad_res}{-0.076 (1.2e-03)} \\
\guidedmean & 0.091 (7.9e-02) & 0.375 (5.7e-109) & \textcolor{bad_res}{-0.032 (1.1e-01)} \\
\guidednorm & \textbf{0.391 (1.7e-98)} & \underline{\textbf{0.432 (3.5e-140)}} & \textcolor{bad_res}{-0.102 (1.7e-06)} \\
\midrule
\multicolumn{4}{c}{\textbf{\lstm}} \\
\rand & 0.018 (3.9e-01) & 0.037 (1.8e-01) & 0.016 (9.2e-03) \\
\shapsamp & 0.398 (3.5e-81) & 0.230 (8.9e-03) & 0.205 (2.1e-16) \\
\lime & \underline{\textbf{0.415 (1.2e-80)}} & 0.079 (8.6e-04) & 0.207 (4.3e-16) \\
\occlusion & 0.363 (1.1e-37) & \textbf{0.429 (7.5e-137)} & \textbf{0.237 (2.9e-29)} \\
\salmean & 0.158 (1.7e-17) & \textcolor{bad_res}{-0.177 (1.6e-10)} & 0.065 (5.8e-03) \\
\salnorm & 0.160 (7.5e-19) & \textcolor{bad_res}{-0.168 (2.0e-15)} & 0.096 (8.2e-03) \\
\inputxmean & 0.142 (3.3e-06) & \textcolor{bad_res}{-0.152 (1.2e-14)} & 0.106 (2.8e-02) \\
\inputxnorm & 0.183 (7.0e-24) & \textcolor{bad_res}{-0.175 (4.7e-17)} & 0.089 (8.4e-03) \\
\guidedmean & 0.163 (1.9e-12) & \textcolor{bad_res}{-0.060 (4.7e-02)} & 0.077 (1.2e-02) \\
\guidednorm & 0.169 (1.8e-12) & \textcolor{bad_res}{-0.214 (5.8e-16)} & 0.115 (4.3e-02) \\
\bottomrule
\end{tabular}
\caption{Dataset Consistency results with Spearman $\rho$; p-value is in brackets. The best results for a dataset and model are in bold and across a dataset are underlined. Correlation lower that the randomly sampled saliency scores are in \textcolor{bad_res}{red}.}
\label{tab:consistency:data}
\end{table*}

\chapter{Diagnostics-Guided Explanation Generation}
\label{chap:diagnostic_guided_explanations}

\section{Introduction}
\begin{figure}[t]
\centering
\includegraphics[width=0.65\columnwidth]{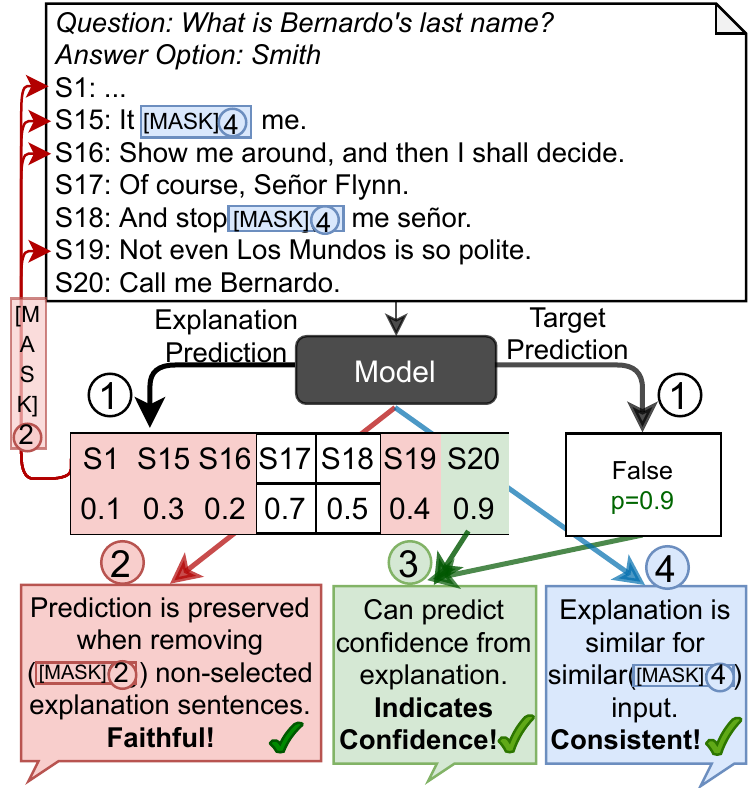}
\caption{Example instance from MultiRC with predicted target and explanation (Step 1), where sentences with confidence $\!\geq\!$ 0.5 are selected as explanations (S17, S18, S20). Steps 2-4 illustrate the use of \faithfulness, \consistency, and \confidence\ diagnostic properties as additional learning signals. `[MASK](2)' is used in Step 2 for sentences (in red) that are not explanations, and `[MASK](4)'--for random words in Step 4.}
\label{figure:example} 
\end{figure}
% url for the figure in draw.io : https://drive.google.com/file/d/10NE9dBXSkzcVzMgkOQhJKgQdCfAibITM/view?usp=sharing

Explanations are an important complement to the predictions of a ML model. 
% when considering its adoption by a broad user group. 
They unveil the decisions of a model that lead to a particular prediction, which increases user trust in the automated system and can help find its vulnerabilities. Moreover, ``The right $\dots$ to obtain an explanation of the decision reached'' is enshrined in the European law~\cite{regulation2016regulation}.

In NLP, %the efforts to provide explanations of a model's predictions have been supported by 
research on explanation generation has spurred the release of datasets \cite{zaidan-eisner-piatko-2008:nips,thorne-etal-2018-fever,khashabi-etal-2018-looking} containing human rationales for the correct predictions of downstream tasks in the form of word- or sentence-level selections of the input text. Such datasets are particularly beneficial for knowledge-intensive tasks~\cite{petroni-etal-2021-kilt} with long sentence-level explanations, e.g., question answering and fact-checking, where identifying the required information is an important prerequisite for a correct prediction. They can be used to supervise and evaluate whether a model employs the correct rationales for its predictions~\cite{deyoung-etal-2020-eraser,thorne-etal-2018-fever,Augenstein2021Doctoral}. The goal of this paper is to improve the sentence-level explanations generated for such complex reasoning tasks.

% During training, explanations can be generated in a pipeline approach - first selecting rationales, then using them for prediction, or can be trained jointly with the task at hand. 
%Apart from agreement with human rationales, we can also observe other properties of explanations. 
When human explanation annotations are not present, a common approach~\cite{lei-etal-2016-rationalizing,yu-etal-2019-rethinking} is to train models that select regions from the input maximising proximity to original task performance which corresponds to the \textit{\faithfulness}\ property.
\citet{atanasova-etal-2020-diagnostic} propose \faithfulness\ and other \textit{diagnostic properties} to evaluate different characteristics of explanations. These include \textit{\consistency}, which measures the similarity of the explanations between similar instances, and \textit{\confidence}, which evaluates whether the explanation reflects the model's confidence, among others (see Figure \ref{figure:example} for an example).

\textbf{Contributions\footnote{We make an extended version of the manuscript and code available on  \url{https://github.com/copenlu/diagnostic-guided-explanations} .}} We present the first method to \textit{learn the aforementioned diagnostic properties in an unsupervised way}, directly optimising for them to improve the quality of generated explanations. We implement a joint task prediction and explanation generation model, which selects rationales at sentence level. Each property can then be included as an additional training objective in the joint model. With experiments on three complex reasoning tasks, we find that apart from improving the properties we optimised for, diagnostic-guided training also %improves the generated explanations' agreement with human annotations and the performance on the downstream tasks. 
leads to explanations with higher agreement with human rationales, and improved downstream task performance. Moreover, we find that jointly optimising for diagnostic properties %guides the model to avoid 
leads to reduced claim/question-only bias~\cite{schuster-etal-2019-towards} for the target prediction, and means that the model relies more extensively on the provided evidence.
Importantly, we also find that optimising for diagnostic properties of explanations without supervision for explanation generation does not lead to good human agreement. This indicates the need for human rationales to train models that make the right predictions for the right reasons.

\begin{comment}
In summary, the contributions of this work are:
\begin{itemize}%[nosep]
\item We propose the use of explanation diagnostic properties, namely \faithfulness, \consistency, and \confidence, as additional training objectives during supervised explanation generation.
\item We evaluate the proposed objectives on three complex reasoning datasets and find improvements in agreement with human rationales (up to 2 $F_1$ score points), explanation properties, downstream task performance (up to 3 $F_1$ score points), as well as in avoiding claim/question-only bias during prediction.
\item We evaluate the proposed objectives without human explanation supervision and find that human rationales are important to train models that predict the target task based on the correct rationales.
\end{itemize}
\end{comment}

\section{Related Work}
\label{sec:related7}

\textbf{Supervised Explanations.} In an effort to guide ML models to perform human-like reasoning and avoid learning spurious patterns~\cite{zhang-etal-2016-rationale,ghaeini-etal-2019-saliency}, multiple datasets with explanation annotations at the word and sentence level have been proposed~\cite{wiegreffe2021teach}. 
These annotations are also used for supervised explanation generation, e.g., in pipeline models, where the generation task is followed by predicting the target task from the selected rationales only ~\cite{deyoung-etal-2020-evidence,lehman-etal-2019-inferring}. As \citet{wiegreffe-etal-2021-measuring,kumar-talukdar-2020-nile,jacovi2021aligning} point out, pipeline models produce explanations without task-specific knowledge and without knowing the label to explain.
However, for completion, we include the baseline pipeline from ERASER's benchmark \citet{deyoung-etal-2020-evidence} as a reference model for our experiments.
% to regularise model's attention weights~\cite{zhang-etal-2016-rationale,ghaeini-etal-2019-saliency}. 
%Another common approach is to use them 
% Moreover, they are used 

Explanation generation can also be trained jointly with the target task~\cite{atanasova-etal-2020-generating-fact,li-etal-2018-end}, which has been shown to improve the performance for both tasks. Furthermore, ~\citet{wiegreffe-etal-2021-measuring} suggest that self-rationalising models, such as multi-task models, provide more label-informed rationales than pipeline models. Such multi-task models can additionally learn a joint probability of the explanation and the target task prediction on the input. This can be decomposed into first extracting evidence, then predicting the class based on it~\cite{zhao2020transformer-xh,zhou-etal-2019-gear}, or vice versa~\cite{pruthi-etal-2020-weakly}. 
In this work, we also employ joint conditional training. 
% Thus, the prediction follows from the explanation by design. 
It additionally provides \textit{a good testbed for our experiments with ablations of supervised and diagnostic property objectives, which is not possible with a pipeline approach.}

Most multi-task models encode each sentence separately, then combine their representations, e.g., with Graph Attention Layers~\cite{zhao2020transformer-xh,zhou-etal-2019-gear}. \citet{glockner-etal-2020-think} predict the target label from each separate sentence encoding and use the most confident sentence prediction as explanation, which also allows for unsupervised explanation generation. %In this work, 
We consider \citet{glockner-etal-2020-think} as a reference model, as it is the only other work that reports results on generating explanations at the sentence level for three complex reasoning datasets from the ERASER benchmark~\cite{deyoung-etal-2020-eraser}. It also outperforms the baseline pipeline model we include from ERASER. \textit{Unlike related multi-task models, we encode the whole input jointly so that the resulting sentence representations are sensitive to the wider document context. The latter proves to be especially beneficial for explanations consisting of multiple sentences. Furthermore, while the model of \citet{glockner-etal-2020-think} is limited to a fixed and small number (up to two) of sentences per explanation, our model can predict a variable number of sentences depending on the separate instances' rationales.}

% Related work selects explanations sentences also using Graph Neural Networks~\cite{zhou-etal-2019-gear,kgat,zhao2020transformer-xh}, unsupervised learning~\cite{glockner-etal-2020-think}, where the explanations are used for evaluation. 

\textbf{Unsupervised Explanations.} When human explanation annotations are not provided, model's rationales can be explained with post-hoc methods based on gradients~\cite{sundararajan2017axiomatic}, simplifications~\cite{ribeiromodel}, or teacher-student setups~\cite{pruthi-etal-2022-evaluating}. Another approach is to select input tokens that preserve a model's original prediction~\cite{lei-etal-2018-cooperative,yu-etal-2019-rethinking,bastings-etal-2019-interpretable,paranjape-etal-2020-information}, which corresponds to the \faithfulness\ property of an explanation. However, as such explanations are not supervised by human rationales, they do not have high overlap with human annotations~\cite{deyoung-etal-2020-eraser}. Rather, they explain what a model has learned, which does not always correspond to correct rationales and can contain spurious patterns~\cite{wang-culotta-2020-identifying}. 

% We also study how to guide a model to generate explanations \textit{faithful} to its predictions, but do so jointly with learning the target task to ensure that the model takes the generated explanations into account. We use two further properties from \citet{atanasova-etal-2020-diagnostic} as training objectives -- \textit{\consistency}\ and \textit{\confidence}\ -- and find that they improve explanation quality. %as we find that they can also be integrated into the training process to improve the generated explanations.

\section{Method}
\label{sec:method7}
We propose a novel Transformer~\cite{vaswani2017attention} based model to jointly optimise sentence-level explanation generation and downstream task performance. The joint training provides a suitable testbed for our experiments with supervised and diagnostic property objectives for a single model. The joint training optimises two training objectives for the two tasks at the same time. By leveraging information from each task, the model is guided to predict the target task based on correct rationales and to generate explanations based on the model's information needs for target prediction. This provides additional useful information for training each of the tasks. Conducting joint training for these two tasks was shown to improve the performance for each of them~\cite{zhao2020transformer-xh,atanasova-etal-2020-generating-fact}. 

The \textbf{core novelty} is that the model is trained to improve the quality of its explanations by using diagnostic properties of explanations as additional training signals (see Figure~\ref{figure:example}). We select the properties \faithfulness, \consistency, and \confidence, as they can be effectively formulated as training objectives. \faithfulness\ is also employed in explainability benchmarks~\cite{deyoung-etal-2020-eraser} and in related work for unsupervised token-level explanation generation~\cite{lei-etal-2016-rationalizing,lei-etal-2018-cooperative}, whereas we consider it at sentence level. Further, multiple studies~\cite{NEURIPS2019_a7471fdc,alvarez2018robustness} find that explainability techniques are not robust to insignificant and/or adversarial input perturbations, which we address with the \consistency\ property. We do not consider Human Agreement and Rationale Consistency, proposed in \citet{atanasova-etal-2020-diagnostic}. The supervised explanation generation training employs human rationale annotations and thus addresses Human Agreement. Rationale Consistency requires the training of a second model, which is resource-expensive.  Another property to investigate in future work is whether a model's prediction can be simulated by another model trained only on the explanations~\cite{hase-etal-2020-leakage,treviso-martins-2020-explanation,pruthi-etal-2020-learning}, which also requires training an additional model. We now describe each component in detail.

\subsection{Joint Modelling}
Let $D\!=\!\{(x_i, y_i, e_i)|i\in[1, \#(D)]\}$ be a classification dataset. The textual input $x_i\!=\!(q_i, a_i^{[opt]}, s_{i})$ consists of a question or a claim, an optional answer, and several sentences (usually above 10) $s_i\!=\!\{s_{i,j}|j\!\in\![1, \#(s_i)]\}$ used for predicting a classification label $y_i\!\in\![1, N]$. Additionally, D contains human rationale annotations selected from the sentences $s_i$ as a binary vector $e_i\!=\!\{{e_{i,j}\!=\!\{0,1\}|j\!\in\![1, \#(s_i)]}\}$, which defines a binary classification task for explanation extraction.

First, the joint model takes $x_i$ as input and encodes it using a Transformer model, resulting in contextual token representations $h^L = \textit{encode}(x_i)$ from the final Transformer layer $L$. From $h^L$, we select the representations of the CLS token that precedes the question--as it is commonly used for downstream task prediction in Transformer architectures--and the CLS token representations preceding each sentence in $s_i$, which we use for selecting sentences as an explanation. 
% From this point, we select and work with the representations of these tokens. 
The selected representations are then transformed with two separate linear layers - $h^C$ for predicting the target, and $h^E$ for generating the explanations, which have the same hidden size as the size of the contextual representations in $h^L$. 

Given representations from $h^E$, a N-dimensional linear layer predicts the importance $p^E \! \in\! \mathbb{R}^{\#(s_i)}$ of the evidence sentences for the prediction of each class. As a final sentence importance score, we only take the score for the predicted class $p^{E[c]}$ and add a sigmoid layer on top for predicting the binary explanation selection task. Given representations from $h^C$, a N-dimensional linear layer with a soft-max layer on top predicts the target label $p^{C'}\!\in\!\mathbb{R}$. The model then predicts the joint conditional likelihood $L$ of the target task and the generated explanation given the input (Eq.~\ref{eq:joint}). This is factorised further into first extracting the explanations conditioned on the input and then predicting the target label (Eq.~\ref{eq:factor}) based on the extracted explanations (assuming  $\left.y_{i} \perp \mathbf{x}_{i} \mid \mathbf{e}_{i}\right)$).
\begin{gather}
L=\prod_{i=1}^{\#(D)} p\left(y_{i}, \mathbf{e}_{i} \mid \mathbf{x}_{i}\right) ~\label{eq:joint} \\
L =\prod_{i=1}^{\#(D)} p\left(\mathbf{e}_{i} \mid \mathbf{x}_{i}\right) p\left(y_{i} \mid \mathbf{e}_{i}\right)~\label{eq:factor}
\end{gather}
We condition the label prediction on the explanation prediction by multiplying $p^{C'}$ and $p^E$, resulting in the final prediction $p^{C}\! \in\! \mathbb{R}$. The model is trained to optimise jointly the target task cross-entropy loss function ($\mathcal{L}_C$) and the explanation generation cross-entropy loss function ($\mathcal{L}_E$):
\begin{gather}
\mathcal{L} = \mathcal{L}_C(p^C, y) + \mathcal{L}_E(p^{E[c]}, e)
\end{gather}
All loss terms of the diagnostic explainability properties described below are added to $\mathcal{L}$ without additional hyper-parameter weights for the separate loss terms.

\subsection{\faithfulness\ (F)}
The \faithfulness\ property guides explanation generation to select sentences preserving the original prediction, (Step 2,  Fig.~\ref{figure:example}). 
In more detail, we take sentence explanation scores $p^{E[c]}\!\in\![0,1]$ and sample from a Bernoulli distribution the sentences which should be preserved in the input: $c^{E}\!\sim\! \textit{Bern}(p^{E[c]})$. Further, we make two predictions -- one, where only the selected sentences are used as an input for the model, thus producing a new target label prediction $l^{S}$, and one where we use only unselected sentences, producing the new target label prediction $l^{Co}$. The assumption is that a high number $\#(l^{C=S})$ of predictions $l^{S}$ matching the original $l^C$ indicate the sufficiency (S) of the selected explanation. On the contrary, a low number $\#(l^{C=Co})$  of predictions $l^{Co}$ matching the original $l^C$ indicate the selected explanation is complete (Co) and no sentences indicating the correct label are missed. We then use the REINFORCE~\cite{williams1992simple} algorithm to maximise the reward:
\begin{gather}
% \small
\begin{split}\mathcal{R}_F = {\scriptstyle \#}(l^{C=S}) - {\scriptstyle \#}(l^{C=Co})- |{\scriptstyle \%}(c^E) - \lambda|
\end{split}
\end{gather}
The last term is an additional sparsity penalty for selecting more/less than $\lambda$\% of the input sentences as an explanation, $\lambda$ is a hyper-parameter.

\subsection{\consistency\ (DC)}
\consistency\ measures how similar the explanations for similar instances are. Including it as an additional training objective can serve as regularisation for the model to be consistent in the generated explanations. To do so, we mask $K$ random words in the input, where $K$ is a hyper-parameter depending on the dataset. We use the masked text (M) as an input for the joint model, which predicts new sentence scores $p^{EM}$. We then construct an $\mathcal{L}1$ loss term for the property to minimise for the absolute difference between $p^{E}$ and $p^{EM}$:
\begin{gather}
\mathcal{L}_{DC} = |p^{E} - p^{EM}|
\end{gather}
\noindent We use $\mathcal{L}1$ instead of $\mathcal{L}2$ loss as we do not want to penalise for potentially masking important words, which would result in entirely different outlier predictions. 

\subsection{\confidence\ (CI)}
The CI property measures whether generated explanations reflect the confidence of the model's predictions (Step 3, Fig.~\ref{figure:example}). We consider this a useful training objective to re-calibrate and align the prediction confidence values of both tasks.
To learn explanations that indicate prediction confidence, we aggregate the sentence importance scores, taking their maximum, minimum, mean, and standard deviation. 
% We use the max, min, mean, standard deviation as we assume that the confidence for a class should correspond to the highest confidence across the sentences to be selected as explanations for the class and to the lowest confidence for the sentences to be selected as explanations for the other classes. 
We transform the four statistics with a linear layer that predicts the confidence $\hat{p}^C$ of the original prediction. We train the model to minimise $\mathcal{L}1$ loss between $\hat{p}^C$ and $p^C$:
\begin{gather}
\mathcal{L}_{CI} = |p^C - \hat{p}^C|
\end{gather}
\noindent We choose $\mathcal{L}1$ as opposed to $\mathcal{L}2$ loss as we do not want to penalise possible outliers due to sentences having high confidence for the opposite class.
% , which is common for the Movies dataset. 

\section{Experiments}
\subsection{Datasets} 
We perform experiments on three datasets from the ERASER benchmark~\cite{deyoung-etal-2020-eraser} (FEVER, MultiRC, Movies), all of which require complex reasoning and have sentence-level rationales. For FEVER~\cite{thorne-etal-2018-fever}, given a claim and an evidence document, a model has to predict the veracity of a claim$\in$\{support, refute\}. 
The evidence for predicting the veracity has to be extracted as explanation. For MultiRC~\cite{khashabi-etal-2018-looking}, given a question, an answer option, and a document, a model has to predict if the answer is correct. For Movies~\cite{zaidan-eisner-piatko-2008:nips}, the sentiment$\in$\{positive, negative\} of a long movie review has to be predicted. For Movies, as in \citet{glockner-etal-2020-think}, we mark each sentence containing annotated explanation at token level as an explanation. 
Note that, in knowledge-intensive tasks such as fact checking and question answering also explored here, human rationales point to regions in the text containing the information needed for prediction. 
Identifying the required information becomes an important preliminary for the correct prediction rather than a plausibility indicator~\cite{jacovi-goldberg-2020-towards}, and is evaluated as well (e.g., FEVER score, Joint Accuracy). 
% (see further discussion in supplemental material).
%These three datasets are also explored in \cite{glockner-etal-2020-think}. 
% We also include an additional dataset from the ERASER benchmark, namely Evidence Inference~\cite{deyoung-etal-2020-evidence}, where given a statement about the effects of a treatment and a scientific article, the model has to predict the effect $\in$ \{no significant difference, significantly increased, significantly decreased\}. 
% We consider the latter to be a challenge dataset with explanations at the sentence level where the input length is over 4000 tokens. For more information on the datasets see the supplemental material.

\subsection{Metrics} 
We evaluate the effect of using diagnostic properties as additional training objectives for explanation generation. We first measure their effect on selecting human-like explanations by evaluating precision, recall, and macro $F_1$ score against human explanation annotations provided in each dataset (\S\ref{sec:explanation_results}). Second, we compute how generating improved explanations affects the target task performance by computing accuracy and macro $F_1$ score for the target task labels (\S\ref{sec:target_results}). Additionally, as identifying the required information in knowledge-intensive datasets, such as FEVER and MultiRC, is an important preliminary for a correct prediction, and following \citet{thorne-etal-2018-fever, glockner-etal-2020-think}, we evaluate the joint target and explanation performance by considering a prediction as correct only when the whole explanation is retrieved (Acc. Full). In case of multiple possible explanations $e_i$ for one instance (ERASER provides comprehensive explanation annotations for the test sets), selecting one of them counts as a correct prediction. Finally, as diagnostic property training objectives target particular properties, we measure the improvements for each property (\S \ref{sec:properties_results}).

\subsection{Experimental Setting} 
Our core goal is to measure the relative improvement of the explanations generated by the underlying model with (as opposed to without) diagnostic properties. We conduct experiments for the supervised model (Sup.), including separately \faithfulness\ (F), \consistency\ (DC), and \confidence\ (CI), as well as all three (All) as additional training signals (\S\ref{sec:method7}).

Nevertheless, we include results from two other architectures generating sentence-level explanations that serve as a reference for explanation generation performance on the employed datasets. Particularly, we include the best supervised sentence explanation generation results reported in \citet{glockner-etal-2020-think}, and the baseline pipeline model from ERASER, which extracts one sentence as explanation and uses it for target prediction (see \S\ref{sec:related7} for a detailed comparison). We also include an additional baseline comparison for the target prediction task. The BERT Blackbox model predicts the target task from the whole document as an input without being supervised by human rationales. The results are as reported by \citet{glockner-etal-2020-think}. In our experiments, we use BERT~\cite{devlin-etal-2019-bert} base-uncased as our base architecture, following \citet{glockner-etal-2020-think}.

%(Supervised model in the tables). 
% in the training objective
% We include additional reference results in the supplemental material.
% \citet{glockner-etal-2020-think} is the only other work with results for all three datasets.%as a baseline system.
% We include \citet{deyoung-etal-2020-evidence}'s supervised explanation generation results, choosing the best-performing two-sentence architecture for the MultiRC dataset. %, where %two sentences is their current limitation of the models. 
%  For FEVER and Movies, we include \citet{glockner-etal-2020-think}'s single-sentence results, which are the best performing, as the explanations in these datasets need only one sentence. 

\section{Results}
\subsection{Explanation Generation Results}
~\label{sec:explanation_results}

\begin{table*}[t]
\centering
\fontsize{10}{10}\selectfont
\begin{tabular}{lll@{\hskip 0.05in \vline \hskip 0.05in}lll@{\hskip 0.05in \vline \hskip 0.05in}l}
\toprule
 \bf Method & \bf $\mathbf{F_1}$-C & \bf Acc-C & \bf P-E & \bf R-E & \bf $\mathbf{F_1}$-E & \bf Acc-Joint \\
\midrule
\multicolumn{7}{c}{\bf FEVER} \\
Blackbox{\tiny \cite{glockner-etal-2020-think}}& 90.2 {\scriptsize $\pm$0.4} & 90.2 {\scriptsize $\pm$0.4} & & & & \\
Pipeline{\tiny ~\cite{deyoung-etal-2020-eraser}} & 87.7 & 87.8 & 88.3 & 87.7 & 88.0 & 78.1\\
Supervised {\tiny \cite{glockner-etal-2020-think}} & 90.7  {\scriptsize $\pm$0.7} & 90.7  {\scriptsize $\pm$0.7} & 92.3  {\scriptsize $\pm$0.1} & 91.6  {\scriptsize $\pm$0.1} & 91.9  {\scriptsize $\pm$0.1} & 83.9  {\scriptsize $\pm$0.1}\\% & 84.9\\
\cdashline{1-7}[2pt/2pt]
Supervised & 89.3 {\scriptsize $\pm$0.4}& 89.4 {\scriptsize $\pm$0.3} & 94.0 {\scriptsize $\pm$0.1} & 93.8 {\scriptsize $\pm$0.1} & 93.9 {\scriptsize $\pm$0.1} & 80.1 {\scriptsize $\pm$0.4} \\%& 83.5\\
Supervised+\consistency & \bf 89.7 {\scriptsize $\pm$0.5} & \bf 89.7 {\scriptsize $\pm$0.5} &\bf 94.4 {\scriptsize $\pm$0.0} & \bf 94.2 {\scriptsize $\pm$0.0} & \bf 94.4 {\scriptsize $\pm$0.0} & 80.8 {\scriptsize $\pm$0.5}\\% & 84.3\\
Supervised+\faithfulness & 89.5 {\scriptsize $\pm$0.4} & 89.6 {\scriptsize $\pm$0.4}  & 92.8 {\scriptsize $\pm$0.2}& 93.7 {\scriptsize $\pm$0.2}& 93.3 {\scriptsize $\pm$0.2}& 75.4 {\scriptsize $\pm$0.3}\\% & 83.5\\
Supervised+\confidence & 87.9 {\scriptsize $\pm$1.0} & 87.9 {\scriptsize $\pm$1.0} & 93.9 {\scriptsize $\pm$0.1} & 93.7 {\scriptsize $\pm$0.1} & 93.8 {\scriptsize $\pm$0.1} & 78.5 {\scriptsize $\pm$0.9}\\% & 83.1\\
Supervised+All &89.6 {\scriptsize $\pm$0.1}& 89.6 {\scriptsize $\pm$0.1} & 94.4 {\scriptsize $\pm$0.1} & 94.2 {\scriptsize $\pm$0.1}& 94.3 {\scriptsize $\pm$0.1} & \bf 80.9 {\scriptsize $\pm$0.1}\\
\midrule    
% \bf MultiRC & & & & & & \\
\multicolumn{7}{c}{\bf MultiRC} \\
Blackbox{\tiny \cite{glockner-etal-2020-think}} & 67.3 {\scriptsize $\pm$1.3} & 67.7 {\scriptsize $\pm$1.6} & & & & \\
Pipeline{\tiny ~\cite{deyoung-etal-2020-eraser}} & 63.3 & 65.0 & 66.7 & 30.2 & 41.6 & 0.0 \\
Supervised {\tiny \cite{glockner-etal-2020-think}} & 65.5  {\scriptsize $\pm$3.6}& 67.7  {\scriptsize $\pm$1.5} & 65.8  {\scriptsize $\pm$0.2} & 42.3  {\scriptsize $\pm$3.9} & 51.4  {\scriptsize $\pm$2.8}& 7.1  {\scriptsize $\pm$2.6}\\ 
\cdashline{1-7}[2pt/2pt]
Supervised & 71.0 {\scriptsize $\pm$0.3}& 71.4 {\scriptsize $\pm$0.3}& 78.0 {\scriptsize $\pm$0.1}& 78.6 {\scriptsize $\pm$0.5}& 78.3 {\scriptsize $\pm$0.1}& 16.2 {\scriptsize $\pm$0.4}\\
Supervised+\consistency & \bf 71.7 {\scriptsize $\pm$0.6} & \bf 72.2 {\scriptsize $\pm$0.7}& \bf 79.9 {\scriptsize $\pm$0.4}& 79.0 {\scriptsize $\pm$0.8}& 79.4 {\scriptsize $\pm$0.5}& \bf 19.3 {\scriptsize $\pm$0.4}\\
Supervised+\faithfulness & 71.0 {\scriptsize $\pm$0.4}& 71.3 {\scriptsize $\pm$0.4}& 78.2 {\scriptsize $\pm$0.1} & 79.1 {\scriptsize $\pm$0.2}& 78.6 {\scriptsize $\pm$0.1} & 16.1 {\scriptsize $\pm$0.5}\\
Supervised+\confidence & 70.6 {\scriptsize $\pm$0.7}& 71.1 {\scriptsize $\pm$0.6}& 77.9 {\scriptsize $\pm$0.8}& 78.3 {\scriptsize $\pm$0.5}& 78.1  {\scriptsize $\pm$0.5}& 16.5  {\scriptsize $\pm$1.0}\\
Supervised+All & 70.5 {\scriptsize $\pm$1.6} & 71.2 {\scriptsize $\pm$1.3}& 79.7 {\scriptsize $\pm$1.1}& \bf 79.4 {\scriptsize $\pm$0.5}& \bf 79.6 {\scriptsize $\pm$0.7}& 18.8 {\scriptsize $\pm$1.6}\\
\midrule
% \bf Movies & & & & & & \\
\multicolumn{7}{c}{\bf Movies} \\
Blackbox{\tiny \cite{glockner-etal-2020-think}} & 90.1 {\scriptsize $\pm$0.3} & 90.1 {\scriptsize $\pm$0.3}  & & & & \\
Pipeline{\tiny ~\cite{deyoung-etal-2020-eraser}}  & 86.0 & 86.0 & 87.9 & 60.5 & 71.7 & 40.7 \\
Supervised {\tiny \cite{glockner-etal-2020-think}} & 85.6 {\scriptsize $\pm$3.6} & 85.8 {\scriptsize$\pm$3.5} & 86.9 {\scriptsize $\pm$2.5} & 62.4 {\scriptsize $\pm$0.1} & 72.6 {\scriptsize $\pm$0.9} & 43.9 {\scriptsize $\pm$0.6}\\ 
\cdashline{1-7}[2pt/2pt]
Supervised & 87.4 {\scriptsize $\pm$0.4} & 87.4 {\scriptsize $\pm$0.4} & 79.6 {\scriptsize $\pm$0.6} & 68.9 {\scriptsize $\pm$0.5} & 73.8 {\scriptsize $\pm$0.5} & 59.4 {\scriptsize $\pm$0.6}\\
Supervised+\consistency & \bf 90.0 {\scriptsize $\pm$0.7}& \bf 90.0 {\scriptsize $\pm$0.7}& 79.5 {\scriptsize $\pm$0.1} & 69.2 {\scriptsize $\pm$0.7} & 74.0 {\scriptsize $\pm$0.8} & 60.8 {\scriptsize $\pm$1.7}\\
Supervised+\faithfulness & 89.1 {\scriptsize $\pm$0.6}& 89.1 {\scriptsize $\pm$0.6} & \bf 80.9 {\scriptsize $\pm$0.9}& \bf 69.9 {\scriptsize $\pm$1.3} & \bf 74.9 {\scriptsize $\pm$1.1} & \bf 62.6 {\scriptsize $\pm$1.6}\\
Supervised+\confidence & 89.9 {\scriptsize $\pm$0.7}&89.9 {\scriptsize $\pm$0.7}& 79.7 {\scriptsize $\pm$1.4}& 69.5 {\scriptsize $\pm$0.7}& 74.3 {\scriptsize $\pm$1.0} & 60.1 {\scriptsize $\pm$2.6}\\
Supervised+All & 89.9 {\scriptsize $\pm$0.7} & 89.9 {\scriptsize $\pm$0.7} & 80.0 {\scriptsize $\pm$1.0}& 69.5 {\scriptsize $\pm$1.0} & 74.4 {\scriptsize $\pm$1.0} & 60.3 {\scriptsize $\pm$2.2}\\
\bottomrule
\end{tabular}
\caption{Target task prediction ($F_1$-C, Accuracy-C) and explanation generation (Precision-E, Recall-E, $F_1$-E) results (mean and standard deviation over three random seed runs). Last columns measures joint prediction of target accuracy and explanation generation. The property with the best relative improvement over the supervised model is in bold.}
\label{tab:results:supervised}
\end{table*}

In Table~\ref{tab:results:supervised}, we
% include the supervised system of ~\citet{glockner-etal-2020-think} as a baseline. While it encodes each sentence separately, our model encodes the whole input together, enabling the model's representations to leverage from the whole document's context. We
see that our supervised model performs better than
%the baseline system
~\citet{glockner-etal-2020-think,deyoung-etal-2020-evidence}.
For the MultiRC dataset, where the explanation consists of more than one sentence, our model brings an improvement of more than 30 $F_1$ points over the reference models, confirming the importance of the contextual information, which performs better than encoding each explanation sentence separately.

When using the diagnostic properties as additional training objectives, we see further improvements in the generated explanations. The most significant improvement is achieved with the \consistency\ property for all datasets with up to 2.5 $F_1$ points over the underlying supervised model. We assume that the \consistency\ objective can be considered as a regularisation for the model's instabilities at the explanation level. The second highest improvement is achieved with the \faithfulness\ property, increasing $F_1$ by up to 1 $F_1$ point for Movies and MultiRC. We assume that the property does not result in improvements for FEVER as it has multiple possible explanation annotations for one instance, which can make the task of selecting one sentence as a complete explanation ambiguous. \confidence\ results in improvements only on Movies. We conjecture that \confidence\ is the least related to promoting similarity to human rationales in the generated explanations. Moreover, the re-calibration of the prediction confidence for both tasks possibly leads to fewer prediction changes, explaining the low scores w.r.t. human annotations. We look into how \confidence\ affects the selected annotations in \S\ref{sec:properties_results}, and \S\ref{sec:discussion7}. Finally, combining all diagnostic property objectives, results in a performance close to the best performing property for each dataset.

\subsection{Target Prediction Results}
~\label{sec:target_results}
In Table~\ref{tab:results:supervised}, the Supervised model, without additional property objectives, consistently improves target task performance by up to 4 points in $F_1$, compared to the two reference models that also generate explanations, except for FEVER, where the models already achieve high results. This can be due to the model encoding all explanation sentences at once, which allows for a more informed prediction of the correct target class. Our model trained jointly with the target task and explanation prediction objective also has similar performance to the BERT Blackbox model and even outperforms it by 4.4 $F_1$ points for the MultiRC dataset. Apart from achieving high target prediction performance ($F_1$-C) on the target task, our supervised model also learns which parts of the input are most important for the prediction, which is an important prerequisite for knowledge-intensive tasks. 

We see further improvements in downstream task performance when using the diagnostic properties as additional training objectives. Improvements of the generated explanations usually lead to improved target prediction as they are conditioned on the extracted evidence. Here, we again see that \consistency\ steadily improves the target task's performance with up to 2.5 $F_1$ points. We also see improvements in $F_1$ with \faithfulness\ for FEVER and MultiRC. Finally, we find that improvements in \confidence\ lead to an improvement for target prediction of 2.5 $F_1$ points for Movies. Combining all objectives, results in performance close the performance of the other properties.

%In Table~\ref{tab:results:supervised}, w
We also show joint prediction results for target task and evidence. For MultiRC and Movies, the improvements of our supervised model over~\citet{glockner-etal-2020-think} are very considerable with up to 9 accuracy points; using diagnostic properties increases results further up to 4 points in accuracy. Apart from improving the properties of the generated explanations, this could be due to the architecture conditioning the prediction on the explanation. The only dataset we do not see improvements for is FEVER, where again the performance is already high, and the target prediction of our model performs worse than \citet{glockner-etal-2020-think}.

\subsection{Explanations Property Results}
~\label{sec:properties_results}
So far, we concentrate on the relative performance improvements compared to human annotations. However, the diagnostic properties' additional training objectives are directed at generating explanations that exhibit these properties to a larger degree. Here, we demonstrate the improvements over the explanation properties themselves for unseen instances in the test splits. Note that this is a control experiment as we expect the properties we optimise for to be improved.

\begin{table}[t]
\centering
% \fontsize{10}{10}\selectfont
\begin{tabular}{llrr}
\toprule
\textbf{Dataset} & \textbf{Method} & \textbf{Suff. $\uparrow$} & \textbf{Compl. $\downarrow$}\\
\midrule
% \multicolumn{3}{l}{\bf FEVER}\\ 
\multirow{2}{*}{\bf FEVER}  & Supervised & 85.1 & 85.1 \\ 
& Supervised+F & 97.4 & 83.6 \\
\midrule
% \multicolumn{3}{l}{\bf MultiRC}\\
\multirow{2}{*}{\bf MultiRC}  & Supervised & 81.7 & 69.2\\
& Supervised+F & 82.3 & 67.0\\ \midrule
% \multicolumn{3}{l}{\bf Movies}\\
\multirow{2}{*}{\bf Movies}  & Supervised & 94.8 & 92.2\\
& Supervised+F & 96.6 & 91.3\\
\bottomrule
\end{tabular}
\caption{Sufficiency and Completeness as proportions of the instances that preserve their prediction when evaluated on only the selected (Suff.) or the unselected (Compl.) explanation sentences, accordingly, for training with and without the \faithfulness\ objective.}~\label{tab:results:faith}
\end{table}

\textbf{\faithfulness.} In Table~\ref{tab:results:faith}  we see that supervision from the \faithfulness\ property leads to generating explanations that preserve the original label of the instance for all datasets. For FEVER, the label is even preserved in 12\% of the instances more than with the supervised objective only. The least faithful explanations are those generated for MultiRC, which can be explained by the low joint performance of both tasks. We also see that even when removing the selected explanations, it is still possible to predict the same label based on the remaining evidence. Such cases are decreased when including the \faithfulness\ property. The latter phenomenon can be explained by the fact that FEVER and Movies' instances contain several possible explanations. We conjecture that this might also be due to the model learning spurious correlations. We further study this in Sec.~\ref{sec:bias}.

\begin{table}[t]
\centering
% \fontsize{10}{10}\selectfont
\begin{tabular}{llrr}
\toprule
\textbf{Dataset} &\textbf{Method} & \textbf{Pred.} & \textbf{Expl.}\\
\midrule
% \multicolumn{3}{l}{\bf FEVER}\\
\multirow{2}{*}{\bf FEVER} & Sup. & 0.03 (9.9e-8) & 3.68 (1.80)\\
 & Sup.+DC & 0.02 (9.1e-8) & 2.56 (0.97)\\ \midrule
% \multicolumn{3}{l}{\bf MultiRC}\\
\multirow{2}{*}{\bf MultiRC} & Sup. & 0.09 (5.6e-8) & 7.83(2.87)\\
 & Sup.+DC & 0.05 (4.9e-8) & 3.01(0.89)\\ \midrule
% \multicolumn{3}{l}{\bf Movies}\\
\multirow{2}{*}{\bf Movies} & Sup. & 0.04 (7.1e-8)& 2.34 (1.38)\\
 & Sup.+DC & 0.01 (6.2e-8) & 1.72 (0.90)\\
\bottomrule
\end{tabular}
\caption{Mean and standard deviation (in brackets) of the difference between target (Pred.) and explanation (Expl.) prediction confidence for similar (masked) instances.}
\label{tab:results:stability}
\end{table}
\begin{table}[t]
\centering
% \fontsize{10}{10}\selectfont
\begin{tabular}{lrrr}
\toprule
\textbf{Method} & \textbf{FEVER} & \textbf{MultiRC} & \textbf{Movies}\\
\midrule
Sup. & 0.10 (0.17) & 0.05 (0.10) & 0.12 (0.09)\\
Sup.+CI & 0.05 (0.09) & 0.04 (0.09) & 0.05 (0.10) \\
\bottomrule
\end{tabular}
\caption{Mean and standard deviation (in brackets) difference between the model's confidence and the confidence of the generated explanations.}
\label{tab:results:confidence}
\end{table}

\textbf{\consistency.} Using \consistency\ as an additional training objective aims to regularise the model to select similar explanations for similar instances. In Table~\ref{tab:results:stability}, we find the variance of downstream task prediction confidence decreases for all datasets with up to 0.04 points. Furthermore, the variance of generated explanation probabilities for similar instances is decreased as well. The largest improvements are for MultiRC and Movies, where the property brings the highest performance improvement w.r.t. human annotations as well. We also find that the Movies dataset, which has the longest inputs, has the smallest variance in explanation predictions. This suggests that the variance in explanation prediction is more pronounced for shorter inputs as in FEVER and MultiRC, where the property brings more improvement w.r.t. human annotations. The variance could also depend on the dataset's nature.

\textbf{\confidence.} Table~\ref{tab:results:confidence} shows the difference between the confidence of the predicted target label and the confidence of the explanation sentence with the highest importance. Including \confidence\ as a training objective indeed decreases the distance between the confidence of the two tasks, making it easier to judge the confidence of the model only based on the generated explanation's confidence. The confidence is most prominently improved for the Movies dataset, where it is also the dataset with the largest improvements for supervised explanation generation with \confidence\ objective. 

\subsection{Unsupervised Rationale Generation}
~\label{sec:unsup}
We explore how well explanations can be generated without supervision from human explanation annotations.
Table~\ref{tab:results:usupervised} shows that the performance of the unsupervised rationales is limited with an up to 47 $F_1$ point decrease for FEVER compared to the supervised model. 
We assume that as our model encodes the whole input together, this leads to a uniform importance of all sentences as they share information through their context. While joint encoding improves the target prediction for complex reasoning datasets especially with more than one explanation sentence, this also limits the unsupervised learning potential of our architecture. As the model is not supervised to select explanations close to human ones, improving the diagnostic properties has a limited effect in improving the results w.r.t. human annotations.

\begin{table}[t]
\centering
% \fontsize{10}{10}\selectfont
\begin{tabular}{lrrr}
\toprule
\textbf{Method} & \textbf{FEVER} & \textbf{MultiRC} & \textbf{Movies}\\ \midrule
Sup. & 93.9 {\scriptsize $\pm$0.1} & 78.3 {\scriptsize $\pm$0.1} & 73.8 {\scriptsize $\pm$0.5}\\
UnS.   & 56.1 {\scriptsize $\pm$0.4}& 34.8 {\scriptsize $\pm$7.6}& 50.0 {\scriptsize $\pm$1.8}\\ 
UnS.+DC & 46.9 {\scriptsize $\pm$0.4}& 38.1 {\scriptsize $\pm$3.2} & 63.8 {\scriptsize $\pm$1.2}\\
UnS.+F & 51.6 {\scriptsize $\pm$0.3}& 24.4 {\scriptsize $\pm$5.2} & 64.6 {\scriptsize $\pm$0.4}\\
UnS.+CI & 57.5 {\scriptsize $\pm$0.4}& 25.4 {\scriptsize $\pm$3.4}& 60.0 {\scriptsize $\pm$1.6}\\ 
UnS.+All & 57.3 {\scriptsize $\pm$0.2}& 37.4 {\scriptsize $\pm$6.4}& 63.6 {\scriptsize $\pm$0.3}\\ 
\bottomrule
\end{tabular}
\caption{Performance on the explanation generation task without human annotation supervision (UnS.).}
\label{tab:results:usupervised}
\end{table}

\section{Discussion}
\label{sec:discussion7}
\subsection{Question/Claim Only Bias}
\label{sec:bias}
Prior work has found that models can learn spurious correlations between the target task and portions of the input text, e.g., predicting solely based on the claim to be fact checked ~\cite{schuster-etal-2019-towards}, regardless of the provided evidence. In our experiments, the input for FEVER and MultiRC also contains two parts - a claim or a question-answer pair and evidence text, where the correct prediction of the target always depends on the evidence.  
Suppose the models do not consider the second part of the input when predicting the target task. In that case, efforts to improve the generated explanations will not affect the target task prediction as it does not rely on that part of the input.

\begin{table}[t]
\centering
% \fontsize{10}{10}\selectfont
\begin{tabular}{llll}
\toprule
\textbf{Dataset} & \textbf{Method} & \textbf{$\mathbf{F_1}$-C} & \textbf{Acc-C}\\
\midrule
% \multicolumn{3}{l}{\bf FEVER}\\
\multirow{7}{*}{\bf FEVER} & Random & 26.1 {\scriptsize $\pm$4.3}& 37.1 {\scriptsize $\pm$5.6}\\
& Sup. & 75.6 {\scriptsize $\pm$0.3} & 75.7 {\scriptsize $\pm$0.3} \\ 
& Sup.+DC & 68.2 {\scriptsize $\pm$0.2} & 75.6 {\scriptsize $\pm$0.3} \\
& Sup.+F & 73.4 {\scriptsize $\pm$0.4} & 73.9 {\scriptsize $\pm$0.3} \\
& Sup.+CI & 73.2 {\scriptsize $\pm$0.4} & 73.7 {\scriptsize $\pm$0.4}\\ 
& Sup.+All &  73.5 {\scriptsize $\pm$0.2} & 73.8 {\scriptsize $\pm$0.4} \\
& Sup. on whole input & 89.3  {\scriptsize $\pm$0.4}& 89.4  {\scriptsize $\pm$0.3}\\ 
\midrule
% \multicolumn{3}{l}{\bf MultiRC}\\
\multirow{7}{*}{\bf MultiRC} & Random & 26.1 {\scriptsize $\pm$5.5} & 31.6 {\scriptsize $\pm$5.9}\\
& Sup. & 59.4 {\scriptsize $\pm$0.8} & 63.5 {\scriptsize $\pm$0.9} \\
& Sup.+DC & 54.5 {\scriptsize $\pm$0.9}& 61.3 {\scriptsize $\pm$1.2}\\
& Sup.+F & 57.8 {\scriptsize $\pm$0.8} & 61.4 {\scriptsize $\pm$0.6}\\
& Sup.+CI & 49.7 {\scriptsize $\pm$0.8}& 60.1 {\scriptsize $\pm$0.2}\\
& Sup.+All & 59.0 {\scriptsize $\pm$0.3} & 61.0 {\scriptsize $\pm$0.2}\\
& Sup. on whole input & 71.0 {\scriptsize $\pm$0.3}& 71.4 {\scriptsize $\pm$0.3}\\ 
\bottomrule
\end{tabular}
\caption{Performance of the models for the downstream task when provided with the query-answer part only.}
\label{tab:results:adversarial}
\end{table}

Table~\ref{tab:results:adversarial} shows target task performance of models trained on the whole input, but using only the first part of the input at test time. We find that, given the limited input, the performance is still considerable compared to a random prediction. For FEVER, the performance drops only with 14 $F_1$ score points to 75.6 $F_1$ score. This could explain the small relative improvements for FEVER when including diagnostic properties as training objectives, where the prediction does not rely on the explanation to a large extent.

Another interesting finding is that including diagnostic properties as training objectives decreases models' performance when a supporting document is not provided. We assume this indicates the properties guide the model to rely more on information in the document than to learn spurious correlations between the question/claim and the target only. The \consistency\ and \confidence\ property lead to the largest decrease in model's performance on the limited input. This points to two potent objectives for reducing spurious correlations.

\subsection{Explanation Examples}
\begin{table}[t]
% \fontsize{10}{10}\selectfont
\small
\centering
\begin{tabular}{p{425pt}}
\toprule
\textbf{Question:} What colors are definitely used in the picture Lucy drew?; \textbf{Answer:} Yellow and purple; \textbf{Label:} True\\
\textbf{Predicted: Sup} True, p=.98; \textbf{Sup+DC} True, p=.99 \\
\textbf{E-Sup:} She draws a picture of her family. She makes sure to draw her mom named Martha wearing a purple dress, because that is her favorite. She draws many yellow feathers for her pet bird named Andy.\\
\textbf{E-Sup+S:} She makes sure to draw her mom named Martha wearing a purple dress, because that is her favorite. She draws many yellow feathers for her pet bird named Andy.\\
\midrule
\textbf{Claim:} Zoey Deutch did not portray Rosemarie Hathaway in Vampire Academy.; \textbf{Label:} REFUTE\\
\textbf{Predicted: Sup} refute, p=.99; \textbf{Sup+F} refute, p=.99\\
\textbf{E-Sup:} Zoey Francis Thompson Deutch (born November 10, 1994) is an American actress. \\
\textbf{E-Sup+F:} She is known for portraying Rosemarie ``Rose'' Hathaway in Vampire Academy(2014), Beverly in the Richard Link later film Everybody Wants Some!! \\ \midrule
\textbf{E-Sup/E-Sup+CI}: For me, they calibrated my creativity as a child; they are masterful, original works of art that mix moving stories with what were astonishing special effects at the time (and they still hold up pretty well).; \textbf{Label:} Positive\\
\textbf{Predicted: Sup} negative, p=.99 \textbf{Sup+CI} positive, p=.99\\

\bottomrule
\end{tabular}
\caption{Example explanation predictions changed by including the diagnostic properties as training objectives.}

\label{tab:examples7}
\end{table}

Table~\ref{tab:examples7} illustrates common effects of the diagnostic properties. We find \consistency\ commonly improves explanations by removing sentences unrelated to the target prediction, as in the first example from MultiRC. This is particularly useful for MultiRC, which has multiple gold explanation sentences. For FEVER and Movies, where one  sentence is needed, the property brings smaller improvements w.r.t. human explanation annotations. 
%In some cases, this effect of the property also removes sentences that are part of the explanation, which we attribute to the fact that each sentence also contains contextual information about other sentences allowing the removal of some explanation sentences without hurting the target task prediction.

The second example from FEVER illustrates the effect of including \faithfulness\ as an objective. Naturally, for instances classified correctly by the supervised model, their generated explanation is improved to reflect the rationale used to predict the target. However, when the prediction is incorrect, the effect of the \faithfulness\ property is limited.

Finally, we find \confidence\ often re-calibrates the prediction probabilities of generated explanations and predicted target tasks, which does not change many target predictions. This explains its limited effect as an additional training objective. The re-calibration also influences downstream task prediction confidence, as in the last example from the Movies dataset. This is a side effect of optimising the property while training the target task% as well
, where both explanation and target prediction confidence can be changed to achieve better alignment.

\section{Conclusion}
In this paper, we study the use of diagnostic properties for improving the quality of generated explanations. % for downstream tasks. 
We find that including them as additional training objectives improves downstream task performance and generated explanations w.r.t. human rationale annotations. Moreover, using only the diagnostic properties as training objectives does not lead to a good performance compared to only using human rationale annotations. The latter indicates the need for human rationale annotations for supervising a model to base its predictions on the correct rationales. In future, we plan to experiment with application tasks with longer inputs, where current architectures have to be adjusted to make it computationally possible to encode longer inputs.

\section*{Acknowledgments}
$\begin{array}{l}\includegraphics[width=1cm]{euflag2.png} \end{array}$
The research documented in this paper has received funding from the European Union's Horizon 2020 research and innovation programme under the Marie Sk\l{}odowska-Curie grant agreement No 801199. Isabelle Augenstein's research is further partially funded by a DFF Sapere Aude research leader grant.

\clearpage
\bibliography{anthology,references}
\bibliographystyle{acl_natbib}
% -------------------------- 
% Back matter
% --------------------------
% \input{frontbackmatter/bibliography.tex}

\end{document}